\PassOptionsToPackage{left=2.5cm,right=2.2cm,top=2cm,bottom=3cm}{geometry}
\documentclass[pdflatex,sn-mathphys-num]{sn-jnl-arxiv}
\usepackage{utils/utils}

\begin{document}
\begin{bibunit}[sn-mathphys-num]
    \title[Article Title]{Agentic Fusion of Large Atomic and Language Models to Accelerate Superconductor Discovery}

\author[1]{Mingze Li}
\equalcont{These authors contributed equally to this work.}
\author*[2,3]{Yu Rong}\email{yu.rong@hotmail.com}
\equalcont{These authors contributed equally to this work.}
\author[1]{Songyou Li}
\equalcont{These authors contributed equally to this work.}
\author[4]{Lihong Wang}
\equalcont{These authors contributed equally to this work.}
\author[1]{Jiacheng Cen}
\author[1]{Liming Wu}
\author[1]{Anyi Li}
\author[1]{Zongzhao Li}
\author[4]{Qiuliang Liu}
\author[5]{Rui Jiao}
\author[2,3]{Tian Bian}
\author[2]{Pengju Wang}
\author[1]{Hao Sun}
\author[2,3]{Jianfeng Zhang}
\author[1]{Ji-Rong Wen}
\author*[2,3]{Deli Zhao}\email{zhaodeli@gmail.com}
\author*[4]{Shifeng Jin}\email{shifengjin@iphy.ac.cn}
\author*[2,3]{Tingyang Xu}\email{xuty\_007@hotmail.com}
\author*[1]{Wenbing Huang}\email{hwenbing@ruc.edu.cn}

\affil[1]{Gaoling School of Artificial Intelligence, Renmin University of China, Beijing, China}

\affil[2]{DAMO Academy, Alibaba Group, Hangzhou, China}

\affil[3]{Hupan Lab, Hangzhou, China}

\affil[4]{Institution of Physics, University of the Chinese Academy of Sciences, Beijing, China}

\affil[5]{Department of Computer Science and Technology, Tsinghua University, Beijing, China}

\abstract{
Artificial intelligence has accelerated materials discovery through high-throughput prediction and generation, yet the decision problem remains a formidable bottleneck. While current AI systems readily propose millions of candidates, navigating the decision regarding a viable experimental target requires resolving multi-dimensional judgments across atomic-scale numerical computation and high-level semantic reasoning. 
Here we present \agentname{}, an agentic framework for materials discovery that orchestrates 
a suite of Large Atomic Model (LAM) tools finetuned from our proposed 1-billion-parameter model \modelname{} for 
numerical computation, while leveraging Large Language Models (LLMs) for semantic reasoning. 
Applied to superconductors, \agentname{} rediscovers 66 experimentally verified superconductors that are absent from the standard SuperCon3D database. Scaling to 2.4 million equilibrium crystals, \agentname{} identifies 68,000 high-confidence candidates in just 28 GPU hours\footnote{The complete dataset of screened superconductors is available at \url{https://developer.damo-academy.com/material}.}, expanding known superconducting space by orders of magnitude compared to datasets curated over decades. 
Guided by the agent's reasoning, we experimentally synthesize and verify four novel superconductors: the motif-guided Zr$_3$ScRe$_8$ ($\tc$ = 6.5 K), the de novo generated HfZrRe$_4$ ($\tc$ = 5.9 K), the structurally reinterpreted Zr$_4$VRe$_7$ ($\tc$ = 3.5 K), and the database-latent Hf$_{21}$Re$_{25}$ ($\tc$ = 2.5 K). Together, our results establish a knowledge integrated, autonomously orchestrated, and experimentally grounded paradigm for materials discovery.}

\maketitle

    \section{Introduction}\label{sec1}

The imperative to accelerate materials discovery for the global energy and quantum technology transitions has driven a profound paradigm shift in the physical sciences~\citep{para_shift,material_energy}. Traditionally, discovery was dictated by the interplay between Edisonian intuition and high-fidelity yet computationally prohibitive first-principles calculations~\citep{para_analysis,mater_compute_weak,expert_DFT_cata}. Deep learning has fundamentally reshaped this landscape, offering a data-driven framework to reconcile classical speed-accuracy trade-offs. Key milestones include machine-learning-based property predictors~\citep{CGCNN,alignn,MEGNet,DimeNet,SchNet,coGN_coNGN,mattersim}, machine-learning-based force fields~\citep{genome,GNO,EquiformerV2,NequIP,allergo,Mace,Mace-MP-0,DPA-2}, and deep generative models for both structure prediction~\citep{cdvae,diffcsp,FlowMM,CrysFlow,CrysBFN} and \textit{de novo} design~\citep{mattergen}. Broadly, the AI-for-materials landscape has transitioned from predictive paradigms like GNoME~\citep{genome}, which aggressively scale stability assessments, to generative paradigms like MatterGen~\citep{mattergen}, which enable the inverse design of crystals. While current predictive or generative models can propose millions of candidates, a central bottleneck remains: the decision problem. Navigating the decision regarding a viable experimental target requires reconciling multi-dimensional judgments across atomic-scale numerical computation (\emph{e.g.}, structure generation, property prediction, and thermodynamic stability evaluation) and high-level semantic reasoning (\emph{e.g.}, novelty and synthetic feasibility verification).

To overcome these limitations, an agentic paradigm offers a vital path to elevate materials discovery from isolated processes into integrated frameworks. In contrast to isolated predictors or generators, a scientific agent should coordinate specialized numerical models, literature synthesis and human guidance to formulate and test experimentally actionable hypotheses. Such a system must decide what to predict, where to retrieve literature, how to create a new predictive skill, when to generate structures beyond existing databases, and when to reject a candidate as unstable or synthetically inaccessible. By coupling quantitative precision with semantic reasoning, agentic discovery opens access to vast chemical landscapes that remain impenetrable to traditional models. Complementary to closed-loop laboratory platforms such as A-Lab \citep{A-lab}, which focus on automated synthesis, this paper emphasizes upstream scientific decision-making spanning from target selection to experimental prioritization.

\begin{figure}[t!]
    \centering
    \includegraphics[width=\textwidth]{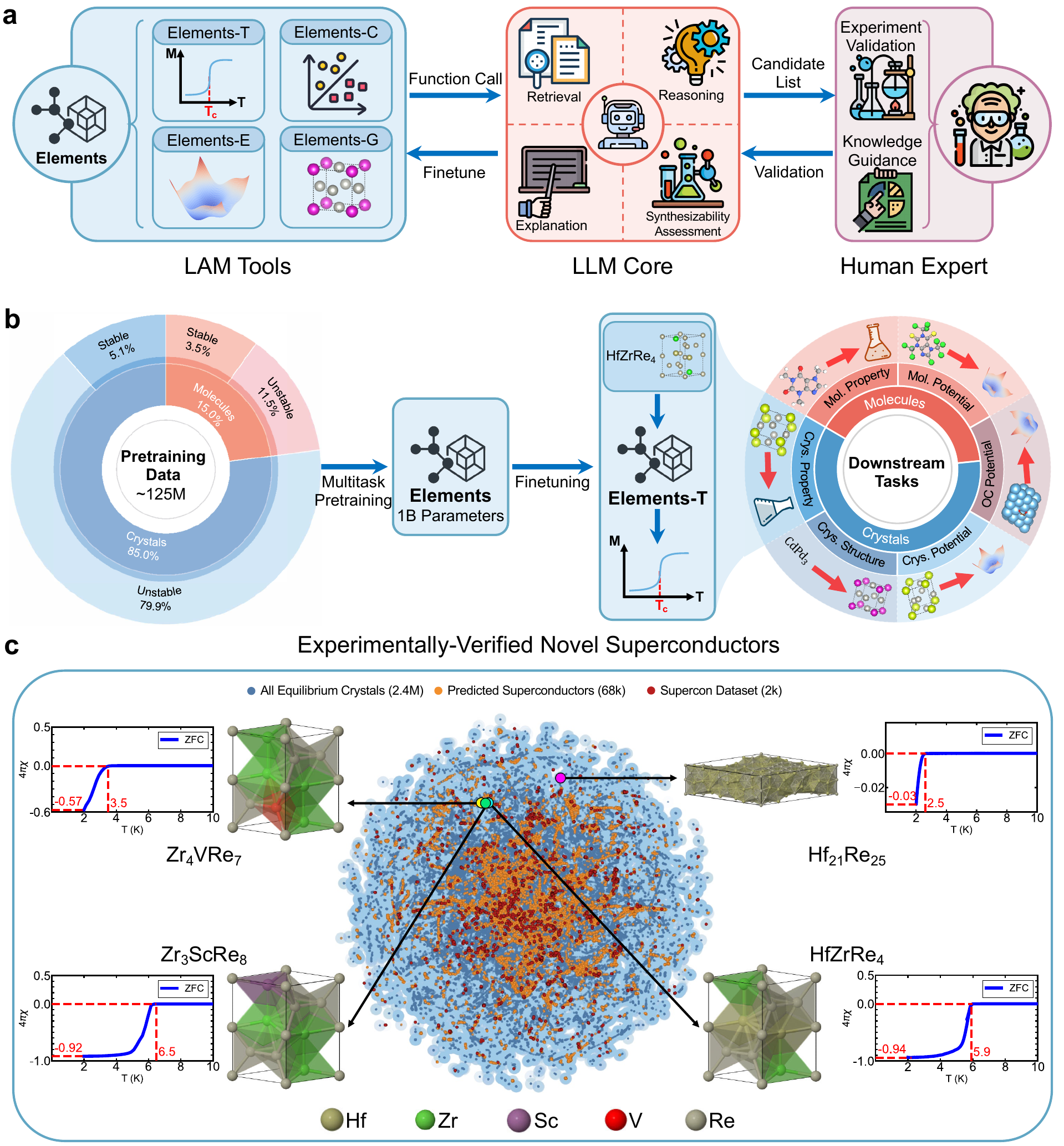}
    \caption{Overview of our methodologies and results. \textbf{a}, The agentic framework. \agentname{} integrates specialized \modelname{} variants (\modelname{}-\textsf{T/C/E/G}) and LLMs' reasoning to identify candidates for experimental validation. \textbf{b}, Omni-domain large atomic model development. \modelname{} is pretrained on $125$ million structures, enabling seamless adaptation across diverse downstream tasks. \textbf{c}, Large-scale superconductor discovery. By leveraging \modelname{T} (predicted $\tc>4~\text{K}$) and \modelname{C} (positive output), \agentname{} screens across 2.4 million distinct stable crystals and identifies a repository of $68\text{k}$ potential superconductors, significantly enriching the existing SuperCon database, which contains just about 2k ordered crystals in its deduplicated version. Highlighted are four experimentally validated novel superconductors:  $\text{Zr}_3\text{ScRe}_8$ ($\tc = 6.5~\text{K}$), $\text{HfZrRe}_4$ ($\tc = 5.9~\text{K}$), $\text{Zr}_4\text{VRe}_7$ ($\tc = 3.5~\text{K}$), and $\text{Hf}_{21}\text{Re}_{25}$ ($\tc = 2.5~\text{K}$). Flanking the central feature space map are their predicted crystal structures and temperature-dependent magnetic susceptibility measurements, directly confirming the superconducting transitions at the indicated $\tc$ values.}
    \label{fig:elements_overview}
\end{figure}

Here, we present \agentname{}, an agentic framework for materials discovery that orchestrates a suite of Large Atomic Models (LAMs) finetuned from our proposed model \modelname{} for atomic-scale numerical computation, while leveraging Large Language Models (LLMs) for high-level semantic reasoning (\cref{fig:elements_overview}a). The entire process remains harmonized through human oversight and prompting.
Pretrained on an extensive corpus of 125 million structures, the 1-billion-parameter \modelname{} encodes a unified representation across a diverse chemical landscape, bridging equilibrium phases with non-equilibrium configurations and linking periodic crystals with molecular systems (\cref{fig:elements_overview}b). Leveraging this large-scale omni-domain pretraining, \modelname{} serves as the foundation for the specialized tools of \agentname{} across various capabilities such as property and structure prediction.
Transcending mere tool invocation, \agentname{} achieves self-evolution by finetuning \modelname{} to create new tools using insights distilled from literature. While contemporary ``AI Scientist'' automate broad scientific workflows~\citep{Nature_AI_sci,lu2026towards,agenticscience}, \agentname{} distinguishes itself through the meticulously designed architecture of prompts and evolving tools, delivering the superior fidelity essential for rigorous materials discovery.

As a demanding proof-of-concept, we apply this framework to superconductor discovery, a field traditionally hindered by extreme chemical complexity~\cite{chemical_complexity_1, chemical_complexity_2} and data scarcity~\cite{data_scarcity}. 
We rigorously validate the agent's decision-making across both retrospective and prospective discovery regimes. In retrospective knowledge reconstruction (\cref{fig:supercon_screen}), \agentname{} systematically screens existing databases and literature to classify verified superconductors and non-superconductors. Through literature reasoning, it achieves a rediscovery success rate exceeding $40\%$ for literature-verified superconductors with high predicted $\tc$, while simultaneously identifying 66 superconductors that are entirely absent from the standard SuperCon3D~\cite{sodnet} database. This demonstrates that the agent can reconstruct dispersed experimental knowledge to resolve systemic database incompleteness.
In prospective discovery (\cref{fig:elements_overview}c and \cref{fig:supercon_discovery}), \agentname{} first achieves self-evolution by finetuning \modelname{} to create the new tool \modelname{C} using insights distilled from literature. With the extended tools and skills, \agentname{} screens over 2.4 million crystals to yield 68,000 high-confidence superconductor candidates within only 28 H20 GPU hours, vastly expanding the known superconducting space of SuperCon~\cite{supercon} which contains only 2,000 ordered crystals collected over decades. Guided by the agent's reasoning, we experimentally synthesize and verify four novel superconductors: the motif-guided Zr$_3$ScRe$_8$ ($\tc$ = 6.5 K), the de novo generated HfZrRe$_4$ ($\tc$ = 5.9 K), the structurally reinterpreted Zr$_4$VRe$_7$ ($\tc$ = 3.5 K), and the database-latent Hf$_{21}$Re$_{25}$ ($\tc$ = 2.5 K). Crucially, our framework extends readily to other complex material classes as the agent
can in principle create desired tools by finetuning \modelname{} with domain-specific data.

    \section{Results}

\subsection{The Multi-Stage Agentic Discovery Process of \agentname}

\agentname{} is an LLM-based agentic system designed to autonomously execute multi-stage exploration strategies. To capture complex interatomic interactions with high fidelity, its core capabilities are built upon \modelname{}, which serves as the foundation for a suite of specialized functional tools. These include \modelname{T} for superconducting property prediction, \modelname{C} for superconductivity classification, \modelname{E} for thermodynamic stability assessment, and \modelname{G} for generative crystal structure prediction. Beyond these internal modules, \agentname{} integrates open-source toolkits, such as \texttt{pymatgen}, to leverage standardized computational protocols. Crucially, the system does not merely invoke these tools in isolation; rather, it adaptively coordinates both internal and external modules, composing task-specific toolchains to address the unique requirements of each discovery stage. 

\begin{figure}[!t]
    \centering
    \includegraphics[width=\textwidth]{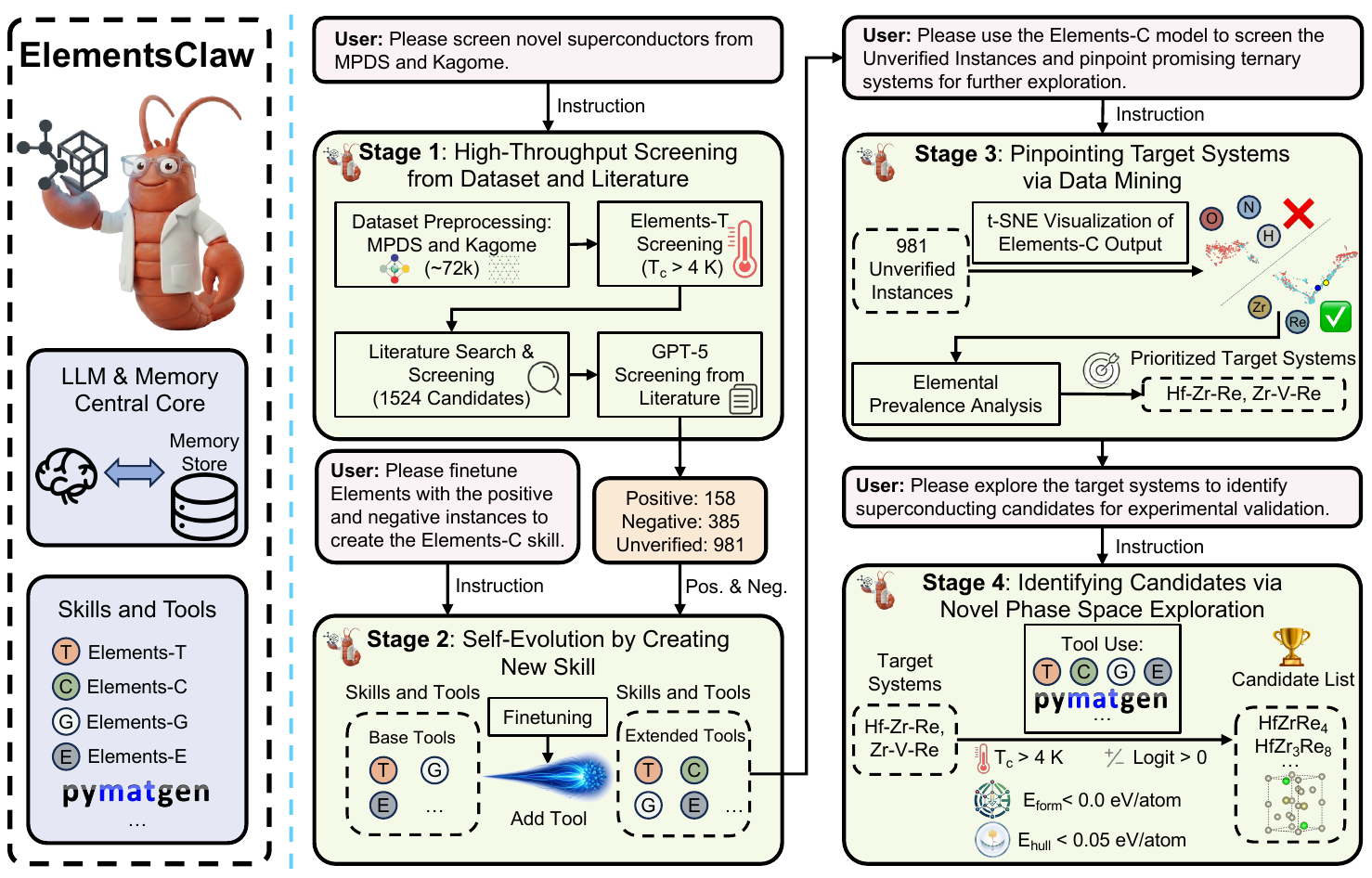}
    \caption{The Multi-Stage Agentic Discovery Process of \agentname{}. Guided by human expertise, the agent first curates labeled superconducting instances via \modelname{T} screening and GPT-5-driven literature synthesis from existing datasets (Stage 1). It then creates the \modelname{C} classification skill through finetuning \modelname{} (Stage 2), enabling the systematic screening of unverified instances to pinpoint promising ternary systems (Stage 3). Finally, \agentname{} identifies high-probability phases based on these identified systems for downstream experimental validation (Stage 4). Notably, although we illustrate only a single representative dialogue for each stage, the actual process involves granular, multi-turn interactions between the user and \agentname{}. Through this workflow, \agentname{} manifests critical capabilities in high-throughput screening, autonomous self-evolution, deep data mining, and novel phase-space exploration.}
    \label{fig:ElementsClaw}
\end{figure}

As illustrated in \cref{fig:ElementsClaw}, \agentname{} conducts a four-stage materials discovery pipeline, autonomously planning and executing a sequence of actions in response to user instructions. The agent first performs large-scale \modelname{T} screening and GPT-5-driven literature synthesis across the MPDS~\citep{MPDS, pauling} and Kagome~\citep{kagome} datasets---comprising 72,000 materials in total---to curate a labeled dataset of 158 positive, 385 negative, and 981 unverified instances (Stage 1). Subsequently, \agentname{} creates the \modelname{C} skill by finetuning \modelname{} on these positive and negative instances, specializing the foundation model for high-fidelity classification (Stage 2). This refined capability is then deployed to systematically screen unverified instances and pinpoint promising ternary systems, Zr--V--Re and Hf--Zr--Re, for further exploration (Stage 3). Finally, based on these identified systems, \agentname{} prioritizes high-probability candidate phases for downstream experimental synthesis and validation (Stage 4). While demonstrated here within the context of superconductor discovery, each stage illustrates a distinct facet of our agent's capabilities. Stage 1 establishes a rigorous protocol for high-throughput screening across diverse datasets and scientific literature. Stage 2 exemplifies the agent's capacity for autonomous self-evolution via targeted skill acquisition. Stage 3 underscores the use of sophisticated data mining to extract actionable insights from unverified domains, while Stage 4 showcases \agentname{}'s ability to navigate the novel phase space under specific target constraints. Collectively, these stages define a versatile paradigm applicable across a broad spectrum of material classes.

The subsequent sections are organized as follows: we first reveal the emergent abilities of \modelname{} in \cref{sec:elements}, followed by its specialization for critical temperature prediction in \cref{sec:Elements-T}. We then detail the outcomes for each discovery stage of \cref{fig:ElementsClaw} in \cref{sec:stage-1,sec:stage-23,sec:experiments}. More details related to the methodologies are provided in \cref{sec:methods}.

\subsection{\modelname~as the Foundational Tool Base for \agentname}
\label{sec:elements}

The superior performance of \modelname{} as an omni-domain atomic foundation model stems from its vast pretraining corpus, optimized equivariant architecture, and a novel multitask pretraining strategy. We curate the Molecule-Crystal DataBase (\databasename{}), a massive-scale repository encompassing 125.21 million configurations. To achieve structural universality, the dataset maintains a strategic balance between periodic crystal structures (85.1\%) and non-periodic molecular geometries (14.9\%). Crucially, to capture the intricate landscape of interatomic potentials, we include both equilibrium ``stable'' states and high-energy ``unstable'' configurations. This diverse manifold provides the essential gradient information required to learn robust non-equilibrium behaviors across disparate chemical domains. 
Built upon an EquiformerV2 backbone~\citep{EquiformerV2}, \modelname{} incorporates targeted innovations to facilitate unified geometric representation. %
The pretraining of \modelname{} utilizes a multitask strategy designed to bridge structural denoising and force field modeling on \databasename{}. For equilibrium inputs, \modelname{} is tasked with a denoising objective, in which specialized output heads predict the synthetic noise added to atomic coordinates (for all systems) and lattice parameters (for crystals only). For non-equilibrium inputs, the model operates as a neural interatomic potential, concurrently predicting total energies and atomic forces. This multitask pretraining framework ensures that \modelname{} internalizes both the static structural signatures of stable matter and the dynamic force fields governing structural evolution. A comprehensive description of the \modelname{} architecture is provided in the Methods (\cref{sec:arch} and \cref{fig:ext_model}).

\begin{figure}[!t]
    \centering
    \includegraphics[width=\textwidth]{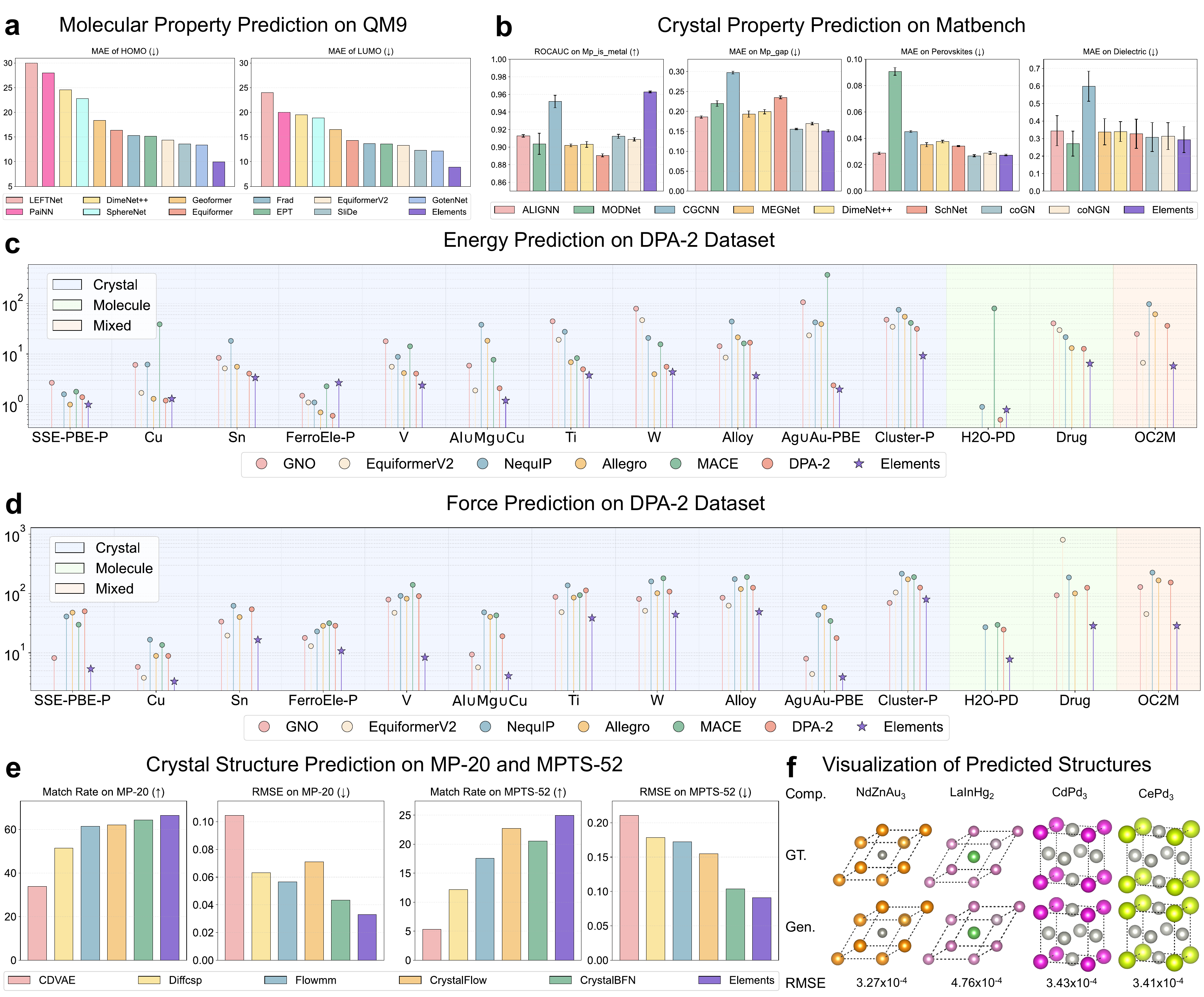}
    \caption{Comparisons between \modelname{} and SOTA methods across $22$ downstream tasks in terms of property prediction, interatomic potential estimation, and structure prediction. The results of all compared methods are directly copied from the corresponding benchmarks.  \textbf{a}, MAE results of molecular property prediction (HOMO and LUMO) on the QM9 dataset containing stable molecules. \textbf{b}, MAE results of crystal property prediction (MP\_is\_metal, Mp\_gap, Perovskites, and Dielectric) on Matbench containing stable crystals. \textbf{c-d}, RMSEs of energy (c) and force prediction (d) on the DPA-2 dataset containing unstable molecules and crystals. \textbf{e}, Match Rate and RMSE results of crystal structure prediction on MP-20 and MPTS-52, both of which contain stable crystals. \textbf{f}, Visualization of generated structures (Gen.) compared with Ground-Truth structures (GT.) given the same composition, on the the MP-20 dataset.}
    \label{fig:Fig3}
\end{figure}

To evaluate the versatility of \modelname{}, we finetune the pretrained model on $22$ downstream tasks covering three fundamental pillars of materials discovery: property prediction of stable systems, interatomic potential estimation of non-equilibrium systems, and stable crystal structure prediction.
For property prediction, we assess the performance of \modelname{} on the QM9 molecular dataset~\citep{qm9}, and on selected Matbench crystal datasets~\citep{matbench} specifically relevant to superconductivity. As shown in \cref{fig:Fig3}a, \modelname{} achieves SOTA performance on two key electronic properties, HOMO and LUMO, outperforming the previous leading method, GotenNet~\citep{gotennet}, by approximately 30\%. On the Matbench benchmark (\cref{fig:Fig3}b), \modelname{} attains SOTA results in predicting ``MP\_is\_metal'' and ``Mp\_gap''. Accurate prediction of these two properties is intrinsically tied to evaluating superconducting potential: establishing metallicity is the crucial first step, as non-metallic ground states are highly unlikely to host superconductivity, while the band gap fundamentally dictates the electrical nature of a material (\ie, whether it behaves as a conductor, semiconductor, or insulator). Furthermore, \modelname{} is ranked as the runner-up for the prediction of Perovskites and Dielectric properties. This is highly relevant to our discovery workflow, as many high-$\tc$ superconductors (\eg, cuprates) adopt perovskite-like structural motifs, and dielectric properties often correlate with the electron-phonon coupling and structural instabilities that drive superconducting phases. Notably, while some specialized methods show inconsistent performance across different benchmarks, \modelname{} maintains robust generalization throughout all evaluated properties.

To examine the interatomic potential estimation, we evaluate \modelname{} on the DPA-2 dataset~\citep{DPA-2}, an extensive benchmark for energy and atomic force prediction across molecules, crystals, and their mixed interfaces. As illustrated in \cref{fig:Fig3}c and \cref{fig:Fig3}d, \modelname{} demonstrates a clear advantage over competing methods across $14$ diverse categories of atomic systems. Specifically, in energy prediction, our model shows superior robustness, particularly in challenging systems (such as ``Ag$\cup$Au-PBE'' and ``Cluster-P''), where baseline methods exhibit high error margins. In terms of atomic force prediction, \modelname{} consistently achieves the highest accuracy across all evaluated categories, underscoring its precision in capturing fine-grained structural gradients.

Finally, we extend \modelname{} to generative tasks by integrating it into a structural prediction pipeline. Following the DiffCSP framework~\citep{diffcsp}, we replace the original invariant message-passing module with \modelname{} and utilize our pretrained denoising heads for coordinate and lattice refinement. Evaluated on the MP-20 and MPTS-52 datasets (\cref{fig:Fig3}e), this approach achieves the best performance, with the Match Rate on MPTS-52 more than doubling that of the original DiffCSP. Such a substantial improvement underscores the transformative impact of employing a high-capacity pretrained foundation model as a backbone for complex generative materials discovery. The high-fidelity nature of these generated structures is further exemplified in \cref{fig:Fig3}f, which visualizes the generated crystalline structures on the MP-20 benchmark.
Collectively, the results in \cref{fig:Fig3} establish \modelname{} as a robust foundation model for capturing complex interatomic interactions, providing a high-capacity backbone that significantly empowers the autonomous discovery of novel superconductors.

\begin{figure}[t!]
 \centering
 \includegraphics[width=\textwidth]{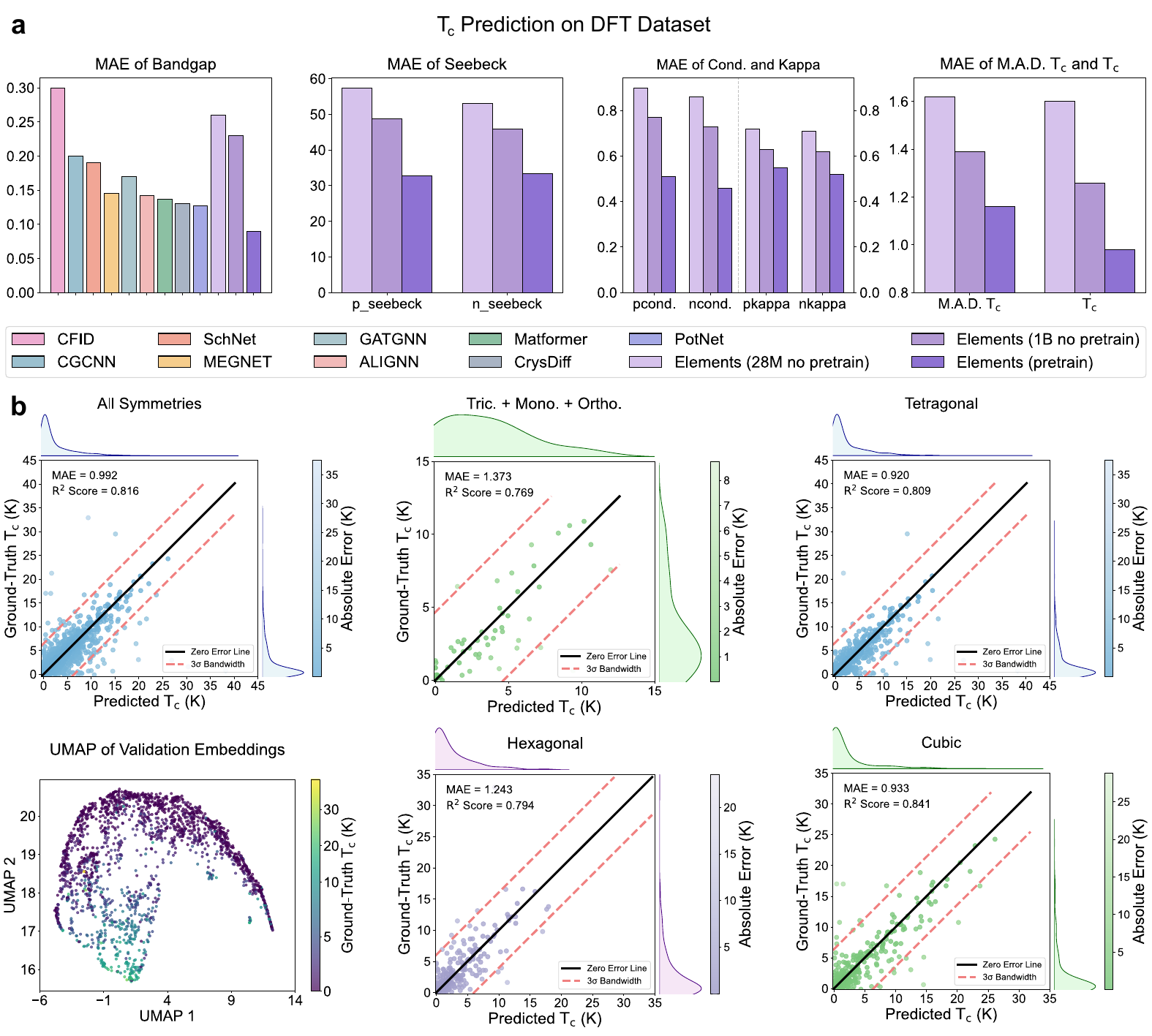}
\vskip -0.in
    \caption{The performance of \modelname-T after finetuning on the DFT dataset. \textbf{a}, MAE results for key superconductivity-related properties on the DFT dataset. From left to right, the four panels report the MAE for the bandgap (eV), Seebeck coefficients ($\mu V/K$), electrical conductivity and electronic thermal conductivity (log-scaled $\sigma/\tau$ in $1/(\Omega\cdot m\cdot s)$ and $\kappa_e/\tau$ in $W/(m\cdot K\cdot s)$, respectively), and critical temperature $\tc$ (K). For the electrical conductivity (p/n\,Cond.) and electronic thermal conductivity (p/n\,Kappa), we predict the log-transformed values due to their wide dynamic range spanning several orders of magnitude. In the fourth panel, the M.A.D.\ $\tc$ denotes the calculation via the McMillan--Allen--Dynes formula, using the electron--phonon coupling $\lambda$ and logarithmic average phonon frequency $\omega_{\log}$ predicted by our model. Besides SOTA methods, we also compare \modelname{} with its variant without pretraining and with its small-scale version.
    \textbf{b}, Visualization of our model's performance on the validation set of the DFT dataset. The top-left panel shows predicted vs.\ true scatter plot with marginal distributions on the entire validation set. The four plots on the right provide the performance stratified by six crystal families (Triclinic, Monoclinic, Orthorhombic, Tetragonal, Hexagonal, Cubic). The bottom-left panel displays the UMAP embeddings of our model on the validation set.}
    \label{fig:supercon_results}
\end{figure}

\subsection{Finetuning \modelname~for Accurate Critical Temperature Prediction}
\label{sec:Elements-T}

Building upon the universal modeling capabilities of \modelname{}, we now address the formidable challenges inherent in the domain of superconductivity. We curate a high-quality database derived from DFT calculations, which pairs crystalline structures with their corresponding $\tc$ values. Beyond simple $\tc$ regression, we employ a multi-objective joint training strategy to capture the complex physical dependencies of superconductivity. Specifically, \modelname{T} simultaneously predicts electronic, transport, and phonon-mediated properties, including bandgaps, Seebeck coefficients, and conductivities from JARVIS~\citep{jarvis}, alongside electron–phonon coupling ($\lambda$) and phonon frequencies ($\omega_{\text{log}}$) from DFT-EPC~\citep{dfttc}. This multidimensional approach compels the model to discern intrinsic physical correlations, yielding a robust, physically consistent representation that significantly enhances $\tc$ prediction accuracy. The predictive performance across all evaluated properties is summarized in \cref{fig:supercon_results}a. For Bandgap prediction, the results of current SOTA methods, such as Matformer~\citep{matformer} and PotNet~\citep{potnet}, are directly adopted from~\citep{potnet}. To isolate the effects of model scale and pretraining, we evaluate two distinct variants: models with 28M and 1B parameters, both without pretraining. A comparative analysis reveals that while increasing parameter volume inherently improves performance, the large-scale variant consistently underperforms relative to \modelname{T}, underscoring the critical role of our massive-scale pretraining in capturing complex atomic interactions. Notably, \modelname{T} significantly outperforms all baseline methods in Bandgap prediction, demonstrating the robustness of its foundational representations. Furthermore, \cref{fig:supercon_results}c provides a detailed assessment of $\tc$ prediction on the DFT-derived dataset. Overall, the model achieves a MAE of 0.992 and an $\mathrm{R^2}$ score of 0.816, marking a significant advancement in predictive fidelity. To understand the geometric dependencies of the model, we visualize $\tc$ predictions across various crystal systems. The results indicate that the model performs optimally on cubic systems, while prediction errors slightly increase for triclinic, monoclinic, and orthorhombic systems. This trend suggests that the model effectively exploits crystalline symmetry, with higher-symmetry lattices facilitating more accurate property mapping. To probe the learned feature space, we visualize the embeddings from \modelname{T} using Uniform Manifold Approximation and Projection (UMAP) in the lower-left panel of \cref{fig:supercon_results}a. Crystalline structures with high $\tc$ values exhibit clear clustering within a specific manifold. This distinct spatial segregation confirms that \modelname{T} has learned a physically meaningful representation where superconductivity-related features are highly separable, effectively capturing the structural distribution characteristics of superconducting materials.

\begin{figure}[t!]
    \centering
\includegraphics[width=\textwidth]{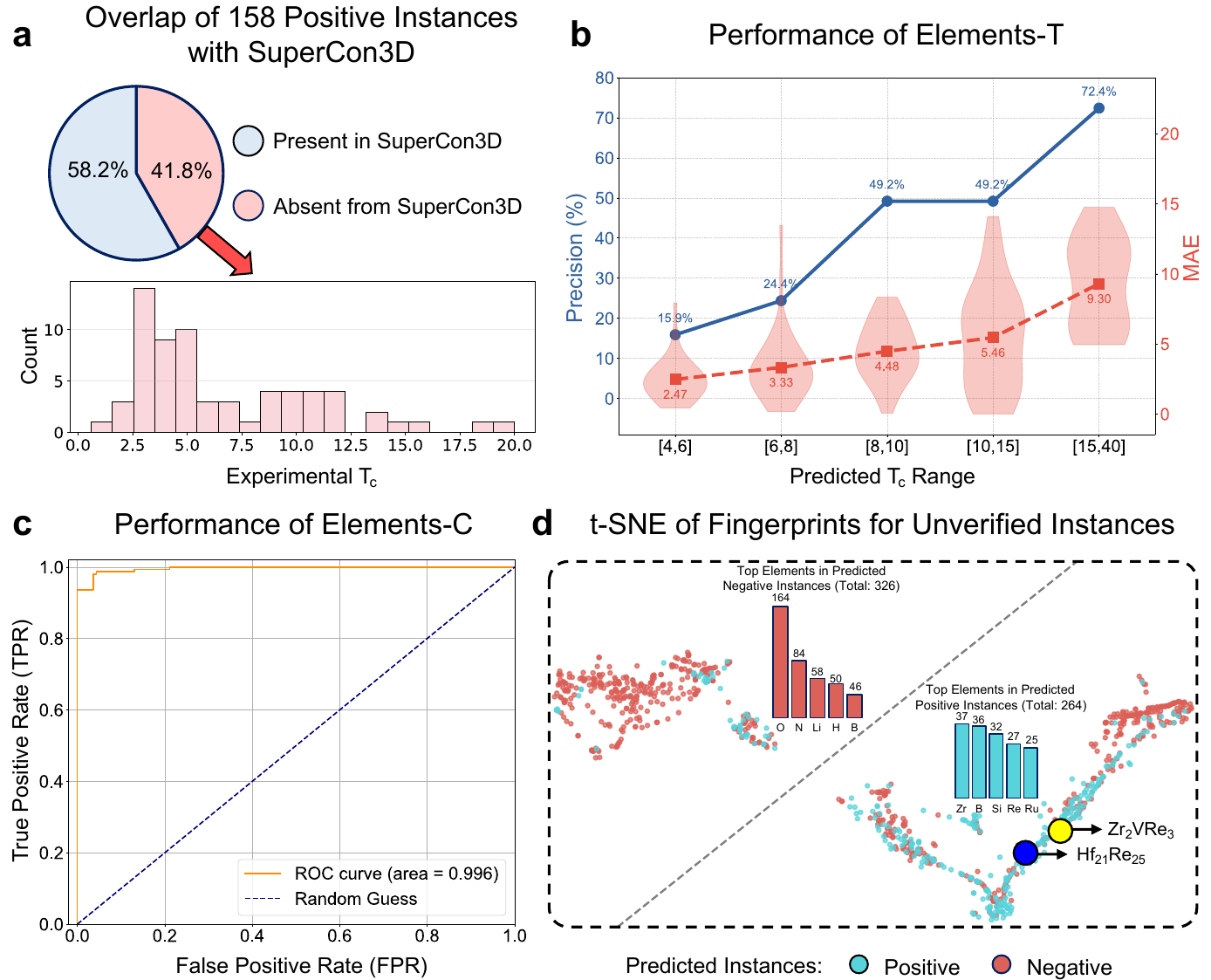}
\vskip 0.0in
    \caption{Superconductor screening from existing databases and literature by \agentname{}. \textbf{a}, Overlap between positive instances and the SuperCon3D database, revealing that $41.8\%$ of the identified materials represent novel entries absent from the SuperCon3D database. The experimental $\tc$ distribution of these undocumented materials is illustrated in the histogram below. \textbf{b}, Performance evaluation of \modelname{T}. The plot illustrates the classification precision (blue line) and the MAE of $\tc$ predictions (red violin plots) across varying predicted $\tc$ intervals. \textbf{c}, ROC curve for the classification model \modelname{C}, achieving an AUC of $0.996$. \textbf{d}, t-SNE visualization of crystal fingerprints for the unverified instances. Data points are color-coded based on \modelname{C} predictions (blue for positive, red for negative). Insets highlight the elemental prevalence in predicted positive clusters versus negative clusters. Guided by this clustering, $\text{Hf}_{21}\text{Re}_{25}$ and $\text{Zr}_{2}\text{VRe}_{3}$ are explicitly marked as the targeted candidates selected for subsequent experimental synthesis and characterization.}
\label{fig:supercon_screen}
\end{figure}

\subsection{Screening Superconductors Hidden in Existing Dataset and Literature}
\label{sec:stage-1}

Leveraging \modelname{T} as a high-throughput triage tool, \agentname{} screens a deduplicated pool of approximately $72{,}000$ structures aggregated from MPDS~\citep{MPDS,pauling} and the Kagome database~\citep{kagome}. Rather than treating model-predicted $\tc$ as final evidence, the agent follows a research process closer to human practice: it uses prediction to prioritize candidates, then returns to the literature to verify whether superconductivity has been experimentally reported for the same composition and crystal structure. Applying a predicted $\tc > 4~\text{K}$ threshold reduces the search space to $1{,}524$ candidates, which are subsequently examined through automated literature retrieval and GPT-5-assisted semantic reasoning. For each candidate, \agentname{} evaluates superconductivity evidence, structural consistency, synthesis feasibility and toxicity risk, with prompt design detailed in \cref{sec:prompt} and hallucination-control procedures in \cref{manual}. After manual verification, the candidates are classified into $158$ literature-verified superconductors, $385$ verified non-superconductors and $981$ unverified instances.

This literature-aware screening independently validates \agentname{}'s ability to recover superconducting knowledge from existing databases and scattered experimental reports. Cross-referencing the $158$ verified positives against SuperCon3D~\citep{sodnet}, which links SuperCon records~\citep{supercon} to three-dimensional structures, shows that only $58.2\%$ are present in SuperCon3D, whereas $41.8\%$---$66$ crystals in total---are absent from this structured database but recovered by \agentname{} (\cref{fig:supercon_screen}a and \cref{tab:supercond}). These SuperCon3D-missing entries demonstrate that the agent does not merely reproduce curated superconductivity labels; it can connect structural records with dispersed literature evidence and thereby expand the corpus of superconductors with confirmed crystal structures. Several recovered entries exhibit experimental $\tc$ values above $10~\text{K}$, indicating that the missing knowledge is not limited to marginal low-temperature cases.

We further assess \modelname{T} on the $543$ literature-verified instances, including both positive and negative examples. The precision of superconductor identification increases with predicted $\tc$, rising from $15.9\%$ in the low-predicted-$\tc$ regime to $72.4\%$ for candidates with predicted $\tc > 15~\text{K}$ (\cref{fig:supercon_screen}b). Although absolute $\tc$ errors increase for high-$\tc$ outliers, this trend shows that \modelname{T} is effective as a ranking model for enriching superconducting candidates. Together, these results establish the database-and-literature stage as an independent demonstration of agentic discovery: \agentname{} not only identifies candidate materials from structural databases, but also recovers missing experimental knowledge and converts it into verified labels for the self-refinement step described below.

\subsection{Creating New Skills to Pinpoint Target Ternary Systems}
\label{sec:stage-23}

Using the 158 verified superconductors and 385 verified non‑superconductors from the literature‑aware screen, \agentname{} converts extracted experimental knowledge into a new decision tool by finetuning \modelname{} into a binary superconductivity classifier,  \modelname{C}, using the literature-verified positive and negative samples. The motivation for introducing this classification stage is to leverage the experimental data extracted from the literature to further refine and enhance the performance of our prediction ability. The model demonstrates exceptional discriminative performance, achieving an Area Under the Curve (AUC) of $0.99$ (\cref{fig:supercon_screen}c). Subsequently, \agentname{} extracts compositional features using the \texttt{matminer} library to compute \texttt{ElementProperty} fingerprints with the Magpie preset and then assigns binary labels identifying each candidate as a superconductor or non-superconductor. Visualization via t-SNE dimensionality reduction reveals a clear decision boundary in \cref{fig:supercon_screen}d, where the upper-left cluster is dominated by materials containing non-metallic elements (\eg, O, N), which are theoretically less likely to exhibit superconductivity. In contrast, the lower-right cluster is populated by metallic compounds, showing a significant prevalence of Zr. By focusing on this metallic region, \agentname{} implements a multi-step automated filtering pipeline to exclude toxic, radioactive, or unstable phases alongside previously known superconductors or non-superconductors. Through this screening (\cref{fig:supp_appendix_fig2_stage3}), \agentname{} prioritizes Re-rich Zr-containing intermetallics as a promising chemical neighborhood. Further ranking by Elements-C confidence and elemental co-occurrence nominate the Hf–Zr–Re and Zr–V–Re ternary systems for prospective exploration. Consequently, the following sections detail the experimental synthesis and structural characterization of these targeted ternary systems: $\text{Zr--V--Re}$ and  $\text{Hf--Zr--Re}$, in which database-latent phases, generated structures, structural reinterpretations and targeted negative controls can be tested systematically.

\begin{figure}[!thp]
    \centering
    \includegraphics[width=\textwidth]{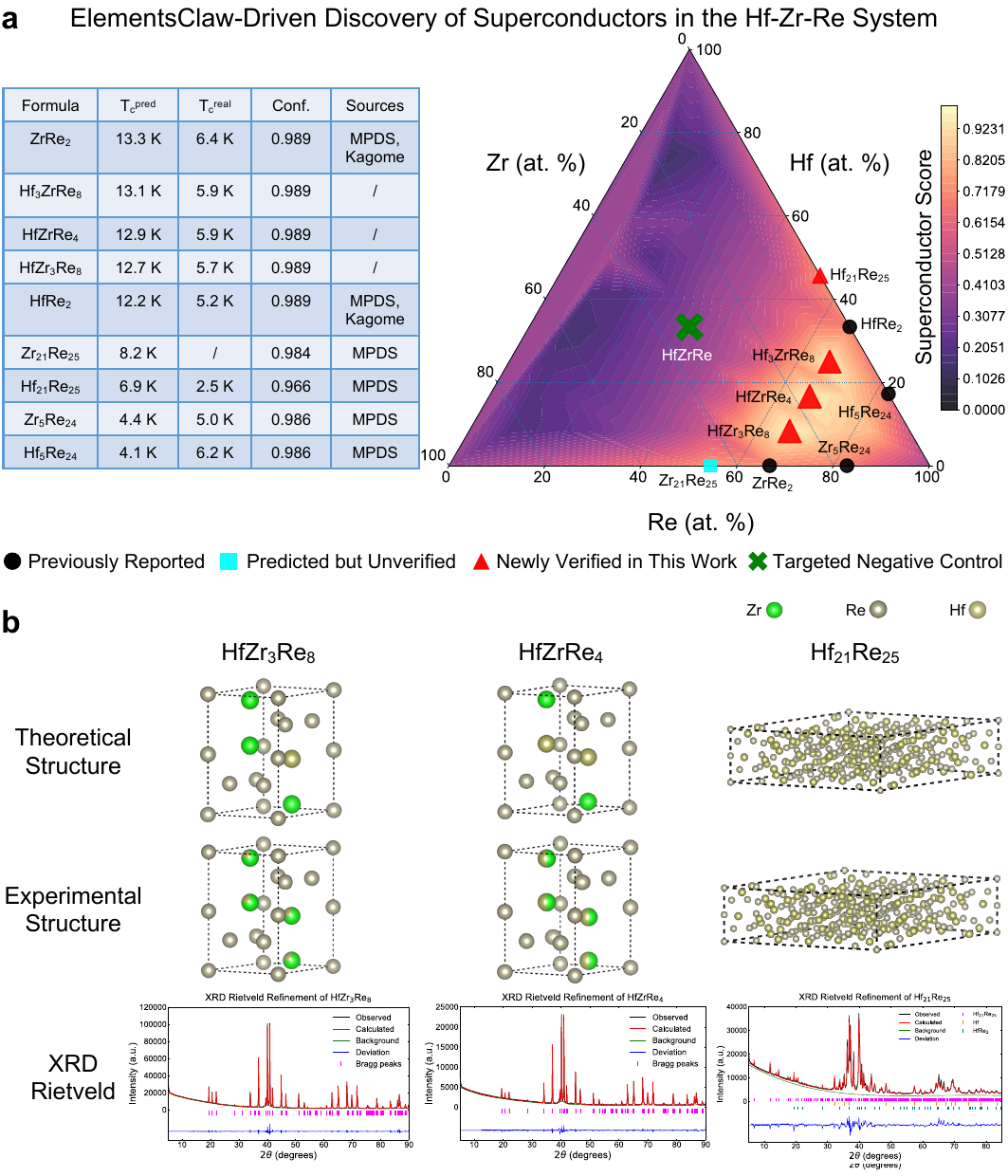}
    \caption{\textbf{Identifying superconducting phases in the Hf--Zr--Re system for experimental verification.} \textbf{a}, (Left) Table of the successful discovery of experimentally-verified superconductors; (Right) Ternary phase diagram illustrating the distribution of explored materials. Within the diagram, \textbf{black circles} denote previously reported superconducting phases, \textbf{cyan squares} represent predicted superconductors without experimental verification, and \textbf{red triangles} highlight candidates successfully predicted and experimentally confirmed as novel superconductors in this work, including $\text{Hf}_{21}\text{Re}_{25}$ sourced from the MPDS dataset, alongside $\text{HfZr}_3\text{Re}_8$, $\text{HfZrRe}_4$, and $\text{Hf}_3\text{ZrRe}_8$. The \textbf{green cross} indicates the targeted negative control $\text{HfZrRe}$, a composition that exhibits phase separation during experimental synthesis. The continuous surface results from the interpolation of candidate points generated by \agentname{} during its systematic traversal of the entire $\text{Hf}_a\text{Zr}_b\text{Re}_c$ ($a+b+c=12$) compositional manifold, where the background heatmap represents a composite superconductor score defined as $0.5 \cdot \tc - 0.25 \cdot E_{\text{form}} - 0.25 \cdot E_{\text{hull}}$. In this scoring function, the predicted $\tc$, formation energy ($E_{\text{form}}$), and energy above hull ($E_{\text{hull}}$) are processed using Min-Max normalization. \textbf{b}, Structural and experimental validation. Comparison of the theoretical crystal structures (top row) with the experimentally determined structures (middle row), and the corresponding PXRD Rietveld refinement profiles (bottom row) for the three newly verified superconductors: $\text{HfZr}_3\text{Re}_8$, $\text{HfZrRe}_4$, and $\text{Hf}_{21}\text{Re}_{25}$. The PXRD Rietveld refinement of $\text{Hf}_3\text{ZrRe}_8$ is provided in \cref{fig:supp_fig6_appendix}.}
\label{fig:supercon_discovery}
\end{figure}

\subsection{Identifying High-Probability Candidates for Experimental Verification}
\label{sec:experiments}

A central challenge in automated materials discovery lies not only in generating candidates, but in reliably selecting experimentally viable compounds from a large and often contradictory design space. To evaluate this capability, we perform prospective experimental validation using candidates identified within the Hf--Zr--Re and Zr--V--Re compositional spaces. Candidate selection integrates predicted $\tc$, thermodynamic stability (energy above hull $E_{\text{hull}}$), and consistency with available literature, ensuring both high-confidence candidates and targeted negative controls.

Within the Hf--Zr--Re system (\cref{fig:supercon_discovery}a), \agentname{} first surveys existing structural databases and recovers several reported superconductors, including HfRe$_2$, Hf$_5$Re$_{24}$, Zr$_5$Re$_{24}$ and ZrRe$_2$. It also identifies Hf$_{21}$Re$_{25}$ and Zr$_{21}$Re$_{25}$ as database-latent candidates whose superconductivity have not been incorporated into structured superconductivity databases. Because Zr$_{21}$Re$_{25}$ is associated with known synthetic difficulty~\citep{Zr21Re25Hf21Re25}, we select Hf$_{21}$Re$_{25}$ as a representative database-latent validation case. Arc-melting synthesis followed by PXRD refinement confirms the target phase with high purity ($>95\%$) and $R_{\text{wp}} \approx 6\%$ (\cref{fig:supercon_discovery}b). Electrical transport measurements reveal a superconducting transition with $\tc^{\mathrm{onset}} = 3.0~\text{K}$ and $\tc^{\mathrm{zero}} = 2.0~\text{K}$  (\cref{fig:supp_Elements_SuppFigE6} ), while magnetic susceptibility further supports superconductivity in this phase (\cref{fig:elements_overview}c, \cref{fig:supp_fig6_appendix} ). Although its low $\tc$ indicates that Hf$_{21}$Re$_{25}$ is not a high-performance discovery, this result validates the agent's ability to recover experimentally accessible superconducting phases hidden in structural databases and literature.

To explore compositional regions not represented in existing databases, \agentname{} activates its generative module (\modelname{G}) to construct candidate structures across the Hf$_a$Zr$_b$Re$_c$ ($a + b + c = 12$) grid (\cref{fig:supercon_discovery}a). Generated candidates are filtered by thermodynamic constraints using \modelname{E} ($E_{\text{form}} < 0.0~\text{eV atom}^{-1}$, $E_{\text{hull}} < 0.05~\text{eV atom}^{-1}$), and then prioritized using \modelname{T} ($\tc > 4~\text{K}$) and \modelname{C}. This workflow selects HfZrRe$_4$, HfZr$_3$Re$_8$ and Hf$_3$ZrRe$_8$ as prospective superconducting candidates (\cref{fig:supercon_discovery}a). All three compositions are synthesized as near-single-phase materials, and PXRD/Rietveld refinement confirms close agreement with the predicted structures (\cref{fig:supercon_discovery}b and \cref{fig:supp_fig6_appendix}). In particular, HfZrRe$_4$ exhibits a clear transport transition with $\tc^{\mathrm{onset}} = 6.7~\text{K}$ and $\tc^{\mathrm{zero}} = 6.1~\text{K}$, together with a magnetic onset near $5.9~\text{K}$.  HfZr$_3$Re$_8$ and Hf$_3$ZrRe$_8$ show magnetic superconducting onsets near $5.9~\text{K}$ and $5.7~\text{K}$, respectively (\cref{fig:supp_fig6_appendix}). These results establish the generative branch of \agentname{} as the central prospective discovery step: the agent uses structure generation, $\tc$ prediction and physical stability filters to identify experimentally realizable superconductors beyond the original MPDS/Kagome entries.
To further assess the reliability of the predicted thermodynamic landscape, we synthesize HfZrRe, a composition excluded by \agentname{} due to its predicted instability ($E_{\text{hull}} > \SI{0.1}{eV\,atom^{-1}}$). Under identical synthesis conditions, this composition exhibits pronounced phase separation rather than forming a single-phase compound. This outcome is consistent with the predicted convex-hull landscape, supporting the framework's ability to exclude unstable regions of the phase diagram.

We next examine the Zr--V--Re system to evaluate performance in a chemically distinct environment. \agentname{} identifies Zr$_4$VRe$_7$, previously reported in the GNoME~\cite{genome} dataset~(ID: mp-3274319) with an orthorhombic structure, but predicted instead a lower-energy hexagonal configuration. Experimental synthesis followed by PXRD and Rietveld refinement confirms the formation of the hexagonal phase ($R_{\text{wp}} \approx 5\%$), in agreement with the AI-predicted $P6/mmm$ structural model. Magnetic measurements reveal bulk superconductivity, with an onset $\tc$ of 3.5 K, a shielding fraction exceeding $70\%$ at 2 K (\cref{fig:elements_overview}c and \cref{fig:supp_fig6_appendix}). This result highlights the capability of the framework to identify and correct structural inconsistencies in existing datasets.
In contrast, Zr$_2$VRe$_3$, selected for its high predicted $\tc$, does not exhibit bulk superconductivity despite successful synthesis of the target phase. Only a weak diamagnetic response (shielding fraction $\approx 0.3\%$) is observed (\cref{fig:supp_fig6_appendix_wrong}). This discrepancy is attributed to the presence of magnetic V atoms, which introduce pair-breaking effects not explicitly captured in the underlying density functional theory-based training data. This limitation highlights the need to incorporate magnetic interactions in future model development.

To evaluate whether the validated chemical motifs could be extended beyond the initially explored ternary systems, \agentname{} next performs a global mining of all 2.4 million equilibrium crystals in the pretraining corpus. Applying \modelname{T} and \modelname{C} yielded $68{,}000$ potential superconducting candidates, providing a broad map of superconductivity-enriched regions within the equilibrium-crystal space. Guided by the preceding validation experiments, we focus on the agent's search using two empirical criteria: retention of the P6/mmm Re-rich framework and preservation of the Re sublattice. This targeted exploration identifies several promising analogues, including Zr$_{3}$ScRe$_{8}$, ZrScRe$_4$, LuZrRe$_4$, ZrSc$_3$Re$_8$ and TmZrRe$_4$, all with predicted $\tc > 9~\text{K}$. Among them, Zr$_{3}$ScRe$_{8}$ is selected as the primary experimental target because it combines the highest predicted $\tc$ within the Zr--Sc--Re family with preservation of the Re-rich hexagonal framework. Experimentally, Zr$_{3}$ScRe$_{8}$ is synthesized as a near-single-phase material, and PXRD/Rietveld refinement confirmed the predicted P6/mmm structure with Sc occupying the Zr site rather than disrupting the Re sublattice (\cref{fig:supp_fig6_appendix}). Transport measurements revealed $\tc^{\mathrm{onset}} = 6.8~\text{K}$ and $\tc^{\mathrm{zero}} = 6.0~\text{K}$, while magnetic susceptibility showed a bulk superconducting transition near $6.5~\text{K}$ (\cref{fig:elements_overview}c and \cref{fig:supp_fig6_appendix}). This result connects global superconductivity mining with focused experimental selection, demonstrating that \agentname{} can reduce a $68{,}000$-candidate landscape to a chemically interpretable and experimentally validated superconductor.

Collectively, these results demonstrate that the \agentname{} framework enables coordinated candidate selection across database retrieval, structural reinterpretation, and generative design, while also providing reliable exclusion of unstable compositions. The agreement between predicted thermodynamic stability, structural models, and experimental outcomes---including both successful synthesis and controlled failure cases---supports its effectiveness for navigating complex compositional spaces in superconducting materials discovery.

    \section{Discussion}

In this work, we present \agentname{}, which transitions AI-driven materials science from isolated predictions toward agentic orchestration. The efficacy of \agentname{} is predicated on the deep synergy between Large Atomic Models (LAMs) and Large Language Models (LLMs). Serving as the foundation for a suite of specialized functional tools, the proposed LAM \modelname{} is pretrained on an extensive corpus of 125 million structures and anchors the physical engine. This model demonstrates superior performance across 22 distinct tasks, establishing a new benchmark for $\tc$ prediction. To complement these universal modeling capabilities, \agentname{} leverages LLM reasoning to navigate the heuristic complexities of materials design, including the evaluation of synthetic feasibility, toxicity screening, and the distillation of literature-derived insights. Crucially, this agentic orchestration transcends static tool invocation by enabling autonomous self-evolution. By using empirical data extracted by the LLM to refine specialized tools such as the \modelname{C} classifier, \agentname{} continuously forges customized predictive capabilities, establishing a highly transferable paradigm for the exploration of uncharted chemical spaces.

In the demanding domain of superconductors, \agentname{} conducts a four-stage materials discovery pipeline by autonomously planning and executing actions in response to user instructions. Specifically, the framework achieves a high rediscovery success rate for literature-verified superconductors, while simultaneously identifying 66 literature-reported superconductors absent from the standard SuperCon database. Experimental validation across the Hf--Zr--Re and V--Zr--Re systems underscores the capacity of the agent to navigate complex chemical landscapes, successfully realizing unverified superconducting phases and generating novel candidates with transition temperatures reaching 6~K. Beyond discovery, \agentname{} performs structural reinterpretation to correct inconsistent database entries and executes heuristic searches through site-selective substitution to preserve critical structural motifs such as the Re Kagome lattice. By screening over 2.4 million stable crystals to yield 68,000 high-confidence candidates, the framework vastly expands the known superconducting space and establishes a robust, physics-informed paradigm for identifying experimentally viable and high-performance materials.

Despite these capabilities, we acknowledge several limitations of the current framework. First, reliance on standard DFT training data constrains predictive fidelity for strongly correlated unconventional superconductors, such as cuprate and iron-based families, where these functionals are inherently inadequate. Expanding the corpus with higher-level electronic structure methods remains a necessary but computationally demanding frontier. Second, while LLMs accelerate literature extraction, they remain susceptible to hallucinations and corpus biases, potentially overlooking nuances in preprints or non-English publications. Finally, our synthesis screening assumes ambient-pressure conditions, precluding the exploration of high-pressure hydrides or extreme-condition phases without tailored thermodynamic modules.

Looking forward, the transformative impact of \agentname{} extends well beyond superconductor discovery. The principal strength of this agentic framework lies in its universal adaptability: rather than manually re-engineering models for new domains, researchers can simply direct the agent to autonomously mine domain-specific literature, construct novel empirical datasets, and forge highly specialized predictive tools. This capability to self-evolve task-specific toolchains means the framework can seamlessly pivot to conquer other strategically critical material classes, such as solid-state battery electrolytes, heterogeneous catalysts, and thermoelectric materials. As agentic systems of this kind are progressively integrated with self-driving laboratory hardware, we anticipate the emergence of increasingly autonomous discovery loops---from hypothesis generation through property prediction to automated synthesis and characterization---that will substantially accelerate the pace of materials innovation.

    \section{Methods}\label{sec:methods}

\subsection{Unified Geometric Representation}
\label{sec:notation}

We represent an atomic system (either a finite molecule or a periodic crystal) as a geometric graph $\gG = (\mA, {\mX}, \gE)$. Here, $\mA = [a_1, a_2, \dots, a_N] \in \mathbb{N}^{1 \times N}$ denotes the vector of atomic numbers; ${\mX} = [{\vx}_1, {\vx}_2, \dots, {\vx}_N] \in \mathbb{R}^{3 \times N}$ denotes the matrix of  3D Cartesian coordinates; $\gE$ denotes the set of edges connecting atom pairs. For crystalline materials, the periodicity is depicted by the lattice matrix ${\mL} = [{\vl}_1, {\vl}_2, {\vl}_3] \in \R^{3 \times 3}$. The Cartesian position of any atom $i$ in a periodic image of the cell, translated by an integer vector $\vz \in \mathbb{Z}^{3 \times 1}$, is given by: ${\vx}_i + {\mL}\vz$.

For molecular graph construction, we connect atom pairs within a predefined radial cutoff distance $r_c = 12\text{\AA}$, leading to:
\begin{equation}
    \gE_{\text{mol}} = \{ (i, j, {\vr}_{ij}) \mid {\vr}_{ij} = {\vx}_i - {\vx}_j, \|{\vr}_{ij}\|_2 \leq r_c, \, i \neq j \},
\end{equation}
where ${\vr}_{ij}$ denotes the relative position vector between atoms $i$ and $j$.

For crystal graph construction, we utilize a multi-edge scheme to capture interactions across periodic boundaries~\citep{CGCNN}:
\begin{equation}
    \gE_{\text{cut}} = \{ (i, j, {\vr}_{ij}) \mid {\vr}_{ij} = {\vx}_i - {\vx}_j + {\mL}\vz, \|{\vr}_{ij}\|_2 \leq r_c, \, i \neq j, \vz \in \{-1,0,1\}^{3 \times 1} \}.
\end{equation}
Additionally, to explicitly encode the lattice information ${\mL}$ into the graph message passing, we connect each atom to its periodic images in neighboring unit cells~\citep{matformer}. Unlike the original six-neighbor approach, our method incorporates only three additional Self-Loops (SL) for each atom to reduce edge density and accelerate training, balancing computational efficiency with structural encoding:
\begin{equation}
    \gE_{\text{SL}} = \{ (i, i, {\vr}_{ii}) \mid {\vr}_{ii} = {\mL}\vz, \vz \in \{\ve_1, \ve_2, \ve_3\}\},
\end{equation}
where $\ve_1, \ve_2, \ve_3$ represent unit vectors $[1, 0, 0], [0, 1, 0], [0, 0, 1]$, respectively.
Finally, by merging  the periodic multiple edges and SL, we obtain the whole crystal edge set as $\gE_{\text{crys}} = \gE_{\text{cut}} \cup \gE_{\text{SL}}$.

In the domain of crystal generation, it is common practice to model structures using fractional (or scaled) coordinates $\mS \in [0, 1)^{3 \times N}$, which are related to Cartesian coordinates $\mX$ via the transformation $\mS = \mL^{-1}\mX$. This approach decouples atomic positions from the lattice $\mL$, thereby simplifying the independent generation of $\mS$ and $\mL$. However, such a strategy is suboptimal for force-field-related tasks, where target quantities---such as atomic forces $\mF$ and equilibrium positions---are physically defined in Cartesian space and necessitate strict rotational equivariance. Recent frameworks like MatterGen~\citep{mattergen} have explored Cartesian-based generation, yet they still rely on fractional coordinates for score computation. We instead opt for a fully Cartesian formulation, ensuring seamless alignment with our pretraining strategy, where noise is injected directly into $\mL$ and $\mX$ within Cartesian space.

\subsection{Dataset Construction}

To cultivate an omni-domain foundation for atomic systems, we curated the \databasename{} dataset, a massive-scale repository comprising $125.21$ million atomic configurations. The dataset maintains a strategic balance between periodic and non-periodic systems, consisting of $106.55$ million crystal structures (85.1\%) and $18.66$ million molecular geometries (14.9\%) (\cref{fig:supp_fig_1_appendix}a). Critically, to ensure the model captures the full complexity of interatomic potentials, we compiled both  ``stable'' (a.k.a. equilibrium) states and ``unstable'' (a.k.a. non-equilibrium) configurations. The \textbf{stable subset} comprises approximately $5.75$ million crystals and $4.16$ million molecules, which contain only structural information without corresponding force fields. In contrast, the significantly larger \textbf{unstable subset}---serving to provide the necessary gradient information to learn robust out-of-equilibrium behaviors---contains $100.8$ million crystals and $14.5$ million molecules, all of which are fully annotated with energy and force labels.

\databasename{} aggregates and harmonizes data from several premier public repositories. The unstable geometries, typically derived from MD trajectories or relaxation paths, are sourced from Transition-1x~\citep{Transition1x} and ANI-1x~\citep{ani-1x} for molecules, and exclusively from OMAT-24~\citep{OMAT-24} for crystals. For the stable systems, molecular structures are aggregated from PCQM4Mv2 and a subset of Transition-1x, rigorously filtered to retain only configurations whose mean atomic force norm is lower than $20\,\text{meV}/\text{\AA}$. Stable crystal structures are curated from a diverse aggregation including GNoME~\citep{genome}, NOMAD~\citep{nomad}, Alexandria~\citep{Alex}, OQMD~\citep{oqmd}, MPF~\citep{M3GNet}, and JARVIS-QETB~\citep{jarvis_qetb}. To derive sufficiently equilibrated subsets from these repositories, we apply strict filters based on interatomic forces ($\leq 20\,\text{meV}/\text{\AA}$) and energy above the hull ($E_{\text{hull}}\leq \SI{0.08}{eV\,atom^{-1}}$). 
A comprehensive breakdown of data sources and preprocessing steps is provided in \cref{app:data_details}.

This synthesis grants \modelname{} expansive coverage of the periodic table (\cref{fig:supp_fig_1_appendix}c), encompassing nearly all chemically relevant species except for heavy radioactive elements (Z = 84--118) and Actinides (Z = 95--103). Such chemical breadth is essential for a model intended to navigate the vast and often unexplored composition space of potential superconductors. We characterize the statistical distributions of key physical quantities to validate the dataset's coverage of the PES. The total energy distributions for the molecular and crystal subsets exhibit distinct, broad profiles, indicating that the model is exposed to diverse energetic scales---ranging from the $-5$ eV/atom peak of the OMAT-24 crystal data to varying molecular regimes (\cref{fig:supp_fig_1_appendix}b). Analysis of the force norm distributions reveals an exponential-like decay for crystals and clustering near zero for molecules (\cref{fig:supp_fig_1_appendix}d). While low-force configurations provide a baseline for stability, the extensive ``long tail'' of high-force samples ensures that \modelname{} learns to accurately resolve the steep repulsive and attractive regions of the interatomic potential.

\subsection{Architecture of \modelname{}} \label{sec:arch}
Our model, \modelname{}, builds upon the foundational architecture of EquiformerV2~\citep{EquiformerV2}, which we substantially adapt for the comprehensive modeling of both molecular and crystalline structures. As illustrated in the architecture diagram (\cref{fig:ext_model}a), the overall pipeline consists of Graph Construction, Embedding, stacked Equivariant Message Passing layers, and task-specific output heads.

\textbf{Input and Graph Construction.} 
For a 3D atomistic system, we construct its geometric graph as detailed in \cref{sec:notation}. While molecular systems utilize standard cutoff-based edges, periodic crystals incorporate multi-edge cutoffs and SL across periodic boundary conditions to capture infinite lattice interactions accurately. 

\textbf{Embedding.} 
We denote by $\vh_{i,l}^{(t)}\in\R^{(2l+1)\times C_t}$ the $l$-th degree steerable feature of atom $i$ in layer $t$, with the channel count as $C_t$. These features transform according to the $l$-th Wigner-D matrix $\mD^{(l)} \in \sR^{(2l + 1) \times (2l + 1)}$ under rotation. Specifically, atomic numbers and coordinates are treated as $0$th-degree (scalar) and $1$st-degree (vector) features, respectively. For brevity, we denote the collection of all features across degrees $l \in \sL = \{0,1,\dots,L\}$ as $\vh_{i, \sL}^{(t)}$.

The initial feature $\vh_{i, l}^{(0)}$ is constructed by integrating of the \textit{atom embedding} and the \textit{edge-degree embedding}. The \textit{atom embedding} is defined via a linear projection of the one-hot-encoded atomic number:
\begin{equation}
\vh^{(\text{atom})}_{i,l} =
\begin{cases}
\mW \texttt{OneHot}(a_i) + \vb, & l = 0 \\
\mathbf{0}, & l > 0
\end{cases},
\end{equation}
where $\mW \in \mathbb{R}^{C \times C_a}$ and $\vb \in \mathbb{R}^{C}$ denote the learnable weight matrix and bias vector, respectively, with $C$ and $C_a$ representing the number of hidden channels and the maximum atomic number.
The \textit{edge-degree embedding} encodes the local geometric environment. First, an edge message is constructed by passing the interatomic distance and atomic numbers through a radial function $\phi$:
\begin{equation}
g_{ij,l,m} =
\begin{cases}
\phi(\|\vr_{ij}\|, a_i, a_j), & m = 0 \\
0, & m \neq 0
\end{cases}.
\end{equation}
Subsequently, the rotated edge messages are aggregated to generate the \textit{edge-degree embedding} for each atom:
\begin{equation}
\vh_{i,l}^{(\text{edge})} = \frac{1}{\bar{d}} \sum_{j \in \mathcal{N}(i)} \left(\mD^{(l)}(\mR_{ij})\right)^{-1} \vg_{ij,l},
\end{equation}
where the rotation frame $\mR_{ij}$ is constructed from the cross product of the edge direction $\vr_{ij}$ and a random vector, and the term $\bar{d}$ denotes the average node degree used for rescaling.

Finally, the initial features are obtained by aggregating the \textit{atom embedding} and the \textit{edge-degree embedding} for each degree:
\begin{equation}
\vh_{i,l}^{(0)} = \vh_{i,l}^{(\text{atom})} + \vh_{i,l}^{(\text{edge})}.
\end{equation}

\textbf{Equivariant Message Passing with Long-Range Residual Connection (LRC).} The backbone of the network comprises 12 equivariant message passing layers. Each layer updates the atomic feature $\vh_{i, \sL}^{(t)}$ by employing depth-wise tensor products and $SO(2)$ linear operations, which are derived from eSCN~\citep{escn}. To enhance feature propagation and mitigate information loss of fundamental atomic identities in deep layers, we adapt the Long-Range Residual Connection (LRC) from GROVER~\citep{grover}. Let $\mathcal{F}^t, t \in \{0, 1, 2, \ldots, T\}$ denote the composite operations within the $t$-th layer, which include Layer Norm, Equivariant Graph Attention, and a Feed Forward Network. While a standard residual connection follows $\vh_{i,l}^{(t)} = \vh_{i,l}^{(t-1)} + \mathcal{F}^{(t)}(\vh_{i,\sL}^{(t-1)}, \gE)$~\citep{resnet}, our LRC explicitly injects the initial atomic feature $\vh_{i,l}^{(0)}$ into the outputs of the last two layers ($t \in \{11, 12\}$):
\begin{equation}
    \vh_{i,l}^{(t)} = \vh_{i,l}^{(t-1)} + \mathcal{F}_{l}^{(t)}(\vh_{i,\sL}^{(t-1)}, \gE) + \vh_{i,l}^{(0)},
\end{equation}
where $\mathcal{F}_{l}^{(t)}$ denotes the $l$-th degree component of the $t$-th layer output.
We observe that this topological modification enhances the overall performance by anchoring deep features to the original atomic features.

\textbf{Grid Activation and Resolution Reduction.} 
Unlike scalar features, nonlinear activation is not readily applicable to steerable features. To address this limitation, \citep{sphcnn} introduced the $\sS^2$ activation, which has been widely adopted in equivariant architectures~\citep{escn, EquiformerV2}. We adopt the same strategy in our model. Specifically, the steerable features $\vh_{i,l}^{(t)}$ in the spherical harmonic domain are first reconstructed as spatial signals on the sphere:
\begin{equation}
    \psi_i^{(t)}(\theta, \phi) = \sum_{l=0}^{L} \sum_{m=-l}^{l} \vh_{i,l,m}^{(t)} Y_{l,m}(\theta, \phi),
    \qquad (\theta, \phi)\in \gQ_R,
\end{equation}
where $\gQ_R\coloneqq\{((i+1/2)\cdot(\pi/R),j\cdot(2\pi/R))\mid i,j=0,1,\dots,R-1\}$ denotes the uniformly discretized spherical grid with angular resolution $R$ and $Y_{l,m}:\sS^2\to\sR$ denotes the spherical harmonic basis of the $l$-th degree and $m$-th order. The resulting signal $\psi_i^{(t)}(\theta, \phi)$ can then be passed through a nonlinear activation (\eg, $\texttt{SiLU}(\cdot)$) in the spatial domain. The activated signal is then projected back to the spherical harmonic domain as
\begin{equation}
    h^{(t)}_{i, l, m} = \int_{\sS^2} \texttt{SiLU}(\psi_i^{(t)}(\theta, \phi))\, Y_{l, m}(\theta, \phi)\,\mathrm{d}\Omega.
\end{equation}
In practice, this integral is approximated numerically over the discretized spherical grid $\gQ_R$. Following EquiformerV2~\citep{EquiformerV2}, the channels for degree $l=0$ are partitioned into two groups. One group undergoes direct SiLU activation, while the other is processed through the $\sS^2$ pathway alongside all $l > 0$ features. The resulting outputs are then concatenated along the channel dimension. This separation stabilizes the training process while preserving the expressive cross-degree mixing inherent in $\sS^2$ activations. Standard implementations adopt $R=18$ with $L=6$, resulting in $324$ grid points. Here, we reduce the grid resolution to $R = 2$ (4 points). As demonstrated in \cref{tab:grid_test}, for a model with $30$M parameters, this reduction decreases memory consumption by approximately $30\%$ and accelerates training by 20\%. Notably, the prediction accuracy slightly improves rather than degrading, despite the reduced resolution. This suggests that the benefit of the $\sS^2$ pathway may not depend on a high-resolution spherical representation, but instead arises from the constrained nonlinear mixing induced by the spectral-spatial-spectral transformation, which may act as an implicit regularizer. The detailed architectural hyperparameters of the \modelname{} model are presented in \cref{sec:supp_arch}.

Following the last Layer Norm and Grid Activation, the final features $\vh_{i, \sL}^{(T)}$ are processed by task-specific output heads. These include an Energy and Property Head for scalar predictions, a Lattice Denoising Head and a Coordinate Denoising Head for crystal structure generation, as well as a Force Head for atomic force prediction. The detailed architectures of these heads and corresponding training objectives are presented in the next subsection.

\subsection{Pretraining and Finetuning Process of \modelname{}} \label{sec:pretraining_finetuning}

We scale the capacity of \modelname{} to 1 billion parameters and adopt a two-stage training paradigm: task-agnostic pretraining followed by task-specific finetuning.

\textbf{Pretraining Phase.} The pretraining protocol remains consistent across all downstream applications and leverages \databasename{} comprising both stable and unstable structures for molecules and crystals. The stable subset contains purely structural information including atom types, atomic coordinates, and lattice vectors, and is utilized for unsupervised denoising tasks. Conversely, the unstable subset provides ground-truth labels for potential energies and atomic forces, facilitating supervised force-field training. To accommodate these dual objectives, the model is equipped with four specialized prediction heads: coordinate denoising, lattice denoising, energy prediction, and force prediction.

\textbf{Finetuning Phase.} For downstream tasks, we initialize the backbone and relevant prediction heads from the pre-trained checkpoint, restricting task-specific modifications primarily to the output heads and the global aggregation strategy. Specifically, for superconducting critical temperature prediction, we shift the global aggregation from sum pooling to mean pooling. This critical adjustment preserves the intensive nature of $\tc$, preventing predictions from unphysically scaling with the system size and ensuring proper generalization. For crystal structure generation, we implement a prior-informed diffusion strategy inspired by MatterGen~\citep{mattergen}, wherein the limit noise distribution is derived from dataset statistics rather than a standard Gaussian prior. Furthermore, the entire forward and reverse diffusion processes operate directly in Cartesian coordinate space, entirely bypassing the need for fractional coordinates. 

Further technical details regarding both training stages are provided in~\cref{sec:supp_training_strategies}.

\subsection{LAM Tools of \agentname{}}
Our superconductor discovery pipeline is orchestrated by \agentname{}, an LLM-based agentic system that autonomously plans and executes multi-step screening strategies. Built upon the OpenClaw framework~\citep{openclaw}, \agentname{} operates by selectively invoking a suite of specialized tools, each derived from finetuning \modelname{} for a distinct task. We describe each tool and the agent's literature screening capability below. More details are available in \cref{sec:supp_sft}.

\medskip

\noindent\textbf{Tool 1: \modelname{T} (Superconducting Property Predictor).} 
Accurate identification of superconducting materials necessitates a nuanced understanding beyond simple $\tc$ regression. To capture the complex physical dependencies underlying superconductivity, we finetune \modelname{} to jointly predict $\tc$ alongside a diverse set of electronic and transport properties.
We curate a high-quality database derived from DFT calculations, designated as SuperConducting Properties (SCP). This database pairs crystal structures with multifaceted properties associated with superconductivity and comprises three specialized subsets: DFT $\tc$, JARVIS-DFT, and DFT-EPC\footnote{While the SuperCon dataset~\citep{supercon} is widely recognized in the field, we observe significant limitations including data redundancy, labeling errors, and a pervasive lack of three-dimensional structural information; consequently, we utilize it only for auxiliary validation (\textbf{\cref{sec:exp_dataset}}).}.  
The DFT $\tc$ subset encompasses $1,227$ superconducting materials with $\tc$ labels from JARVIS~\citep{jarvis}. The JARVIS-DFT subset provides bandgaps, Seebeck coefficients, and electrical/thermal conductivities. To reinforce the model’s physical grounding, we explicitly incorporate phonon-mediated mechanisms by predicting the Electron–Phonon Coupling (EPC) strength ($\lambda$) and the logarithmic average phonon frequency ($\omega_{\text{log}}$) using the DFT-EPC dataset, which consists of $8,241$ structures from high-throughput screenings of conventional superconductors~\citep{dfttc}. For the JARVIS-DFT subset, we adopt the data splits from PotNet~\citep{potnet} to ensure consistency with established benchmarks. For DFT $\tc$ and DFT-EPC, we randomly partition the data into an 8:2 ratio for training and validation. To manage the heterogeneous nature of the combined data, we construct a unified property vector for each crystal structure; present properties are assigned their numerical values, while missing entries are zero-padded. During joint training, the loss is computed exclusively on available labels, masking out undefined properties. Architecturally, we initialize the property head using pretrained energy weights and reinitialize only the final linear projection layer to match the total dimensionality of the target properties.

\medskip

\noindent\textbf{Tool 2: \modelname{C} (Superconductor Classifier).} \modelname{C} is a binary classifier trained to distinguish superconductors from non-superconductors. The training set comprises positive and negative instances mined from the literature via the screening procedure described in \cref{fig:supercon_screen}a. Training is conducted with a batch size of 64 for 20 epochs, optimizing a Binary Cross-Entropy (BCE) loss. We select the checkpoint exhibiting the lowest validation BCE loss for downstream candidate screening.

\medskip

\noindent\textbf{Tool 3: \modelname{E} (Thermodynamic Stability Evaluator).}
\modelname{E} evaluates the thermodynamic stability of candidate structures. It is finetuned on the MPtrj and sAlex datasets to predict formation energies ($E_{\text{form}}$). During training, we first fit elemental reference energies from the training set distribution, then derive the target $E_{\text{form}}$ by subtracting these references from total energies. The network is trained to predict this intermediate quantity directly; the total energy can be recovered by adding back the pre-calculated elemental references.
To assess thermodynamic stability, we compute the energy above the convex hull ($E_{\text{hull}}$) for each candidate structure. Given a predicted formation energy and the candidate's chemical composition, we query our filtered comprehensive superconductor dataset to retrieve all known competing phases within the same chemical system (\ie, all entries whose elements are a subset of the candidate's constituent elements). These reference entries, together with our candidate, are used to construct a convex hull in composition--energy space via \texttt{pymatgen}~\citep{pymatgen}. The $E_{\text{hull}}$ is then computed as the energy difference (per atom) between the candidate and the lowest-energy linear combination of stable phases at the same composition. We consider a candidate structure to be thermodynamically stable if it satisfies two criteria: (1) $E_{\text{form}} < \SI{0.0}{eV\,atom^{-1}}$, indicating that the compound is energetically favorable relative to its constituent elements, and (2) $E_{\text{hull}} < \SI{0.05}{eV\,atom^{-1}}$, indicating that the structure lies on or very close to the convex hull and is unlikely to decompose into competing phases.

\medskip
 
\noindent\textbf{Tool 4: \modelname{G} (Crystal Structure Generator).}
\modelname{G} is the generative variant of our architecture, developed by finetuning our foundation model on the crystal structure generation task using the MP-20 dataset. When invoked by \agentname{}, it generates novel candidate structures conditioned on a specified composition, expanding the search space beyond known databases.

\medskip
\noindent\textbf{Skills Creation (Superconductor Identification).}
Leveraging these tools, \agentname{} have configured two primary skills. First, when a user seeks to determine whether a specific structure is superconducting, \agentname{} directly invokes \modelname{T} to predict its $\tc$, \modelname{C} to output the confidence score, and \modelname{G} to calculate its formation energy. If $\tc$ $> 4$~K and the confidence score $> 0.5$, \agentname{} classifies the material as a high-confidence superconductor. 
Second, if the user provides only a chemical formula, \agentname{} initially calls \modelname{G} to generate the corresponding crystal structure, and then employs \modelname{E} to evaluate its formation energy and energy above hull. Provided that the formation energy $E_{\text{form}} < \SI{0.0}{eV\,atom^{-1}}$ and $E_{\text{hull}} < \SI{0.05}{eV\,atom^{-1}}$, \agentname{} subsequently invokes \modelname{T} and \modelname{C} to further assess its superconductivity.

\subsection{Literature and Condition Screening Skills of \agentname{}}
\label{sec:prompt}

Beyond the \modelname{}-based tools, \agentname{} integrates an automated literature mining and feasibility assessment module. For candidate materials, the agent retrieves the corresponding original research articles and processes them using an LLM equipped with a specialized system prompt. The model rigorously evaluates the literature against the specific material structure (provided as a POSCAR file from the experimental database MPDS). It extracts and evaluates several critical dimensions, distilling the analysis into a structured 7-tuple (\eg, \texttt{y,3\textasciitilde5,y,n,y,y,NaCl-type cubic}):

\begin{enumerate}
    \item \textbf{Superconductivity Verification} (\texttt{is\_supercond}) \textbf{\&} $\tc$ (\texttt{tc})\textbf{:} The model performs a strict structural match, ensuring the reported superconductivity belongs to the exact polymorph (atomic coordinates) of the candidate, not just matching the chemical formula. It differentiates among confirmed superconductors (\texttt{y}), those explicitly tested and found non-superconducting (\texttt{n}), and those lacking experimental proof (\texttt{have not been proved}). Furthermore, the prompt integrates a ``common sense'' insulator check to automatically rule out stable binary oxides, ternary fluorides, and known salts (\texttt{n(common sense)}), avoiding false positives. If confirmed, the corresponding transition temperature ($\tc$) or range is extracted.
    
    \item \textbf{Synthesis Feasibility Check} (\texttt{is\_easy\_to\_synthesize})\textbf{:} To ensure that identified candidates can be synthesized using standard laboratory equipment, the prompt imposes strict processing limits. A material is deemed experimentally feasible (\texttt{y}) if synthesized at ambient pressure and within the following equipment thresholds:
    \begin{itemize}
        \item Long-term heat preservation in open air: $\leq 1600^\circ$C.
        \item Oxygen-flow/oxygen-rich environments: $\leq 1150^\circ$C.
        \item Oxygen-free sealed environments (quartz tube): $\leq 1200^\circ$C.
        \item Instantaneous heating (arc melting): $\leq 3000^\circ$C.
        \item Hydrothermal intercalation (solution reaction): $\leq 210^\circ$C.
    \end{itemize}
    Materials requiring high pressure ($>5$~GPa) are flagged as unfeasible (\texttt{n}), while those lacking reported synthesis conditions in the literature are marked as \texttt{do not provide}.
    
    \item \textbf{Safety and Toxicity Screening} (\texttt{is\_toxic})\textbf{:} To adhere to safety protocols, the system screens for hazardous constituents. Materials containing specific toxic elements, namely Beryllium (Be), Mercury (Hg), or Thallium (Tl), are flagged as \texttt{y}.
    
    \item \textbf{Experimental Provenance} (\texttt{is\_experimental})\textbf{:} Differentiates between structures sourced directly from the experimental MPDS database (\texttt{y}) versus non-experimental sources (\texttt{n}).
    
    \item \textbf{Chemical Formula Matching} (\texttt{formula\_match})\textbf{:} Verifies whether the simplified integer ratio of the provided POSCAR formula perfectly aligns with the chemical formula reported in the literature (\texttt{y} or \texttt{n}).
    
    \item \textbf{Structural Annotation} (\texttt{structure\_note})\textbf{:} Generates a concise summary (under 50 words) of the verified crystal structure, such as the prototype name or specific lattice properties (\eg, ``NaCl-type cubic'').
\end{enumerate}

We detail the two prompt categories below. To uphold rigorous extraction and classification standards, the LLM analyzes each paper via the Single-Article Analysis Prompt and subsequently executes final aggregation across the processed literature using the Final Synthesis Prompt.

\medskip
\noindent\textbf{Single-Article Analysis Prompt}

\textit{Description: This prompt is designed for analyzing a single academic paper. It instructs the model to compare the provided structural information (POSCAR) of a material with the literature to extract precise data regarding its superconductivity, structural matching, and synthesis conditions.}

You are an expert in superconducting materials and condensed matter physics. I will provide you with the structural information of a material (POSCAR from the experimental database MPDS) and a piece of literature about this material.

\textbf{Important}: The POSCAR chemical formula represents the number of atoms in a unit cell, which might be an integer multiple of the simplest (empirical) chemical formula. For example, Nb$_6$Sn$_2$ = Nb$_3$Sn ($\times$2), Mo$_4$N$_4$ = MoN ($\times$4), Fe$_4$Se$_4$ = FeSe ($\times$4). The literature usually uses the simplest chemical formula. Please reduce the POSCAR chemical formula to its simplest ratio before comparing it with the literature.

Please read the literature carefully and answer the following questions:
\begin{enumerate}[before=\vspace{-0.1em}, topsep=0pt]
    \item \textbf{Superconductivity}: Does this literature report that this material (note: the material corresponding to this specific chemical formula / reduced chemical formula, not other materials) has been experimentally verified as a superconductor?
    \begin{itemize}
        \item Note the distinction: SQUID/MPMS are magnetic measurement devices (mentioning them does not mean the sample is superconducting); the literature might mention the Tc of other materials as a reference (e.g., MgB$_2$ 39K, YBCO 92K); magnetic transitions (Curie/Neel temperatures) $\neq$ superconducting transition temperature.
        \item If it is superconducting, tell me the Tc (superconducting transition temperature), and confirm this Tc belongs to this material.
        \item If the literature explicitly states it is not superconducting, tell me.
        \item If the literature did not test for superconductivity, say ``not tested for superconductivity''.
    \end{itemize}

    \item \textbf{Structure Match}: Does the structure discussed in the literature (lattice parameters, space group, atomic coordinates / structure type) match the given POSCAR structure?
    \begin{itemize}
        \item The same chemical formula may have multiple crystal structures (polymorphs), and different crystal structures may have different superconducting properties.
        \item Pay attention to the conversion between supercells and primitive cells when comparing lattice parameters.
        \item \textbf{Important}: Please determine the structure type based on the atomic coordinates in the POSCAR (such as NaCl-type, NiAs-type, MnP-type, CsCl-type, ZnS-type, etc.). Relying solely on lattice parameters and space groups is not enough to distinguish structure types (for example, under the Pnma space group, the atomic coordinates of MnP-type and FeB-type are completely different).
    \end{itemize}

    \item \textbf{Synthesis Conditions}: How was this material synthesized? Does it require high temperature and high pressure? What are the temperature and pressure ranges?
\end{enumerate}

Please answer concisely in the following format:
\begin{itemize}[before=\vspace{-0.1em}, topsep=0pt]
    \item Superconductivity: [Yes/No/Not tested] Tc=[Temperature]K (if any)
    \item Structure Match: [Yes/Partial/No] [Brief description]
    \item Synthesis: [Brief description of conditions]
\end{itemize}

\medskip

\noindent\textbf{Final Synthesis Prompt (Multiple Articles)}

\textit{Description: This prompt serves as the final aggregation step after multiple articles have been analyzed. It guides the model to make a definitive, overall judgment on the material's properties using strict structural matching rules and fundamental chemical common sense to filter out non-superconductors.}

\vspace{1em}

You are an expert in superconducting materials and condensed matter physics. I have provided you with the structural information of a material and the analysis results from multiple papers.

Now, please synthesize all the literature and provide a final judgment.

\textbf{Important Notes:}
\begin{itemize}[before=\vspace{-0.1em}, topsep=0pt]
    \item Only judge the superconductivity of this specific chemical formula and specific structure (the crystal form given by POSCAR, and the structure type determined by atomic coordinates).
    \item \textbf{EXTREMELY IMPORTANT: The structure must match to judge as `y'!} If the structure of the superconductor reported in the literature (space group, lattice parameters, structure type) is significantly different from the POSCAR (e.g., different crystal system, lattice parameter difference $>$5\%, different structure type), even if the chemical formula is the same, you \textbf{ABSOLUTELY CANNOT} judge it as `y'. Different polymorphs of the same chemical formula have completely different superconducting properties (e.g., a superconducting tetragonal phase does not mean the cubic phase is also superconducting). In this case, it should be judged as ``have not been proved'' (the superconductivity of this POSCAR structure has not been verified).
    \item If the Tc comes from other reference materials (e.g., 39K for MgB$_2$, 92K for YBCO, 200K for H$_3$S, 260K for LaH$_{10}$) instead of this material, it does not count.
    \item Curie temperature, Neel temperature, structural phase transition temperature $\neq$ superconducting transition temperature.
    \item SQUID/MPMS are merely magnetic measurement instruments; mentioning them does not mean the sample is superconducting.
    \item The word ``superconductivity'' appearing in the introduction/review section does not mean this material is superconducting.
    \item If it is a theoretical prediction (DFT calculations, etc.) but lacks experimental verification, count it as ``have not been proved''.
    \item \textbf{EXTREMELY IMPORTANT: If all literature says ``not tested for superconductivity'', then judge as ``have not been proved'', not ``n''. You can only judge as ``n'' when the literature explicitly tested for superconductivity and found it is not superconducting. ``Not tested'' $\neq$ ``Not a superconductor''.}
    \item \textbf{HOWEVER}: If the material is analyzed by chemical valence states and \textbf{belongs to a typical insulator/ionic crystal}, even if the literature did not test for superconductivity, it should be judged as ``n(common sense)''. Please follow these steps strictly to judge:
    \begin{itemize}
        \item[] \textbf{Step 1: Simplify the chemical formula.} Simplify the POSCAR chemical formula to its simplest integer ratio (e.g., Al$_{12}$O$_{18}$ $\rightarrow$ Al$_2$O$_3$, Na$_6$O$_{12}$Sb$_2$Zn$_4$ $\rightarrow$ Na$_3$O$_6$SbZn$_2$).  
        \item[] \textbf{Step 2: Check the following rules one by one (if any rule is met, judge as n(common sense)):}
        \begin{itemize}
            \item[]  \textbf{Rule 1 - Binary stable valence oxides}: If the material is a binary oxide (contains only one metal/non-metal element + oxygen), and the element is in its most stable oxidation state, judge as n(common sense). Examples: TiO$_2$ (Ti most stable at +4), SiO$_2$ (Si most stable at +4), Al$_2$O$_3$ (Al most stable at +3), MnO$_2$ (Mn most stable at +4), MgO (Mg most stable at +2), Fe$_2$O$_3$ (Fe most stable at +3). Counter-examples: Cu$_2$O (Cu has +1/+2, +1 is not the most stable) $\rightarrow$ rule does not apply; VO$_2$ (V has +2/+3/+4/+5, might be metallic) $\rightarrow$ exercise caution.
            \item[] \textbf{Rule 2 - F-containing ternary compounds (all most stable oxidation states)}: If the material is a ternary compound containing F, and all elements except F(-1) are in their most stable (most common) oxidation states with balanced positive and negative charges, judge as n(common sense). Examples: Na$_3$AlF$_6$ (Na+1, Al+3, F-1, all most stable valences), CaF$_2$ (Ca+2, F-1), BaF$_2$. Counter-examples: Contains transition metals with multiple common valences (e.g., CuF$_2$, Cu+2 is not necessarily the most stable) $\rightarrow$ does not apply. 
            \item[] \textbf{Rule 3 - Acid radical salts}: If the material can be identified as a salt containing known acid radical ions, judge as n(common sense). Known acid radicals include: phosphate PO$_4^{3-}$, sulfate SO$_4^{2-}$, nitrate NO$_3^-$, carbonate CO$_3^{2-}$, silicate SiO$_4^{4-}$/Si$_2$O$_7^{6-}$/SiO$_3^{2-}$, borate BO$_3^{3-}$/B$_4$O$_7^{2-}$, chromate CrO$_4^{2-}$, manganate MnO$_4^-$, molybdate MoO$_4^{2-}$, tungstate WO$_4^{2-}$, vanadate VO$_4^{3-}$, aluminate AlO$_2^-$, chlorate ClO$_3^-$/ClO$_4^-$, and other oxoacid radicals formed by halogens/pnictogens/chalcogens. Examples: AlPO$_4$ (aluminum phosphate), Ca$_3$(PO$_4$)$_2$, NaNO$_3$ (sodium nitrate), K$_2$SO$_4$ (potassium sulfate), CaCO$_3$ (calcium carbonate), BaSO$_4$, Li$_2$SiO$_3$. Counter-examples: Contains transition metal oxides but lacks clear acid radicals (e.g., LaCoO$_3$ perovskite) $\rightarrow$ does not apply.
            \item[] \textbf{Rule 4 - Multi-component compounds with all most stable valences and balanced charges}: If the material contains 3 or more elements, all elements are in their most stable (most common) oxidation states, and positive and negative charges are perfectly balanced, judge as n(common sense). Examples: Na$_3$SbZn$_2$O$_6$ (Na+1$\times$3, Sb+5$\times$1, Zn+2$\times$2, O-2$\times$6 $\rightarrow$ +3+5+4-12=0, all most stable valences), MgAl$_2$O$_4$ (Mg+2, Al+3, O-2, spinel but all most stable valences). Counter-examples: YBa$_2$Cu$_3$O$_7$ (Cu has +2/+3 mixed valence states, not all most stable) $\rightarrow$ does not apply; LaFeAsO (Fe+2 is not most stable) $\rightarrow$ does not apply; MgB$_2$ (B has no clear ionic valence, metallic boride) $\rightarrow$ does not apply. 
        \end{itemize}
    \end{itemize}
\end{itemize}

\medskip
\textbf{Important Exclusions}: The following types of materials \textbf{CANNOT} be judged as n(common sense) even if they meet the above rules, because there are many superconductors among them:
\begin{itemize}[before=\vspace{-0.1em}, topsep=0pt]
    \item Elemental metals, alloys, intermetallic compounds (e.g., A15 phase Nb$_3$Sn, Laves phases, etc.)
    \item Metal nitrides/carbides/borides (e.g., NbN, MoC, MgB$_2$)
    \item Layered chalcogenides (e.g., FeSe, NbSe$_2$, TaS$_2$)
    \item Cu-containing oxides (e.g., YBCO, LSCO and other cuprate superconductors)
    \item Heavy fermion compounds (e.g., CeCoIn$_5$, UPt$_3$)
    \item Mixed valence/charge unbalanced compounds (implying metallicity/conductivity)
\end{itemize}

\medskip
Please return 7 results separated by English commas, without spaces:
\begin{enumerate}[before=\vspace{-0.1em}, topsep=0pt]
    \item \texttt{is\_supercond}: Is this material and this structure superconducting? (y/n/n(common sense)/have not been proved)
    \item \texttt{tc}: Superconducting temperature, in K (number, use $\sim$ for range like 3$\sim$5; write n if not superconducting; write n(common sense) if common sense dictates non-superconducting; write have not been proved if unverified)
    \item \texttt{is\_easy\_to\_synthesize}: Is it easy to synthesize? Judgment criteria:
    \begin{itemize}
        \item If existence marker contains hp/hthp (high pressure) $\rightarrow$ n
        \item If synthesis requires high pressure ($>$5GPa) $\rightarrow$ n
        \item Synthesized at normal pressure and meets one of the following conditions $\rightarrow$ y:
        Open environment $\leq$1600$^\circ$C; oxygen flow $\leq$1150$^\circ$C; sealed in quartz tube $\leq$1200$^\circ$C; arc melting $\leq$3000$^\circ$C; hydrothermal $\leq$210$^\circ$C
        \item If the literature does not mention it $\rightarrow$ do not provide
    \end{itemize}
    \item \texttt{is\_toxic}: Does it contain toxic elements (Be, Hg, Tl) $\rightarrow$ y/n
    \item \texttt{is\_experimental}: y if POSCAR is from the experimental database (MPDS), otherwise n
    \item \texttt{formula\_match}: Does the CSV chemical formula match the POSCAR chemical formula? (y/n)
    \item \texttt{structure\_note}: Structural description (brief, e.g., ``NaCl-type cubic a=4.24\AA, Tc for this phase'' or ``A15 Cr$_3$Si-type'', max 50 words)
\end{enumerate}

Output examples: y,3$\sim$5,y,n,y,y,NaCl-type cubic. Or: have not been proved,have not been proved,y,n,y,y,hexagonal WC-type.
Or: n,n,n,n,y,n,high-pressure phase only.Or: n(common sense),n(common sense),y,n,y,y,ionic insulator Al$_2$O$_3$ stable oxide

\subsection{LLM Hallucination Prevention and Manual Verification}
\label{manual}

To categorize candidates into verified positive, verified negative, and unverified instances, we execute the GPT-5.4 extraction pipeline over three independent times and take the union of the identified positive and negative instances. We manually verify all extracted positive instances. 
For the remaining unverified instances, we employ Opus-4.6 for a secondary review. The extraction accuracies across the three individual GPT-5.4 runs are 143/158, 144/158, and 145/158, respectively. However, taking the union of these extractions yields an improved overall accuracy of 154/158, demonstrating the necessity and effectiveness of a multi-pass extraction strategy. The four additional positive instances initially missed by GPT-5.4 are BW, Zr$_2$Ir, V$_3$Pb, and CaSi$_2$.

Our analysis reveals that these omissions largely stem from the model's inadequate spatial reasoning regarding crystal representations. Specifically, while the reference POSCAR files provide primitive cells, the source literature reports the structures of BW, Zr$_2$Ir, and CaSi$_2$ as conventional cells. Despite representing the identical underlying structure, the differing coordinate systems prevent GPT-5.4 from accurately recognizing the match. Additionally, in the case of V$_3$Pb, the compound is explicitly listed in Table~1.2 of the source text~\cite{V3Pb}, yet GPT-5.4 fails to extract it.

\subsection{Experimental Synthesis and Verification}

In this section, we detail the experimental conditions and synthesis methods, to comprehensively validate the promising candidates identified by our agentic screening workflow.

\textbf{Sample Synthesis and Optimization.} Polycrystalline samples of the selected candidates Hf$_{21}$Re$_{25}$, HfZrRe$_4$, HfZr$_3$Re$_8$, Hf$_3$ZrRe$_8$, HrZrRe, Zr$_2$VRe$_3$, Zr$_4$VRe$_7$, and Zr$_3$ScRe$_8$, are synthesized utilizing the arc melting method. High-purity elemental powders of Hf, Zr, Re, and V (99.99\%) serve as the starting materials. All handling and weighing procedures are meticulously conducted within an argon-filled glove box to prevent oxidation. The raw materials are mixed and pressed into 3~g pellets under a pressure of 2~t/cm$^2$, then subsequently melted on a water-cooled copper hearth under a high-purity argon atmosphere (4N). To ensure macroscopic compositional homogeneity, each ingot is flipped and remelted a minimum of eight times.

While other samples are synthesized successfully using their exact stoichiometric ratios, the preparation of single-phase Hf$_{21}$Re$_{25}$ requires systematic optimization. Direct arc melting of Hf and Re at the stoichiometric ratio (21:25) results in significant phase separation, yielding a mixture of Hf$_{21}$Re$_{25}$ and HfRe$_2$. Because the secondary phase HfRe$_2$ is a known superconductor with a relatively high transition temperature, its presence interferes with the intrinsic property measurements of our target phase. To successfully suppress the formation of HfRe$_2$, we systematically vary the starting Hf:Re molar ratio (1:1, 1.1:1, 1.2:1, 1.3:1, and 1.4:1). We find that a starting ratio of 1.2:1 yields the optimal phase purity.

\textbf{Structural and Magnetic Characterization.} The crystal structures and phase purities of all as-synthesized samples are characterized by Powder X-Ray Diffraction (PXRD) utilizing a Rigaku diffractometer equipped with Cu K$\alpha$ radiation. For the optimized Hf$_{21}$Re$_{25}$ sample, PXRD analysis confirms that the final product consists predominantly of the target phase, with only a minor trace of elemental Hf impurity. Crucially, elemental Hf exhibits a $\tc$ of only 0.12~K, which is drastically lower than the transition observed in our measurements, confirming that the observed superconductivity arises intrinsically from the Hf$_{21}$Re$_{25}$ main phase. The macroscopic superconducting properties of all validated samples are then rigorously investigated through AC susceptibility measurements performed on a Quantum Design DynaCool Magnetic Properties Measurement System (MPMS). More details including both magnetic and electrical characterizations are provided in \cref{sec:X-Ray_Diffraction}.

\subsection{Evaluation Metrics}
To comprehensively train and evaluate \modelname{} across property prediction, interatomic potential estimation, and structure prediction, we adopt the following metrics throughout this work.

\textbf{Mean Absolute Error (MAE).}
The MAE quantifies the average magnitude of prediction errors. We distinguish two formulations depending on the physical nature of the target quantity:
\begin{itemize}
    \item \textbf{Energy MAE.} For total energy predictions, the error is normalized by the number of atoms $N$ in each system, yielding a per-atom MAE (typically expressed in meV/atom):
    \begin{equation}
        \text{MAE}_{E} = \frac{1}{M} \sum_{i=1}^{M} \frac{1}{N^{(i)}} | \hat{E}_i - E_i|,
    \end{equation}
    where $M$ denotes the total number of test samples, $\hat{E}_i$ and $E_i$ are the predicted and ground-truth energies for the $i$-th sample, and $N^{(i)}$ is its atom count.
    
    \item \textbf{Property MAE.} For other intensive or invariant physical properties (\eg, band gap, $\tc$), the MAE is computed directly without per-atom normalization:
    \begin{equation}
        \text{MAE}_{\text{prop}} = \frac{1}{M} \sum_{i=1}^{M} \left| \hat{y}_i - y_i \right|,
    \end{equation}
    where $\hat{y}_i$ and $y_i$ denote the predicted and ground-truth property values, respectively.
\end{itemize}

\textbf{Coefficient of Determination ($R^{2}$).}
To assess the goodness-of-fit for regression tasks, we report the $R^{2}$ score, which measures the proportion of target variance captured by the model:
\begin{equation}
    R^{2} = 1 - \frac{\sum_{i=1}^{M} (y_i - \hat{y}_i)^2}{\sum_{i=1}^{M} (y_i - \bar{y})^2},
\end{equation}
where $\bar{y} = \frac{1}{M}\sum_{i=1}^{M} y_i$ is the mean of the ground-truth values. An $R^{2}$ approaching unity indicates near-perfect predictive fidelity.

\textbf{Match Rate (MR).}
For generative structure prediction, MR measures the fraction of ground-truth structures in the test set that are successfully recovered. A generated structure $\hat{\gG}$ is deemed a match to the ground truth $\gG$ if it satisfies the crystallographic tolerances enforced by the \texttt{StructureMatcher} algorithm from \texttt{pymatgen}~\citep{pymatgen}: (1) Site tolerance (\texttt{stol}): $0.5$; (2) Angle tolerance (\texttt{angle\_tol}): $10^{\circ}$; (3) Lattice length tolerance (\texttt{ltol}): $0.3$. 
MR is then defined as the proportion of test samples for which the generated candidate satisfies these criteria.

\textbf{Root Mean Square Error (RMSE).}
We use RMSE to evaluate both structural geometric fidelity, and the accuracy of energy/force predictions on the DPA-2 dataset, with distinct formulations below:
\begin{itemize}
    \item \textbf{Structure Prediction RMSE:} To quantify geometric fidelity, we calculate the RMSE of the structural difference between the ground truth and the predicted structure for successfully matched pairs. To account for varying cell sizes, this RMSE is normalized by the cube root of the volume per atom, $\sqrt[3]{V/N}$.
    
    \item \textbf{Energy and Force RMSE:}  This RMSE is computed by averaging the squared differences over all individual scalar components across the dataset:
\begin{equation}
    \text{RMSE} = \sqrt{\frac{1}{M \cdot D} \sum_{i=1}^{M} \sum_{d=1}^{D} \left( T_{i,d} - P_{i,d} \right)^2},
\end{equation}
where $M$ denotes the total number of evaluated items (\eg, the total number of structures for macroscopic properties, or the total number of atoms for atomic-level properties), and $D$ represents the dimensionality of the target property ($D=1$ for energy, $D=3$ for force). $T_{i,d}$ and $P_{i,d}$ are the ground truth and predicted values for the $d$-th dimension of the $i$-th item, respectively.
\end{itemize}

    \section{Author Contributions Statement}\label{sec:author}
M.L., Y.R., and S.L. conceived the model and agentic framework, performed the training and experiments, developed the software, and drafted the manuscript, under the supervision of W.H. 
L.Wa. conducted materials synthesis and experimental validation under the supervision of S.J. 
T.B. contributed to the engineering of the agentic system under the supervision of T.X. 
D.Z. provided computational resources and technical guidance. 
J.C., L.Wu., and A.L. contributed to the exploration of the model architecture, data cleaning, and figure drawing. 
Q.L., and P.W. assisted in dataset construction. 
Z.L., R.J., H.S., J.Z., and J.-R.W. provided technical support. 
W.H., Y.R., D.Z., S.J., and T.X. provided overall research supervision. 
All authors contributed to manuscript preparation and reviewed the final manuscript.

    \putbib[Reference]
    \clearpage

\setcounter{figure}{0}
\renewcommand{\figurename}{Extended Data Figure}
\setcounter{table}{0}
\renewcommand{\tablename}{Extended Data Table}

\renewcommand{\theHfigure}{Ext.\arabic{figure}}
\renewcommand{\theHtable}{Ext.\arabic{table}}

\crefalias{figure}{ext:figure}
\crefalias{table}{ext:table}

\crefname{ext:figure}{Extended Data Figure}{Extended Data Figures}
\Crefname{ext:figure}{Extended Data Figure}{Extended Data Figures}

\crefname{ext:table}{Extended Data Table}{Extended Data Tables}
\Crefname{ext:table}{Extended Data Table}{Extended Data Tables}

\begin{figure}[htbp!]
\centering
\includegraphics[width=0.99\textwidth]{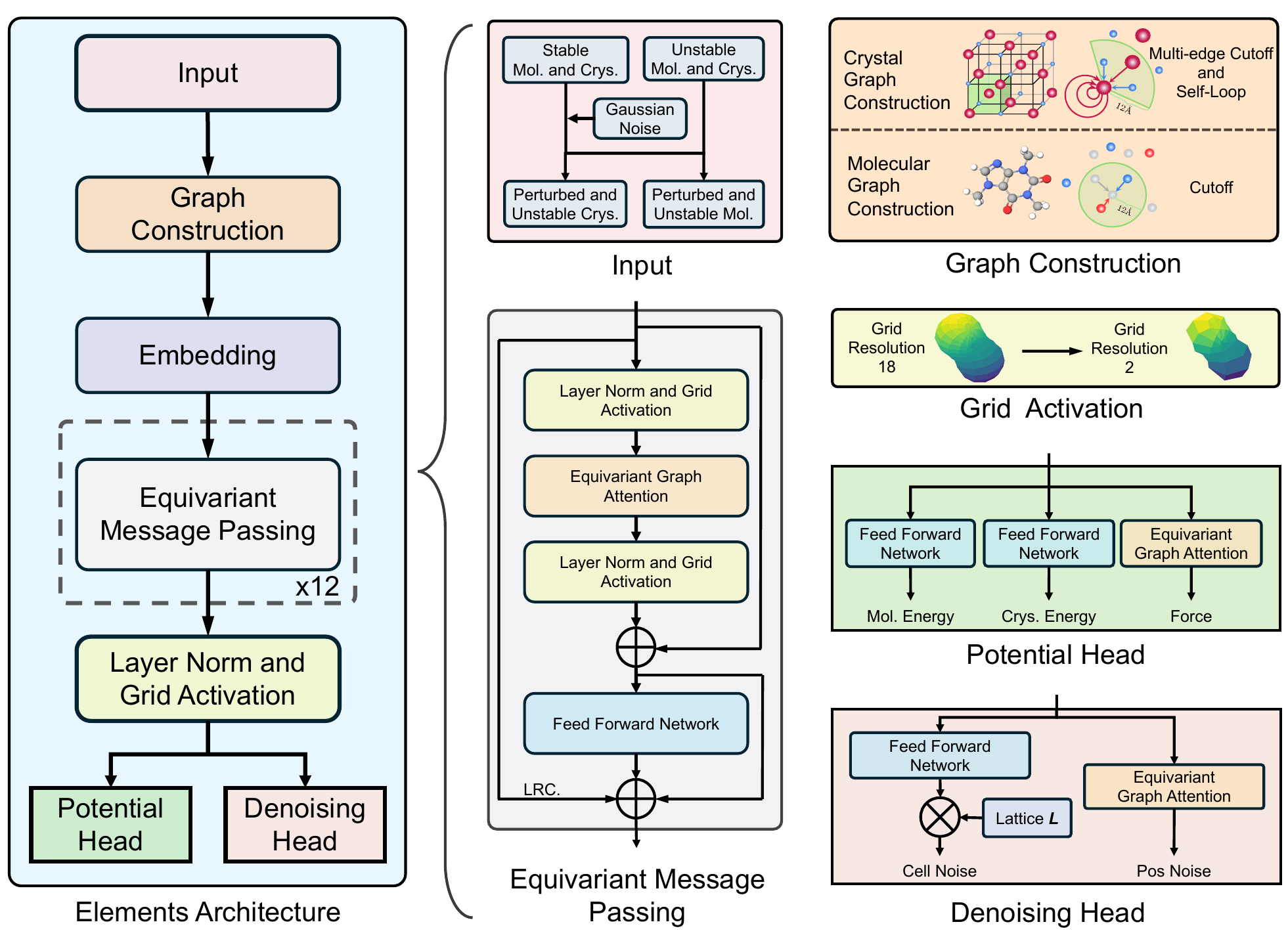}
\caption{The architecture of \modelname{}. Our model leverage  EquiformerV2 as the backbone, with several key innovations introduced in the input processing, model architecture, and output head. The detailed description is provided in~\ref{para:model_ablation}.}
\label{fig:ext_model}
\end{figure}

\begin{figure}[htbp!]
\centering
\includegraphics[width=0.99\textwidth]{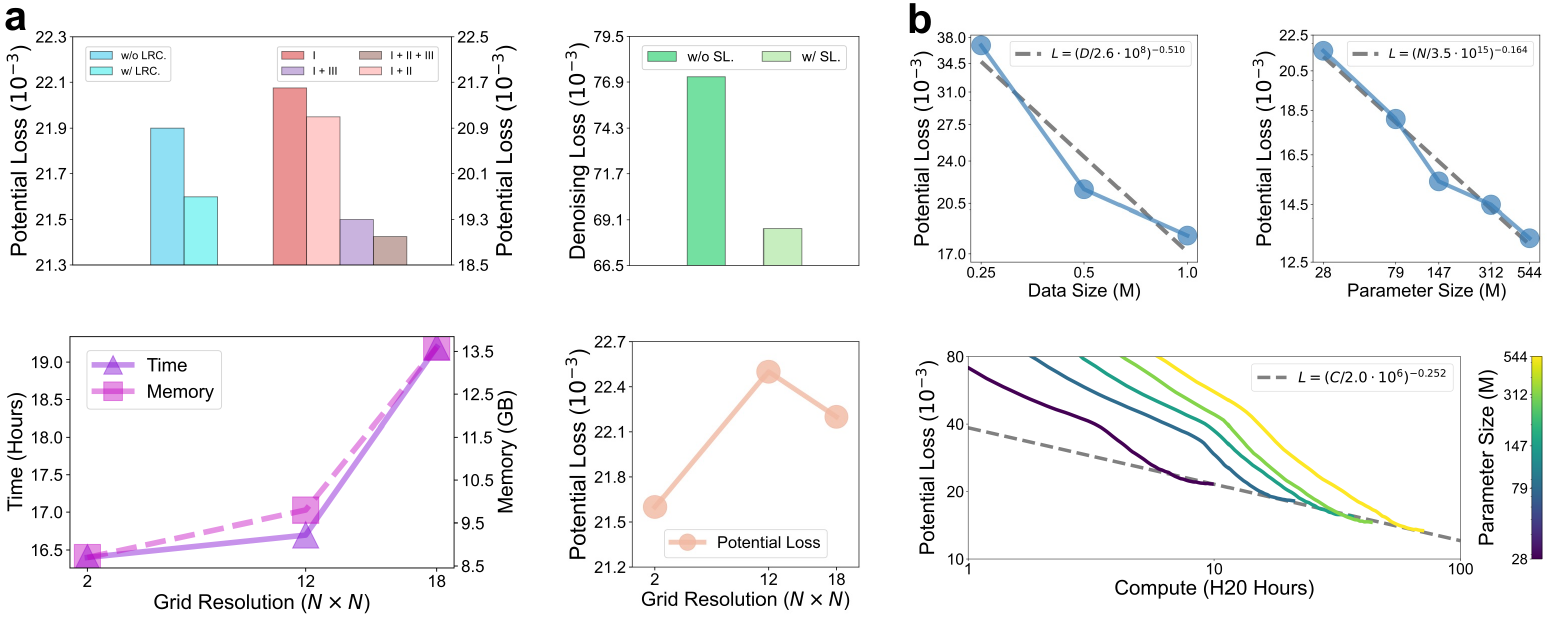}
\caption{
Ablation studies and the scaling law of \modelname{}. 
\textbf{a}, The ablation results of the key innovations on the validation set, including Long-Range connection (LRC), training data composition, Self-Loop (SL), and grid resolution. We reported the denoising loss for the SL technique, and potential loss of other innovations. For training data composition, we tested our model trained on distinct compositions of unstable crystals (I), unstable molecules (II), and stable structures (III). For grid resolution, we also demonstrated the time/memory cost.     
\textbf{b}, Scaling Laws of \modelname{}. Model performance follows predictable power-law relationships with respect to dataset size, parameter size and training compute power. In all panels, Solid lines denote empirical measurements and dashed lines represent power-law fits.
}
\label{fig:ext_ab_scale}
\end{figure}

\end{bibunit}

\setappendix
\begin{bibunit}[sn-mathphys-num]
    \section{Preliminary}\label{sec:preliminary}

\subsection{Geometric Graphs and Symmetry}

We model an atomic system as a geometric graph $\gG = (\mA, \mX, \gE)$, following the notation established in \cref{sec:notation}. A central requirement for modeling atomic systems is to respect the symmetries of the Euclidean group $E(3)$, which comprises translations, rotations, and reflections. Let $g \in E(3)$ denote a transformation acting on the graph as $g \cdot \gG = (\mA, g \cdot \mX, \gE)$. That is, $g$ acts on the spatial coordinates while leaving the atomic identities invariant. Within this framework, two classes of mappings are fundamental:
\begin{itemize}
    \item \textbf{$E(3)$-Invariant} mappings satisfy $\phi(g \cdot \gG) = \phi(\gG)$. Such mappings are essential for predicting scalar properties, including total energy $E_{\text{total}}$, formation energy $E_{\mathrm{form}}$, and the superconducting critical temperature $\tc$, whose values must remain independent of the coordinate frame.
    \item \textbf{$E(3)$-Equivariant} mappings satisfy $\phi(g \cdot \gG) = \mD_g \cdot \phi(\gG)$, where $\mD_g$ denotes the representation of $g$ in the output space. These mappings are crucial for predicting vector or tensor fields, such as atomic forces $\mF$ and denoising directions $\hat{\boldsymbol{\epsilon}}_{\text{pos}}$, that must transform consistently with the coordinate system.
\end{itemize}

\subsection{Equivariant Graph Neural Networks}

To encode geometric graphs while preserving physical symmetries, numerous equivariant graph neural networks (GNNs) have been proposed, achieving remarkable success across a wide range of scientific applications~\citep{supp_han2025survey,supp_zhang2025artificial,supp_huang2026geometric}. We briefly review two principal design paradigms and highlight the architectural family adopted in \modelname{}.

\textbf{Message Passing Neural Networks.}
The majority of equivariant GNNs are developed within the Message Passing Neural Network (MPNN) framework~\citep{supp_mpnn}. Depending on the operators employed, these architectures can be broadly categorized into two families. \textit{Scalarization-based} models, such as EGNN~\citep{supp_EGNN}, PaiNN~\citep{supp_painn}, and HEGNN~\citep{supp_hegnn}, convert geometric vectors into invariant scalars (\eg, interatomic distances $\|\vx_i - \vx_j\|$ or angular inner products $(\vx_j - \vx_i)^\top (\vx_k - \vx_i)$) before message passing. While computationally efficient, this projection discards directional information. In contrast, \textit{tensor-product-based} methods, including TFN~\citep{supp_tfn}, MACE~\citep{supp_Mace}, and NequIP~\citep{supp_NequIP}, perform tensor products over irreducible representations (irreps), enabling direct interactions between features of different angular degrees $l$. Although computationally more demanding, these designs enable richer equivariant basis construction~\citep{supp_uniegnn,supp_xie2025price} and more faithful mappings between feature spaces~\citep{supp_dym2021on,supp_lin2026reducing}, thereby providing stronger expressivity supported by both theoretical analyses and empirical studies.

\textbf{Geometric Graph Transformers.}
Inspired by the success of Transformers~\citep{supp_vaswani2017attention} and their graph-structured variants~\citep{supp_yuan2025survey}, geometric graph Transformers have emerged as powerful architectures for modeling geometric data. Notable examples include SE(3)-Transformer~\citep{supp_SE3_Transformer}, Equiformer~\citep{supp_Equiformer}, and EquiformerV2~\citep{supp_EquiformerV2}. Among these, we adopt \textbf{EquiformerV2} as the backbone of \modelname{} due to two key advantages:
\begin{enumerate}
    \item \textbf{Scalability.} EquiformerV2 enhances training stability through an equivariant attention mechanism coupled with a normalization layer, enabling reliable optimization of large parameterizations. Building on this foundation, we introduce the Long-Range Residual Connection (LRC) mechanism to further improve scalability, enabling models with up to 1B parameters.
    \item \textbf{Efficiency.} EquiformerV2 introduces the eSCN convolution~\citep{supp_escn}, which reduces the computational complexity of tensor products from $\gO(L^6)$ to $\gO(L^3)$, thereby enabling higher-degree steerable representations (\eg, $L_{\max} = 6$). Within this framework, we further reduce the grid resolution $R$ of the $\sS^2$ activations in the eSCN convolution, decreasing both computational cost and GPU memory consumption without sacrificing prediction accuracy (\cref{tab:grid_test}).
\end{enumerate}

\subsection{Generative Models for Crystal Structure Prediction}

Crystal structure prediction requires the joint generation of a lattice matrix $\mL \in \R^{3 \times 3}$, atomic types $\mA \in \mathbb{N}^{1 \times N}$, and fractional coordinates $\mS \in [0, 1)^{N \times 3}$. The generation process is typically formulated as a joint diffusion over these components.

DiffCSP~\citep{supp_diffcsp} pioneered the joint diffusion of lattice parameters and fractional coordinates. However, its denoising network operates directly in \textit{fractional coordinate space}, which is non-Euclidean. Since distances depend on the dynamic lattice $\mL$, performing convolution in this space can lead to physically inconsistent interaction modeling.

To address this limitation, MatterGen~\citep{supp_mattergen} proposed a refinement: while the diffusion process (noising) remains in fractional space to satisfy periodic boundary conditions, the denoising backbone projects atoms into \textit{Cartesian space} ($\mX = \mS \cdot \mL$) to compute interactions. This design ensures that the network learns from physically valid distances and angles.

\section{Dataset Description}
\subsection{Pretrained Datasets}\label{app:data_details}
\begin{figure}[htbp]
    \centering
    \includegraphics[width=\textwidth]{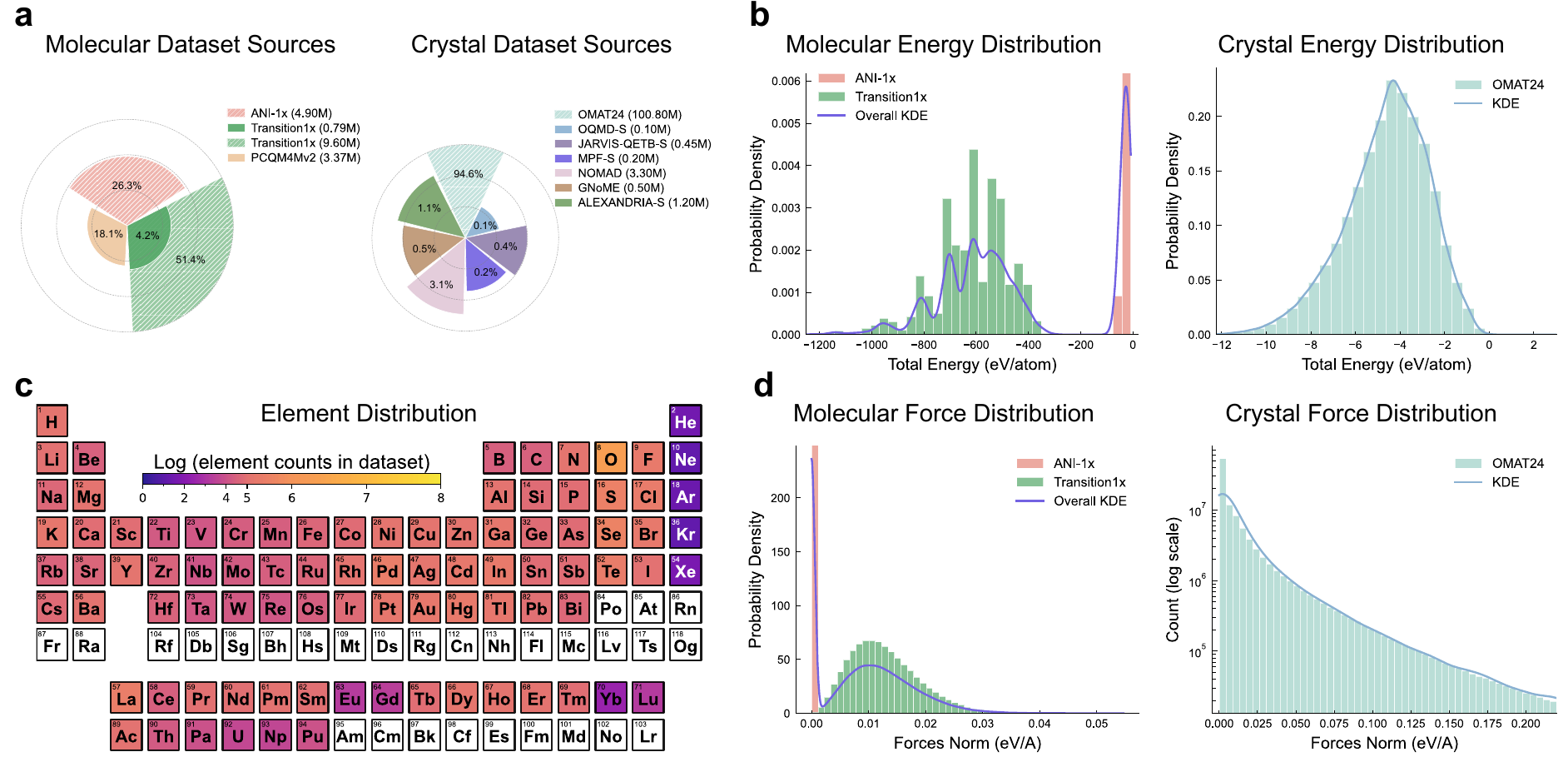}
    \caption{\textbf{a}, Proportions of publicly-sourced datasets for molecules and crystals. \textbf{Hatched sectors denote datasets of unstable structures; all others represent stable structures.} \textbf{b}, Energy distributions of the unstable structures in the publicly-sourced datasets for molecules and crystals. \textbf{c}, Elemental distribution mapped onto the periodic table, illustrating the broad chemical coverage achieved. \textbf{d}, Force distributions of the unstable structures in the publicly-sourced datasets for molecules and crystals.}
    \label{fig:supp_fig_1_appendix}
\end{figure}

\subsubsection{Non-equilibrium Datasets}

The construction of non-equilibrium datasets aims to sample the configurational space as broadly as possible, specifically targeting regions characterized by high energies and significant forces. For the training of Universal Interatomic Potentials (UIPs), the model must learn the restoring forces that arise when atoms deviate from their equilibrium positions. This necessitates that the training data extends beyond perfect lattices or molecules to include a substantial volume of ``perturbed'' states.

\textbf{Crystal Datasets (Non-Equilibrium).} In the domain of crystalline materials, \textbf{OMAT-24~\citep{supp_OMAT-24}} represents the current state-of-the-art for open-source datasets in terms of scale and diversity. Released by Meta FAIR, this dataset aims to address the limitations of traditional datasets (such as the Materials Project~\citep{supp_mp-20mpts-52}), which primarily focus on relaxed structures and lack dynamical information. OMAT-24 contains over 110 million single-point Density Functional Theory (DFT) calculations, covering the chemical space of the vast majority of inorganic materials in the periodic table. To serve force field prediction tasks, OMAT-24 employs a highly targeted data generation strategy. In addition to standard AIMD (Ab Initio Molecular Dynamics) trajectories, the dataset incorporates a substantial volume of ``Rattled'' structures. Specifically, researchers applied random Gaussian noise to the atomic positions and lattice vectors of equilibrium crystal structures to generate non-equilibrium configurations. This method efficiently samples gradient information near the bottom of the potential energy surface wells, enabling models to learn robust atomic forces. Furthermore, OMAT-24 includes intermediate trajectories from structural relaxation processes, which naturally provide gradient paths evolving from high-energy to low-energy states. Statistically, the volume of OMAT-24 data utilized in this pre-training set reaches 100.8 million samples, all derived from the official source. Its immense scale makes it the premier data source for training universal force-field prediction and energy prediction models, significantly enhancing the model's generalization capabilities across unseen chemical compositions.

\textbf{Molecular Datasets (Non-Equilibrium).}
To comprehensively map the potential energy surfaces of organic molecules, ranging from near-equilibrium fluctuations to high-energy reaction pathways, we integrate the following key datasets: 
\begin{itemize}
    \item \textbf{ANI-1x~\citep{supp_ani-1x}.} In the realm of organic small molecules, ANI-1x stands as a prime example of constructing non-equilibrium datasets through active learning strategies. It specifically targets vibrational modes and geometric distortions in the vicinity of equilibrium, ensuring the model captures the subtle energetics of thermal fluctuations. This dataset is designed to train the ANI series of neural network potentials. In contrast to traditional grid sampling or random sampling, the construction of ANI-1x integrates a dynamic feedback loop: it utilizes a preliminarily trained potential ensemble to predict new configurations, specifically screening for those where the prediction variance (\ie, uncertainty) among ensemble members is highest, to undergo high-precision DFT calculations (at the $\omega$B97x/6-31G(d) level). This strategy ensures that the data points in ANI-1x are highly concentrated in ``hard-to-learn'' regions of chemical space, such as bond dissociation, atypical dihedral rotations, and high-energy regions characterized by steric clashes (atoms in close proximity). Consequently, although primarily composed of just four elements (H, C, N, O), the dataset encompasses extremely high conformational diversity, effectively preventing unphysical model collapse during simulations. In this pre-training collection, ANI-1x contributes 4.9 million  molecular conformations, derived from the official source. These non-equilibrium conformations provide precise atomic force labels for the model, serving as the foundation for realizing organic molecular dynamics simulations.
    \item \textbf{Transition1x~\citep{supp_Transition1x}.} While ANI-1x concentrates on near-equilibrium physics, Transition1x extends the energy spectrum to the extreme by focusing on chemical reaction processes. By capturing high-energy configurations along reaction coordinates, it provides essential data for modeling bond breaking and formation, significantly expanding the scope of non-equilibrium sampling. This dataset contains reaction pathways for approximately 10,000 organic chemical reactions. The data generation employs the Nudged Elastic Band (NEB) method, which inserts a series of intermediate ``Images'' between reactants and products and simultaneously optimizes them to locate the Minimum Energy Path (MEP). The vast majority of configurations in Transition1x are situated on the ascending or descending slopes of reaction barriers, possessing extremely high potential energies and significant atomic forces. This data is critical for training machine learning potentials capable of describing chemical reactivity, as traditional equilibrium datasets (such as QM9~\citep{supp_qm9}) completely lack information regarding the Transition State (TS) region. The volume of Transition1x data utilized in this collection stands at 9.6 million, obtained from the official source. Its high-energy, non-equilibrium nature makes it a challenging yet highly valuable supplement for force field prediction tasks, endowing models with the ``reaction intuition'' necessary to describe bond breaking and formation.
\end{itemize}

\subsubsection{Equilibrium Datasets}
In contrast to force field prediction, structure prediction (position denoising) and lattice parameter prediction (cell denoising) require the model to learn the ``stability distribution'' of matter. Consequently, the training data must consist of equilibrium structures obtained via Geometry Optimization (GO), specifically, configurations where atomic forces have converged below a threshold (typically $\|\mF\| < 0.02$ eV/\AA). These data define the valid manifold within chemical space.

\textbf{Crystal Datasets (Equilibrium).}
To construct an equilibrium dataset covering the global materials space, we integrate multiple high-confidence databases, totaling approximately 5.75 million (5.75M) structures. These data were primarily acquired via the \texttt{jarvis\_tools~\citep{supp_jarvis}} interface, while also including directly downloaded official sources. The specific composition is as follows:
\begin{itemize}
    \item \textbf{ALEXANDRIA-S.} 
    Here, we present ALEXANDRIA-S, a high-fidelity subset extracted from the ALEXANDRIA database ~\citep{supp_Alex}, currently the most extensive open repository of DFT calculations, encompassing over 5 million entries for periodic 1D, 2D, and 3D compounds. While the parent ALEXANDRIA dataset provides an unprecedented coverage of chemical space (spanning 83 elements and 2.6 million unique compositions ), it contains a significant portion of high-energy configurations sampled during machine-learning-accelerated discovery rounds. We curated ALEXANDRIA-S by downloading the database via the \texttt{jarvis-tools} and implementing a stringent thermodynamic stability filter. By selecting only structures with a distance to the convex hull ($E_{hull}$) $\le$ 0.08 eV/atom, we filtered the collection down to 1.4 million equilibrium or near-equilibrium structures.
    \item \textbf{GNoME~\citep{supp_genome}.} Released by Google DeepMind, the GNoME (Graph Networks for Materials Exploration) dataset represents a milestone in AI-assisted materials discovery. This dataset generates candidate structures via graph neural networks, verified by DFT, greatly expanding the number of known stable crystals. This collection utilizes 0.5 million structures, mainly corresponding to its publicly released stable and metastable subsets. The strict convex hull stability screening applied to these structures positions the dataset as a gold standard for training generative models to learn thermodynamic stability.
    \item \textbf{OQMD-S} 
    Here, we introduce OQMD-S, a curated subset derived from the Open Quantum Materials Database (OQMD)~\citep{supp_oqmd}, one of the most established repositories of DFT calculated thermodynamic and structural properties, currently encompassing over 1.4 million materials. While the parent OQMD offers a vast mapping of the inorganic chemical space, it includes a significant density of metastable and high-energy phases essential for exploring potential synthesis pathways. We first utilize \texttt{jarvis-tools} to download OQMD. By imposing a rigorous energetic constraint, specifically selecting structures with a distance to the convex hull ($E_{\text{hull}}$) $\le$ 0.08 eV/atom, we isolated a collection of 0.1 million equilibrium or near-equilibrium structures to form OQMD-S.
    \item \textbf{JARVIS-QETB-S.} 
    Here, we present JARVIS-QETB-S, a curated near-equilibrium subset derived from universal three-body Tight-Binding (TB) database~\citep{supp_jarvis_qetb} covering 65 elements and 2,080 binary combinations. 
    While the parent JARVIS-QETB dataset comprises over 0.8 million DFT calculations, many configurations involve high-energy structures during recursive active-learning cycles. 
    We download JARVIS-QETB from \texttt{jarvis\_tools} and extract JARVIS-QETB-S by applying a stringent force-based filter:
    \begin{equation}
    \bar{F} = \frac{1}{N} \sum_{i=1}^{N} \|\mF_i\| \leq 20 \text{ meV/\AA},
    \label{Eq:force_filter}
    \end{equation}
    where $N$ is the atom number. This filtering yields 0.45 million equilibrium crystal configurations.
    \item \textbf{MPF-S.}
    Here, we present MPF-S, a curated large-scale subset derived from the Materials Project Force (MPF) database~\citep{supp_M3GNet}, the most comprehensive collection of DFT relaxation trajectories for inorganic crystals to date. While the parent MPF dataset (version 2021.2.8) encompasses a vast chemical space of 89 elements with over 187,000 structural snapshots, its broad coverage includes numerous non-equilibrium configurations sampled during early-stage ionic relaxation steps. We download MPF from \texttt{jarvis-tools} and apply force-based filter defined in \cref{Eq:force_filter}. This force-based filtering results in 0.2 million equilibrium structures that form MPF-S.
    \item \textbf{NOMAD~\citep{supp_nomad}.} As the world's largest repository for computational materials science data, NOMAD aggregates calculation results from various codes (VASP~\citep{supp_hafner2008ab}, QUANTUM ESPRESSO~\citep{supp_giannozzi2009quantum}, etc.). This collection screened 3.3 million equilibrium structures. The introduction of NOMAD greatly enriches the diversity of elemental combinations and crystal structure prototypes, enabling the generative model to encounter extremely rare or complex crystal configurations.
\end{itemize}

\textbf{Molecular Datasets (Equilibrium).}
In molecular denoising tasks, the objective of the data is to provide accurate 3D equilibrium conformers:
\begin{itemize}
    \item \textbf{PCQM4Mv2~\citep{supp_pcq}.} This dataset originates from the OGB-LSC~\citep{supp_pcq} challenge and is built upon the PubChemQC project~\citep{supp_nakata2017pubchemqc}. PCQM4Mv2 provides not only the graph topology information (SMILES) of molecules but also the corresponding 3D coordinates situated at the minima of the potential energy surface. For denoising diffusion models, PCQM4Mv2 is the most critical data for learning the mapping from noisy graphs to 3D molecules. This collection utilized 3.37 million molecules.
    \item  \textbf{Transition1x-S.} 
    While Transition1x~\citep{supp_Transition1x} describes reaction dynamics, it encompasses critical stationary points including reactants, products, and transition states. Thus, we extracted a near equilibrium subset, Transition1x-S, from original Transition1x dataset by the same force-based filter as \cref{Eq:force_filter}.
    This filtering yields 0.79 million equilibrium molecular configurations.
\end{itemize}

\subsection{Downstream Datasets}

\textbf{QM9~\citep{supp_qm9}.} To evaluate molecular property prediction, we utilize the QM9 dataset, which consists of approximately 134,000 stable small organic molecules made up of (C,H,O,N,F) atoms. Following standard benchmarks, we focus on two electronic properties critical for chemical reactivity and stability: the energy of the Highest Occupied Molecular Orbital (HOMO) and the energy of the Lowest Unoccupied Molecular Orbital (LUMO).

\textbf{Matbench~\citep{supp_matbench}.} For crystalline materials, we employ selected tasks from the Matbench benchmark, specifically focusing on properties relevant to electronic and superconducting behavior. We utilize the following subsets:
\begin{itemize}
    \item \textbf{matbench\_is\_metal:} A classification task to distinguish metals from non-metals (approx. 106k entries).
    \item \textbf{matbench\_dielectric:} Regression of the refractive index (approx. 4.7k entries).
    \item \textbf{matbench\_mp\_gap:} Prediction of the DFT band gap (approx. 106k entries).
    \item \textbf{matbench\_perovskites:} Prediction of formation energy for ABO$_3$ perovskites (approx. 18.9k entries), a structural family highly relevant to high-temperature superconductivity.
\end{itemize}

\textbf{DPA-2 Datasets~\citep{supp_DPA-2}.} To assess the model's capability in handling unstable states and potential energy surfaces (PES) across diverse chemical spaces, we utilize the large-scale datasets curated for the DPA-2 project. These datasets cover both molecular and periodic systems, including:
\begin{itemize}
    \item \textbf{Crystals:} Contains multi-principal element alloys and pure bulk metals (\eg, Vanadium, Tungsten) under various thermal conditions and covers complex ionic transport and polarization phenomena, enriching the model's understanding of ionic lattices.
    \item \textbf{Molecules:} Dynamics of water (H$_2$O) and drug-like molecules (Drug).
    \item \textbf{Adsorbates/Mixtures:} The OC2M subset~\citep{supp_chanussot2021open}, focusing on catalyst-adsorbate interactions.
\end{itemize}

\textbf{MP-20 and MPTS-52~\citep{supp_mp-20mpts-52}.} For the generative tasks, we evaluate the model's ability to generate chemically valid and structurally stable crystals.
\begin{itemize}
    \item \textbf{MP-20:} A standard benchmark consisting of stable crystals with at most 20 atoms in the unit cell (45,231 structures).
    \item \textbf{MPTS-52:} A complementary dataset focusing on larger unit cells containing 40,476 structures.
\end{itemize}

\textbf{JARVIS-DFT (Transport).} We select electronic and thermal transport properties from the JARVIS-DFT database~\citep{supp_jarvis}. Specifically, we predict the Seebeck coefficient, thermal conductivity ($\kappa$), and electrical conductivity, which provide insight into the electronic transport behavior of materials.

\textbf{Custom DFT $\tc$.} To directly target superconductivity, we curated a dataset of Critical Temperatures ($\tc$). This dataset aggregates data from two primary sources:
\begin{enumerate}
    \item Sampling the Materials Space for Conventional Superconducting Compounds~\citep{supp_dfttc}: A high-throughput screening of conventional superconductors (8241 structures).
    \item JARVIS Superconductors: Calculated superconducting materials from the JARVIS-DFT database (1227 structures)~\citep{supp_jarvis}.
\end{enumerate}

\textbf{SuperCon3D.}
The SuperCon3D dataset~\citep{supp_sodnet} addresses a critical gap in superconducting materials research: the lack of 3D structural information in legacy databases. While the original NIMS SuperCon database~\citep{supp_supercon} contains over 33,000 experimental critical temperature ($\tc$) records, it provides only chemical formulas (\eg, YBa$_2$Cu$_3$O$_7$) without atomic coordinates. SuperCon3D bridges this gap by systematically aligning these formulas with high-fidelity DFT relaxed structures from the Materials Project~\citep{supp_mp-20mpts-52}. This alignment enables the application of geometric deep learning and graph neural networks that require precise atomic positions and bond lengths as input. For the purposes of this study, we exclusively utilize the \textbf{Ordered} subset of SuperCon3D.

\textbf{Positive Instances and Negative Instances: Newly Constructed Dataset from our agentic system.} 
To train and validate the classification model (\modelname{C}), we construct a specialized dataset derived from our agentic literature survey, explicitly termed Positive Instances and Negative Instances.
Initially, the literature mining process identifies $158$ unique positive compositions (confirmed superconductors) and $385$ unique negative compositions (confirmed non-superconductors). To enhance the model's robustness and capture the realistic variance in experimental reporting, we implement a \textit{structural augmentation strategy}. Since different studies often report slight variations in lattice parameters or atomic coordinates for the same nominal composition due to synthesis conditions or measurement precision, we retain all distinct structural entries reported across the retrieved literature. This expansion results in a total dataset size of $1,138$ positive structural entries and $2,026$ negative structural entries.

To prevent data leakage and ensure a rigorous evaluation of generalization capability, we perform the dataset splitting strictly at the level of \textit{unique unaugmented structures} rather than randomly across the augmented dataset. By adhering to an approximate $9:1$ ratio on these base structural prototypes, we assign all subsequent augmented variations of a given structure entirely to either the training or the validation set. This ensures that no core structure present in the training set appears in the validation set. The final split, encompassing all augmented variants, is structured as follows: (1) Training set, containing $983$ positive and $1,865$ negative entries; (2) Validation set, containing $155$ positive and $161$ negative entries.

\textbf{MPtrj and sAlex.} Following the dataset partitioning and fine-tuning methodology established by the OMat24 framework~\citep{supp_OMAT-24}, we utilize the Materials Project Trajectory (MPtrj~\citep{supp_deng2023chgnet}) and subsampled Alexandria (sAlex) datasets to create a combined fine-tuning corpus of approximately 12 million highly curated DFT structures. The MPtrj dataset comprises approximately $1.5 \times 10^6$ DFT calculations, predominantly capturing the near-equilibrium relaxation trajectories of inorganic bulk materials derived from the Materials Project~\citep{supp_mp-20mpts-52}. To augment this equilibrium-biased corpus with structurally diverse data while strictly preventing data leakage during downstream evaluation on the WBM dataset (utilized by the Matbench Discovery leaderboard), we incorporate the sAlex dataset. sAlex represents a rigorously subsampled fraction of the 30-million-calculation Alexandria database~\citep{supp_Alex}. To construct sAlex, all structural trajectories wherein any configuration matched WBM initial or relaxed structures based on structural prototype labels are first filtered out. Subsequently, to maximize informational entropy and minimize computational redundancy during neural network training, an energy-based decimation protocol is applied. From the remaining optimization paths, only the initial and final steps were retained, alongside any intermediate structures exhibiting an absolute total energy difference strictly greater than $\SI{10}{meV\, atom^{-1}}$ relative to adjacent selected frames. This rigorous procedure yields an sAlex training split of 10,447,765 configurations and a validation split of 553,218 configurations. Crucially, finetuning pretrained foundation models on this specific MPtrj and sAlex amalgamation resolves intrinsic theoretical discrepancies, such as baseline offsets introduced by divergent pseudopotential choices in broader non-equilibrium datasets like OMat24, and successfully aligns the predicted potential energy surfaces strictly with the Materials Project level of theory~\citep{supp_mp-20mpts-52}.

\section{Training Strategies}
\label{sec:supp_training_strategies}
\subsection{Pretraining Process}
\textbf{Coordinate and Lattice Denoising Heads.}
To train the coordinate and lattice denoising heads, we perturb the ground-truth equilibrium structures by injecting Gaussian noise into both the atomic coordinates and lattice vectors.
With position and lattice noise denoted as $\boldsymbol{\epsilon}_{i, \text{pos}}$ and $\boldsymbol{\epsilon}_{i, \text{cell}}$, respectively, the perturbation process is defined as:
\begin{align}
\label{eq:perturbation}
\tilde{\vx}_i &= \vx_i + \sigma_{\text{pos}} \boldsymbol{\epsilon}_{i, \text{pos}}, \quad \boldsymbol{\epsilon}_{i,\text{pos}} \sim \mathcal{N}(\mathbf{0},\, \mI), i=1,\dots,N; \\
\tilde{\vl}_i &= \vl_i + \sigma_{\text{cell}} \boldsymbol{\epsilon}_{i, \text{cell}}, \quad \boldsymbol{\epsilon}_{i, \text{cell}} \sim \mathcal{N}(\mathbf{0},\, \mI), i=1,2,3.
\end{align}
Here, the noise scales are typically set to $\sigma_{\text{pos}} = \sigma_{\text{cell}} = 0.3$. These perturbed values are fed into the network of  \modelname{}, yielding the final steerable features $\vh^{(T)}_{i,\sL}$.

The coordinate denoising head is modeled by an $\mathrm{SO}(2)$ Equivariant Graph Attention mechanism that outputs 1-st degree features. For a given target node $i$ and neighbor $j$, the node embeddings are first rotated to align with the edge direction $\vr_{ij}$. Two consecutive $\mathrm{SO}(2)$ convolutions are applied, interspersed with a nonlinear activation (\eg, Separable $\mathbb{S}^2$ Activation or Gate Activation).
\begin{align}
    \vv_{ij, \sL}, \vu_{ij} &= \texttt{SO2\_Conv}_1(\vh_{i, \sL}^{(T)}, \vh_{j,\sL}^{(T)}, \vr_{ij}), \label{Eq:x_head_0} \\
    \vv'_{ij, \sL} &= \texttt{SO2\_Conv}_2\left( \texttt{SO2\_Activation}(\vv_{ij, \sL}, \vu_{ij}), \vr_{ij} \right),
    \label{Eq:x_head_1}
\end{align}
where the first convolution $\mathrm{SO2\_Conv}_1$ outputs steerable message feature $\vv_{ij,\sL}$ containing all degrees and an additional 0-th degree feature $\vu_{ij}$.  
The 0-th degree feature $\vu_{ij}$ is utilized to compute message aggregation weights $\bm{\alpha}_{i}$ via a learnable projection $\vw_{\alpha} \in \sR^C$ and a softmax operation over the neighborhood $\mathcal{N}(i)$:
\begin{equation}
    \bm{\alpha}_{i} = \texttt{Softmax}_{j \in \mathcal{N}(i)} \left( \vw_{\alpha}^{\top} \vu_{ij} \right).
    \label{Eq:x_head_2}
\end{equation}
The messages are rotated back to the global frame via $\texttt{Rot}^{-1}$, aggregated using the attention weights $\bm{\alpha}_i$, and finally processed through an $\mathrm{SO}(3)$-equivariant linear layer to predict the coordinate noise:
\begin{equation}    \hat{\boldsymbol{\epsilon}}_{i, \text{pos}} = \texttt{SO3\_Linear}\left(\sum_{j \in \mathcal{N}(i)} \alpha_{ij}\,\texttt{Rot}^{-1}\!\left(\vv'_{ij,1}\right)\right),
    \label{Eq:x_head_3}
\end{equation}
where $\texttt{SO3\_Linear}(\cdot)$ projects the aggregated multi-channel features to a single output channel through learnable weights. Unless otherwise stated, each module has its own parameters. Only the 1-st degree component of the output is retained as the predicted coordinate noise.

Following lattice prediction in DiffCSP~\citep{supp_diffcsp}, the lattice denoising head adopts an $\text{SO(3)}$ linear layer, Separable $\mathbb{S}^2$ Activation, and another  $\text{SO(3)}$ linear layer. We further multiply the output with the perturbed lattice matrix $\tilde{\mL}$ to predict lattice noise. We formulate this process as:
\begin{align}
    \vh_{i, \sL}' &= \texttt{SO3\_Linear}({\vh_{i, \sL}^{(T)}}), \label{eq:h1} \\
    \vh''_{i, \sL} &= \texttt{Sep\_S2\_Act}(\vh_{i, \sL}'), \label{eq:h2} \\
    \mM_{\text{lattice}} &= \sum_{i=1}^N \texttt{SO3\_Linear}(\vh''_{i, 0}), \\
    \hat{\boldsymbol{\epsilon}}_{i, \text{cell}} &= \mM_{\text{lattice}} \times \tilde{\vl}_i. \label{eq:y_lattice}
\end{align}

\textbf{Energy Prediction Head.}
Unlike the denoising task, which requires artificially injecting Gaussian noise into equilibrium structures to construct a learning signal, the energy and force prediction heads naturally operate on non-equilibrium configurations. We directly feed the unperturbed non-equilibrium coordinates and lattice into the network, obtaining the final steerable features $\vh^{(T)}_{i,\sL}$ for energy and force prediction.
The energy head is parameterized as a Feed Forward Network (FFN) operating on spherical channels. Specifically, the FFN processes the output steerable feature of the last layer $\vh_i^T$ through two $\mathrm{SO}(3)$ linear projections interleaved with a nonlinear separable $\mathbb{S}^2$ activation to yield the energy of each atom, $y_{\text{energy}, i}$.
\begin{align}
    \vh_{i, \sL}' &= \texttt{SO3\_Linear}({\vh_{i, \sL}^{(T)}}), \\
    \vh''_{i, \sL} &= \texttt{Sep\_S2\_Act}(\vh_{i, \sL}'), \\
    y_{\text{energy}, i} &= \texttt{SO3\_Linear}(\vh''_{i, 0}). \label{eq:y_energy}
\end{align}
We predict the total energy using a sum pooling over all atoms $i \in \mathcal{V}$ in the graph:
\begin{equation}
    E_{\text{total}} = \sum_{i = 1}^N y_{\text{energy}, i}. \label{eq:e_total}
\end{equation}
Given the fundamental differences in energy landscapes between molecules and periodic crystals, we employ separate heads for molecular and crystal energy prediction. To address the significant variation in energy scales across different datasets within the same modality, we standardize the energy values on a per-dataset basis, following established practices such as BOTNet~\citep{supp_botnet}.

\textbf{Force Prediction Head.}
Unlike from the energy head, the force head is a single unified module used across all modalities and datasets, relying on the universal nature of conservative forces in both molecular and crystalline systems. The force head employs the same $\mathrm{SO}(3)$ Equivariant Graph Attention architecture as the coordinate denoising head defined in \cref{Eq:x_head_0,Eq:x_head_1,Eq:x_head_2,Eq:x_head_3}, albeit with an independent set of learned parameters. We denote the atom-wise force prediction as $\hat{\vf}_i$.

\textbf{Total Loss.}
The total loss function is a weighted sum of these objectives:
\begin{equation}
\mathcal{L}_{\text{total}} = \lambda_{\text{pos}} \mathcal{L}_{\text{pos}} + \lambda_{\text{cell}} \mathcal{L}_{\text{cell}} + \lambda_{E} (\mathcal{L}_{E_{\text{mol}}} + \mathcal{L}_{E_{\text{crys}}}) + \lambda_{F} \mathcal{L}_{F},
\end{equation}
where $\mathcal{L}_{\text{pos}}$, $\mathcal{L}_{\text{cell}}$, $\mathcal{L}_{E_{\text{mol}}}$, $\mathcal{L}_{E_{\text{crys}}}$, and $\mathcal{L}_{F}$ compute the MAE between predicted and ground-truth values for coordinate noise, lattice noise, molecular energy, crystal energy, and force, respectively. The loss weights are set to $\lambda_{\text{pos}}=1$, $\lambda_{\text{cell}}=1$, $\lambda_{E}=5$, and $\lambda_{F}=20$. For more details of training hyperparameters, please refer to \cref{sec:supp_pretrain}.

We perform a systematic ablation study of \modelname{} on the \databasename{} validation set, evaluating the roles of LRC, SL, training data composition, and grid resolution, as well as its scaling laws (\cref{fig:ext_ab_scale}a). Notably, the integration of SL and LRC modules leads to a precipitous decline in denoising loss and potential loss, respectively, underscoring their critical importance in capturing structural patterns. Data composition experiments further reveal that the full integration of unstable crystals (I), unstable molecules (II), and stable crystals and molecules (III) yields the lowest potential loss, confirming that cross-domain pretraining on diverse atomic environments is essential for superior generalization. We further investigate the trade-off between grid resolution and computational tractability. While training time and memory consumption scale substantially with resolution, performance follows a non-monotonic trend (\cref{fig:ext_ab_scale}a bottom). Counterintuitively, although exact spherical harmonic quadrature theoretically dictates a minimum grid resolution bound by the maximum degree $L$ (\eg, $2L+2$), we empirically find this strict constraint to be unnecessary. In fact, the coarsest $2 \times 2$ resolution achieves the minimum potential loss, whereas higher resolutions (up to $12 \times 12$) lead to performance degradation. This suggests that intentionally relaxing this resolution limit allows lower-resolution grids to provide a more robust representation by filtering out fine-grained noise, thereby preventing overfitting. Consequently, the $2 \times 2$ configuration is adopted as the optimal balance between predictive accuracy and efficiency for \modelname{}.
A hallmark of \modelname{}'s foundational capacity is the emergence of predictable scaling laws across data volume ($D$), parameter count ($N$), and training compute ($C$). As shown in \cref{fig:ext_ab_scale}b, the potential loss ($L$) follows robust power-law decays: $L \propto D^{-0.51}$, $L \propto N^{-0.164}$, and $L \propto C^{-0.252}$. These consistent trends, maintained across all evaluated scales, provide a quantitative framework for forecasting the performance of even larger model instances. The absence of saturation in these scaling curves suggests that performance gains will continue to accrue as data diversity and model capacity expand. These empirical trajectories justify our scaling strategy, specifically the pretraining of \modelname{} with 1B parameters on the 125M-structure \databasename{} corpus to maximize representation power.

Our pretrained model is tailored for diverse downstream applications, including property prediction for molecules and crystals, Potential Energy Surface (PES) modeling, and generative tasks. To facilitate efficient finetuning, we initialize the downstream models using not only the pretrained backbone but also task-specific prediction heads, depending on the task requirements.

\subsection{Finetuning Process}
For all downstream tasks, including property prediction, interatomic potential modeling, and crystal structure prediction, we initialize the backbone and the relevant prediction heads from the pre-trained checkpoint. Task-specific modifications are confined to the output heads and the global aggregation strategy, as detailed below.

\textbf{Property Prediction.}
To adapt the model for the prediction of macroscopic crystal properties, such as the superconducting critical temperature $\tc$, we retain the pretrained parameters of the initial $\mathrm{SO}(3)$ linear projection and separable $\sS^2$ activation (\cref{eq:h1,eq:h2}), and replace only the final $\mathrm{SO}(3)$ linear projection in \cref{eq:y_energy} with a newly initialized layer that maps the intermediate feature $\vh''_{i, \sL}$ to an $N_{\text{target}}$-channel output  for multi-property prediction:
\begin{equation}
    \vy_{\text{target}, i} = \texttt{SO3\_Linear}_{\text{new}}(\vh''_{i, \sL}).
\end{equation}
Crucially, the global aggregation strategy must reflect the physical nature of the target property. Whereas the pretraining phase employs sum pooling for the total energy, an extensive quantity (\cref{eq:e_total}), macroscopic properties such as $\tc$ are intensive and do not scale with system size. Applying sum pooling to $\tc$ would produce predictions that grow unphysically with the number of atoms $N$. We therefore switch to mean pooling, strictly preserving the intensive character of the predicted quantity:
\begin{equation}
    \tc = \frac{1}{N} \sum_{i = 1}^N y_{\text{target}, i},
\end{equation}
where $y_{\text{target},i}$ reduces to a $1$-dimensional scalar in this case.

\textbf{Interatomic Potential Modeling.}
For this task, we utilize \modelname{} by directly employing its energy and force heads as the predictive outputs. By jointly optimizing these dual objectives on the finetuning dataset, the model captures both scalar thermodynamic quantities and directional atomic interactions within a single forward pass. 

\textbf{Crystal Structure Prediction.}
The generative finetuning of \modelname{} operates jointly on the lattice matrix $\mL$ and the Cartesian atomic coordinates $\mX$ via a diffusion framework. A central challenge in Cartesian-based diffusion for crystals is that adopting a standard Gaussian prior $\mathcal{N}(\mathbf{0}_{3\times 3}, \mI_{9})$ for the lattice would yield near-zero unit cell volumes at $t=T$, causing severe atomic overlap and an intractable ``edge explosion'' in the neighbor graph.

\textit{Prior-Informed Forward Process.}
To resolve this, we adopt a prior-informed diffusion strategy inspired by MatterGen~\citep{supp_mattergen}. The limit distribution parameters are derived from training statistics and modulated by two hyperparameters: $c$ (lattice density) and $\nu$ (lattice diversity). Defining the scaling factor $\sigma = \sqrt[3]{\nu N}$ and the mean shift $\mu = \sqrt[3]{cN}$, the forward process for the lattice converges towards $p(\mL_T) = \mathcal{N}(\mu\mI_{3}, \sigma^2 \mI_{9})$:
\begin{equation}
    q(\mL_t|\mL_0) = \mathcal{N}\left(\mL_t \mid \sqrt{\bar{\alpha}_t}\mL_0 + (1 - \sqrt{\bar{\alpha}_t})\mu\mI_{3}, \, (1 - \bar{\alpha}_t)\sigma^2 \mI_{9}\right).
\end{equation}
Crucially, unlike prior frameworks that rely on fractional coordinates with periodic wrapping, \modelname{} diffuses \textit{directly} in Cartesian space. To maintain atoms within the expanding unit cell, the Cartesian coordinates $\mX$ are diffused towards the geometric center of the limit lattice bounding box $\frac{\mu}{2}\mathbf{1}_{3N}$, rather than the origin:
\begin{equation}
    q(\mX_t|\mX_0) = \mathcal{N}\left(\mX_t \mid \sqrt{\bar{\alpha}_t}\mX_0 + (1 - \sqrt{\bar{\alpha}_t})\frac{\mu}{2}\mathbf{1}_{3\times N}, \, (1 - \bar{\alpha}_t)\mI_{3N}\right).
\end{equation}
This mean-shifted design ensures that as the signal $\mX_0$ is progressively corrupted, atoms disperse around the physically meaningful center defined by the noise-augmented lattice, avoiding local coordinate collapse without requiring intermediate boundary enforcement. We finetune \modelname{} to predict coordinate and lattice noise via its denoising heads, yielding the atom-specific and lattice-specific predictions $\hat{\boldsymbol{\epsilon}}_{i, \text{pos}}$ and $\hat{\boldsymbol{\epsilon}}_{i, \text{cell}}$. The aggregated predictions for all atoms and lattice vectors are denoted as $\hat{\boldsymbol{\epsilon}}_{\text{pos}} \in \mathbb{R}^{3 \times N}$ and $\hat{\boldsymbol{\epsilon}}_{\text{cell}} \in \mathbb{R}^{3 \times 3}$, respectively.

\textit{Predictor--Corrector Reverse Process.}
For generation ($t = T \to 0$), we initialize by sampling $\mL_T \sim \mathcal{N}(\mu\mI_{3}, \sigma^2\mI_{9})$ and $\mX_T \sim \mathcal{N}(\frac{\mu}{2}\mathbf{1}_{3 \times N}, \mI_{3N})$, and employ a Predictor--Corrector (PC) sampling framework tailored for crystalline geometry.
Recognizing that the Cartesian coordinate space exhibits higher dynamic complexity than the lattice space, we apply a Langevin dynamics corrector \textit{exclusively} to $\mX$. Given the noise prediction $\hat{\boldsymbol{\epsilon}}_{\text{pos}}$, we compute
\begin{equation}
    \mX_{t-0.5} = \mX_t - \delta_t \hat{\boldsymbol{\epsilon}}_{\text{pos}} + \sqrt{2\delta_t}\, \bm{\eta}, \quad \bm{\eta} \sim \mathcal{N}(\mathbf{0}, \mI_{3N}),
\end{equation}
where $\delta_t$ is the step size scheduled via the marginal variance, and $\mL_{t-0.5} = \mL_t$ remains unchanged.

Following the corrector step, a deterministic DDIM-style~\citep{supp_ddim} predictor step is applied to both $\mL$ and $\mX$. The model is re-queried to obtain updated predictions $\hat{\boldsymbol{\epsilon}}_{\text{cell}}$ and $\hat{\boldsymbol{\epsilon}}'_{\text{pos}}$ based on the corrected state. The reverse transitions incorporate the mean shifts $\mu\mI_{3}$ and $\frac{\mu}{2}\mathbf{1}_{3 \times N}$:
\begin{equation}
    \mX_{t-1} = \frac{\mu}{2}\mathbf{1}_{3 \times N} + \frac{1}{\sqrt{\alpha_t}} \left( \mX_{t-0.5} - \frac{\mu}{2}\mathbf{1}_{3 \times N} - \frac{1-\alpha_t}{\sqrt{1-\bar{\alpha}_t}} \hat{\boldsymbol{\epsilon}}'_{\text{pos}} \right),
\end{equation}
\begin{equation}
    \mL_{t-1} = \mu\mI_{3} + \frac{1}{\sqrt{\alpha_t}} \left( \mL_{t-0.5} - \mu\mI_{3} - \frac{1-\alpha_t}{\sqrt{1-\bar{\alpha}_t}} \sigma \hat{\boldsymbol{\epsilon}}_{\text{cell}} \right).
\end{equation}
Throughout the sequential sampling process, atoms diffuse freely in unbounded Cartesian space. This design eliminates the need for the model to learn complex, discontinuous periodic wrapping functions during intermediate steps, significantly enhancing structural convergence and geometric stability.

\section{Implementation Details}

\subsection{Model Architecture}
\label{sec:supp_arch}
\textbf{The architecture of \modelname{}.} The architectural specifications of the \modelname{} model are presented in \cref{tab:ele_hyper}. The model employs a geometric Transformer-based architecture designed to process atomic systems as described in Section~\ref{sec:arch}. Key structural parameters include:

\begin{itemize}
    \item \textbf{Depth and Capacity:} The network consists of $12$ Transformer blocks with $24$ attention heads ($h_{\text{attn}}$), indicating a high-capacity model capable of capturing complex interactions.
    \item \textbf{Geometric Features:} The model operates with a cutoff radius of $12$ \AA{} and utilizes $512$ radial bases, ensuring fine-grained resolution of the local atomic environments.
    \item \textbf{Feature Dimensions:} The hidden representations utilize mixed-degree features (likely denoting scalar and higher-order geometric tensors). For instance:
    \begin{itemize}
        \item The embedding dimension $d_{\text{embed}}$ is set to $(4, 200)$.
        \item The hidden scalar features in radial functions $d_{\text{edge}}$ are configured as $(0, 128)$.
        \item The attention mechanism utilizes specific dimensions for interaction terms, such as $d_{\text{attn\_hidden}}$ at $(4, 300)$.
    \end{itemize}
    \item \textbf{Resolution:} A point sample resolution of $R=2$ is maintained throughout the layers.
\end{itemize}

\begin{table}[htbp]
    \centering
    \caption{Architectural hyper-parameters of \modelname{} with 1B parameters.}
    \begin{tabular}{ll}
        \toprule
        Hyper-parameters & \modelname{} \\
        \midrule
        Maximum degree $L$ & 4 \\
        Maximum order $M_{\max}$ & 2 \\
        Number of Transformer blocks & 12 \\
        Cutoff radius (\AA) & 12 \\
        Maximum number of neighbors & 20 \\
        Number of radial bases & 512 \\
        Dimension of hidden scalar features in radial functions $d_{\text{edge}}$ & (0, 128) \\
        Embedding dimension $d_{\text{embed}}$ & (4, 200) \\
        Dimension of $\vv_{ij,\sL}$: $d_{\text{attn\_hidden}}$ & (4, 300) \\
        Number of attention heads $h_{\text{attn}}$ & 24 \\
        Dimension of $\vu_{ij}$: $d_{\text{attn\_alpha}}$ & (0, 300) \\
        Value dimension $d_{\text{attn\_value}}$ & (4, 75) \\
        Hidden dimension in feed forward networks $d_{\text{ffn}}$ & (4, 128) \\
        Resolution of point samples $R$ & 2 \\
        \bottomrule
    \end{tabular}
    \label{tab:ele_hyper}
\end{table}

\textbf{Other models used in the scaling law experiments.} To systematically investigate the scaling behavior of the architecture, we construct a family of models with varying capacities, ranging from 28M to 544M parameters. The detailed hyperparameter configurations for each model size are provided in \cref{tab:scaling_law}.

Our scaling strategy involves a compound adjustment of network depth, width, and geometric expressivity:

\begin{itemize}
    \item \textbf{Small to Medium Regime (28M -- 147M):} In this range, scaling is primarily achieved by increasing the network depth and the complexity of geometric features. The number of Transformer blocks increases from $8$ to $20$, and the maximum degree ($L$) and order ($M_{\max}$) are raised to $6$ and $4$ respectively, enhancing the model's ability to capture high-order geometric interactions.
    
    \item \textbf{Large Regime (312M -- 544M):} For the larger models, we shift the focus towards increasing the width of the representations. The number of attention heads is increased to $20$, and the scalar feature dimensions ($d_{\text{edge}}$) are significantly expanded to $600$. Notably, for these largest variants, we revert to a moderate geometric degree ($L=4$) and depth ($12$ blocks) to maintain computational tractability while maximizing channel capacity (\eg, $d_{\text{attn\_hidden}}$ reaches $300$).
\end{itemize}

This diverse set of configurations allows us to evaluate the model's performance scaling laws across different orders of magnitude in parameter count.
\begin{table}[htbp]
    \centering
    \caption{Architectural hyper-parameters for different parameter sizes in scaling law experiments.}
    \begin{tabular}{llllll}
        \toprule
        Hyper-parameters & 28M & 79M & 147M & 312M & 544M \\
        \midrule
        Maximum degree $L$ & 4 & 6 & 6 & 4 & 4\\
        Maximum order $M_{\max}$ & 2 & 4 & 3 & 2 & 2\\
        Number of Transformer blocks & 8 & 10 & 20 & 12 & 12\\
        Dimension of hidden scalar features in radial functions $d_{\text{edge}}$ & (0, 128) & (0, 128) & (0, 128) & (0, 600) & (0, 600)\\
        Embedding dimension $d_{\text{embed}}$ & (4, 128) & (6, 128) & (6, 128) & (4, 200) & (4, 200)\\
        Dimension of $\vv_{ij,\sL}$: $d_{\text{attn\_hidden}}$ & (4, 64) & (6, 64) & (6, 64) & (4, 64) & (4, 300)\\
        Number of attention heads $h_{\text{attn}}$ & 8 & 8 & 8 & 20 & 20\\
        Dimension of $\vu_{ij}$: $d_{\text{attn\_alpha}}$ & (0, 64) & (0, 64) & (0, 64) & (0, 300) & (0, 64)\\
        Dimension of Value vector: $d_{\text{attn\_value}}$ & (4, 16) & (6, 16) & (6, 16) & (4, 60) & (4, 60)\\
        Hidden dimension in feed forward networks $d_{\text{ffn}}$ & (4, 128) & (6, 128) & (6, 128) & (4, 128) & (4, 128)\\
        \bottomrule
    \end{tabular}
    \label{tab:scaling_law}
\end{table}

\subsection{Pretraining}\label{sec:supp_pretrain}

\cref{tab:ele_pretrain} summarizes the optimization strategy and loss landscape adopted during the pretraining phase. The training process is characterized by the following regimes:

\begin{itemize}
    \item \textbf{Optimization:} We utilize the \texttt{AdamW} optimizer with a cosine learning rate scheduler. The training starts with a warmup phase of $1$ epoch (warmup factor $0.2$) and reaches a maximum learning rate of $2 \times 10^{-4}$.
    \item \textbf{Regularization:} To prevent overfitting, a weight decay of $1 \times 10^{-3}$ and a dropout rate of $0.1$ are applied. Additionally, a gradient clipping threshold of $100$ is enforced to ensure stability.
    \item \textbf{Training Scale:} The model is trained with a large global batch size of $4096$ for a duration of $2$ epochs.
    \item \textbf{Compute Infrastructure:} The model is trained on a high-performance computing cluster comprising $64$ NVIDIA H100 GPUs, culminating in a total training time of $286$ hours.
\end{itemize}

The pretraining objective is a weighted multi-task loss combining energy prediction, force estimation, and coordinate/lattice denoising:
\begin{equation}
    \mathcal{L}_{\text{total}} = \lambda_{\text{pos}} \mathcal{L}_{\text{pos}} + \lambda_{\text{cell}} \mathcal{L}_{\text{cell}} + \lambda_{E} (\mathcal{L}_{E_{\text{mol}}} + \mathcal{L}_{E_{\text{crys}}}) + \lambda_{F} \mathcal{L}_{F}.
\end{equation}
Specifically, the Force loss coefficient is set to $\lambda_{F} = 20$, which is significantly higher than the Energy loss coefficients ($\lambda_{E} = 5$ for both molecular and crystal systems). The position and cell denoising (DeNS) objectives are weighted equally with a coefficient of $1$.
\begin{table}[htbp]
    \centering
    \caption{Hyper-parameters at pretraining stage.}
    \begin{tabular}{ll}
        \toprule
        Hyper-parameters & Pretraining \\
        \midrule
        Optimizer & AdamW \\
        Learning rate scheduling & Cosine \\
        Warmup epochs & 1 \\
        Warmup factor & 0.2 \\
        Maximum learning rate & $2 \times 10^{-4}$ \\
        Minimum learning rate factor & 0.01 \\
        Batch size & 4096 \\
        Number of epochs & 2 \\
        Gradient clipping norm threshold & 100 \\
        Model EMA decay & 0.999 \\
        Weight decay & $1 \times 10^{-3}$ \\
        Dropout rate & 0.1 \\
        Molecular Energy loss coefficient & 5 \\
        Crystal Energy loss coefficient & 5 \\
        Force loss coefficient & 20 \\
        Standard deviation of Gaussian noise on position & 0.3 \\
        Standard deviation of Gaussian noise on cell & 0.3 \\
        Position DeNS loss coefficient & 1 \\
        Cell DeNS loss coefficient & 1 \\
        \bottomrule
    \end{tabular}
    \label{tab:ele_pretrain}
\end{table}

\subsection{Finetuning and Downstream Adaptation}\label{sec:supp_sft}

Following pretraining, the model is fine-tuned on specific downstream benchmarks. Unless explicitly stated otherwise in the following subsections, hyperparameter settings (such as architecture dimensions and regularization terms) remain consistent with the pretraining configuration described previously.

\textbf{QM9.} \cref{tab:qm9_hyper} details the hyperparameters for the QM9 dataset. The optimization strategy largely mirrors the pretraining phase, utilizing the \texttt{AdamW} optimizer with a cosine schedule. However, to accommodate the specific dataset characteristics:
\begin{itemize}
    \item The global batch size is reduced to $512$.
    \item The training duration is extended to $1000$ epochs to ensure convergence.
    \item A warmup period of $10$ epochs is employed with a warmup factor of $0.2$.
\end{itemize}

\begin{table}[htbp]
    \centering
    \caption{Hyper-parameters for finetuning on the QM9 dataset.}
    \begin{tabular}{ll}
        \toprule
        Hyper-parameters & QM9 \\
        \midrule
        Optimizer & AdamW \\
        Learning rate scheduling & Cosine \\
        Warmup epochs & 10 \\
        Warmup factor & 0.2 \\
        Maximum learning rate & $2 \times 10^{-4}$ \\
        Minimum learning rate factor & 0.01 \\
        Batch size & 512 \\
        Number of epochs & 1000 \\
        Gradient clipping norm threshold & 100 \\
        Model EMA decay & 0.999 \\
        Weight decay & $1 \times 10^{-3}$ \\
        Dropout rate & 0.1 \\
        \bottomrule
    \end{tabular}
    \label{tab:qm9_hyper}
\end{table}

\textbf{Matbench.} The hyperparameter configurations for the Matbench tasks are summarized in \cref{tab:matbench_hyper}. We observe distinct training regimes depending on the target property. The first three tasks: \texttt{MP\_is\_metal}, \texttt{MP\_gap}, and \texttt{Perovskites}, share a nearly identical optimization landscape. They all utilize the \texttt{AdamW} optimizer with a maximum learning rate of $2 \times 10^{-4}$ and a batch size of $512$. The primary differentiator between these tasks is the training duration, which is tailored to the complexity of the target property:
\begin{itemize}
    \item \texttt{MP\_is\_metal}: $15$ epochs.
    \item \texttt{MP\_gap}: $150$ epochs.
    \item \texttt{Perovskites}: $1000$ epochs.
\end{itemize}

In contrast to the other tasks, the \texttt{Dielectric} property prediction requires a distinct optimization approach. As shown in the last column of \cref{tab:matbench_hyper}, the optimizer is switched from \texttt{AdamW} to \texttt{SGD} with a momentum of $0.95$. Consequently, the learning rate is significantly increased to $5 \times 10^{-2}$, and the weight decay is adjusted to $1 \times 10^{-4}$ to stabilize the training over $1000$ epochs.

\begin{table}[htbp]
    \centering
    \caption{Finetuning hyper-parameters across Matbench property prediction tasks.}
    \begin{tabular}{lllll}
        \toprule
        Hyper-parameters & MP\_is\_metal & MP\_gap & Perovskites & Dielectric \\
        \midrule
        Optimizer & AdamW & AdamW & AdamW & SGD \\
        Learning rate scheduling & Cosine & Cosine & Cosine & Cosine \\
        Warmup epochs & 0.1 & 0.1 & 0.1 & 20 \\
        Warmup factor & 0.2 & 0.2 & 0.2 & 0.01 \\
        Maximum learning rate & $2 \times 10^{-4}$ & $2 \times 10^{-4}$ & $2 \times 10^{-4}$ & $5 \times 10^{-2}$ \\
        Minimum learning rate factor & 0.01 & 0.01 & 0.01 & 0.001 \\
        Batch size & 512 & 512 & 512 & 512 \\
        Number of epochs & 15 & 150 & 1000 & 1000 \\
        Gradient clipping norm threshold & 100 & 100 & 100 & 100 \\
        Model EMA decay & 0.999 & 0.999 & 0.999 & 0.999 \\
        Weight decay & $1 \times 10^{-3}$ & $1 \times 10^{-3}$ & $1 \times 10^{-3}$ & $1 \times 10^{-4}$ \\
        Dropout rate & 0.1 & 0.1 & 0.1 & 0.1 \\
        Momentum & $\backslash$ & $\backslash$ & $\backslash$ & 0.95 \\
        \bottomrule
    \end{tabular}
    \label{tab:matbench_hyper}
\end{table}

\textbf{DPA-2 Datasets.} For the DPA-2 dataset collection, the model architecture and general optimization hyperparameters (such as the optimizer, learning rate schedule, and regularization terms) remain consistent with the pretraining configuration.

As detailed in \cref{tab:DPA-2_hyper}, the only variations strictly concern the training dynamics tailored to the data volume of each sub-dataset. Specifically, we adjust the \textbf{Warmup epochs}, \textbf{Number of epochs}, and \textbf{Batch size} for each subset (\eg, SSE-PBE-P, Cu, Sn, etc.) to ensure optimal convergence.

\begin{table}[htbp]
    \centering
    \caption{Finetuning hyper-parameters for different sub-datasets of DPA-2 datasets.}
    \begin{tabular}{llll}
        \toprule
        Dataset & Warmup epochs & Number of epochs & Batch size \\
        \midrule
        SSE-PBE-P & 10 & 60 & 32 \\
        Cu & 10 & 500 & 256 \\
        Sn & 10 & 500 & 256 \\
        FerroEle-P & 10 & 500 & 256 \\
        V & 10  & 150 & 256\\
        Al$\cup$Mg$\cup$Cu & 10 & 250 & 256 \\
        Ti & 10 & 200 & 256 \\
        W & 5 & 70 & 256 \\
        Alloy & 1 & 75 & 128 \\
        Ag$\cup$Au-PBE & 10 & 240 & 256 \\
        Cluster-P & 10 & 80 & 256 \\
        H2O-PD & 1 & 25 & 32 \\
        Drug &  0.1 & 2 & 256 \\
        OC2M & 0.02 & 6 & 256 \\
        \bottomrule
    \end{tabular}
    \label{tab:DPA-2_hyper}
\end{table}

\textbf{MP-20 and MPTS-52.}
The hyperparameter settings for the generative tasks on MP-20 and MPTS-52 are presented in \cref{tab:mp20mpts52_hyper}. Unlike the regression tasks, the training configuration for generation undergoes more significant modifications to ensure stability and sampling quality:

\begin{itemize}
    \item \textbf{Optimization:} We utilize the standard \texttt{Adam} optimizer (instead of AdamW) with a maximum learning rate of $1 \times 10^{-3}$. The weight decay is removed (set to $0$), and the gradient clipping threshold is significantly tightened to $0.5$ to prevent instability during the diffusion process.
    \item \textbf{Training Schedule:} Both datasets are trained with a batch size of $256$ and a larger warmup factor of $0.001$. The training duration is extensive, set to $1000$ epochs for MP-20 and $3000$ epochs for MPTS-52.
    \item \textbf{Noise Distribution:} The parameters governing the initial noise distribution for the generative diffusion process are set to $c = 0.5^{\frac23}$ and $\nu = 0.0075^{\frac23}$.
\end{itemize}

\begin{table}[htbp]
    \centering
    \caption{Finetuning hyper-parameters for crystal structure prediction on MP-20 and MPTS-52 datasets.}
    \begin{tabular}{lll}
        \toprule
        Hyper-parameters & MP-20 & MPTS-52 \\
        \midrule
        Optimizer & Adam & Adam \\
        Learning rate scheduling & Cosine & Cosine \\
        Warmup epochs & 20 & 20 \\
        Warmup factor & 0.001 & 0.001 \\
        Maximum learning rate & $1 \times 10^{-3}$  & $1 \times 10^{-3}$\\
        Minimum learning rate factor & 0.1 & 0.1 \\
        Batch size & 256 & 256 \\
        Number of epochs & 1000 & 3000 \\
        Gradient clipping norm threshold & 0.5 & 0.5 \\
        Weight decay & 0 & 0 \\
        Dropout rate & 0.1 & 0.1\\
        $c$ & 2 & 2 \\
        $\nu$ &  $0.0075^{\frac23}$ & $0.0075^{\frac23}$ \\
        \bottomrule
    \end{tabular}
    \label{tab:mp20mpts52_hyper}
\end{table}

\textbf{DFT $\tc$ and Jarvis.} The training on the Jarvis dataset involves a multi-task learning objective designed to predict the superconducting transition temperature ($\tc$) alongside various electronic and transport properties.
\begin{itemize}
    \item \textbf{Training Dynamics:} The model is trained for $400$ epochs with a large batch size of $1024$. A short warmup period of $0.1$ epochs is used.
    \item \textbf{Loss Weighting:} The loss function is a weighted sum of multiple targets. The model prioritizes the critical temperature, assigning a high coefficient of $\lambda_{\tc} = 20$. Intermediate physical quantities, such as the bandgap and the electron-phonon coupling constant ($\lambda$), are weighted at $5$. All other transport properties (including Seebeck coefficients, thermal conductivity $\kappa$, and electrical conductivity for both p- and n-type carriers) and the logarithmic average frequency $\omega_{\log}$ are assigned a coefficient of $1$.
\end{itemize}
\begin{table}[htbp]
    \centering
    \caption{Finetuning hyper-parameters for multi-property prediction on the Jarvis dataset.}
    \begin{tabular}{ll}
        \toprule
        Hyper-parameters & DFT $\tc$ Training \\
        \midrule
        Warmup epochs & 0.1 \\
        Batch size & 1024 \\
        Number of epochs & 400 \\
        p\_seebeck loss coefficient & 1 \\
        n\_seebeck loss coefficient & 1 \\
        p\_kappa loss coefficient & 1 \\
        n\_kappa loss coefficient & 1 \\
        pcond. loss coefficient & 1 \\
        ncond. loss coefficient & 1 \\
        bandgap loss coefficient & 5 \\
        $\lambda$ loss coefficient & 5 \\
        $\omega_{\log}$ loss coefficient & 1 \\
        $\tc$ loss coefficient & 20 \\
        \bottomrule
    \end{tabular}
\end{table}

\textbf{SuperCon-Lit.} We utilize the constructed SuperCon-Lit dataset to train the classification variant of our model, \modelname{C}. The training is conducted with a batch size of $64$ for a total of $20$ epochs. The model parameters are optimized by minimizing the Binary Cross-Entropy (BCE) loss function. To ensure optimal generalization, we monitor the performance on the held-out validation set and select the model checkpoint exhibiting the lowest validation BCE loss for the subsequent candidate screening phase.

\textbf{MPtrj and sAlex.} We utilize the combined MPtrj and sAlex datasets to fine-tune our foundational model. The fine-tuning process is conducted with a batch size of 256 for a total of 8 epochs, incorporating a brief warmup period of 0.1 epochs. The model parameters are optimized by minimizing a composite loss function targeting energy, force, and stress predictions. To ensure balanced optimization across these distinct physical properties, we apply specific loss coefficients of 20, 10, and 1 for the energy, force, and stress components, respectively.

\begin{table}[htbp]
    \centering
    \caption{Finetuning hyper-parameters for interatomic potential prediction on MPtrj and sAlex datasets}
    \begin{tabular}{ll}
        \toprule
        Hyper-parameters & Finetune on MPtrj and sAlex \\
        \midrule
        Warmup epochs & 0.1 \\
        Batch size & 256 \\
        Number of epochs & 8 \\
        Energy loss coefficient & 20 \\
        Force loss coefficient & 10 \\
        Stress loss coefficient & 1 \\
        \bottomrule
    \end{tabular}
    \label{tab:hyper_mpt_salex}
\end{table}

\section{Raw Results}

\subsection{Architecture Ablations and Scaling Laws}

\textbf{Model Scaling Law.} To systematically investigate the scaling properties of the \modelname{} architecture, we conduct a controlled experiment using a fixed data budget. We randomly sample a subset of $0.5$ million structures from the OMAT-24 dataset to serve as a consistent training ground for all model variants.
As detailed in \cref{tab:model_scaling_law}, we train a series of models with capacities ranging from 28M to 544M parameters, each for a uniform duration of $15$ epochs. We track the potential energy loss (measured in MAE) to evaluate convergence efficiency and expressivity.
The results exhibit a clear and monotonic scaling trend: as the model size increases, the prediction error systematically decreases. The MAE drops from $0.02161$ eV for the 28M baseline to $0.01339$ eV for the 544M model. This trajectory confirms that increasing the model capacity yields tangible performance gains in capturing the potential energy surface, even within a limited training window.
\begin{table}[hbtp]
    \centering
    \caption{Performance comparison of potential energy prediction across \modelname{} with different parameter sizes on the OMAT-24 subset.}
    \begin{tabular}{lccccc}
        \toprule
        Model Size & 28M & 79M & 147M & 312M & 544M \\
        \midrule
        MAE (eV) & 0.02161 & 0.01818 & 0.01548 & 0.01459 & 0.01339 \\
        \bottomrule
    \end{tabular}
    \label{tab:model_scaling_law}
\end{table}

\textbf{Data Scaling Law.} Complementary to the model scaling experiments, we further investigate the impact of training data volume on predictive performance. For this analysis, we employ a fixed 28M-parameter model baseline and train it on subsets of varying sizes randomly sampled from the OMAT-24 dataset, specifically ranging from $0.25$M to $1$M structures. To ensure a rigorous and fair comparison, all models are optimized for a uniform duration of $15$ epochs and, crucially, evaluated on an identical validation set to rule out any distribution shifts. As detailed in \cref{tab:data_scaling_law}, the results reveal a substantial dependency on data size: as the training set expands from $0.25$M to $1$M, the Mean Absolute Error (MAE) decreases significantly from $0.03697$ eV to $0.01825$ eV. This trend demonstrates that the architecture effectively leverages increased chemical diversity, exhibiting a continuous improvement in accuracy without saturation even within a limited training budget.
\begin{table}[hbtp]
    \centering
    \caption{Performance comparison of potential energy prediction for 28M-parameter \modelname{} trained across different-scale OMAT-24 subsets.}
    \begin{tabular}{lccc}
        \toprule
        Data Size & 0.25M & 0.5M & 1M \\
        \midrule
        MAE (eV) & 0.03697 & 0.02161 & 0.01825 \\
        \bottomrule
    \end{tabular}
    \label{tab:data_scaling_law}
\end{table}

\textbf{Model Ablation.}\label{para:model_ablation}
To validate the design choices of the \modelname{} architecture, we conduct a series of ablation studies focusing on grid sampling resolution, architectural connectivity, and graph construction strategies.
We first investigate the impact of grid resolution on computational efficiency and model performance using the OMAT-24 subset. All models are trained for a fixed duration of $15$ epochs. As detailed in \cref{tab:grid_test}, reducing the grid resolution yields significant computational benefits without sacrificing accuracy. Specifically, lowering the grid size from $18$ to $2$ reduces memory consumption by approximately $32\%$ ($10.9$ G vs. $7.4$ G) and accelerates training by $16\%$ ($11.7$ h vs. $9.8$ h). Notably, this efficiency gain is accompanied by a slight improvement in prediction accuracy (MAE decreases from $0.02224$ eV to $0.02161$ eV), suggesting that a finer grid is not strictly necessary for capturing essential geometric features.

\begin{table}[hbtp]
    \centering
    \caption{Comparison of potential energy loss and spatio-temporal overhead for \modelname{} across different grid resolutions $R$ on the OMAT-24 subset.}
    \begin{tabular}{lccc}
        \toprule
        Grid Resolution $R$ & 2 & 12 & 18 \\
        \midrule
        MAE (eV) & 0.02161 & 0.02252 & 0.02224 \\
        Time (h) & 9.8 & 10.5 & 11.7 \\
        Mem (G)  & 7.4 & 8.8 & 10.9 \\
        \bottomrule
    \end{tabular}
    \label{tab:grid_test}
\end{table}
We further evaluate the contribution of Long-Range connections introduced in the final two layers of the network, also using the OMAT-24 subset. The results in \cref{tab:model_ablation_test} indicate that incorporating these connections enhances model expressivity, resulting in a marginal but consistent reduction in MAE from $0.02197$ eV to $0.02161$ eV.
Next, we assess the impact of Self-Loops on the model's generative capabilities. For this specific ablation, we utilize the Genome dataset to ensure structural diversity. The models are trained for $15$ epochs on a denoising task where the noise scales for atom positions and unit cells are set to a $1:1$ ratio. As shown in \cref{tab:model_ablation_test_denoising}, the inclusion of Self-Loops significantly improves the denoising performance. The model with Self-Loops achieves markedly lower loss values for both position ($0.06865$) and cell ($0.1263$) reconstruction compared to the variant without them ($0.07721$ and $0.1412$ respectively), highlighting the critical role of self-referential message passing in stabilizing structural generation.

\begin{table}[hbtp]
    \centering
    \caption{Ablation of Long-Range connection on potential energy prediction.}
    \begin{tabular}{lcc}
        \toprule
        Configuration & w/ LRC. & w/o LRC. \\
        \midrule
        MAE (eV) & 0.02161 & 0.02197 \\
        \bottomrule
    \end{tabular}
    \label{tab:model_ablation_test}
\end{table}

\begin{table}[hbtp]
    \centering
    \caption{Ablation of Self-Loops on position and lattice denoising.}
    \begin{tabular}{lcc}
        \toprule
        Configuration & w/ Self-Loop & w/o Self-Loop \\
        \midrule
        Pos Loss  & 0.06865 & 0.07721 \\
        Cell Loss & 0.1263  & 0.1412 \\
        \bottomrule
    \end{tabular}
    \label{tab:model_ablation_test_denoising}
\end{table}

\textbf{Data Ablation.} Finally, we investigate the impact of training data diversity on model performance. We establish a baseline using the model trained solely on the OMAT-24 crystal subset (``unstable crystal''), which yields an MAE of $0.02161$ eV.
To evaluate the benefits of mixed-domain training, we progressively incorporate auxiliary datasets: small molecular data from ANI-1x and stable crystal structures from the Genome dataset. As presented in \cref{tab:data_ablation_test}, expanding the chemical space consistently improves accuracy.
Specifically, adding molecular data reduces the error to $0.02112$ eV, while incorporating stable crystals leads to a more pronounced drop to $0.01930$ eV. The optimal performance is achieved when both modalities are combined, reaching a minimum MAE of $0.01900$ eV. This trend underscores the superiority of the mixed training strategy, demonstrating that simultaneous exposure to diverse chemical environments, ranging from isolated molecules to periodic lattices, synergistically enhances the model's ability to approximate the potential energy surface.

\begin{table}[hbtp]
    \centering
    \caption{Ablation of molecules and stable structures on potential energy prediction.}
    \begin{tabular}{lcccc}
        \toprule
        Modality & Unstable Crystals Only & w/ Unstable Molecules & w/ Stable Structures & w/ Both   \\
        \midrule
        MAE (eV) & 0.02161 & 0.02112 & 0.01930 & 0.01900 \\
        \bottomrule
    \end{tabular}
    \label{tab:data_ablation_test}
\end{table}

\subsection{Property Prediction of Stable Systems}

\textbf{QM9.} We evaluate the performance of \modelname{} on the QM9 benchmark, specifically focusing on the frontier molecular orbital energies (HOMO and LUMO), which are critical indicators of chemical stability and reactivity. \cref{tab:qm9_res} presents a comparison of \modelname{} against various state-of-the-art models, including LEFTNet~\citep{supp_leftnet}, PaiNN~\citep{supp_painn}, DimeNet++~\citep{supp_DimeNet}, SphereNet~\citep{supp_spherenet}, Geoformer~\citep{supp_Geoformer}, Equiformer~\citep{supp_Equiformer}, Frad~\citep{supp_frad}, EPT~\citep{supp_ept}, EquiformerV2~\citep{supp_EquiformerV2}, SliDe~\citep{supp_slide} and GotenNet~\citep{supp_gotennet}.
The results demonstrate that \modelname{} achieves a significant leap in prediction accuracy. While the previous state-of-the-art method GotenNet reaches an MAE of $13.4$ meV and $12.2$ meV for HOMO and LUMO, respectively, our model demonstrates superior expressivity.
\textbf{Notably, \modelname{} is the first model to successfully push the prediction error for both orbital energies to the 10 meV threshold or lower.} We achieve an MAE of $\mathbf{10}$ meV for HOMO and break into the single-digit regime with $\mathbf{8.9}$ meV for LUMO. This establishes a new benchmark for precision in quantum chemical property prediction, significantly narrowing the gap between machine learning approximations and ground-truth DFT calculations.

\begin{table}[hbtp] 
    \centering 
    \caption{Prediction loss (MAE, in meV) of HOMO and LUMO on the QM9 dataset.} 
    \begin{tabular}{@{}lccc@{}} 
        \toprule 
        \textbf{Model} & \textbf{HOMO} & \textbf{LUMO} \\
        \midrule 
        LEFTNet & 30 & 24 \\
        PaiNN & 28 & 20 \\
        DimeNet++ & 24.6 & 19.5 \\
        SphereNet & 22.8 & 18.9 \\
        Geoformer & 18.4 & 16.5 \\
        Equiformer & 16.4 & 14.3 \\
        Frad & 15.3 & 13.7 \\
        EPT & 15.2 & 13.6 \\
        EquiformerV2 & 14.4 & 13.3 \\
        SliDe & 13.6 & 12.3 \\
        GotenNet & \underline{13.4} & \underline{12.2} \\
        \modelname{} & \textbf{10} & \textbf{8.9} \\
        \bottomrule 
    \end{tabular}
    \label{tab:qm9_res} 
\end{table}

\noindent\textbf{Matbench.} \cref{tab:matbench_res} summarizes the performance on four diverse tasks from the Matbench benchmark. We compare \modelname{} with SOTA models on the Matbench leaderboard, including ALIGNN~\citep{supp_alignn}, MODNet~\citep{supp_MODNet}, CGCNN~\citep{supp_CGCNN}, MEGNet~\citep{supp_MEGNet}, DimeNet++~\citep{supp_DimeNet}, SchNet~\citep{supp_SchNet}, coGN/coNGN family~\citep{supp_coGN_coNGN}. \modelname{} demonstrates exceptional robustness and generalization capabilities, consistently ranking either \textbf{1st or 2nd} across all tasks.
Specifically, our model achieves state-of-the-art performance on the \texttt{Mp\_is\_metal} (classification) and \texttt{Mp\_gap} (regression) tasks, significantly outperforming previous architectures. In the \texttt{Perovskites} and \texttt{Dielectric} tasks, it remains highly competitive, trailing the top-performing specific models by only narrow margins.
A key observation is the variance in baseline performance: no other competing model maintains consistent excellence. For instance, while MODNet achieves the best result on the Dielectric task, its performance drops significantly on other properties like band gap or metallicity. In contrast, \modelname{} proves to be a robust universal approximator, delivering reliable predictions across disparate physical properties.

\begin{table}[hbtp] 
    \centering 
    \caption{Performance comparison on the Matbench benchmark across classification (\texttt{Mp\_is\_metal}) and regression (\texttt{Mp\_gap}, \texttt{Perovskites}, \texttt{Dielectric}) tasks.} 
    \label{tab:matbench_res} 
    \begin{tabular}{@{}lcccc@{}} 
        \toprule 
        \textbf{Model} & \textbf{Mp\_is\_metal ($\uparrow$)} & \textbf{Mp\_gap (eV, $\downarrow$)} & \textbf{Perovskites (eV/unit cell, $\downarrow$)} & \textbf{Dielectric ($\downarrow$)} \\
        \midrule 
        ALIGNN & 0.9128$\pm$0.0014 & 0.1861$\pm$0.0029 & 0.0288$\pm$0.0009 & 0.3449$\pm$0.0859 \\
        MODNet & 0.9038$\pm$0.0120 & 0.2199$\pm$0.0070 & 0.0908$\pm$0.0029 & \textbf{0.2711$\pm$0.0714} \\
        CGCNN & \underline{0.9520$\pm$0.0071} & 0.2972$\pm$0.0035 & 0.0452$\pm$0.0007 & 0.5988$\pm$0.0855 \\
        MEGNet & 0.9032$\pm$0.0016 & 0.1934$\pm$0.0080 & 0.0352$\pm$0.0015 & 0.3391$\pm$0.0749 \\
        DimeNet++ & 0.8907$\pm$0.0033 & 0.1993$\pm$0.0057 & 0.0342$\pm$0.0010 & 0.3277$\pm$0.0566 \\ 
        SchNet & 0.9124$\pm$0.0017 & 0.2352$\pm$0.0035 & 0.0269$\pm$0.0004 & 0.3088$\pm$0.0834 \\
        coGN & 0.9089$\pm$0.0023 & \underline{0.1559$\pm$0.0017} & \textbf{0.0269$\pm$0.0008} & 0.3088$\pm$0.0829 \\
        coNGN & 0.9089$\pm$0.0019 & 0.1697$\pm$0.0032 & 0.0290$\pm$0.0011 & 0.3142$\pm$0.0769 \\
        \modelname{} & \textbf{0.9628$\pm$0.0010} & \textbf{0.1514$\pm$0.0030} & \underline{0.0274$\pm$0.0006} & \underline{0.2936$\pm$0.0736} \\ 
        \bottomrule 
    \end{tabular}
\end{table}

\subsection{Interatomic Potential Estimation of Non-equilibrium Systems}

\textbf{DPA-2 Datasets.} We further validate the model's scalability and precision on the DPA-2 dataset, which encompasses a wide variety of systems including crystals, molecules, and mixed states. The results for Energy and Force RMSE are detailed in \cref{tab:DPA-2_res}.
We compare \modelname{} with baselines in the DPA-2 original paper, including GNO (GemNet-OC~\citep{supp_gemnet-oc}), EFV2
(EquiformerV2~\citep{supp_EquiformerV2}), NequIP~\citep{supp_NequIP}, Allegro~\citep{supp_Allegro}, MACE~\citep{supp_Mace} and DPA-2~\citep{supp_DPA-2}. \modelname{} achieves state-of-the-art performance on the majority of the subsets. Notably, on complex systems such as \texttt{Al$\cup$Mg$\cup$Cu} and \texttt{High Entropy Alloys}, our model reduces the simulation error significantly compared to the DPA-2 baseline and other equivariant models. 
While there are a few specific subsets (\eg, \texttt{FerroEle-P} energy) where our model performs slightly below the absolute best specialized baseline, the difference is marginal. Overall, the results confirm that \modelname{} effectively captures the potential energy surface across diverse chemical spaces, maintaining high accuracy in both energy and force predictions.

\begin{table}[htbp]
    \centering
    \caption{Performance comparison of energy and force prediction across DPA-2 sub-datasets in diverse domains.}
    \fontsize{6}{7}\selectfont
     \begin{tabular}{l @{\hspace{4.0pt}} l @{\hspace{4.0pt}} r @{\hspace{4.0pt}} r @{\hspace{4.0pt}} r @{\hspace{4.0pt}} r @{\hspace{4.0pt}} r 
     @{\hspace{4.0pt}} r @{\hspace{4.0pt}} r @{\hspace{4.0pt}} r @{\hspace{4.0pt}} r @{\hspace{4.0pt}} r @{\hspace{4.0pt}} r @{\hspace{4.0pt}} r @{\hspace{4.0pt}} r
     @{\hspace{4.0pt}} r}
        \toprule
        & & \multicolumn{7}{c}{\textbf{Energy RMSE [meV/atom]}} & \multicolumn{7}{c}{\textbf{Force RMSE [meV/\AA]}} \\
        \cmidrule(lr){3-9} \cmidrule(lr){10-16}
        Domain & Dataset & {GNO} & {EFV2} & {NequIP} & {Allegro} & {MACE} & {DPA-2} & {\modelname{}} & {GNO} & {EFV2} & {NequIP} & {Allegro} & {MACE} & {DPA-2} & {\modelname{}} \\
        \midrule
        \multirow{11}{*}{Crystal} & SSE-PBE-P & 2.7 & {{OOM}} & 1.6 & \textbf{1.0} & 1.8 & \underline{1.4} & \textbf{1.0} & \underline{8.2} & {{OOM}} & 41.1 & 47.8 & 29.9 & 50.3 & \textbf{5.4}\\
        & Cu & 6.1 & 1.7 & 6.2 & \underline{1.3} & 38.8 & \textbf{1.2} & \underline{1.3} & 5.8 & \underline{3.8} & 16.7 & 8.9 & 13.6 & 8.9 & \textbf{3.3} \\
        & Sn & 8.4 & 5.2 & 18.2 & 5.6 & {/} & \underline{4.1} & \textbf{3.4} & 33.7 & \underline{19.6} & 62.2 & 40.2 & {/} & 54.4 & \textbf{16.6} \\
        & FerroEle-P & 1.5 & 1.1 & 1.1 & \underline{0.7} & 2.3 & \textbf{0.6} & 2.7 & 17.9 & \underline{13.0} & 23.0 & 28.6 & 31.7 & 28.7 & \textbf{10.8}\\
        & V & 17.9 & 5.6 & 8.8 & 4.2 & 14.2 & \underline{4.1} & \textbf{2.4} & 79.3 & \underline{47.4} & 91.6 & 82.1 & 140.4 & 90.8 & \textbf{8.4} \\
        & Al$\cup$Mg$\cup$Cu & 5.9 & \underline{1.9} & 38.0 & 18.3 & 7.7 & 2.1 & \textbf{1.2} & 9.4 & \underline{5.7} & 48.3 & 40.6 & 42.9 & 19.1 & \textbf{4.1}\\
        & Ti & 44.5 & 19.1 & 27.6 & 6.9 & 8.3 & \underline{5.0} & \textbf{3.8}  & 87.9 & \underline{48.6} & 137.4 & 85.6 & 94.2 & 113.1 & \textbf{38.7} \\
        & W & 79.1 & 46.8 & 20.8 & \textbf{4.0} & 15.6 & 5.6 & \underline{4.4} & 81.2 & \underline{51.3} & 160.4 & 101.6 & 181.2 & 108.1 & \textbf{44.4}\\
        & Alloy & 14.3 & \underline{8.5} & 44.0 & 21.4 & 16.2 & 16.8 & \textbf{3.7} & 85.1 & \underline{62.7} & 175.6 & 119.4 & 190.2 & 125.7 & \textbf{49.1}\\
        & Ag$\cup$Au-PBE & 106.0 & 23.4 & 42.3 & 39.2 & 369.1 & \underline{2.4} & \textbf{2.0} & 8.0 & \underline{4.4} & 43.8 & 58.9 & 34.5 & 17.8 & \textbf{3.9} \\
        & Cluster-P & 47.7 & 34.6 & 75.1 & 54.8 & 41.3 & \underline{31.5} & \textbf{9.1} & \textbf{69.6} & 104.4 & 216.6 & 174.1 & 189.7 & 126.0 & \underline{78.6}\\
        \midrule
        \multirow{2}{*}{Molecule} & H2O-PD & {{OOM}} & {{OOM}} & 0.9 & {{OOM}} & 79.9 & \textbf{0.5} & \underline{0.8} & {{OOM}} & {{OOM}} & 27.1 & {{OOM}} & 29.7 & \underline{24.7} & \textbf{7.8} \\
        & Drug & 40.5 & 29.8 & 21.6 & 13.1 & {/} & \underline{12.7} & \textbf{6.5} & \underline{93.6} & 807.4 & 187.2 & 100.8 & {/} & 125.5 & \textbf{28.6}\\
        \midrule
        \multirow{1}{*}{Mixed} & OC2M & 25.0 & \underline{6.7} & 97.4 & 61.3 & {/} & 36.2 & \textbf{5.8} & 129.1 & \underline{45.2} & 226.1 & 166.8 & {/} & 154.0 & \textbf{28.6}\\
        \bottomrule
    \end{tabular}
    \label{tab:DPA-2_res}
\end{table}

\subsection{Crystal Structure Prediction}

\textbf{MP-20 and MPTS-52.} \cref{tab:mp20mpts52_res} compares the generation performance of \modelname{} against established baselines on the MP-20 and MPTS-52 benchmarks. These baselines include CDVAE~\citep{supp_cdvae}, DiffCSP~\citep{supp_diffcsp}, FlowMM~\citep{supp_FlowMM}, 
CrystalFlow~\citep{supp_CrystalFlow}, CrysBFN~\citep{supp_CrysBFN}. For these experiments, we apply our generative framework follows the diffusion process used in DiffCSP, enhancing it with our pretrained representations.
The results show that \modelname{} consistently outperforms all competing methods in both structural validity (Match Rate) and reconstruction accuracy (RMSE).
A particularly striking improvement is observed on the more challenging MPTS-52 dataset. While the original DiffCSP model achieves a Match Rate of $12.19\%$, our enhanced model reaches $\mathbf{24.95\%}$. \textbf{This represents a more than two-fold improvement over the DiffCSP baseline}, demonstrating that integrating our pretrained encoder significantly boosts the model's ability to generate valid and accurate crystal structures in complex chemical spaces.
\begin{table}[htbp]
    \centering
    \caption{Performance comparison of crystal structure prediction on MP-20 and MPTS-52 datasets.}
    \begin{tabular}{l c c c c}
        \toprule
        \multirow{2}{*}{\textbf{Method}} & \multicolumn{2}{c}{\textbf{MP-20}} & \multicolumn{2}{c}{\textbf{MPTS-52}} \\
        \cmidrule(lr){2-3} \cmidrule(lr){4-5}
        & \textbf{Match Rate (\%)} $\uparrow$ & \textbf{RMSE} $\downarrow$ & \textbf{Match Rate (\%)} $\uparrow$ & \textbf{RMSE} $\downarrow$ \\
        \midrule
        CDVAE & 33.9 & 0.1045 & 5.34 & 0.2106 \\
        DiffCSP & 51.49 & 0.0631 & 12.19 & 0.1786 \\
        FlowMM & 61.39 & 0.0566 & 17.54 & 0.1726 \\
        CrystalFlow & 62.02 & 0.071 & \underline{22.71} & 0.1548 \\
        CrysBFN & \underline{64.35} & \underline{0.0433} & 20.52 & \underline{0.1038} \\
        \modelname{} & \textbf{66.4} & \textbf{0.0329} & \textbf{24.95} & \textbf{0.0908} \\
        \bottomrule
    \end{tabular}
    \label{tab:mp20mpts52_res}
\end{table}

\subsection{Superconductivity Validation}

\textbf{SCP Database.} We further assess the model's capability in predicting complex electronic properties through two distinct evaluation setups: a self-constructed benchmark for DFT-calculated critical temperatures and the standard public leaderboard for the Jarvis dataset.
To rigorously analyze the impact of model scaling versus pretraining on superconductivity prediction, we establish a specialized benchmark for DFT-calculated properties. As detailed in \cref{tab:dft_tc_res}, we compare three distinct model configurations: a lightweight baseline (28M, trained from scratch), a large-scale model (1B, trained from scratch), and our proposed pretrained model.
The results illustrate that increasing model capacity alone yields performance gains, reducing the error on the M.A.D. $\tc$ (calculated via the McMillan--Allen--Dynes formula) from $1.62$ K to $1.39$ K. However, it is insufficient to reach optimal performance. The introduction of pretraining provides a decisive advantage, significantly outperforming the massive 1B parameter model. The pretrained \modelname{} model achieves the lowest error rates across all transport metris (including Seebeck coefficients and thermal conductivity) and reaches a prediction error of just $\mathbf{0.98}$ K for the direct $\tc$ and $\mathbf{1.16}$ K for the M.A.D. $\tc$.

\begin{table}[hbtp] 
    \centering 
    \caption{Performance comparison on the DFT-calculated $\tc$ benchmark to evaluate the impact of model scaling and pretraining.} 
    \fontsize{6}{7}\selectfont
    
    \begin{tabular}{@{}lcccccccccc@{}} 
        \toprule 
        \textbf{Model} & p\_seebeck & n\_seebeck & pcond. & ncond. &pkappa & nkappa&$\lambda$ &$\omega_{\log}$ & $\tc$ & M.A.D. $\tc$\\
        \midrule 
        \modelname{} (28M no pretrain) & 60.82 & 56.23 & 0.90 & 0.86 & 0.72 & 0.71 & 0.13 & 29.82 & 1.60 & 1.62 \\
        \modelname{} (1B no pretrain) & \underline{48.83} & \underline{45.93} & \underline{0.77} & \underline{0.73} & \underline{0.63} & \underline{0.62} & \underline{0.11} & \underline{26.17} & \underline{1.26} & \underline{1.39} \\
        \modelname{} (pretrain) & \textbf{32.80} & \textbf{33.45} & \textbf{0.51} & \textbf{0.46} & \textbf{0.55} & \textbf{0.52} & \textbf{0.08} & \textbf{23.76} & \textbf{0.98} & \textbf{1.16} \\
        \bottomrule 
    \end{tabular}
    \label{tab:dft_tc_res} 
\end{table}

For the electronic bandgap prediction on the public Jarvis dataset, we adhere to the strict evaluation protocols and data splits established by the current SOTA model PotNet.
\cref{tab:jarvis_bandgap} highlights the critical role of pretraining in achieving state-of-the-art results. When trained from scratch, both the 28M and 1B variants fail to surpass strong baselines like ALIGNN or PotNet. However, leveraging our pretrained representations dramatically reduces the prediction error. \modelname{} achieves an MAE of $\mathbf{0.09}$ eV, setting a new record by significantly outperforming the previous SOTA, PotNet ($0.127$ eV). This result confirms that pretraining is essential for unlocking the full potential of the architecture on this challenging benchmark.

\begin{table}[hbtp] 
    \centering 
    \caption{Performance comparison of Jarvis bandgap prediction.} 
    \begin{tabular}{@{}lc@{}} 
        \toprule 
        \textbf{Model} & \textbf{MAE  (eV)} \\
        \midrule 
        CFID & 0.3\\
        CGCNN & 0.2\\
        SchNet & 0.19\\
        MEGNet & 0.145\\
        GATGNN & 0.17\\ 
        ALIGNN & 0.142\\
        Matformer & 0.137\\
        CrysDiff & 0.131\\
        PotNet & \underline{0.127}\\
        \modelname{} (28M w/o pretrain) & 0.24\\
        \modelname{} (1B w/o pretrain) & 0.21\\ 
        \modelname{} (pretrain) & \textbf{0.09}   \\
        \bottomrule
    \end{tabular}
    \label{tab:jarvis_bandgap}
\end{table}

\textbf{Positive Instances and Negative Instances.} We evaluate the classification performance of the trained \modelname{C} model on the held-out validation set. The resulting confusion matrix is detailed in \cref{tab:confusion_matrix}.
Based on these results, the model achieves a \textbf{Precision of 95.6\%}, a \textbf{Recall of 98.7\%}, and an \textbf{F1 Score of 0.971}. 
Crucially, the model demonstrates an exceptionally high recall (with only 2 false negatives out of 155 actual positive instances) alongside a strong precision (with just 7 false positives). In the context of material discovery, missing a highly promising candidate can be a significant setback. Our model ensures that nearly all genuine superconductors are successfully identified, while maintaining a high enough precision to avoid overwhelming experimentalists with false leads. The excellent F1 score confirms that the candidates flagged as "Positive" possess a remarkably high probability of being genuine superconductors, making \modelname{C} a highly reliable and comprehensive filter for identifying high-priority candidates.

\begin{table}[ht]
    \centering
    \caption{Confusion matrix of \modelname{C} on the Positive Instances and Negative Instances validation set.}
    \label{tab:confusion_matrix}
    \renewcommand{\arraystretch}{1.2}
    \begin{tabular}{cccc}
    \toprule
    & & \multicolumn{2}{c}{\textbf{Ground Truth}} \\
    & & \textbf{Positive} & \textbf{Negative} \\
    \midrule
    \multirow{2}{*}{\textbf{Predicted}} & \textbf{Positive} & 153 & 7 \\
    & \textbf{Negative} & 2 & 154 \\
    \bottomrule
    \end{tabular}
\end{table}

\section{More Results}

\subsection{The Dialogue with \agentname{} for Superconductor Recommendation}
Here, we detail the step-by-step prompting process for \agentname{} in Stage 3 of \cref{fig:ElementsClaw}. This interactive workflow is designed to identify a pool of candidate materials and ultimately recommend those with the highest probability of superconductivity that are also highly feasible for experimental synthesis.

\begin{figure}[t!]
    \centering
    \includegraphics[width=\textwidth]{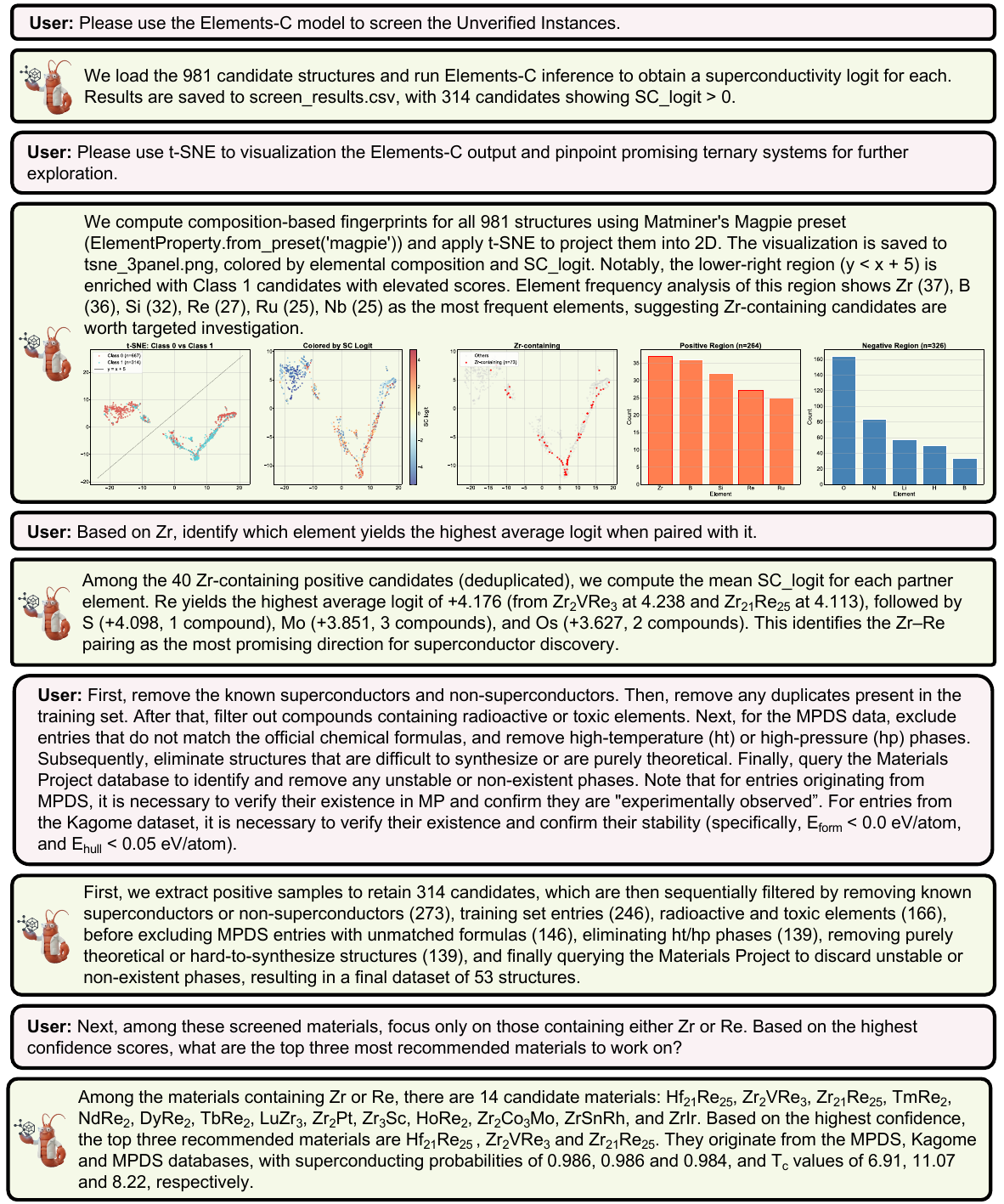}
    \caption{Step-by-step dialogue with \agentname{} to identify and recommend candidate superconductors for experimental synthesis. The prompting process includes invoking the \modelname{C} model for initial classification, performing t-SNE clustering analysis to visualize the material space, and applying comprehensive rule-based screening criteria. Through this guided interaction, the agent systematically filters the dataset and outputs the most promising superconductors for experimental validation. The 53 structures identified in the penultimate response by \agentname{} are summarized in \cref{tab:screening}.}
    \label{fig:supp_appendix_fig2_stage3}
\end{figure}

\begin{table}[htbp]
\centering
\caption{Known superconductors matched in predictions (excluding materials already present in the Supercon3D~\citep{supp_sodnet}). $\tcpred$ is the model-predicted $\tc$ (K), $\tcreal$ is the experimental value (midpoint for ranges). Space groups are determined from the matched MPDS~\citep{supp_mpds} structural entries.}\label{tab:supercond}
\small
\setlength{\tabcolsep}{4pt}
\begin{tabular}{@{} r l r r l !{\hspace{0.8em}\vrule\hspace{0.8em}} r l r r l @{}}
\toprule
No. & Formula & $\tcpred$ & $\tcreal$ & Space Group & No. & Formula & $\tcpred$ & $\tcreal$ & Space Group \\
\midrule
  1 & $\mathrm{MoN}$~\cite{supp_MoN_1} & 33.84 & 5 & $P\overline{6}m2$ & 34 & $\mathrm{V_{3}Pb}$~\cite{supp_V3Pb} & 7.453 & 3.7 & $Pm\overline{3}n$ \\
  2 & $\mathrm{MoN}$~\cite{supp_MoN_2} & 26.83 & 13.8 & $P6_3/mmc$ & 35 & $\mathrm{Mo_{2}C}$~\cite{supp_Mo2C} & 7.4 & 4.5 & $P\overline{3}1m$ \\
  3 & $\mathrm{CaC_{6}}$~\cite{supp_CaC6} & 23.81 & 11.5 & $R\overline{3}m$ & 36 & $\mathrm{KOs_3O_2}$~\cite{supp_KOs3O2} & 7.242 & 9.9 & $F\overline{4}3m$ \\
  4 & $\mathrm{SrC_{6}}$~\cite{supp_SrC6} & 21.28 & 1.6 & $P6_3/mmc$ & 37 & $\mathrm{CaSi_{2}}$~\cite{supp_CaSi2} & 6.9 & 1.6 & $I4_1/amd$ \\
  5 & $\mathrm{PrH_{9}}$~\cite{supp_PrH9} & 20.72 & 9 & $P6_3/mmc$ & 38 & $\mathrm{B_{3}Ru_{7}}$~\cite{supp_B3Ru7} & 6.82 & 3.0 & $Cmc2_1$ \\
  6 & $\mathrm{VRu}$~\cite{supp_VRu_TaRu_NbRu} & 19 & 4.2 & $Pm\overline{3}m$ & 39 & $\mathrm{Rb_2Cr_3As_3}$~\cite{supp_Rb2Cr3As3} & 6.8 & 4.8 & $Amm2$ \\
  7 & $\mathrm{MgAlB_{4}}$~\cite{supp_MgAlB4} & 18.14 & 3 & $P6/mmm$ & 40 & $\mathrm{TaBRu}$~\cite{supp_TaBRu} & 6.754 & 4.0 & $Pbam$ \\
  8 & $\mathrm{MoN}$~\cite{supp_MoN_3} & 17.5 & 12.1 & $P3m1$ & 41 & $\mathrm{AlSb}$~\cite{supp_AlSb} & 6.69 & 2.8 & $Fmmm$ \\
  9 & $\mathrm{ZrP}$~\cite{supp_ZrP} & 16.19 & 4.5 & $Fm\overline{3}m$ & 42 & $\mathrm{Re_{2}W_{3}C}$~\cite{supp_Re2W3C} & 6.676 & 2.9 & $P2_13$ \\
  10 & $\mathrm{BeNb_{3}}$~\cite{supp_BeNb3} & 16.08 & 10 & $Pm\overline{3}n$ & 43 & $\mathrm{BaH_{12}}$~\cite{supp_BaH12} & 6.555 & 20 & $Fm\overline{3}m$ \\
  11 & $\mathrm{Nb_{3}Si}$~\cite{supp_Nb3Si} & 15.58 & 8.9 & $Pm\overline{3}n$ & 44 & $\mathrm{HfAsRu}$~\cite{supp_HfAsRu} & 6.406 & 4.7 & $P\overline{6}2m$ \\
  12 & $\mathrm{Ta_{5}N_{6}}$~\cite{supp_Ta5N6} & 14.59 & 7 & $Cm$ & 45 & $\mathrm{Re_{7}B_{3}}$~\cite{supp_Re7B3} & 6.406 & 3.3 & $Cm$ \\
  13 & $\mathrm{NbN}$~\cite{supp_NbN} & 14.57 & 13.7 & $Fm\overline{3}m$ & 46 & $\mathrm{Nb_{5}Re_{24}}$~\cite{supp_Nb5Re24} & 6.23 & 8.8 & $I\overline{4}3m$ \\
  14 & $\mathrm{Cr_{3}Os}$~\cite{supp_Cr3Os} & 14.03 & 4.0 & $Pm\overline{3}n$ & 47 & $\mathrm{ThTc_{2}}$~\cite{supp_ThTc2} & 6.145 & 5.3 & $Cmcm$ \\
  15 & $\mathrm{NbC}$~\cite{supp_NbC} & 13.99 & 11.8 & $Fm\overline{3}m$ & 48 & $\mathrm{ZrPRu}$~\cite{supp_ZrPRu_2_ZrPOs} & 6.113 & 3.7 & $Pnma$ \\
  16 & $\mathrm{MoC}$~\cite{supp_MoC} & 12.35 & 9.3 & $P6_3/mmc$ & 49 & $\mathrm{AlV_{2}N}$~\cite{supp_AlV2N} & 6.023 & 15.7 & $Fm\overline{3}m$ \\
  17 & $\mathrm{V_{3}Ga}$~\cite{supp_V3Ga} & 11.68 & 15 & $Pm\overline{3}n$ & 50 & $\mathrm{Rb_2Mo_3As_3}$~\cite{supp_Rb2Mo3As3} & 5.99 & 10.3 & $Amm2$ \\
  18 & $\mathrm{HfN}$~\cite{supp_HfN} & 11.65 & 6.7 & $Fm\overline{3}m$ & 51 & $\mathrm{SrC_{10}}$~\cite{supp_SrC10} & 5.973 & 4 & $Im\overline{3}$ \\
  19 & $\mathrm{Nb_{3}Al}$~\cite{supp_Nb3Al} & 11.46 & 18.5 & $Pm\overline{3}n$ & 52 & $\mathrm{ZrRh}$~\cite{supp_ZrRh_1} & 5.906 & 2.7 & $Pnma$ \\
  20 & $\mathrm{Nb_{2}CS}$~\cite{supp_Nb2CS} & 11.164 & 4 & $P6_3/mmc$ & 53 & $\mathrm{ScS}$~\cite{supp_NbS_ScS} & 5.86 & 5 & $Fm\overline{3}m$ \\
  21 & $\mathrm{CdNi_{3}C}$~\cite{supp_CdNi3C} & 10.34 & 3.0 & $Pm\overline{3}m$ & 54 & $\mathrm{CuNi_{3}N}$~\cite{supp_CuNi3N} & 5.836 & 3.2 & $Pm\overline{3}m$ \\
  22 & $\mathrm{Zr_{4}Tc_{25}}$~\cite{supp_Zr4Tc25} & 10.336 & 9.7 & $I\overline{4}3m$ & 55 & $\mathrm{ZrPOs}$~\cite{supp_ZrPRu_2_ZrPOs} & 5.758 & 7.2 & $P\overline{6}2m$ \\
  23 & $\mathrm{NbRu}$~\cite{supp_VRu_TaRu_NbRu} & 10.19 & 4.7 & $P4/mmm$ & 56 & $\mathrm{InSb}$~\cite{supp_InSb} & 5.562 & 3.4 & $Pmmn$ \\
  24 & $\mathrm{Zr_{2}Co}$~\cite{supp_Zr2Co} & 9.26 & 6 & $I4/mcm$ & 57 & $\mathrm{K_2Cr_3As_3}$~\cite{supp_K2Cr3As3} & 5.52 & 6.1 & $Amm2$ \\
  25 & $\mathrm{TaC}$~\cite{supp_TaC} & 9.05 & 10.3 & $Fm\overline{3}m$ & 58 & $\mathrm{NbSiOs}$~\cite{supp_NbSiOs_TaSiOs} & 5.42 & 3.4 & $Pnma$ \\
  26 & $\mathrm{BW}$~\cite{supp_BW_B5W2} & 8.94 & 4.3 & $Cmcm$ & 59 & $\mathrm{Mo_{3}Se}$~\cite{supp_Mo3Se} & 5.324 & 2.2 & $Pm\overline{3}n$ \\
  27 & $\mathrm{ZrPRu}$~\cite{supp_ZrPRu_1} & 8.61 & 11.6 & $P\overline{6}2m$ & 60 & $\mathrm{LaBe_{13}}$~\cite{supp_LaBe13} & 5.188 & 0.6 & $Fm\overline{3}c$ \\
  28 & $\mathrm{NbS}$~\cite{supp_NbS_ScS} & 8.55 & 3.5 & $Pnma$ & 61 & $\mathrm{NdH_{9}}$~\cite{supp_NdH9} & 5.18 & 4.5 & $P6_3/mmc$ \\
  29 & $\mathrm{La_{3}InB}$~\cite{supp_La3InB} & 8.43 & 10 & $Pm\overline{3}m$ & 62 & $\mathrm{ZrP_2Ru_4}$~\cite{supp_ZrP2Ru4} & 5.164 & 11 & $P4_2/mnm$ \\
  30 & $\mathrm{TaRu}$~\cite{supp_VRu_TaRu_NbRu} & 8.13 & 2.8 & $P4/mmm$ & 63 & $\mathrm{ZrRh}$~\cite{supp_ZrRh_2} & 5.027 & 2.5 & $Pm\overline{3}m$ \\
  31 & $\mathrm{B_{5}W_{2}}$~\cite{supp_BW_B5W2} & 7.9 & 5.4 & $R\overline{3}m$ & 64 & $\mathrm{TaSiOs}$~\cite{supp_NbSiOs_TaSiOs} & 5.02 & 5.5 & $Pnma$ \\
  32 & $\mathrm{NbBRu}$~\cite{supp_NbBRu} & 7.9 & 3.1 & $Pmma$ & 65 & $\mathrm{CaAlSi}$~\cite{supp_CaAlSi} & 4.996 & 7.8 & $P\overline{6}m2$ \\
  33 & $\mathrm{AlV_{3}}$~\cite{supp_AlV3} & 7.812 & 10.4 & $Pm\overline{3}n$ & 66 & $\mathrm{LiGa_{2}Ir}$~\cite{supp_LiGa2Ir} & 4.996 & 2.9 & $Fm\overline{3}m$ \\
\bottomrule
\end{tabular}
\end{table}

\clearpage

\begin{table}[htbp]
\centering
\caption{Superconducting candidates from MPDS and Kagome databases. These are materials identified through our interactive dialogue with \agentname{} as potential superconductor candidates that have not been previously reported as superconductors.}\label{tab:screening}
\normalsize
\begin{tabular}{r l c c !{\hspace{0.8em}\vrule\hspace{0.8em}} r l c c}
\toprule
No. & Formula & $\tcpred$ & Confidence & No. & Formula & $\tcpred$ & Confidence \\
\midrule
  1 & $\mathrm{Nb_{5}N_{6}}$ & 13.28 & 0.977 & 28 & $\mathrm{PrPIr}$ & 5.70 & 0.771 \\
  2 & $\mathrm{Zr_{2}VRe_{3}}$ & 11.07 & 0.986 & 29 & $\mathrm{DyOs_{2}}$ & 5.65 & 0.590 \\
  3 & $\mathrm{NbP}$ & 11.05 & 0.905 & 30 & $\mathrm{ZrIr}$ & 5.63 & 0.544 \\
  4 & $\mathrm{NbRu}$ & 9.34 & 0.910 & 31 & $\mathrm{HoOs_{2}}$ & 5.59 & 0.686 \\
  5 & $\mathrm{TbRe_{2}}$ & 8.56 & 0.932 & 32 & $\mathrm{La_{3}Rh_{2}}$ & 5.50 & 0.866 \\
  6 & $\mathrm{Nb_{7}B_{6}C_{3}}$ & 8.55 & 0.798 & 33 & $\mathrm{NdPIr}$ & 5.40 & 0.761 \\
  7 & $\mathrm{YSi_{2}}$ & 8.47 & 0.952 & 34 & $\mathrm{MoS_{2}}$ & 5.40 & 0.527 \\
  8 & $\mathrm{Zr_{21}Re_{25}}$ & 8.22 & 0.984 & 35 & $\mathrm{TmOs_{2}}$ & 5.36 & 0.790 \\
  9 & $\mathrm{TaBRu}$ & 7.92 & 0.801 & 36 & $\mathrm{Nb_{2}Al}$ & 5.34 & 0.972 \\
 10 & $\mathrm{TiIr}$ & 7.80 & 0.872 & 37 & $\mathrm{ZrSnRh}$ & 5.30 & 0.590 \\
 11 & $\mathrm{Nb_{3}B_{3}C}$ & 7.71 & 0.683 & 38 & $\mathrm{LiGa_{2}Rh}$ & 5.26 & 0.894 \\
 12 & $\mathrm{TmRe_{2}}$ & 7.71 & 0.967 & 39 & $\mathrm{NbPt}$ & 5.21 & 0.786 \\
 13 & $\mathrm{TbSi_{2}}$ & 7.55 & 0.978 & 40 & $\mathrm{Zr_{3}Sc}$ & 5.19 & 0.910 \\
 14 & $\mathrm{LaGeIr}$ & 7.46 & 0.942 & 41 & $\mathrm{Nb_{10}Ge_{7}}$ & 5.17 & 0.952 \\
 15 & $\mathrm{YbOs_{2}}$ & 7.43 & 0.916 & 42 & $\mathrm{Zr_{2}Pt}$ & 5.16 & 0.917 \\
 16 & $\mathrm{Nb_{7}B_{4}C_{4}}$ & 7.21 & 0.921 & 43 & $\mathrm{Nb_{9}Ni_{4}Ge}$ & 5.09 & 0.948 \\
 17 & $\mathrm{Hf_{21}Re_{25}}$ & 6.91 & 0.986 & 44 & $\mathrm{LiAl}$ & 5.01 & 0.888 \\
 18 & $\mathrm{PrSi_{2}}$ & 6.82 & 0.885 & 45 & $\mathrm{La_{4}MgRh}$ & 5.00 & 0.877 \\
 19 & $\mathrm{NdRe_{2}}$ & 6.70 & 0.951 & 46 & $\mathrm{BW}$ & 5.00 & 0.928 \\
 20 & $\mathrm{TaRu}$ & 6.69 & 0.777 & 47 & $\mathrm{Mo_{3}Pt_{2}N}$ & 4.99 & 0.929 \\
 21 & $\mathrm{HoRe_{2}}$ & 6.69 & 0.834 & 48 & $\mathrm{Ta_{5}Ge_{3}B}$ & 4.95 & 0.597 \\
 22 & $\mathrm{DyRe_{2}}$ & 6.68 & 0.948 & 49 & $\mathrm{Zr_{2}Co_{3}Mo}$ & 4.87 & 0.818 \\
 23 & $\mathrm{LiGa_{2}Ru}$ & 6.44 & 0.937 & 50 & $\mathrm{Y_{2}NiRu_{3}}$ & 4.76 & 0.971 \\
 24 & $\mathrm{NdSi_{2}}$ & 6.37 & 0.888 & 51 & $\mathrm{LuZr_{3}}$ & 4.30 & 0.930 \\
 25 & $\mathrm{B_{2}Mo_{2}Os}$ & 5.90 & 0.967 & 52 & $\mathrm{Y_{2}CoRu_{3}}$ & 4.16 & 0.972 \\
 26 & $\mathrm{HfIr}$ & 5.88 & 0.942 & 53 & $\mathrm{NdOs_{2}}$ & 4.13 & 0.766 \\
 27 & $\mathrm{TaMoN}$ & 5.76 & 0.955 & & & & \\
\bottomrule
\end{tabular}
\end{table}

\subsection{Visualization of Predicted Superconductors}

\begin{figure}[!thp]
    \centering
    \includegraphics[
        width=\linewidth,
        height=0.925\textheight,
        keepaspectratio
    ]{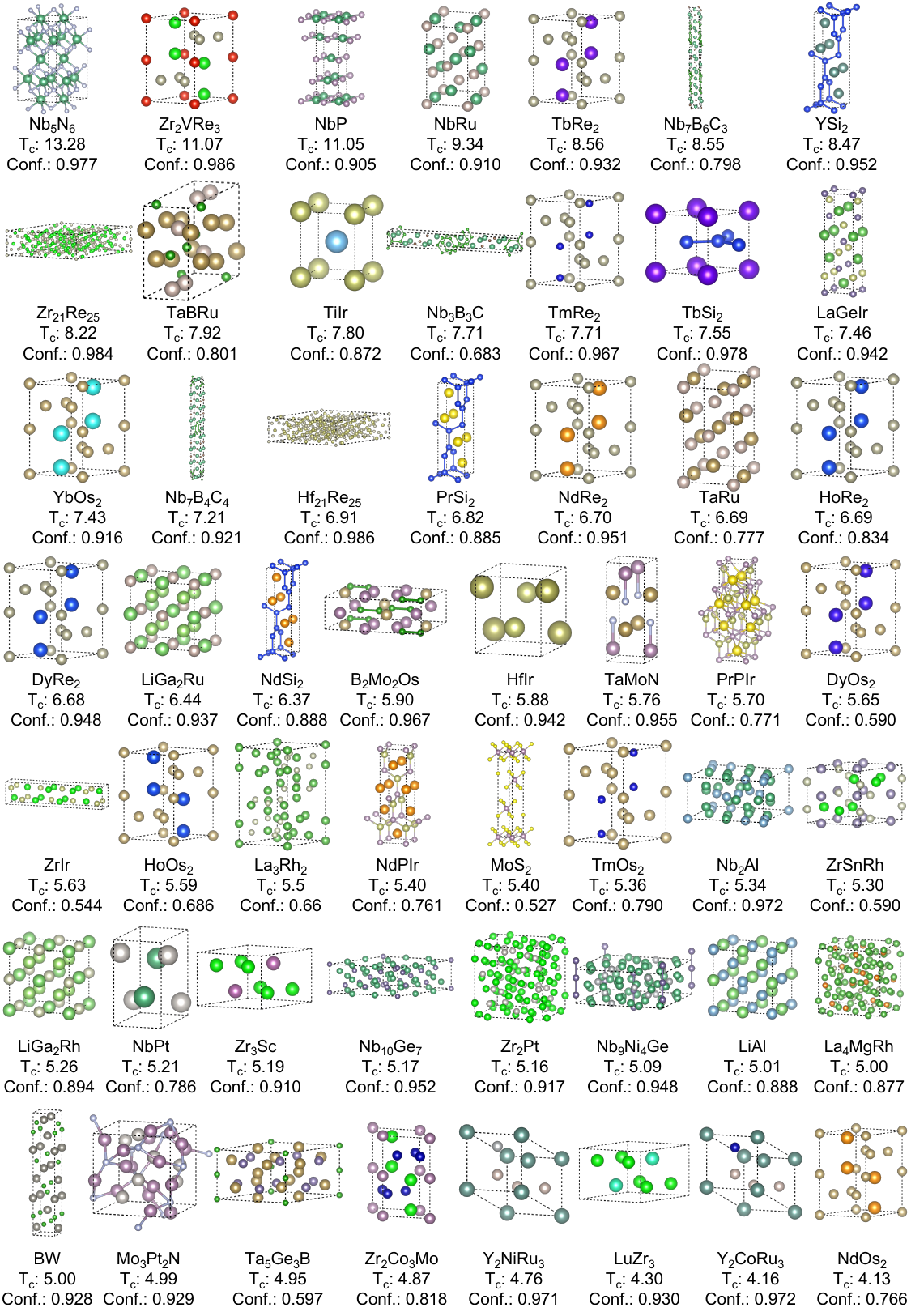}
    \caption{Visualization of the 53 superconducting candidates in \cref{tab:screening}. These positive instances are specifically screened from the unverified MPDS and Kagome databases using \modelname{C}, and ranked according to their highest predicted $\tc$ using \modelname{T}.}
    \label{fig:supp_MPDS_Kagome_supercon}
\end{figure}

\begin{figure}[!htp]
    \centering
    \includegraphics[
        width=\linewidth,
        height=0.9\textheight,
        keepaspectratio
    ]{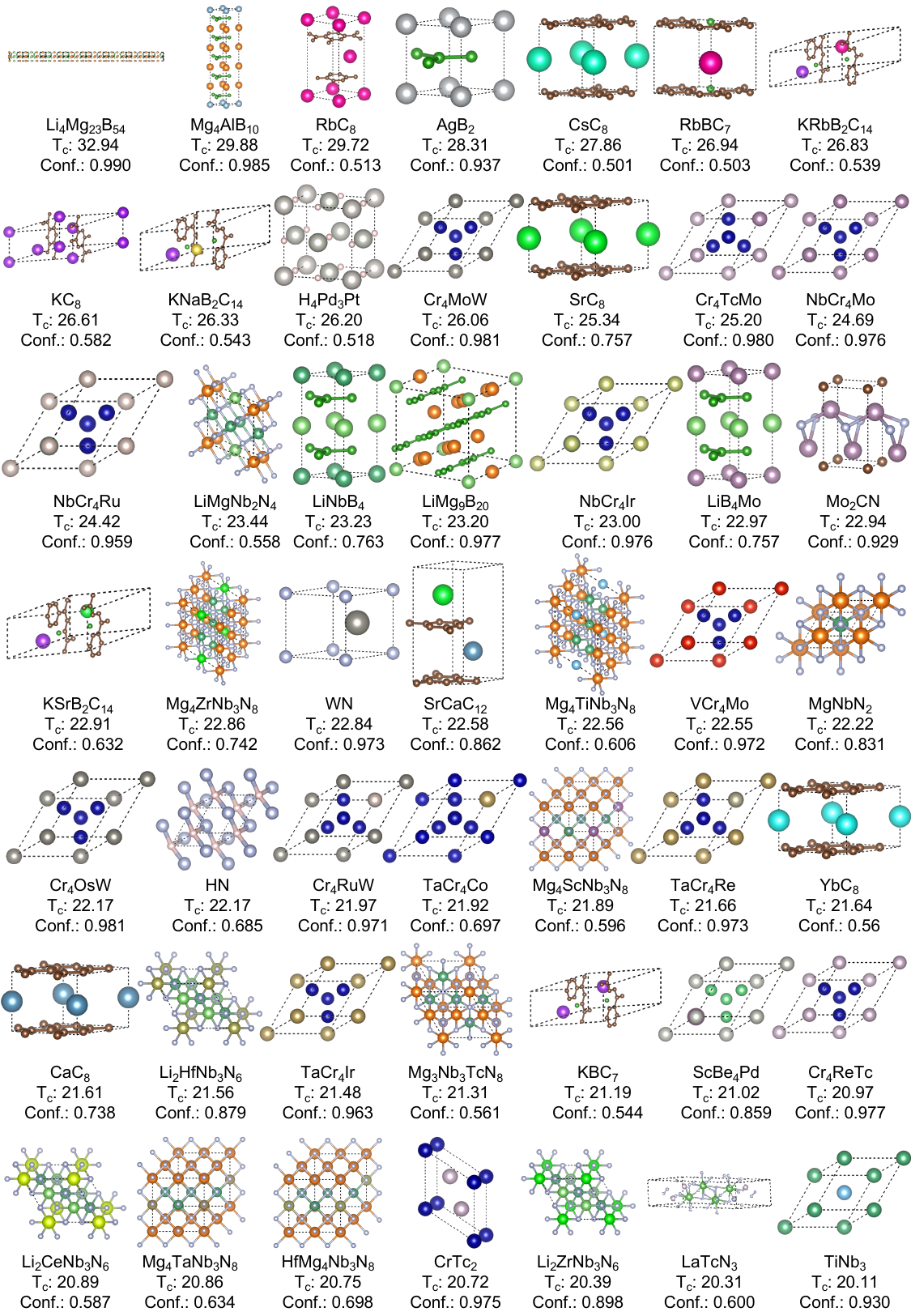}
    \caption{Visualization of the top 49 superconducting candidates selected from all equilibrium crystals in the pretraining dataset. Positive instances are identified using \modelname{C}, ranked by their highest predicted $\tc$ using \modelname{T}, and filtered to exclude hydrogen- and boron-containing compounds.}
    \label{fig:supp_all_supercon}
\end{figure}

To identify viable experimental targets, \agentname{} initially recommended 53 candidates (\cref{fig:supp_MPDS_Kagome_supercon}) via a streamlined screening pipeline (\cref{fig:supp_appendix_fig2_stage3}) that involved deduplication, removing toxic elements, excluding known materials, and verifying phase stability. Narrowing our focus to the Zr--Re system, we ultimately selected Zr$_2$VRe$_3$, Zr$_{21}$Re$_{25}$, and Hf$_{21}$Re$_{25}$ for experimental synthesis. Building on this pipeline, we subsequently extended our screening to all stable crystals within the pretraining dataset. Using \modelname{T} to predict $\tc$ and \modelname{C} to isolate positive instances, this process culminated in the selection of the top 49 candidates with the highest predicted $\tc$ (\cref{fig:supp_all_supercon}).

\subsection{Superconductivity Validation on Experimental Dataset}
\label{sec:exp_dataset}

\textbf{SuperCon3D.} For the experimental $\tc$ prediction, we adapt the training regime to accommodate the sparsity and noise inherent in experimental data compared to DFT calculations.
\begin{itemize}
    \item \textbf{Optimization Dynamics:} To ensure adequate fine-tuning of the pretrained representations, the training duration is extended to $1000$ epochs. We employ a smaller batch size of $256$ and a prolonged warmup phase of $10$ epochs relative to the DFT tasks to stabilize the initial optimization trajectory.
    \item \textbf{Hybrid Loss Function:} A key component of our approach is a hybrid objective function that leverages theoretical priors to guide the learning of experimental properties. The total loss is a weighted sum where the theoretical proxy ($\text{DFT } \tc$) is assigned a dominant coefficient of $10$, while the ground-truth experimental target ($\text{EXP } \tc$) is weighted at $5$. Auxiliary physical constraints, specifically the electron-phonon coupling ($\lambda$) and logarithmic average frequency ($\omega_{log}$), are incorporated with coefficients of $5$ and $1$, respectively.
\end{itemize}

\begin{table}[htbp]
    \centering
    \caption{Training hyper-parameters and hybrid loss function coefficients for experimental $\tc$ prediction on the SuperCon3D dataset.}
    \begin{tabular}{ll}
        \toprule
        Hyper-parameters & EXP $\tc$ Training \\
        \midrule
        Warmup epochs & 10 \\
        Batch size & 256 \\
        Number of epochs & 1000 \\
        $\lambda$ loss coefficient & 5 \\
        $\omega_{\log}$ loss coefficient & 1 \\
        DFT $\tc$ loss coefficient & 10 \\
        EXP $\tc$ loss coefficient & 5 \\
        \bottomrule
    \end{tabular}
\end{table}

To ensure a rigorous comparison, we reproduce several state-of-the-art baselines (marked with * in the results) using their official architectural implementations. We standardize the optimization environment using the \texttt{Adam} optimizer and, for most models, a One-Cycle learning rate scheduler, while adhering to model-specific configurations:
\begin{itemize}
    \item \textbf{Interaction-based Models (SchNet~\citep{supp_SchNet}\footnote{\url{https://github.com/atomistic-machine-learning/SchNet}}, DimeNet++~\citep{supp_DimeNet}\footnote{\url{https://github.com/gasteigerjo/dimenet}}):} These models require an extended training period of $500$ epochs to reach convergence. SchNet utilizes a batch size of $64$ with a cutoff at the 12th nearest neighbor, while DimeNet++ employs a radial cutoff of $8.0$ \AA{} with a larger batch size of $128$.
    \item \textbf{Graph Convolutional Baselines (CGCNN~\citep{supp_CGCNN}\footnote{\url{https://github.com/txie-93/cgcnn}}, MEGNet~\citep{supp_MEGNet}\footnote{\url{https://github.com/materialsvirtuallab/megnet}}):} Both models are trained for $200$ epochs with a batch size of $64$. CGCNN constructs the graph using the $32$ nearest neighbors, whereas MEGNet operates with a fixed radius cutoff of $8.0$ \AA{} and utilizes a Set2Set readout function.
    \item \textbf{Advanced Architectures (ALIGNN~\citep{supp_alignn}\footnote{\url{https://github.com/usnistgov/alignn}}, Matformer~\citep{supp_matformer}\footnote{\url{https://github.com/YKQ98/Matformer}}, SphereNet~\citep{supp_spherenet}\footnote{\url{https://github.com/divelab/DIG}}):} ALIGNN and Matformer are optimized over $150$ epochs using a batch size of $64$, both employing a neighbor-based graph construction ($k=12$). SphereNet, utilizing a multi-graph representation with a cutoff of $6$ \AA{}, is trained for $300$ epochs with a smaller batch size of $32$ and a distinct learning rate strategy involving plateau-based decay.
\end{itemize}

We evaluate the model's performance on predicting the experimental $\tc$ using the SuperCon3D dataset, employing a rigorous \textbf{10-fold cross-validation} scheme to ensure statistical reliability. To facilitate a fair comparison under identical experimental conditions, we locally reproduce several state-of-the-art graph neural networks (\eg, ALIGNN, Matformer), marking these baselines with an asterisk (*) in \cref{tab:supercon3d_res}. Our results indicate that the base \modelname{} model demonstrates superior intrinsic generalization, achieving a Mean Absolute Error (MAE) of $\mathbf{0.732}$ on $\log(\tc)$ and surpassing all competing baselines even without auxiliary guidance. Furthermore, by incorporating the DFT-calculated $\tc$ as an auxiliary supervision task (denoted as \textit{+\text{DFT} $\tc$}), we observe a significant performance boost, lowering the MAE to $\mathbf{0.703}$ and increasing the $\mathrm{R^2}$ score to $\mathbf{0.548}$. This confirms that integrating theoretical physical priors effectively guides the model in navigating the noise and sparsity inherent in experimental datasets.

\begin{table}[hbtp] 
    \centering 
    \caption{Performance comparison on the SuperCon3D dataset for predicting experimental $\tc$ using a 10-fold cross-validation scheme.} 
    \label{tab:supercon3d_res} 
    \begin{tabular}{@{}lcc@{}} 
        \toprule 
        \textbf{Model} & \textbf{MAE  (logK, $\downarrow$)} & \textbf{$\mathrm{R^2}$ Score ($\uparrow$)} \\
        \midrule 
        *SchNet & 0.891$\pm$0.041 & 0.401$\pm$0.032 \\
        *CGCNN & 0.879$\pm$0.047 & 0.405$\pm$0.022 \\
        *DimeNet++ & 0.827$\pm$0.058 & 0.444$\pm$0.061  \\
        *SphereNet & 0.811$\pm$0.058 & 0.434$\pm$0.092  \\
        *ALIGNN & 0.762$\pm$0.048 & 0.467$\pm$0.096 \\ 
        *Matformer & 0.755$\pm$0.049 & 0.479$\pm$0.090  \\
        *MEGNet & 0.770$\pm$0.065 & 0.463$\pm$0.112 \\
        \modelname{} & \underline{0.732$\pm$0.109} & \underline{0.505$\pm$0.145}  \\ 
        \modelname{} (+DFT $\tc$) & \textbf{0.703$\pm$0.075} & \textbf{0.548$\pm$0.111}  \\
        \bottomrule 
    \end{tabular}
\end{table}

\subsection{First-Principles Analysis and Computational Efficiency}
\label{sec:computation_cost}

To complement our experimental investigations and provide a microscopic understanding of the underlying physical mechanisms, we also performed first-principles Density Functional Theory (DFT) calculations on the representative candidate, Zr$_2$VRe$_3$. Geometry optimizations and total-energy calculations are carried out within the framework of DFT using the Vienna \textit{ab initio} Simulation Package (VASP)~\citep{supp_vasp1,supp_vasp2}. The exchange and correlation functional is treated using the Generalized Gradient Approximation (GGA)~\citep{supp_pbe}, and electron-ion interactions are described by the Projector Augmented-Wave (PAW) method~\citep{supp_paw1,supp_paw2}. A plane-wave kinetic-energy cutoff is set to 600~eV. Reliable convergence is ensured by performing Brillouin zone integrations using Monkhorst-Pack \textit{k}-point meshes~\citep{supp_monkhorst_pack} with a maximum grid spacing of $2\pi\times 0.03$~\AA{}. Structural relaxations continue until the residual Hellmann-Feynman forces on each atom are less than $10^{-2}$~eV/\AA~and the total-energy difference between successive ionic steps is smaller than $10^{-5}$~eV.

Furthermore, phonon dispersions and electron-phonon coupling (EPC) calculations for Zr$_2$VRe$_3$ are performed utilizing density functional perturbation theory (DFPT)~\citep{supp_baroni_dfpt} as implemented in the \textsc{Quantum ESPRESSO}~\citep{supp_qe} package. Ultrasoft pseudopotentials~\citep{supp_vanderbilt_uspp} are employed with a kinetic-energy cutoff for the wave-function expansion set to 70~Ry. Dense $\Gamma$-centered Monkhorst-Pack \textit{k}-point meshes are used for Brillouin zone integrations, targeting a reciprocal-space resolution of $2\pi\times 0.02~\mathrm{\mathring{A}}^{-1}$. The EPC matrix elements are evaluated on a uniform \textit{q}-point mesh of $4\times 4\times 4$. Convergence is rigorously checked with respect to both \textit{k}- and \textit{q}-point samplings to ensure the stability of the calculated parameters.

Ultimately, the theoretical $\tc$ is estimated using the Allen-Dynes modified McMillan formula:
\begin{equation}
    T_c = \frac{\omega_{\log}}{1.2} \exp \left[ -\frac{1.04(1+\lambda)}{\lambda-\mu^{\ast}(1+0.62\lambda)} \right],
\end{equation}
where $\lambda$ represents the EPC constant, $\omega_{\log}$ is the logarithmic average phonon frequency, and $\mu^{\ast}$ is the Coulomb pseudopotential, which is set to $\mu^{\ast}=0.1$ for this work.

Interestingly, despite the theoretical potential suggested by these first-principles calculations, our subsequent experimental measurements reveal that the synthesized Zr$_2$VRe$_3$ samples do not exhibit bulk superconductivity, despite showing a transition signal that we attribute to a trace impurity phase. This discrepancy sharply highlights the inherent gap between idealized theoretical predictions and experimental reality, where factors such as complex phase stabilities, unpredictable defect formations, trace impurities, and subtle non-ideal stoichiometries often intervene. It is precisely this profound gap that underscores the critical importance of developing and training data-driven models like \modelname{C}. By implicitly learning from vast, high-dimensional datasets of real-world experimental outcomes rather than relying solely on idealized physics equations, \modelname{C} is equipped to capture the complex, hidden heuristics that govern actual synthesizability and macroscopic properties, thereby effectively bridging the divide between theoretical promise and experimental realization.

Beyond bridging this complex gap between idealized theory and experimental reality, the \modelname{} suite also demonstrates a tremendous advantage in computational efficiency compared to traditional DFT methods. Taking the Zr$_2$VRe$_3$ system as an example, standard DFT calculations typically require approximately 2 days to complete. In stark contrast, synergistically utilizing the \modelname{T}, \modelname{C} and \modelname{E} models for comprehensive property prediction of this material takes less than 2 seconds, achieving a speedup of nearly 80,000 times. Furthermore, even when utilizing \modelname{G} for \textit{ab initio} generation of the Zr$_2$VRe$_3$ crystal structure, the entire process takes only about 5 minutes. By achieving an orders-of-magnitude leap in computational speed while maintaining robust prediction accuracy grounded in real-world data, this breakthrough substantially overcomes the computing power bottlenecks in traditional computational materials science. It significantly accelerates the high-throughput screening pipeline for potential superconductors, marking a crucial step forward in the rapid prediction and discovery of novel materials.

\begin{figure}[!th]
    \centering
    \includegraphics[width=\textwidth]{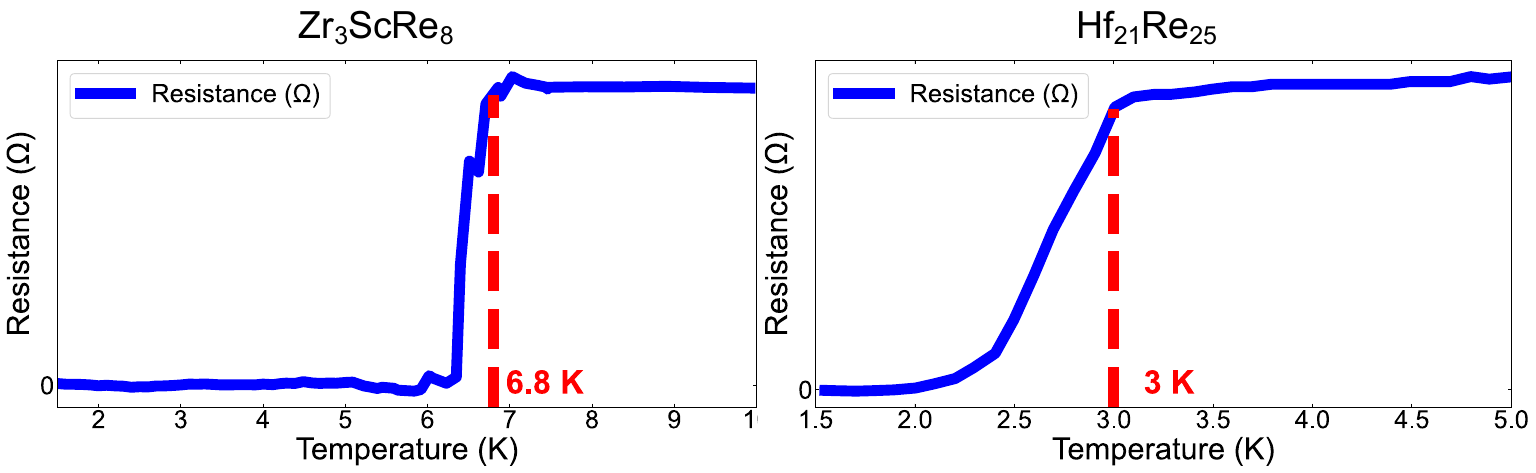}
    \caption{Detailed characterization of six newly discovered superconducting compounds: Zr$_3$ScRe$_8$, HfZrRe$_4$, HfZr$_3$Re$_8$, Hf$_3$ZrRe$_8$, Zr$_4$VRe$_7$, and Hf$_{21}$Re$_{25}$. For each material, we present its refined crystal structure model (left), PXRD pattern with Rietveld refinement results (center), and temperature-dependent magnetic susceptibility ($4\pi\chi$) curve (right). The XRD plots display the observed, calculated, background, difference (deviation), and Bragg peak positions. The $4\pi\chi$ curves provide definitive evidence for the superconducting transitions, with $\tc$ values of approximately 6.5 K, 5.9 K, 5.9 K, 5.7 K, 3.5 K, and 2.5 K, respectively. This detailed validation supports the successful synthesis of these predicted candidate materials.}
    \label{fig:supp_fig6_appendix}
\end{figure}

\begin{figure}[!th]
    \centering
    \includegraphics[width=\textwidth]{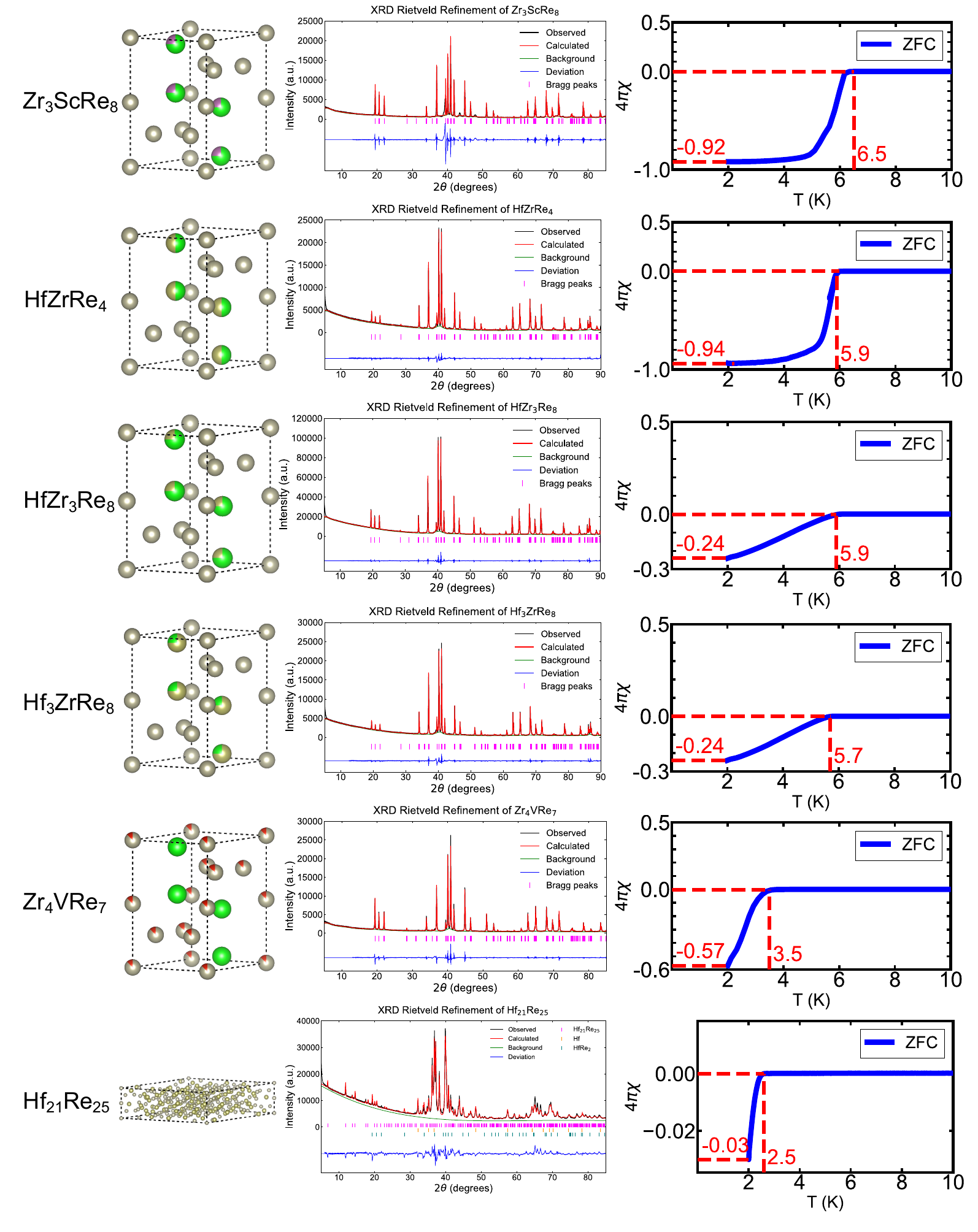}
    \caption{Temperature dependence of electrical resistance for Zr$_3$ScRe$_8$ (left) and Hf$_{21}$Re$_{25}$ (right). The red dashed lines indicate the superconducting transition temperatures ($\tc$) at approximately 6.8 K and 3 K, respectively. Notably, the $T_c$ values obtained from these electrical transport measurements are slightly higher than those derived from the magnetic susceptibility data ($\sim$6.5 K and 2.5 K). This slight difference is physically reasonable, as a zero-resistance state is typically achieved once a continuous superconducting percolation path forms, which generally occurs at slightly higher temperatures than the onset of the bulk diamagnetic response.}
    \label{fig:supp_Elements_SuppFigE6}
\end{figure}

\begin{figure}[!th]
    \centering
    \includegraphics[width=\textwidth]{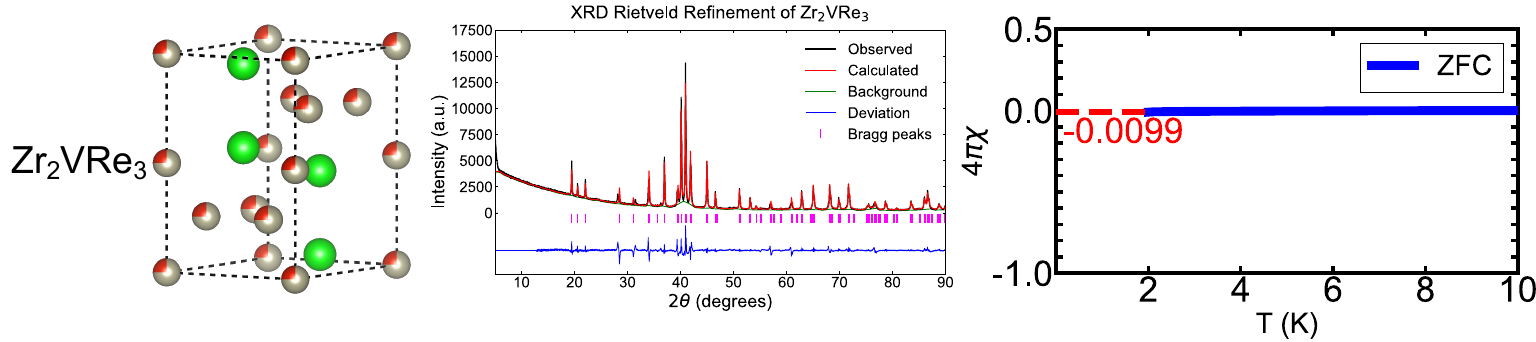}
    \caption{Detailed characterization and magnetic properties of Zr$_{2}$VRe$_{3}$. We present its crystal structure model (left), powder X-ray diffraction (XRD) pattern with Rietveld refinement results (center), and temperature-dependent magnetic susceptibility ($4\pi\chi$) curve (right). The XRD plot displays the observed, calculated, background, difference (deviation), and Bragg peak positions. Although a diamagnetic transition is observed at approximately 8.5 K in the $4\pi\chi$ curve, the extremely small shielding volume fraction ($\sim -9.9 \times 10^{-3}$) indicates the absence of bulk superconductivity in Zr$_{2}$VRe$_{3}$, with the signal likely originating from a trace impurity phase. This discrepancy is attributed to the presence of magnetic V atoms, which introduce pair-breaking effects that are not explicitly captured in the underlying density functional theory-based training data.}
    \label{fig:supp_fig6_appendix_wrong}
\end{figure}

\subsection{Structural Validation, Magnetic, and Electrical Transport Characterization}\label{sec:X-Ray_Diffraction}

To accurately evaluate the superconducting diamagnetic volume fraction of the samples, we adopt the effective demagnetization factor approximation model proposed for fully diamagnetic bodies in the literature~\cite{supp_demag}, and correct the apparent magnetic susceptibility observed by the instrument. Since the polycrystalline powder samples in this study are loaded and compacted into capsules with a finite-length cylindrical cavity, and the external magnetic field is applied parallel to the cylinder axis during the magnetic measurements, the equivalent axial demagnetization factor can be analytically approximated from the geometric dimensions of the sample:

\begin{equation}
N = \frac{1}{1 + 1.6\cdot(\frac{H}{D})},
\end{equation}
where $H$ is the height and $D$ is the diameter of the cylindrical sample (both determined by direct measurement). After obtaining the demagnetization factor $N$ corresponding to the specific geometry, the observed susceptibility is corrected for the demagnetization effect using the following relation to obtain the intrinsic susceptibility of the sample:

\begin{equation}
4\pi\chi_{\mathrm{true}} = \frac{4\pi\chi_{\mathrm{obs}}}{1 - N \cdot 4\pi\chi_{\mathrm{obs}}},
\end{equation}
where $4\pi\chi_{\mathrm{obs}}$ is the apparent volume magnetic susceptibility directly measured and initially converted from the MPMS data, $4\pi\chi_{\mathrm{true}}$ is the intrinsic volume magnetic susceptibility after demagnetization correction, and $N$ is the dimensionless equivalent demagnetization factor obtained above. The volume magnetic susceptibility values before and after the demagnetization correction are summarized in Table~\ref{tab:demag}.

\begin{table}[htbp]
\centering
\caption{Volume magnetic susceptibility ($4\pi\chi$) before and after demagnetization correction for the six superconducting samples.}\label{tab:demag}
\begin{tabular}{@{} l c c @{}}
\toprule
Sample & $4\pi\chi_{\mathrm{obs}}$ & $4\pi\chi_{\mathrm{true}}$ \\
\midrule
$\mathrm{Zr_3ScRe_8}$   & $-1.49$ & $-0.92$ \\
$\mathrm{HfZrRe_4}$      & $-2.00$ & $-0.94$ \\
$\mathrm{HfZr_3Re_8}$    & $-0.27$ & $-0.24$ \\
$\mathrm{Hf_3ZrRe_8}$    & $-0.27$ & $-0.24$ \\
$\mathrm{Zr_4VRe_7}$     & $-0.68$ & $-0.57$ \\
$\mathrm{Hf_{21}Re_{25}}$ & $-0.033$ & $-0.03$ \\
\bottomrule
\end{tabular}
\end{table}

\cref{fig:supp_fig6_appendix} presents the detailed characterization of these six confirmed superconducting compounds ($\mathrm{Zr_3ScRe_8}$, $\mathrm{HfZrRe_4}$, $\mathrm{HfZr_3Re_8}$, $\mathrm{Hf_3ZrRe_8}$, $\mathrm{Zr_4VRe_7}$, and $\mathrm{Hf_{21}Re_{25}}$). For each material, the refined crystal structure model, powder X-ray diffraction (PXRD) patterns with Rietveld refinement results, and temperature-dependent magnetic susceptibility ($4\pi\chi$) curves provide definitive structural validation and robust evidence for the onset of bulk superconductivity.

The temperature-dependent electrical transport properties (resistance--temperature, $R$--$T$) of the $\mathrm{Zr_3ScRe_8}$ and $\mathrm{Hf_{21}Re_{25}}$ samples are further characterized using the standard four-probe method. Bulk metallic ingots prepared by arc melting are first cut into thin slices and polished to obtain smooth surfaces. Four fine copper wires are then attached to the sample surface using conductive silver paste as contact electrodes, effectively eliminating the influence of contact resistance. From the $R$--$T$ data shown in \cref{fig:supp_Elements_SuppFigE6}, clear superconducting transitions are observed: $\mathrm{Zr_3ScRe_8}$ exhibits a $\tc^{\mathrm{onset}}$ of 6.8~K and a $\tc^{\text{Zero}}$ of 6~K, while $\mathrm{Hf_{21}Re_{25}}$ shows a $\tc^{\mathrm{onset}}$ of 3~K and a $\tc^{\text{Zero}}$ of 2~K. Notably, the $\tc$ values obtained from these electrical transport measurements are slightly higher than those derived from the magnetic susceptibility data. This slight difference is physically reasonable, as a zero-resistance state is typically achieved once a continuous superconducting percolation path forms, which generally occurs at slightly higher temperatures than the onset of the bulk diamagnetic response.

In contrast, further analysis of the $\mathrm{Zr_2VRe_3}$ sample revealed an absence of bulk superconductivity, as detailed in \cref{fig:supp_fig6_appendix_wrong}. Although a diamagnetic transition is observed at approximately 8.5~K in the $4\pi\chi$ curve, the extremely small shielding volume fraction ($\sim -9.9 \times 10^{-3}$) indicates that the signal likely originates from a trace impurity phase rather than the target material. This discrepancy is attributed to the presence of magnetic V atoms within the structure, which introduce strong pair-breaking effects that are not explicitly captured in the underlying density functional theory-based training data, ultimately suppressing the bulk superconducting state.

\section{Comparison and Discussion}

In this section, we delineate the specific architectural advancements of \modelname{} compared to existing state-of-the-art models, focusing on both the encoder design and the generative diffusion strategy.

\textbf{Architectural Refinements based on EquiformerV2.}
While our backbone leverages the powerful equivariant representations of \textbf{EquiformerV2}, we introduce several critical structural adaptations to enhance geometric expressivity and multi-task capability:
\begin{itemize}
    \item \textbf{Graph Construction:} Unlike the standard implementation, we explicitly incorporate \textit{periodic self-connections} into the crystal graph to better capture local lattice invariance.
    \item \textbf{Long-Range Connectivity:} To improve information flow across deep networks, we introduce long-range residual connections specifically in the final two Transformer blocks.
    \item \textbf{Multi-Head Design:} We extend the architecture with specialized heads for \textit{Denoising} (predicting coordinate/lattice noise) and \textit{Force Prediction} (vector outputs). Furthermore, for multi-task property prediction, we modify the final projection layer of the energy head. Instead of a standard reduction to a scalar $(d_{\text{ffn}}, 1)$, we utilize a multi-output linear mapping $(d_{\text{ffn}}, N_{\text{target}})$ to simultaneously regress multiple invariant physical quantities.
\end{itemize}

\textbf{Generative Framework and Diffusion Dynamics.}
For the crystal generation tasks, our approach aligns with the diffusion framework established by \textbf{DiffCSP}, but with significant modifications to the encoder and the diffusion process itself:
\begin{itemize}
    \item \textbf{Backbone Integration:} We replace the standard EGNN-like encoder used in DiffCSP with our pretrained \modelname{} backbone, enabling the generative process to leverage chemically rich, pretrained representations.
    \item \textbf{Noise Prior:} Distinct from the standard Gaussian distribution $\mathcal{N}(\vzero, \mI)$, we adopt a dataset-dependent initial noise distribution characterized by specific centering and variance parameters ($c, \nu$), drawing inspiration from the initialization strategies in \textbf{MatterGen}.
    \item \textbf{Cartesian vs. Fractional Diffusion:} A key distinction lies in the coordinate space of the diffusion process. While MatterGen employs a Cartesian-based backbone (GemNet) but performs diffusion and denoising on \textit{fractional} coordinates, our model operates the diffusion dynamics \textbf{directly in the Cartesian coordinate system}. This approach unifies the treatment of atomic positions with lattice deformations. We do not enforce periodic boundary conditions during the intermediate diffusion steps; instead, the conversion from Cartesian to fractional coordinates (and the subsequent wrapping into the unit cell) is performed only at the final sampling step ($t=0$).
\end{itemize}

    \putbib[Supp_Reference]
\end{bibunit}



\begin{thebibliography}{61}
\ifx \bisbn   \undefined \def \bisbn  #1{ISBN #1}\fi
\ifx \binits  \undefined \def \binits#1{#1}\fi
\ifx \bauthor  \undefined \def \bauthor#1{#1}\fi
\ifx \batitle  \undefined \def \batitle#1{#1}\fi
\ifx \bjtitle  \undefined \def \bjtitle#1{#1}\fi
\ifx \bvolume  \undefined \def \bvolume#1{\textbf{#1}}\fi
\ifx \byear  \undefined \def \byear#1{#1}\fi
\ifx \bissue  \undefined \def \bissue#1{#1}\fi
\ifx \bfpage  \undefined \def \bfpage#1{#1}\fi
\ifx \blpage  \undefined \def \blpage #1{#1}\fi
\ifx \burl  \undefined \def \burl#1{\textsf{#1}}\fi
\ifx \doiurl  \undefined \def \doiurl#1{\url{https://doi.org/#1}}\fi
\ifx \betal  \undefined \def \betal{\textit{et al.}}\fi
\ifx \binstitute  \undefined \def \binstitute#1{#1}\fi
\ifx \binstitutionaled  \undefined \def \binstitutionaled#1{#1}\fi
\ifx \bctitle  \undefined \def \bctitle#1{#1}\fi
\ifx \beditor  \undefined \def \beditor#1{#1}\fi
\ifx \bpublisher  \undefined \def \bpublisher#1{#1}\fi
\ifx \bbtitle  \undefined \def \bbtitle#1{#1}\fi
\ifx \bedition  \undefined \def \bedition#1{#1}\fi
\ifx \bseriesno  \undefined \def \bseriesno#1{#1}\fi
\ifx \blocation  \undefined \def \blocation#1{#1}\fi
\ifx \bsertitle  \undefined \def \bsertitle#1{#1}\fi
\ifx \bsnm \undefined \def \bsnm#1{#1}\fi
\ifx \bsuffix \undefined \def \bsuffix#1{#1}\fi
\ifx \bparticle \undefined \def \bparticle#1{#1}\fi
\ifx \barticle \undefined \def \barticle#1{#1}\fi
\bibcommenthead
\ifx \bconfdate \undefined \def \bconfdate #1{#1}\fi
\ifx \botherref \undefined \def \botherref #1{#1}\fi
\ifx \url \undefined \def \url#1{\textsf{#1}}\fi
\ifx \bchapter \undefined \def \bchapter#1{#1}\fi
\ifx \bbook \undefined \def \bbook#1{#1}\fi
\ifx \bcomment \undefined \def \bcomment#1{#1}\fi
\ifx \oauthor \undefined \def \oauthor#1{#1}\fi
\ifx \citeauthoryear \undefined \def \citeauthoryear#1{#1}\fi
\ifx \endbibitem  \undefined \def \endbibitem {}\fi
\ifx \bconflocation  \undefined \def \bconflocation#1{#1}\fi
\ifx \arxivurl  \undefined \def \arxivurl#1{\textsf{#1}}\fi
\csname PreBibitemsHook\endcsname

\bibitem[\protect\citeauthoryear{Butler et~al.}{2018}]{para_shift}
\begin{barticle}
\bauthor{\bsnm{Butler}, \binits{K.T.}},
\bauthor{\bsnm{Davies}, \binits{D.W.}},
\bauthor{\bsnm{Cartwright}, \binits{H.}},
\bauthor{\bsnm{Isayev}, \binits{O.}},
\bauthor{\bsnm{Walsh}, \binits{A.}}:
\batitle{Machine learning for molecular and materials science}.
\bjtitle{Nature}
\bvolume{559}(\bissue{7715}),
\bfpage{547}--\blpage{555}
(\byear{2018})
\end{barticle}
\endbibitem

\bibitem[\protect\citeauthoryear{Tabor et~al.}{2018}]{material_energy}
\begin{barticle}
\bauthor{\bsnm{Tabor}, \binits{D.P.}},
\bauthor{\bsnm{Roch}, \binits{L.M.}},
\bauthor{\bsnm{Saikin}, \binits{S.K.}},
\bauthor{\bsnm{Kreisbeck}, \binits{C.}},
\bauthor{\bsnm{Sheberla}, \binits{D.}},
\bauthor{\bsnm{Montoya}, \binits{J.H.}},
\bauthor{\bsnm{Dwaraknath}, \binits{S.}},
\bauthor{\bsnm{Aykol}, \binits{M.}},
\bauthor{\bsnm{Ortiz}, \binits{C.}},
\bauthor{\bsnm{Tribukait}, \binits{H.}}, \betal:
\batitle{Accelerating the discovery of materials for clean energy in the era of smart automation}.
\bjtitle{Nature reviews materials}
\bvolume{3}(\bissue{5}),
\bfpage{5}--\blpage{20}
(\byear{2018})
\end{barticle}
\endbibitem

\bibitem[\protect\citeauthoryear{Agrawal and Choudhary}{2016}]{para_analysis}
\begin{botherref}
\oauthor{\bsnm{Agrawal}, \binits{A.}},
\oauthor{\bsnm{Choudhary}, \binits{A.}}:
Perspective: Materials informatics and big data: Realization of the “fourth paradigm” of science in materials science.
APL materials
\textbf{4}(5)
(2016)
\end{botherref}
\endbibitem

\bibitem[\protect\citeauthoryear{Curtarolo et~al.}{2013}]{mater_compute_weak}
\begin{barticle}
\bauthor{\bsnm{Curtarolo}, \binits{S.}},
\bauthor{\bsnm{Hart}, \binits{G.L.}},
\bauthor{\bsnm{Nardelli}, \binits{M.B.}},
\bauthor{\bsnm{Mingo}, \binits{N.}},
\bauthor{\bsnm{Sanvito}, \binits{S.}},
\bauthor{\bsnm{Levy}, \binits{O.}}:
\batitle{The high-throughput highway to computational materials design}.
\bjtitle{Nature materials}
\bvolume{12}(\bissue{3}),
\bfpage{191}--\blpage{201}
(\byear{2013})
\end{barticle}
\endbibitem

\bibitem[\protect\citeauthoryear{Greeley et~al.}{2006}]{expert_DFT_cata}
\begin{barticle}
\bauthor{\bsnm{Greeley}, \binits{J.}},
\bauthor{\bsnm{Jaramillo}, \binits{T.F.}},
\bauthor{\bsnm{Bonde}, \binits{J.}},
\bauthor{\bsnm{Chorkendorff}, \binits{I.}},
\bauthor{\bsnm{N{\o}rskov}, \binits{J.K.}}:
\batitle{Computational high-throughput screening of electrocatalytic materials for hydrogen evolution}.
\bjtitle{Nature materials}
\bvolume{5}(\bissue{11}),
\bfpage{909}--\blpage{913}
(\byear{2006})
\end{barticle}
\endbibitem

\bibitem[\protect\citeauthoryear{Xie and Grossman}{2018}]{CGCNN}
\begin{barticle}
\bauthor{\bsnm{Xie}, \binits{T.}},
\bauthor{\bsnm{Grossman}, \binits{J.C.}}:
\batitle{Crystal graph convolutional neural networks for an accurate and interpretable prediction of material properties}.
\bjtitle{Physical review letters}
\bvolume{120}(\bissue{14}),
\bfpage{145301}
(\byear{2018})
\end{barticle}
\endbibitem

\bibitem[\protect\citeauthoryear{Choudhary and DeCost}{2021}]{alignn}
\begin{barticle}
\bauthor{\bsnm{Choudhary}, \binits{K.}},
\bauthor{\bsnm{DeCost}, \binits{B.}}:
\batitle{Atomistic line graph neural network for improved materials property predictions}.
\bjtitle{npj Computational Materials}
\bvolume{7}(\bissue{1}),
\bfpage{185}
(\byear{2021})
\end{barticle}
\endbibitem

\bibitem[\protect\citeauthoryear{Chen et~al.}{2019}]{MEGNet}
\begin{barticle}
\bauthor{\bsnm{Chen}, \binits{C.}},
\bauthor{\bsnm{Ye}, \binits{W.}},
\bauthor{\bsnm{Zuo}, \binits{Y.}},
\bauthor{\bsnm{Zheng}, \binits{C.}},
\bauthor{\bsnm{Ong}, \binits{S.P.}}:
\batitle{Graph networks as a universal machine learning framework for molecules and crystals}.
\bjtitle{Chemistry of Materials}
\bvolume{31}(\bissue{9}),
\bfpage{3564}--\blpage{3572}
(\byear{2019})
\end{barticle}
\endbibitem

\bibitem[\protect\citeauthoryear{Gasteiger et~al.}{2020}]{DimeNet}
\begin{bchapter}
\bauthor{\bsnm{Gasteiger}, \binits{J.}},
\bauthor{\bsnm{Groß}, \binits{J.}},
\bauthor{\bsnm{Günnemann}, \binits{S.}}:
\bctitle{Directional message passing for molecular graphs}.
In: \bbtitle{International Conference on Learning Representations}
(\byear{2020})
\end{bchapter}
\endbibitem

\bibitem[\protect\citeauthoryear{Sch{\"u}tt et~al.}{2017}]{SchNet}
\begin{bchapter}
\bauthor{\bsnm{Sch{\"u}tt}, \binits{K.}},
\bauthor{\bsnm{Kindermans}, \binits{P.-J.}},
\bauthor{\bsnm{Sauceda~Felix}, \binits{H.E.}},
\bauthor{\bsnm{Chmiela}, \binits{S.}},
\bauthor{\bsnm{Tkatchenko}, \binits{A.}},
\bauthor{\bsnm{M{\"u}ller}, \binits{K.-R.}}:
\bctitle{Schnet: A continuous-filter convolutional neural network for modeling quantum interactions}.
In: \bbtitle{Advances in Neural Information Processing Systems},
vol. \bseriesno{30}
(\byear{2017})
\end{bchapter}
\endbibitem

\bibitem[\protect\citeauthoryear{Ruff et~al.}{2024}]{coGN_coNGN}
\begin{barticle}
\bauthor{\bsnm{Ruff}, \binits{R.}},
\bauthor{\bsnm{Reiser}, \binits{P.}},
\bauthor{\bsnm{St{\"u}hmer}, \binits{J.}},
\bauthor{\bsnm{Friederich}, \binits{P.}}:
\batitle{Connectivity optimized nested line graph networks for crystal structures}.
\bjtitle{Digital Discovery}
\bvolume{3}(\bissue{3}),
\bfpage{594}--\blpage{601}
(\byear{2024})
\end{barticle}
\endbibitem

\bibitem[\protect\citeauthoryear{Yang et~al.}{2024}]{mattersim}
\begin{botherref}
\oauthor{\bsnm{Yang}, \binits{H.}},
\oauthor{\bsnm{Hu}, \binits{C.}},
\oauthor{\bsnm{Zhou}, \binits{Y.}},
\oauthor{\bsnm{Liu}, \binits{X.}},
\oauthor{\bsnm{Shi}, \binits{Y.}},
\oauthor{\bsnm{Li}, \binits{J.}},
\oauthor{\bsnm{Li}, \binits{G.}},
\oauthor{\bsnm{Chen}, \binits{Z.}},
\oauthor{\bsnm{Chen}, \binits{S.}},
\oauthor{\bsnm{Zeni}, \binits{C.}}, et al.:
Mattersim: A deep learning atomistic model across elements, temperatures and pressures.
arXiv preprint arXiv:2405.04967
(2024)
\end{botherref}
\endbibitem

\bibitem[\protect\citeauthoryear{Merchant et~al.}{2023}]{genome}
\begin{barticle}
\bauthor{\bsnm{Merchant}, \binits{A.}},
\bauthor{\bsnm{Batzner}, \binits{S.}},
\bauthor{\bsnm{Schoenholz}, \binits{S.S.}},
\bauthor{\bsnm{Aykol}, \binits{M.}},
\bauthor{\bsnm{Cheon}, \binits{G.}},
\bauthor{\bsnm{Cubuk}, \binits{E.D.}}:
\batitle{Scaling deep learning for materials discovery}.
\bjtitle{Nature}
\bvolume{624}(\bissue{7990}),
\bfpage{80}--\blpage{85}
(\byear{2023})
\end{barticle}
\endbibitem

\bibitem[\protect\citeauthoryear{Li et~al.}{2023}]{GNO}
\begin{barticle}
\bauthor{\bsnm{Li}, \binits{Z.}},
\bauthor{\bsnm{Kovachki}, \binits{N.}},
\bauthor{\bsnm{Choy}, \binits{C.}},
\bauthor{\bsnm{Li}, \binits{B.}},
\bauthor{\bsnm{Kossaifi}, \binits{J.}},
\bauthor{\bsnm{Otta}, \binits{S.}},
\bauthor{\bsnm{Nabian}, \binits{M.A.}},
\bauthor{\bsnm{Stadler}, \binits{M.}},
\bauthor{\bsnm{Hundt}, \binits{C.}},
\bauthor{\bsnm{Azizzadenesheli}, \binits{K.}}, \betal:
\batitle{Geometry-informed neural operator for large-scale 3d pdes}.
\bjtitle{Advances in Neural Information Processing Systems}
\bvolume{36},
\bfpage{35836}--\blpage{35854}
(\byear{2023})
\end{barticle}
\endbibitem

\bibitem[\protect\citeauthoryear{Liao et~al.}{2023}]{EquiformerV2}
\begin{bchapter}
\bauthor{\bsnm{Liao}, \binits{Y.-L.}},
\bauthor{\bsnm{Wood}, \binits{B.M.}},
\bauthor{\bsnm{Das}, \binits{A.}},
\bauthor{\bsnm{Smidt}, \binits{T.}}:
\bctitle{Equiformerv2: Improved equivariant transformer for scaling to higher-degree representations}.
In: \bbtitle{The Twelfth International Conference on Learning Representations}
(\byear{2023})
\end{bchapter}
\endbibitem

\bibitem[\protect\citeauthoryear{Batzner et~al.}{2022}]{NequIP}
\begin{barticle}
\bauthor{\bsnm{Batzner}, \binits{S.}},
\bauthor{\bsnm{Musaelian}, \binits{A.}},
\bauthor{\bsnm{Sun}, \binits{L.}},
\bauthor{\bsnm{Geiger}, \binits{M.}},
\bauthor{\bsnm{Mailoa}, \binits{J.P.}},
\bauthor{\bsnm{Kornbluth}, \binits{M.}},
\bauthor{\bsnm{Molinari}, \binits{N.}},
\bauthor{\bsnm{Smidt}, \binits{T.E.}},
\bauthor{\bsnm{Kozinsky}, \binits{B.}}:
\batitle{E (3)-equivariant graph neural networks for data-efficient and accurate interatomic potentials}.
\bjtitle{Nature communications}
\bvolume{13}(\bissue{1}),
\bfpage{2453}
(\byear{2022})
\end{barticle}
\endbibitem

\bibitem[\protect\citeauthoryear{Musaelian et~al.}{2023}]{allergo}
\begin{barticle}
\bauthor{\bsnm{Musaelian}, \binits{A.}},
\bauthor{\bsnm{Batzner}, \binits{S.}},
\bauthor{\bsnm{Johansson}, \binits{A.}},
\bauthor{\bsnm{Sun}, \binits{L.}},
\bauthor{\bsnm{Owen}, \binits{C.J.}},
\bauthor{\bsnm{Kornbluth}, \binits{M.}},
\bauthor{\bsnm{Kozinsky}, \binits{B.}}:
\batitle{Learning local equivariant representations for large-scale atomistic dynamics}.
\bjtitle{Nature Communications}
\bvolume{14}(\bissue{1}),
\bfpage{579}
(\byear{2023})
\end{barticle}
\endbibitem

\bibitem[\protect\citeauthoryear{Batatia et~al.}{2022}]{Mace}
\begin{barticle}
\bauthor{\bsnm{Batatia}, \binits{I.}},
\bauthor{\bsnm{Kovacs}, \binits{D.P.}},
\bauthor{\bsnm{Simm}, \binits{G.}},
\bauthor{\bsnm{Ortner}, \binits{C.}},
\bauthor{\bsnm{Cs{\'a}nyi}, \binits{G.}}:
\batitle{Mace: Higher order equivariant message passing neural networks for fast and accurate force fields}.
\bjtitle{Advances in neural information processing systems}
\bvolume{35},
\bfpage{11423}--\blpage{11436}
(\byear{2022})
\end{barticle}
\endbibitem

\bibitem[\protect\citeauthoryear{Batatia et~al.}{2025}]{Mace-MP-0}
\begin{botherref}
\oauthor{\bsnm{Batatia}, \binits{I.}},
\oauthor{\bsnm{Benner}, \binits{P.}},
\oauthor{\bsnm{Chiang}, \binits{Y.}},
\oauthor{\bsnm{Elena}, \binits{A.M.}},
\oauthor{\bsnm{Kov{\'a}cs}, \binits{D.P.}},
\oauthor{\bsnm{Riebesell}, \binits{J.}},
\oauthor{\bsnm{Advincula}, \binits{X.R.}},
\oauthor{\bsnm{Asta}, \binits{M.}},
\oauthor{\bsnm{Avaylon}, \binits{M.}},
\oauthor{\bsnm{Baldwin}, \binits{W.J.}}, et al.:
A foundation model for atomistic materials chemistry.
The Journal of chemical physics
\textbf{163}(18)
(2025)
\end{botherref}
\endbibitem

\bibitem[\protect\citeauthoryear{Zhang et~al.}{2024}]{DPA-2}
\begin{barticle}
\bauthor{\bsnm{Zhang}, \binits{D.}},
\bauthor{\bsnm{Liu}, \binits{X.}},
\bauthor{\bsnm{Zhang}, \binits{X.}},
\bauthor{\bsnm{Zhang}, \binits{C.}},
\bauthor{\bsnm{Cai}, \binits{C.}},
\bauthor{\bsnm{Bi}, \binits{H.}},
\bauthor{\bsnm{Du}, \binits{Y.}},
\bauthor{\bsnm{Qin}, \binits{X.}},
\bauthor{\bsnm{Peng}, \binits{A.}},
\bauthor{\bsnm{Huang}, \binits{J.}}, \betal:
\batitle{Dpa-2: a large atomic model as a multi-task learner}.
\bjtitle{npj Computational Materials}
\bvolume{10}(\bissue{1}),
\bfpage{293}
(\byear{2024})
\end{barticle}
\endbibitem

\bibitem[\protect\citeauthoryear{Xie et~al.}{2022}]{cdvae}
\begin{bchapter}
\bauthor{\bsnm{Xie}, \binits{T.}},
\bauthor{\bsnm{Fu}, \binits{X.}},
\bauthor{\bsnm{Ganea}, \binits{O.-E.}},
\bauthor{\bsnm{Barzilay}, \binits{R.}},
\bauthor{\bsnm{Jaakkola}, \binits{T.S.}}:
\bctitle{Crystal diffusion variational autoencoder for periodic material generation}.
In: \bbtitle{International Conference on Learning Representations}
(\byear{2022})
\end{bchapter}
\endbibitem

\bibitem[\protect\citeauthoryear{Jiao et~al.}{2023}]{diffcsp}
\begin{barticle}
\bauthor{\bsnm{Jiao}, \binits{R.}},
\bauthor{\bsnm{Huang}, \binits{W.}},
\bauthor{\bsnm{Lin}, \binits{P.}},
\bauthor{\bsnm{Han}, \binits{J.}},
\bauthor{\bsnm{Chen}, \binits{P.}},
\bauthor{\bsnm{Lu}, \binits{Y.}},
\bauthor{\bsnm{Liu}, \binits{Y.}}:
\batitle{Crystal structure prediction by joint equivariant diffusion}.
\bjtitle{Advances in Neural Information Processing Systems}
\bvolume{36},
\bfpage{17464}--\blpage{17497}
(\byear{2023})
\end{barticle}
\endbibitem

\bibitem[\protect\citeauthoryear{Miller et~al.}{2024}]{FlowMM}
\begin{bchapter}
\bauthor{\bsnm{Miller}, \binits{B.K.}},
\bauthor{\bsnm{Chen}, \binits{R.T.}},
\bauthor{\bsnm{Sriram}, \binits{A.}},
\bauthor{\bsnm{Wood}, \binits{B.M.}}:
\bctitle{Flowmm: Generating materials with riemannian flow matching}.
In: \bbtitle{International Conference on Machine Learning},
pp. \bfpage{35664}--\blpage{35686}
(\byear{2024}).
\bcomment{PMLR}
\end{bchapter}
\endbibitem

\bibitem[\protect\citeauthoryear{Luo et~al.}{2025}]{CrysFlow}
\begin{barticle}
\bauthor{\bsnm{Luo}, \binits{X.}},
\bauthor{\bsnm{Wang}, \binits{Z.}},
\bauthor{\bsnm{Wang}, \binits{Q.}},
\bauthor{\bsnm{Shao}, \binits{X.}},
\bauthor{\bsnm{Lv}, \binits{J.}},
\bauthor{\bsnm{Wang}, \binits{L.}},
\bauthor{\bsnm{Wang}, \binits{Y.}},
\bauthor{\bsnm{Ma}, \binits{Y.}}:
\batitle{Crystalflow: a flow-based generative model for crystalline materials}.
\bjtitle{Nature Communications}
\bvolume{16}(\bissue{1}),
\bfpage{9267}
(\byear{2025})
\end{barticle}
\endbibitem

\bibitem[\protect\citeauthoryear{Wu et~al.}{2025}]{CrysBFN}
\begin{bchapter}
\bauthor{\bsnm{Wu}, \binits{H.}},
\bauthor{\bsnm{Song}, \binits{Y.}},
\bauthor{\bsnm{Gong}, \binits{J.}},
\bauthor{\bsnm{Cao}, \binits{Z.}},
\bauthor{\bsnm{Ouyang}, \binits{Y.}},
\bauthor{\bsnm{Zhang}, \binits{J.}},
\bauthor{\bsnm{Zhou}, \binits{H.}},
\bauthor{\bsnm{Ma}, \binits{W.-Y.}},
\bauthor{\bsnm{Liu}, \binits{J.}}:
\bctitle{A periodic bayesian flow for material generation}.
In: \bbtitle{International Conference on Learning Representations}
(\byear{2025})
\end{bchapter}
\endbibitem

\bibitem[\protect\citeauthoryear{Zeni et~al.}{2025}]{mattergen}
\begin{barticle}
\bauthor{\bsnm{Zeni}, \binits{C.}},
\bauthor{\bsnm{Pinsler}, \binits{R.}},
\bauthor{\bsnm{Z{\"u}gner}, \binits{D.}},
\bauthor{\bsnm{Fowler}, \binits{A.}},
\bauthor{\bsnm{Horton}, \binits{M.}},
\bauthor{\bsnm{Fu}, \binits{X.}},
\bauthor{\bsnm{Wang}, \binits{Z.}},
\bauthor{\bsnm{Shysheya}, \binits{A.}},
\bauthor{\bsnm{Crabb{\'e}}, \binits{J.}},
\bauthor{\bsnm{Ueda}, \binits{S.}}, \betal:
\batitle{A generative model for inorganic materials design}.
\bjtitle{Nature}
\bvolume{639}(\bissue{8055}),
\bfpage{624}--\blpage{632}
(\byear{2025})
\end{barticle}
\endbibitem

\bibitem[\protect\citeauthoryear{Szymanski et~al.}{2023}]{A-lab}
\begin{barticle}
\bauthor{\bsnm{Szymanski}, \binits{N.J.}},
\bauthor{\bsnm{Rendy}, \binits{B.}},
\bauthor{\bsnm{Fei}, \binits{Y.}},
\bauthor{\bsnm{Kumar}, \binits{R.E.}},
\bauthor{\bsnm{He}, \binits{T.}},
\bauthor{\bsnm{Milsted}, \binits{D.}},
\bauthor{\bsnm{McDermott}, \binits{M.J.}},
\bauthor{\bsnm{Gallant}, \binits{M.}},
\bauthor{\bsnm{Cubuk}, \binits{E.D.}},
\bauthor{\bsnm{Merchant}, \binits{A.}}, \betal:
\batitle{An autonomous laboratory for the accelerated synthesis of inorganic materials}.
\bjtitle{Nature}
\bvolume{624}(\bissue{7990}),
\bfpage{86}
(\byear{2023})
\end{barticle}
\endbibitem

\bibitem[\protect\citeauthoryear{Castelvecchi}{2024}]{Nature_AI_sci}
\begin{barticle}
\bauthor{\bsnm{Castelvecchi}, \binits{D.}}:
\batitle{Researchers built an ‘ai scientist’—what can it do}.
\bjtitle{Nature}
\bvolume{633}(\bissue{8029}),
\bfpage{266}--\blpage{266}
(\byear{2024})
\end{barticle}
\endbibitem

\bibitem[\protect\citeauthoryear{Lu et~al.}{2026}]{lu2026towards}
\begin{barticle}
\bauthor{\bsnm{Lu}, \binits{C.}},
\bauthor{\bsnm{Lu}, \binits{C.}},
\bauthor{\bsnm{Lange}, \binits{R.T.}},
\bauthor{\bsnm{Yamada}, \binits{Y.}},
\bauthor{\bsnm{Hu}, \binits{S.}},
\bauthor{\bsnm{Foerster}, \binits{J.}},
\bauthor{\bsnm{Ha}, \binits{D.}},
\bauthor{\bsnm{Clune}, \binits{J.}}:
\batitle{Towards end-to-end automation of ai research}.
\bjtitle{Nature}
\bvolume{651}(\bissue{8107}),
\bfpage{914}--\blpage{919}
(\byear{2026})
\end{barticle}
\endbibitem

\bibitem[\protect\citeauthoryear{Wei et~al.}{2025}]{agenticscience}
\begin{botherref}
\oauthor{\bsnm{Wei}, \binits{J.}},
\oauthor{\bsnm{Yang}, \binits{Y.}},
\oauthor{\bsnm{Zhang}, \binits{X.}},
\oauthor{\bsnm{Chen}, \binits{Y.}},
\oauthor{\bsnm{Zhuang}, \binits{X.}},
\oauthor{\bsnm{Gao}, \binits{Z.}},
\oauthor{\bsnm{Zhou}, \binits{D.}},
\oauthor{\bsnm{Wang}, \binits{G.}},
\oauthor{\bsnm{Gao}, \binits{Z.}},
\oauthor{\bsnm{Cao}, \binits{J.}}, et al.:
From ai for science to agentic science: A survey on autonomous scientific discovery.
arXiv preprint arXiv:2508.14111
(2025)
\end{botherref}
\endbibitem

\bibitem[\protect\citeauthoryear{Keimer et~al.}{2015}]{chemical_complexity_1}
\begin{barticle}
\bauthor{\bsnm{Keimer}, \binits{B.}},
\bauthor{\bsnm{Kivelson}, \binits{S.A.}},
\bauthor{\bsnm{Norman}, \binits{M.R.}},
\bauthor{\bsnm{Uchida}, \binits{S.}},
\bauthor{\bsnm{Zaanen}, \binits{J.}}:
\batitle{From quantum matter to high-temperature superconductivity in copper oxides}.
\bjtitle{Nature}
\bvolume{518}(\bissue{7538}),
\bfpage{179}--\blpage{186}
(\byear{2015})
\end{barticle}
\endbibitem

\bibitem[\protect\citeauthoryear{Pickard et~al.}{2020}]{chemical_complexity_2}
\begin{barticle}
\bauthor{\bsnm{Pickard}, \binits{C.J.}},
\bauthor{\bsnm{Errea}, \binits{I.}},
\bauthor{\bsnm{Eremets}, \binits{M.I.}}:
\batitle{Superconducting hydrides under pressure}.
\bjtitle{Annual Review of Condensed Matter Physics}
\bvolume{11}(\bissue{1}),
\bfpage{57}--\blpage{76}
(\byear{2020})
\end{barticle}
\endbibitem

\bibitem[\protect\citeauthoryear{Stanev et~al.}{2018}]{data_scarcity}
\begin{barticle}
\bauthor{\bsnm{Stanev}, \binits{V.}},
\bauthor{\bsnm{Oses}, \binits{C.}},
\bauthor{\bsnm{Kusne}, \binits{A.G.}},
\bauthor{\bsnm{Rodriguez}, \binits{E.}},
\bauthor{\bsnm{Paglione}, \binits{J.}},
\bauthor{\bsnm{Curtarolo}, \binits{S.}},
\bauthor{\bsnm{Takeuchi}, \binits{I.}}:
\batitle{Machine learning modeling of superconducting critical temperature}.
\bjtitle{npj Computational Materials}
\bvolume{4}(\bissue{1}),
\bfpage{29}
(\byear{2018})
\end{barticle}
\endbibitem

\bibitem[\protect\citeauthoryear{Chen et~al.}{2024}]{sodnet}
\begin{barticle}
\bauthor{\bsnm{Chen}, \binits{P.}},
\bauthor{\bsnm{Peng}, \binits{L.}},
\bauthor{\bsnm{Jiao}, \binits{R.}},
\bauthor{\bsnm{Mo}, \binits{Q.}},
\bauthor{\bsnm{Wang}, \binits{Z.}},
\bauthor{\bsnm{Huang}, \binits{W.}},
\bauthor{\bsnm{Liu}, \binits{Y.}},
\bauthor{\bsnm{Lu}, \binits{Y.}}:
\batitle{Learning superconductivity from ordered and disordered material structures}.
\bjtitle{Advances in Neural Information Processing Systems}
\bvolume{37},
\bfpage{108902}--\blpage{108928}
(\byear{2024})
\end{barticle}
\endbibitem

\bibitem[\protect\citeauthoryear{Group}{}]{supercon}
\begin{botherref}
\oauthor{\bsnm{Group}, \binits{M.D.}}:
MDR SuperCon Datasheet Ver.220808.
National Institute for Materials Science
\end{botherref}
\endbibitem

\bibitem[\protect\citeauthoryear{Blokhin and Villars}{2022}]{MPDS}
\begin{botherref}
\oauthor{\bsnm{Blokhin}, \binits{E.}},
\oauthor{\bsnm{Villars}, \binits{P.}}:
Materials Platform for Data Science.
MPDS. Accessed on
(2022)
\end{botherref}
\endbibitem

\bibitem[\protect\citeauthoryear{Villars et~al.}{2004}]{pauling}
\begin{barticle}
\bauthor{\bsnm{Villars}, \binits{P.}},
\bauthor{\bsnm{Berndt}, \binits{M.}},
\bauthor{\bsnm{Brandenburg}, \binits{K.}},
\bauthor{\bsnm{Cenzual}, \binits{K.}},
\bauthor{\bsnm{Daams}, \binits{J.}},
\bauthor{\bsnm{Hulliger}, \binits{F.}},
\bauthor{\bsnm{Massalski}, \binits{T.}},
\bauthor{\bsnm{Okamoto}, \binits{H.}},
\bauthor{\bsnm{Osaki}, \binits{K.}},
\bauthor{\bsnm{Prince}, \binits{A.}}, \betal:
\batitle{The pauling file}.
\bjtitle{Journal of Alloys and Compounds}
\bvolume{367}(\bissue{1-2}),
\bfpage{293}--\blpage{297}
(\byear{2004})
\end{barticle}
\endbibitem

\bibitem[\protect\citeauthoryear{Wang et~al.}{2025}]{kagome}
\begin{barticle}
\bauthor{\bsnm{Wang}, \binits{L.}},
\bauthor{\bsnm{Li}, \binits{Q.}},
\bauthor{\bsnm{Ma}, \binits{K.}},
\bauthor{\bsnm{Yu}, \binits{Y.}},
\bauthor{\bsnm{Jin}, \binits{S.}},
\bauthor{\bsnm{Chen}, \binits{X.}}:
\batitle{Database of superconductors with kagome lattice by high-throughput screening}.
\bjtitle{Chinese Physics B}
\bvolume{34}(\bissue{10}),
\bfpage{106101}
(\byear{2025})
\end{barticle}
\endbibitem

\bibitem[\protect\citeauthoryear{Biamonte et~al.}{2017}]{qm9}
\begin{barticle}
\bauthor{\bsnm{Biamonte}, \binits{J.}},
\bauthor{\bsnm{Wittek}, \binits{P.}},
\bauthor{\bsnm{Pancotti}, \binits{N.}},
\bauthor{\bsnm{Rebentrost}, \binits{P.}},
\bauthor{\bsnm{Wiebe}, \binits{N.}},
\bauthor{\bsnm{Lloyd}, \binits{S.}}:
\batitle{Quantum machine learning}.
\bjtitle{Nature}
\bvolume{549}(\bissue{7671}),
\bfpage{195}--\blpage{202}
(\byear{2017})
\end{barticle}
\endbibitem

\bibitem[\protect\citeauthoryear{Dunn et~al.}{2020}]{matbench}
\begin{barticle}
\bauthor{\bsnm{Dunn}, \binits{A.}},
\bauthor{\bsnm{Wang}, \binits{Q.}},
\bauthor{\bsnm{Ganose}, \binits{A.}},
\bauthor{\bsnm{Dopp}, \binits{D.}},
\bauthor{\bsnm{Jain}, \binits{A.}}:
\batitle{Benchmarking materials property prediction methods: the matbench test set and automatminer reference algorithm}.
\bjtitle{npj Computational Materials}
\bvolume{6}(\bissue{1}),
\bfpage{138}
(\byear{2020})
\end{barticle}
\endbibitem

\bibitem[\protect\citeauthoryear{Aykent and Xia}{2025}]{gotennet}
\begin{bchapter}
\bauthor{\bsnm{Aykent}, \binits{S.}},
\bauthor{\bsnm{Xia}, \binits{T.}}:
\bctitle{Gotennet: Rethinking efficient 3d equivariant graph neural networks}.
In: \bbtitle{The Thirteenth International Conference on Learning Representations}
(\byear{2025})
\end{bchapter}
\endbibitem

\bibitem[\protect\citeauthoryear{Choudhary et~al.}{2020}]{jarvis}
\begin{barticle}
\bauthor{\bsnm{Choudhary}, \binits{K.}},
\bauthor{\bsnm{Garrity}, \binits{K.F.}},
\bauthor{\bsnm{Reid}, \binits{A.C.}},
\bauthor{\bsnm{DeCost}, \binits{B.}},
\bauthor{\bsnm{Biacchi}, \binits{A.J.}},
\bauthor{\bsnm{Hight~Walker}, \binits{A.R.}},
\bauthor{\bsnm{Trautt}, \binits{Z.}},
\bauthor{\bsnm{Hattrick-Simpers}, \binits{J.}},
\bauthor{\bsnm{Kusne}, \binits{A.G.}},
\bauthor{\bsnm{Centrone}, \binits{A.}}, \betal:
\batitle{The joint automated repository for various integrated simulations (jarvis) for data-driven materials design}.
\bjtitle{npj computational materials}
\bvolume{6}(\bissue{1}),
\bfpage{173}
(\byear{2020})
\end{barticle}
\endbibitem

\bibitem[\protect\citeauthoryear{Cerqueira et~al.}{2024}]{dfttc}
\begin{barticle}
\bauthor{\bsnm{Cerqueira}, \binits{T.F.}},
\bauthor{\bsnm{Sanna}, \binits{A.}},
\bauthor{\bsnm{Marques}, \binits{M.A.}}:
\batitle{Sampling the materials space for conventional superconducting compounds}.
\bjtitle{Advanced Materials}
\bvolume{36}(\bissue{1}),
\bfpage{2307085}
(\byear{2024})
\end{barticle}
\endbibitem

\bibitem[\protect\citeauthoryear{Yan et~al.}{2022}]{matformer}
\begin{barticle}
\bauthor{\bsnm{Yan}, \binits{K.}},
\bauthor{\bsnm{Liu}, \binits{Y.}},
\bauthor{\bsnm{Lin}, \binits{Y.}},
\bauthor{\bsnm{Ji}, \binits{S.}}:
\batitle{Periodic graph transformers for crystal material property prediction}.
\bjtitle{Advances in Neural Information Processing Systems}
\bvolume{35},
\bfpage{15066}--\blpage{15080}
(\byear{2022})
\end{barticle}
\endbibitem

\bibitem[\protect\citeauthoryear{Lin et~al.}{2023}]{potnet}
\begin{bchapter}
\bauthor{\bsnm{Lin}, \binits{Y.}},
\bauthor{\bsnm{Yan}, \binits{K.}},
\bauthor{\bsnm{Luo}, \binits{Y.}},
\bauthor{\bsnm{Liu}, \binits{Y.}},
\bauthor{\bsnm{Qian}, \binits{X.}},
\bauthor{\bsnm{Ji}, \binits{S.}}:
\bctitle{Efficient approximations of complete interatomic potentials for crystal property prediction}.
In: \bbtitle{International Conference on Machine Learning},
pp. \bfpage{21260}--\blpage{21287}
(\byear{2023}).
\bcomment{PMLR}
\end{bchapter}
\endbibitem

\bibitem[\protect\citeauthoryear{Cenzual et~al.}{1986}]{Zr21Re25Hf21Re25}
\begin{barticle}
\bauthor{\bsnm{Cenzual}, \binits{K.}},
\bauthor{\bsnm{Parth{\'e}}, \binits{E.}},
\bauthor{\bsnm{Waterstrat}, \binits{R.}}:
\batitle{Zr21re25, a new rhombohedral structure type containing 12 {\aa}-thick infinite mgzn2 (laves)-type columns}.
\bjtitle{Crystal Structure Communications}
\bvolume{42}(\bissue{3}),
\bfpage{261}--\blpage{266}
(\byear{1986})
\end{barticle}
\endbibitem

\bibitem[\protect\citeauthoryear{Schreiner et~al.}{2022}]{Transition1x}
\begin{barticle}
\bauthor{\bsnm{Schreiner}, \binits{M.}},
\bauthor{\bsnm{Bhowmik}, \binits{A.}},
\bauthor{\bsnm{Vegge}, \binits{T.}},
\bauthor{\bsnm{Busk}, \binits{J.}},
\bauthor{\bsnm{Winther}, \binits{O.}}:
\batitle{Transition1x-a dataset for building generalizable reactive machine learning potentials}.
\bjtitle{Scientific Data}
\bvolume{9}(\bissue{1}),
\bfpage{779}
(\byear{2022})
\end{barticle}
\endbibitem

\bibitem[\protect\citeauthoryear{Smith et~al.}{2020}]{ani-1x}
\begin{barticle}
\bauthor{\bsnm{Smith}, \binits{J.S.}},
\bauthor{\bsnm{Zubatyuk}, \binits{R.}},
\bauthor{\bsnm{Nebgen}, \binits{B.}},
\bauthor{\bsnm{Lubbers}, \binits{N.}},
\bauthor{\bsnm{Barros}, \binits{K.}},
\bauthor{\bsnm{Roitberg}, \binits{A.E.}},
\bauthor{\bsnm{Isayev}, \binits{O.}},
\bauthor{\bsnm{Tretiak}, \binits{S.}}:
\batitle{The ani-1ccx and ani-1x data sets, coupled-cluster and density functional theory properties for molecules}.
\bjtitle{Scientific data}
\bvolume{7}(\bissue{1}),
\bfpage{134}
(\byear{2020})
\end{barticle}
\endbibitem

\bibitem[\protect\citeauthoryear{Barroso-Luque et~al.}{2024}]{OMAT-24}
\begin{botherref}
\oauthor{\bsnm{Barroso-Luque}, \binits{L.}},
\oauthor{\bsnm{Shuaibi}, \binits{M.}},
\oauthor{\bsnm{Fu}, \binits{X.}},
\oauthor{\bsnm{Wood}, \binits{B.M.}},
\oauthor{\bsnm{Dzamba}, \binits{M.}},
\oauthor{\bsnm{Gao}, \binits{M.}},
\oauthor{\bsnm{Rizvi}, \binits{A.}},
\oauthor{\bsnm{Zitnick}, \binits{C.L.}},
\oauthor{\bsnm{Ulissi}, \binits{Z.W.}}:
Open materials 2024 (omat24) inorganic materials dataset and models.
arXiv preprint arXiv:2410.12771
(2024)
\end{botherref}
\endbibitem

\bibitem[\protect\citeauthoryear{Scheidgen et~al.}{2023}]{nomad}
\begin{barticle}
\bauthor{\bsnm{Scheidgen}, \binits{M.}},
\bauthor{\bsnm{Himanen}, \binits{L.}},
\bauthor{\bsnm{Ladines}, \binits{A.N.}},
\bauthor{\bsnm{Sikter}, \binits{D.}},
\bauthor{\bsnm{Nakhaee}, \binits{M.}},
\bauthor{\bsnm{Fekete}, \binits{{\'A}.}},
\bauthor{\bsnm{Chang}, \binits{T.}},
\bauthor{\bsnm{Golparvar}, \binits{A.}},
\bauthor{\bsnm{M{\'a}rquez}, \binits{J.A.}},
\bauthor{\bsnm{Brockhauser}, \binits{S.}}, \betal:
\batitle{Nomad: A distributed web-based platform for managing materials science research data}.
\bjtitle{Journal of Open Source Software}
\bvolume{8}(\bissue{90}),
\bfpage{5388}
(\byear{2023})
\end{barticle}
\endbibitem

\bibitem[\protect\citeauthoryear{Schmidt et~al.}{2024}]{Alex}
\begin{barticle}
\bauthor{\bsnm{Schmidt}, \binits{J.}},
\bauthor{\bsnm{Cerqueira}, \binits{T.F.}},
\bauthor{\bsnm{Romero}, \binits{A.H.}},
\bauthor{\bsnm{Loew}, \binits{A.}},
\bauthor{\bsnm{J{\"a}ger}, \binits{F.}},
\bauthor{\bsnm{Wang}, \binits{H.-C.}},
\bauthor{\bsnm{Botti}, \binits{S.}},
\bauthor{\bsnm{Marques}, \binits{M.A.}}:
\batitle{Improving machine-learning models in materials science through large datasets}.
\bjtitle{Materials Today Physics}
\bvolume{48},
\bfpage{101560}
(\byear{2024})
\end{barticle}
\endbibitem

\bibitem[\protect\citeauthoryear{Saal et~al.}{2013}]{oqmd}
\begin{barticle}
\bauthor{\bsnm{Saal}, \binits{J.E.}},
\bauthor{\bsnm{Kirklin}, \binits{S.}},
\bauthor{\bsnm{Aykol}, \binits{M.}},
\bauthor{\bsnm{Meredig}, \binits{B.}},
\bauthor{\bsnm{Wolverton}, \binits{C.}}:
\batitle{Materials design and discovery with high-throughput density functional theory: the open quantum materials database (oqmd)}.
\bjtitle{Jom}
\bvolume{65}(\bissue{11}),
\bfpage{1501}--\blpage{1509}
(\byear{2013})
\end{barticle}
\endbibitem

\bibitem[\protect\citeauthoryear{Chen and Ong}{2022}]{M3GNet}
\begin{barticle}
\bauthor{\bsnm{Chen}, \binits{C.}},
\bauthor{\bsnm{Ong}, \binits{S.P.}}:
\batitle{A universal graph deep learning interatomic potential for the periodic table}.
\bjtitle{Nature Computational Science}
\bvolume{2}(\bissue{11}),
\bfpage{718}--\blpage{728}
(\byear{2022})
\end{barticle}
\endbibitem

\bibitem[\protect\citeauthoryear{Garrity and Choudhary}{2023}]{jarvis_qetb}
\begin{barticle}
\bauthor{\bsnm{Garrity}, \binits{K.F.}},
\bauthor{\bsnm{Choudhary}, \binits{K.}}:
\batitle{Fast and accurate prediction of material properties with three-body tight-binding model for the periodic table}.
\bjtitle{Physical review materials}
\bvolume{7}(\bissue{4}),
\bfpage{044603}
(\byear{2023})
\end{barticle}
\endbibitem

\bibitem[\protect\citeauthoryear{Passaro and Zitnick}{2023}]{escn}
\begin{bchapter}
\bauthor{\bsnm{Passaro}, \binits{S.}},
\bauthor{\bsnm{Zitnick}, \binits{C.L.}}:
\bctitle{Reducing so (3) convolutions to so (2) for efficient equivariant gnns}.
In: \bbtitle{International Conference on Machine Learning},
pp. \bfpage{27420}--\blpage{27438}
(\byear{2023}).
\bcomment{PMLR}
\end{bchapter}
\endbibitem

\bibitem[\protect\citeauthoryear{Rong et~al.}{2020}]{grover}
\begin{bchapter}
\bauthor{\bsnm{Rong}, \binits{Y.}},
\bauthor{\bsnm{Bian}, \binits{Y.}},
\bauthor{\bsnm{Xu}, \binits{T.}},
\bauthor{\bsnm{Xie}, \binits{W.}},
\bauthor{\bsnm{WEI}, \binits{Y.}},
\bauthor{\bsnm{Huang}, \binits{W.}},
\bauthor{\bsnm{Huang}, \binits{J.}}:
\bctitle{Self-supervised graph transformer on large-scale molecular data}.
In: \beditor{\bsnm{Larochelle}, \binits{H.}},
\beditor{\bsnm{Ranzato}, \binits{M.}},
\beditor{\bsnm{Hadsell}, \binits{R.}},
\beditor{\bsnm{Balcan}, \binits{M.F.}},
\beditor{\bsnm{Lin}, \binits{H.}} (eds.)
\bbtitle{Advances in Neural Information Processing Systems},
vol. \bseriesno{33},
pp. \bfpage{12559}--\blpage{12571}.
\bpublisher{Curran Associates, Inc.},
\blocation{Red Hook, NY}
(\byear{2020})
\end{bchapter}
\endbibitem

\bibitem[\protect\citeauthoryear{He et~al.}{2016}]{resnet}
\begin{bchapter}
\bauthor{\bsnm{He}, \binits{K.}},
\bauthor{\bsnm{Zhang}, \binits{X.}},
\bauthor{\bsnm{Ren}, \binits{S.}},
\bauthor{\bsnm{Sun}, \binits{J.}}:
\bctitle{Deep residual learning for image recognition}.
In: \bbtitle{Proceedings of the IEEE Conference on Computer Vision and Pattern Recognition},
pp. \bfpage{770}--\blpage{778}
(\byear{2016})
\end{bchapter}
\endbibitem

\bibitem[\protect\citeauthoryear{Cohen et~al.}{2018}]{sphcnn}
\begin{bchapter}
\bauthor{\bsnm{Cohen}, \binits{T.S.}},
\bauthor{\bsnm{Geiger}, \binits{M.}},
\bauthor{\bsnm{Köhler}, \binits{J.}},
\bauthor{\bsnm{Welling}, \binits{M.}}:
\bctitle{Spherical {CNN}s}.
In: \bbtitle{International Conference on Learning Representations}
(\byear{2018})
\end{bchapter}
\endbibitem

\bibitem[\protect\citeauthoryear{{OpenClaw Contributors}}{2026}]{openclaw}
\begin{botherref}
\oauthor{\bsnm{{OpenClaw Contributors}}}:
OpenClaw.
\url{https://github.com/openclaw/openclaw}
(2026)
\end{botherref}
\endbibitem

\bibitem[\protect\citeauthoryear{Ong et~al.}{2013}]{pymatgen}
\begin{barticle}
\bauthor{\bsnm{Ong}, \binits{S.P.}},
\bauthor{\bsnm{Richards}, \binits{W.D.}},
\bauthor{\bsnm{Jain}, \binits{A.}},
\bauthor{\bsnm{Hautier}, \binits{G.}},
\bauthor{\bsnm{Kocher}, \binits{M.}},
\bauthor{\bsnm{Cholia}, \binits{S.}},
\bauthor{\bsnm{Gunter}, \binits{D.}},
\bauthor{\bsnm{Chevrier}, \binits{V.L.}},
\bauthor{\bsnm{Persson}, \binits{K.A.}},
\bauthor{\bsnm{Ceder}, \binits{G.}}:
\batitle{Python materials genomics (pymatgen): A robust, open-source python library for materials analysis}.
\bjtitle{Computational Materials Science}
\bvolume{68},
\bfpage{314}--\blpage{319}
(\byear{2013})
\end{barticle}
\endbibitem

\bibitem[\protect\citeauthoryear{Tilley}{2019}]{V3Pb}
\begin{bbook}
\bauthor{\bsnm{Tilley}, \binits{D.R.}}:
\bbtitle{Superfluidity and Superconductivity}.
\bpublisher{Routledge},
\blocation{London}
(\byear{2019})
\end{bbook}
\endbibitem

\end{thebibliography}



\begin{thebibliography}{139}
\ifx \bisbn   \undefined \def \bisbn  #1{ISBN #1}\fi
\ifx \binits  \undefined \def \binits#1{#1}\fi
\ifx \bauthor  \undefined \def \bauthor#1{#1}\fi
\ifx \batitle  \undefined \def \batitle#1{#1}\fi
\ifx \bjtitle  \undefined \def \bjtitle#1{#1}\fi
\ifx \bvolume  \undefined \def \bvolume#1{\textbf{#1}}\fi
\ifx \byear  \undefined \def \byear#1{#1}\fi
\ifx \bissue  \undefined \def \bissue#1{#1}\fi
\ifx \bfpage  \undefined \def \bfpage#1{#1}\fi
\ifx \blpage  \undefined \def \blpage #1{#1}\fi
\ifx \burl  \undefined \def \burl#1{\textsf{#1}}\fi
\ifx \doiurl  \undefined \def \doiurl#1{\url{https://doi.org/#1}}\fi
\ifx \betal  \undefined \def \betal{\textit{et al.}}\fi
\ifx \binstitute  \undefined \def \binstitute#1{#1}\fi
\ifx \binstitutionaled  \undefined \def \binstitutionaled#1{#1}\fi
\ifx \bctitle  \undefined \def \bctitle#1{#1}\fi
\ifx \beditor  \undefined \def \beditor#1{#1}\fi
\ifx \bpublisher  \undefined \def \bpublisher#1{#1}\fi
\ifx \bbtitle  \undefined \def \bbtitle#1{#1}\fi
\ifx \bedition  \undefined \def \bedition#1{#1}\fi
\ifx \bseriesno  \undefined \def \bseriesno#1{#1}\fi
\ifx \blocation  \undefined \def \blocation#1{#1}\fi
\ifx \bsertitle  \undefined \def \bsertitle#1{#1}\fi
\ifx \bsnm \undefined \def \bsnm#1{#1}\fi
\ifx \bsuffix \undefined \def \bsuffix#1{#1}\fi
\ifx \bparticle \undefined \def \bparticle#1{#1}\fi
\ifx \barticle \undefined \def \barticle#1{#1}\fi
\bibcommenthead
\ifx \bconfdate \undefined \def \bconfdate #1{#1}\fi
\ifx \botherref \undefined \def \botherref #1{#1}\fi
\ifx \url \undefined \def \url#1{\textsf{#1}}\fi
\ifx \bchapter \undefined \def \bchapter#1{#1}\fi
\ifx \bbook \undefined \def \bbook#1{#1}\fi
\ifx \bcomment \undefined \def \bcomment#1{#1}\fi
\ifx \oauthor \undefined \def \oauthor#1{#1}\fi
\ifx \citeauthoryear \undefined \def \citeauthoryear#1{#1}\fi
\ifx \endbibitem  \undefined \def \endbibitem {}\fi
\ifx \bconflocation  \undefined \def \bconflocation#1{#1}\fi
\ifx \arxivurl  \undefined \def \arxivurl#1{\textsf{#1}}\fi
\csname PreBibitemsHook\endcsname

\bibitem[\protect\citeauthoryear{Han et~al.}{2025}]{supp_han2025survey}
\begin{barticle}
\bauthor{\bsnm{Han}, \binits{J.}},
\bauthor{\bsnm{Cen}, \binits{J.}},
\bauthor{\bsnm{Wu}, \binits{L.}},
\bauthor{\bsnm{Li}, \binits{Z.}},
\bauthor{\bsnm{Kong}, \binits{X.}},
\bauthor{\bsnm{Jiao}, \binits{R.}},
\bauthor{\bsnm{Yu}, \binits{Z.}},
\bauthor{\bsnm{Xu}, \binits{T.}},
\bauthor{\bsnm{Wu}, \binits{F.}},
\bauthor{\bsnm{Wang}, \binits{Z.}}, \betal:
\batitle{A survey of geometric graph neural networks: Data structures, models and applications}.
\bjtitle{Frontiers of Computer Science}
\bvolume{19}(\bissue{11}),
\bfpage{1911375}
(\byear{2025})
\end{barticle}
\endbibitem

\bibitem[\protect\citeauthoryear{Zhang et~al.}{2025}]{supp_zhang2025artificial}
\begin{barticle}
\bauthor{\bsnm{Zhang}, \binits{X.}},
\bauthor{\bsnm{Wang}, \binits{L.}},
\bauthor{\bsnm{Helwig}, \binits{J.}},
\bauthor{\bsnm{Luo}, \binits{Y.}},
\bauthor{\bsnm{Fu}, \binits{C.}},
\bauthor{\bsnm{Xie}, \binits{Y.}},
\bauthor{\bsnm{Liu}, \binits{M.}},
\bauthor{\bsnm{Lin}, \binits{Y.}},
\bauthor{\bsnm{Xu}, \binits{Z.}},
\bauthor{\bsnm{Yan}, \binits{K.}}, \betal:
\batitle{Artificial intelligence for science in quantum, atomistic, and continuum systems}.
\bjtitle{Foundations and Trends{\textregistered} in Machine Learning}
\bvolume{18}(\bissue{4}),
\bfpage{385}--\blpage{849}
(\byear{2025})
\end{barticle}
\endbibitem

\bibitem[\protect\citeauthoryear{Huang and Cen}{2026}]{supp_huang2026geometric}
\begin{botherref}
\oauthor{\bsnm{Huang}, \binits{W.}},
\oauthor{\bsnm{Cen}, \binits{J.}}:
Geometric graph learning for drug design.
Deep Learning in Drug Design,
133--151
(2026)
\end{botherref}
\endbibitem

\bibitem[\protect\citeauthoryear{Gilmer et~al.}{2017}]{supp_mpnn}
\begin{bchapter}
\bauthor{\bsnm{Gilmer}, \binits{J.}},
\bauthor{\bsnm{Schoenholz}, \binits{S.S.}},
\bauthor{\bsnm{Riley}, \binits{P.F.}},
\bauthor{\bsnm{Vinyals}, \binits{O.}},
\bauthor{\bsnm{Dahl}, \binits{G.E.}}:
\bctitle{Neural message passing for quantum chemistry}.
In: \bbtitle{International Conference on Machine Learning},
pp. \bfpage{1263}--\blpage{1272}
(\byear{2017}).
\bcomment{Pmlr}
\end{bchapter}
\endbibitem

\bibitem[\protect\citeauthoryear{Satorras et~al.}{2021}]{supp_EGNN}
\begin{bchapter}
\bauthor{\bsnm{Satorras}, \binits{V.G.}},
\bauthor{\bsnm{Hoogeboom}, \binits{E.}},
\bauthor{\bsnm{Welling}, \binits{M.}}:
\bctitle{E (n) equivariant graph neural networks}.
In: \bbtitle{International Conference on Machine Learning},
pp. \bfpage{9323}--\blpage{9332}
(\byear{2021}).
\bcomment{PMLR}
\end{bchapter}
\endbibitem

\bibitem[\protect\citeauthoryear{Sch{\"u}tt et~al.}{2021}]{supp_painn}
\begin{botherref}
\oauthor{\bsnm{Sch{\"u}tt}, \binits{K.}},
\oauthor{\bsnm{Unke}, \binits{O.}},
\oauthor{\bsnm{Gastegger}, \binits{M.}}:
Equivariant message passing for the prediction of tensorial properties and molecular spectra
(2021)
\end{botherref}
\endbibitem

\bibitem[\protect\citeauthoryear{Cen et~al.}{2024}]{supp_hegnn}
\begin{barticle}
\bauthor{\bsnm{Cen}, \binits{J.}},
\bauthor{\bsnm{Li}, \binits{A.}},
\bauthor{\bsnm{Lin}, \binits{N.}},
\bauthor{\bsnm{Ren}, \binits{Y.}},
\bauthor{\bsnm{Wang}, \binits{Z.}},
\bauthor{\bsnm{Huang}, \binits{W.}}:
\batitle{Are high-degree representations really unnecessary in equivariant graph neural networks?}
\bjtitle{Advances in Neural Information Processing Systems}
\bvolume{37},
\bfpage{26238}--\blpage{26266}
(\byear{2024})
\end{barticle}
\endbibitem

\bibitem[\protect\citeauthoryear{Thomas et~al.}{2018}]{supp_tfn}
\begin{botherref}
\oauthor{\bsnm{Thomas}, \binits{N.}},
\oauthor{\bsnm{Smidt}, \binits{T.}},
\oauthor{\bsnm{Kearnes}, \binits{S.}},
\oauthor{\bsnm{Yang}, \binits{L.}},
\oauthor{\bsnm{Li}, \binits{L.}},
\oauthor{\bsnm{Kohlhoff}, \binits{K.}},
\oauthor{\bsnm{Riley}, \binits{P.}}:
Tensor field networks: Rotation-and translation-equivariant neural networks for 3d point clouds.
arXiv preprint arXiv:1802.08219
(2018)
\end{botherref}
\endbibitem

\bibitem[\protect\citeauthoryear{Batatia et~al.}{2022}]{supp_Mace}
\begin{barticle}
\bauthor{\bsnm{Batatia}, \binits{I.}},
\bauthor{\bsnm{Kovacs}, \binits{D.P.}},
\bauthor{\bsnm{Simm}, \binits{G.}},
\bauthor{\bsnm{Ortner}, \binits{C.}},
\bauthor{\bsnm{Cs{\'a}nyi}, \binits{G.}}:
\batitle{Mace: Higher order equivariant message passing neural networks for fast and accurate force fields}.
\bjtitle{Advances in neural information processing systems}
\bvolume{35},
\bfpage{11423}--\blpage{11436}
(\byear{2022})
\end{barticle}
\endbibitem

\bibitem[\protect\citeauthoryear{Batzner et~al.}{2022}]{supp_NequIP}
\begin{barticle}
\bauthor{\bsnm{Batzner}, \binits{S.}},
\bauthor{\bsnm{Musaelian}, \binits{A.}},
\bauthor{\bsnm{Sun}, \binits{L.}},
\bauthor{\bsnm{Geiger}, \binits{M.}},
\bauthor{\bsnm{Mailoa}, \binits{J.P.}},
\bauthor{\bsnm{Kornbluth}, \binits{M.}},
\bauthor{\bsnm{Molinari}, \binits{N.}},
\bauthor{\bsnm{Smidt}, \binits{T.E.}},
\bauthor{\bsnm{Kozinsky}, \binits{B.}}:
\batitle{E (3)-equivariant graph neural networks for data-efficient and accurate interatomic potentials}.
\bjtitle{Nature communications}
\bvolume{13}(\bissue{1}),
\bfpage{2453}
(\byear{2022})
\end{barticle}
\endbibitem

\bibitem[\protect\citeauthoryear{Cen et~al.}{2025}]{supp_uniegnn}
\begin{bchapter}
\bauthor{\bsnm{Cen}, \binits{J.}},
\bauthor{\bsnm{Li}, \binits{A.}},
\bauthor{\bsnm{Lin}, \binits{N.}},
\bauthor{\bsnm{Xu}, \binits{T.}},
\bauthor{\bsnm{Rong}, \binits{Y.}},
\bauthor{\bsnm{Zhao}, \binits{D.}},
\bauthor{\bsnm{Wang}, \binits{Z.}},
\bauthor{\bsnm{Huang}, \binits{W.}}:
\bctitle{Universally invariant learning in equivariant {GNN}s}.
In: \bbtitle{The Thirty-ninth Annual Conference on Neural Information Processing Systems}
(\byear{2025})
\end{bchapter}
\endbibitem

\bibitem[\protect\citeauthoryear{Xie et~al.}{2025}]{supp_xie2025price}
\begin{bchapter}
\bauthor{\bsnm{Xie}, \binits{Y.}},
\bauthor{\bsnm{Daigavane}, \binits{A.}},
\bauthor{\bsnm{Kotak}, \binits{M.}},
\bauthor{\bsnm{Smidt}, \binits{T.}}:
\bctitle{The price of freedom: Exploring expressivity and runtime tradeoffs in equivariant tensor products}.
In: \bbtitle{International Conference on Machine Learning},
pp. \bfpage{68599}--\blpage{68625}
(\byear{2025}).
\bcomment{PMLR}
\end{bchapter}
\endbibitem

\bibitem[\protect\citeauthoryear{Dym and Maron}{2021}]{supp_dym2021on}
\begin{bchapter}
\bauthor{\bsnm{Dym}, \binits{N.}},
\bauthor{\bsnm{Maron}, \binits{H.}}:
\bctitle{On the universality of rotation equivariant point cloud networks}.
In: \bbtitle{International Conference on Learning Representations}
(\byear{2021})
\end{bchapter}
\endbibitem

\bibitem[\protect\citeauthoryear{Lin et~al.}{2026}]{supp_lin2026reducing}
\begin{bchapter}
\bauthor{\bsnm{Lin}, \binits{N.}},
\bauthor{\bsnm{Cen}, \binits{J.}},
\bauthor{\bsnm{Li}, \binits{A.}},
\bauthor{\bsnm{Huang}, \binits{W.}},
\bauthor{\bsnm{Sun}, \binits{H.}}:
\bctitle{Reducing symmetry increase in equivariant neural networks}.
In: \bbtitle{The Fourteenth International Conference on Learning Representations}
(\byear{2026})
\end{bchapter}
\endbibitem

\bibitem[\protect\citeauthoryear{Vaswani et~al.}{2017}]{supp_vaswani2017attention}
\begin{botherref}
\oauthor{\bsnm{Vaswani}, \binits{A.}},
\oauthor{\bsnm{Shazeer}, \binits{N.}},
\oauthor{\bsnm{Parmar}, \binits{N.}},
\oauthor{\bsnm{Uszkoreit}, \binits{J.}},
\oauthor{\bsnm{Jones}, \binits{L.}},
\oauthor{\bsnm{Gomez}, \binits{A.N.}},
\oauthor{\bsnm{Kaiser}, \binits{{\L}.}},
\oauthor{\bsnm{Polosukhin}, \binits{I.}}:
Attention is all you need.
Annual Conference on Neural Information Processing Systems
\textbf{30}
(2017)
\end{botherref}
\endbibitem

\bibitem[\protect\citeauthoryear{Yuan et~al.}{2025}]{supp_yuan2025survey}
\begin{botherref}
\oauthor{\bsnm{Yuan}, \binits{C.}},
\oauthor{\bsnm{Zhao}, \binits{K.}},
\oauthor{\bsnm{Kuruoglu}, \binits{E.E.}},
\oauthor{\bsnm{Wang}, \binits{L.}},
\oauthor{\bsnm{Xu}, \binits{T.}},
\oauthor{\bsnm{Huang}, \binits{W.}},
\oauthor{\bsnm{Zhao}, \binits{D.}},
\oauthor{\bsnm{Cheng}, \binits{H.}},
\oauthor{\bsnm{Rong}, \binits{Y.}}:
A survey of graph transformers: Architectures, theories and applications.
arXiv preprint arXiv:2502.16533
(2025)
\end{botherref}
\endbibitem

\bibitem[\protect\citeauthoryear{Fuchs et~al.}{2020}]{supp_SE3_Transformer}
\begin{barticle}
\bauthor{\bsnm{Fuchs}, \binits{F.}},
\bauthor{\bsnm{Worrall}, \binits{D.}},
\bauthor{\bsnm{Fischer}, \binits{V.}},
\bauthor{\bsnm{Welling}, \binits{M.}}:
\batitle{Se (3)-transformers: 3d roto-translation equivariant attention networks}.
\bjtitle{Advances in neural information processing systems}
\bvolume{33},
\bfpage{1970}--\blpage{1981}
(\byear{2020})
\end{barticle}
\endbibitem

\bibitem[\protect\citeauthoryear{Liao and Smidt}{2023}]{supp_Equiformer}
\begin{bchapter}
\bauthor{\bsnm{Liao}, \binits{Y.-L.}},
\bauthor{\bsnm{Smidt}, \binits{T.}}:
\bctitle{Equiformer: Equivariant graph attention transformer for 3d atomistic graphs}.
In: \bbtitle{The Eleventh International Conference on Learning Representations}
(\byear{2023})
\end{bchapter}
\endbibitem

\bibitem[\protect\citeauthoryear{Liao et~al.}{2024}]{supp_EquiformerV2}
\begin{bchapter}
\bauthor{\bsnm{Liao}, \binits{Y.-L.}},
\bauthor{\bsnm{Wood}, \binits{B.M.}},
\bauthor{\bsnm{Das}, \binits{A.}},
\bauthor{\bsnm{Smidt}, \binits{T.}}:
\bctitle{Equiformerv2: Improved equivariant transformer for scaling to higher-degree representations}.
In: \bbtitle{The Twelfth International Conference on Learning Representations}
(\byear{2024})
\end{bchapter}
\endbibitem

\bibitem[\protect\citeauthoryear{Passaro and Zitnick}{2023}]{supp_escn}
\begin{bchapter}
\bauthor{\bsnm{Passaro}, \binits{S.}},
\bauthor{\bsnm{Zitnick}, \binits{C.L.}}:
\bctitle{Reducing so (3) convolutions to so (2) for efficient equivariant gnns}.
In: \bbtitle{International Conference on Machine Learning},
pp. \bfpage{27420}--\blpage{27438}
(\byear{2023}).
\bcomment{PMLR}
\end{bchapter}
\endbibitem

\bibitem[\protect\citeauthoryear{Jiao et~al.}{2023}]{supp_diffcsp}
\begin{barticle}
\bauthor{\bsnm{Jiao}, \binits{R.}},
\bauthor{\bsnm{Huang}, \binits{W.}},
\bauthor{\bsnm{Lin}, \binits{P.}},
\bauthor{\bsnm{Han}, \binits{J.}},
\bauthor{\bsnm{Chen}, \binits{P.}},
\bauthor{\bsnm{Lu}, \binits{Y.}},
\bauthor{\bsnm{Liu}, \binits{Y.}}:
\batitle{Crystal structure prediction by joint equivariant diffusion}.
\bjtitle{Advances in Neural Information Processing Systems}
\bvolume{36},
\bfpage{17464}--\blpage{17497}
(\byear{2023})
\end{barticle}
\endbibitem

\bibitem[\protect\citeauthoryear{Zeni et~al.}{2025}]{supp_mattergen}
\begin{barticle}
\bauthor{\bsnm{Zeni}, \binits{C.}},
\bauthor{\bsnm{Pinsler}, \binits{R.}},
\bauthor{\bsnm{Z{\"u}gner}, \binits{D.}},
\bauthor{\bsnm{Fowler}, \binits{A.}},
\bauthor{\bsnm{Horton}, \binits{M.}},
\bauthor{\bsnm{Fu}, \binits{X.}},
\bauthor{\bsnm{Wang}, \binits{Z.}},
\bauthor{\bsnm{Shysheya}, \binits{A.}},
\bauthor{\bsnm{Crabb{\'e}}, \binits{J.}},
\bauthor{\bsnm{Ueda}, \binits{S.}}, \betal:
\batitle{A generative model for inorganic materials design}.
\bjtitle{Nature}
\bvolume{639}(\bissue{8055}),
\bfpage{624}--\blpage{632}
(\byear{2025})
\end{barticle}
\endbibitem

\bibitem[\protect\citeauthoryear{Barroso-Luque et~al.}{2024}]{supp_OMAT-24}
\begin{botherref}
\oauthor{\bsnm{Barroso-Luque}, \binits{L.}},
\oauthor{\bsnm{Shuaibi}, \binits{M.}},
\oauthor{\bsnm{Fu}, \binits{X.}},
\oauthor{\bsnm{Wood}, \binits{B.M.}},
\oauthor{\bsnm{Dzamba}, \binits{M.}},
\oauthor{\bsnm{Gao}, \binits{M.}},
\oauthor{\bsnm{Rizvi}, \binits{A.}},
\oauthor{\bsnm{Zitnick}, \binits{C.L.}},
\oauthor{\bsnm{Ulissi}, \binits{Z.W.}}:
Open materials 2024 (omat24) inorganic materials dataset and models.
arXiv preprint arXiv:2410.12771
(2024)
\end{botherref}
\endbibitem

\bibitem[\protect\citeauthoryear{Jain et~al.}{2013}]{supp_mp-20mpts-52}
\begin{botherref}
\oauthor{\bsnm{Jain}, \binits{A.}},
\oauthor{\bsnm{Ong}, \binits{S.P.}},
\oauthor{\bsnm{Hautier}, \binits{G.}},
\oauthor{\bsnm{Chen}, \binits{W.}},
\oauthor{\bsnm{Richards}, \binits{W.D.}},
\oauthor{\bsnm{Dacek}, \binits{S.}},
\oauthor{\bsnm{Cholia}, \binits{S.}},
\oauthor{\bsnm{Gunter}, \binits{D.}},
\oauthor{\bsnm{Skinner}, \binits{D.}},
\oauthor{\bsnm{Ceder}, \binits{G.}}, et al.:
Commentary: The materials project: A materials genome approach to accelerating materials innovation.
APL materials
\textbf{1}(1)
(2013)
\end{botherref}
\endbibitem

\bibitem[\protect\citeauthoryear{Smith et~al.}{2020}]{supp_ani-1x}
\begin{barticle}
\bauthor{\bsnm{Smith}, \binits{J.S.}},
\bauthor{\bsnm{Zubatyuk}, \binits{R.}},
\bauthor{\bsnm{Nebgen}, \binits{B.}},
\bauthor{\bsnm{Lubbers}, \binits{N.}},
\bauthor{\bsnm{Barros}, \binits{K.}},
\bauthor{\bsnm{Roitberg}, \binits{A.E.}},
\bauthor{\bsnm{Isayev}, \binits{O.}},
\bauthor{\bsnm{Tretiak}, \binits{S.}}:
\batitle{The ani-1ccx and ani-1x data sets, coupled-cluster and density functional theory properties for molecules}.
\bjtitle{Scientific data}
\bvolume{7}(\bissue{1}),
\bfpage{134}
(\byear{2020})
\end{barticle}
\endbibitem

\bibitem[\protect\citeauthoryear{Schreiner et~al.}{2022}]{supp_Transition1x}
\begin{barticle}
\bauthor{\bsnm{Schreiner}, \binits{M.}},
\bauthor{\bsnm{Bhowmik}, \binits{A.}},
\bauthor{\bsnm{Vegge}, \binits{T.}},
\bauthor{\bsnm{Busk}, \binits{J.}},
\bauthor{\bsnm{Winther}, \binits{O.}}:
\batitle{Transition1x-a dataset for building generalizable reactive machine learning potentials}.
\bjtitle{Scientific Data}
\bvolume{9}(\bissue{1}),
\bfpage{779}
(\byear{2022})
\end{barticle}
\endbibitem

\bibitem[\protect\citeauthoryear{Biamonte et~al.}{2017}]{supp_qm9}
\begin{barticle}
\bauthor{\bsnm{Biamonte}, \binits{J.}},
\bauthor{\bsnm{Wittek}, \binits{P.}},
\bauthor{\bsnm{Pancotti}, \binits{N.}},
\bauthor{\bsnm{Rebentrost}, \binits{P.}},
\bauthor{\bsnm{Wiebe}, \binits{N.}},
\bauthor{\bsnm{Lloyd}, \binits{S.}}:
\batitle{Quantum machine learning}.
\bjtitle{Nature}
\bvolume{549}(\bissue{7671}),
\bfpage{195}--\blpage{202}
(\byear{2017})
\end{barticle}
\endbibitem

\bibitem[\protect\citeauthoryear{Choudhary et~al.}{2020}]{supp_jarvis}
\begin{barticle}
\bauthor{\bsnm{Choudhary}, \binits{K.}},
\bauthor{\bsnm{Garrity}, \binits{K.F.}},
\bauthor{\bsnm{Reid}, \binits{A.C.}},
\bauthor{\bsnm{DeCost}, \binits{B.}},
\bauthor{\bsnm{Biacchi}, \binits{A.J.}},
\bauthor{\bsnm{Hight~Walker}, \binits{A.R.}},
\bauthor{\bsnm{Trautt}, \binits{Z.}},
\bauthor{\bsnm{Hattrick-Simpers}, \binits{J.}},
\bauthor{\bsnm{Kusne}, \binits{A.G.}},
\bauthor{\bsnm{Centrone}, \binits{A.}}, \betal:
\batitle{The joint automated repository for various integrated simulations (jarvis) for data-driven materials design}.
\bjtitle{npj computational materials}
\bvolume{6}(\bissue{1}),
\bfpage{173}
(\byear{2020})
\end{barticle}
\endbibitem

\bibitem[\protect\citeauthoryear{Schmidt et~al.}{2024}]{supp_Alex}
\begin{barticle}
\bauthor{\bsnm{Schmidt}, \binits{J.}},
\bauthor{\bsnm{Cerqueira}, \binits{T.F.}},
\bauthor{\bsnm{Romero}, \binits{A.H.}},
\bauthor{\bsnm{Loew}, \binits{A.}},
\bauthor{\bsnm{J{\"a}ger}, \binits{F.}},
\bauthor{\bsnm{Wang}, \binits{H.-C.}},
\bauthor{\bsnm{Botti}, \binits{S.}},
\bauthor{\bsnm{Marques}, \binits{M.A.}}:
\batitle{Improving machine-learning models in materials science through large datasets}.
\bjtitle{Materials Today Physics}
\bvolume{48},
\bfpage{101560}
(\byear{2024})
\end{barticle}
\endbibitem

\bibitem[\protect\citeauthoryear{Merchant et~al.}{2023}]{supp_genome}
\begin{barticle}
\bauthor{\bsnm{Merchant}, \binits{A.}},
\bauthor{\bsnm{Batzner}, \binits{S.}},
\bauthor{\bsnm{Schoenholz}, \binits{S.S.}},
\bauthor{\bsnm{Aykol}, \binits{M.}},
\bauthor{\bsnm{Cheon}, \binits{G.}},
\bauthor{\bsnm{Cubuk}, \binits{E.D.}}:
\batitle{Scaling deep learning for materials discovery}.
\bjtitle{Nature}
\bvolume{624}(\bissue{7990}),
\bfpage{80}--\blpage{85}
(\byear{2023})
\end{barticle}
\endbibitem

\bibitem[\protect\citeauthoryear{Saal et~al.}{2013}]{supp_oqmd}
\begin{barticle}
\bauthor{\bsnm{Saal}, \binits{J.E.}},
\bauthor{\bsnm{Kirklin}, \binits{S.}},
\bauthor{\bsnm{Aykol}, \binits{M.}},
\bauthor{\bsnm{Meredig}, \binits{B.}},
\bauthor{\bsnm{Wolverton}, \binits{C.}}:
\batitle{Materials design and discovery with high-throughput density functional theory: the open quantum materials database (oqmd)}.
\bjtitle{Jom}
\bvolume{65}(\bissue{11}),
\bfpage{1501}--\blpage{1509}
(\byear{2013})
\end{barticle}
\endbibitem

\bibitem[\protect\citeauthoryear{Garrity and Choudhary}{2023}]{supp_jarvis_qetb}
\begin{barticle}
\bauthor{\bsnm{Garrity}, \binits{K.F.}},
\bauthor{\bsnm{Choudhary}, \binits{K.}}:
\batitle{Fast and accurate prediction of material properties with three-body tight-binding model for the periodic table}.
\bjtitle{Physical review materials}
\bvolume{7}(\bissue{4}),
\bfpage{044603}
(\byear{2023})
\end{barticle}
\endbibitem

\bibitem[\protect\citeauthoryear{Chen and Ong}{2022}]{supp_M3GNet}
\begin{barticle}
\bauthor{\bsnm{Chen}, \binits{C.}},
\bauthor{\bsnm{Ong}, \binits{S.P.}}:
\batitle{A universal graph deep learning interatomic potential for the periodic table}.
\bjtitle{Nature Computational Science}
\bvolume{2}(\bissue{11}),
\bfpage{718}--\blpage{728}
(\byear{2022})
\end{barticle}
\endbibitem

\bibitem[\protect\citeauthoryear{Scheidgen et~al.}{2023}]{supp_nomad}
\begin{barticle}
\bauthor{\bsnm{Scheidgen}, \binits{M.}},
\bauthor{\bsnm{Himanen}, \binits{L.}},
\bauthor{\bsnm{Ladines}, \binits{A.N.}},
\bauthor{\bsnm{Sikter}, \binits{D.}},
\bauthor{\bsnm{Nakhaee}, \binits{M.}},
\bauthor{\bsnm{Fekete}, \binits{{\'A}.}},
\bauthor{\bsnm{Chang}, \binits{T.}},
\bauthor{\bsnm{Golparvar}, \binits{A.}},
\bauthor{\bsnm{M{\'a}rquez}, \binits{J.A.}},
\bauthor{\bsnm{Brockhauser}, \binits{S.}}, \betal:
\batitle{Nomad: A distributed web-based platform for managing materials science research data}.
\bjtitle{Journal of Open Source Software}
\bvolume{8}(\bissue{90}),
\bfpage{5388}
(\byear{2023})
\end{barticle}
\endbibitem

\bibitem[\protect\citeauthoryear{Hafner}{2008}]{supp_hafner2008ab}
\begin{barticle}
\bauthor{\bsnm{Hafner}, \binits{J.}}:
\batitle{Ab-initio simulations of materials using vasp: Density-functional theory and beyond}.
\bjtitle{Journal of computational chemistry}
\bvolume{29}(\bissue{13}),
\bfpage{2044}--\blpage{2078}
(\byear{2008})
\end{barticle}
\endbibitem

\bibitem[\protect\citeauthoryear{Giannozzi et~al.}{2009}]{supp_giannozzi2009quantum}
\begin{barticle}
\bauthor{\bsnm{Giannozzi}, \binits{P.}},
\bauthor{\bsnm{Baroni}, \binits{S.}},
\bauthor{\bsnm{Bonini}, \binits{N.}},
\bauthor{\bsnm{Calandra}, \binits{M.}},
\bauthor{\bsnm{Car}, \binits{R.}},
\bauthor{\bsnm{Cavazzoni}, \binits{C.}},
\bauthor{\bsnm{Ceresoli}, \binits{D.}},
\bauthor{\bsnm{Chiarotti}, \binits{G.L.}},
\bauthor{\bsnm{Cococcioni}, \binits{M.}},
\bauthor{\bsnm{Dabo}, \binits{I.}}, \betal:
\batitle{Quantum espresso: a modular and open-source software project for quantum simulations of materials}.
\bjtitle{Journal of physics: Condensed matter}
\bvolume{21}(\bissue{39}),
\bfpage{395502}
(\byear{2009})
\end{barticle}
\endbibitem

\bibitem[\protect\citeauthoryear{Hu et~al.}{2021}]{supp_pcq}
\begin{botherref}
\oauthor{\bsnm{Hu}, \binits{W.}},
\oauthor{\bsnm{Fey}, \binits{M.}},
\oauthor{\bsnm{Ren}, \binits{H.}},
\oauthor{\bsnm{Nakata}, \binits{M.}},
\oauthor{\bsnm{Dong}, \binits{Y.}},
\oauthor{\bsnm{Leskovec}, \binits{J.}}:
Ogb-lsc: A large-scale challenge for machine learning on graphs.
arXiv preprint arXiv:2103.09430
(2021)
\end{botherref}
\endbibitem

\bibitem[\protect\citeauthoryear{Nakata and Shimazaki}{2017}]{supp_nakata2017pubchemqc}
\begin{barticle}
\bauthor{\bsnm{Nakata}, \binits{M.}},
\bauthor{\bsnm{Shimazaki}, \binits{T.}}:
\batitle{Pubchemqc project: a large-scale first-principles electronic structure database for data-driven chemistry}.
\bjtitle{Journal of chemical information and modeling}
\bvolume{57}(\bissue{6}),
\bfpage{1300}--\blpage{1308}
(\byear{2017})
\end{barticle}
\endbibitem

\bibitem[\protect\citeauthoryear{Dunn et~al.}{2020}]{supp_matbench}
\begin{barticle}
\bauthor{\bsnm{Dunn}, \binits{A.}},
\bauthor{\bsnm{Wang}, \binits{Q.}},
\bauthor{\bsnm{Ganose}, \binits{A.}},
\bauthor{\bsnm{Dopp}, \binits{D.}},
\bauthor{\bsnm{Jain}, \binits{A.}}:
\batitle{Benchmarking materials property prediction methods: the matbench test set and automatminer reference algorithm}.
\bjtitle{npj Computational Materials}
\bvolume{6}(\bissue{1}),
\bfpage{138}
(\byear{2020})
\end{barticle}
\endbibitem

\bibitem[\protect\citeauthoryear{Zhang et~al.}{2024}]{supp_DPA-2}
\begin{barticle}
\bauthor{\bsnm{Zhang}, \binits{D.}},
\bauthor{\bsnm{Liu}, \binits{X.}},
\bauthor{\bsnm{Zhang}, \binits{X.}},
\bauthor{\bsnm{Zhang}, \binits{C.}},
\bauthor{\bsnm{Cai}, \binits{C.}},
\bauthor{\bsnm{Bi}, \binits{H.}},
\bauthor{\bsnm{Du}, \binits{Y.}},
\bauthor{\bsnm{Qin}, \binits{X.}},
\bauthor{\bsnm{Peng}, \binits{A.}},
\bauthor{\bsnm{Huang}, \binits{J.}}, \betal:
\batitle{Dpa-2: a large atomic model as a multi-task learner}.
\bjtitle{npj Computational Materials}
\bvolume{10}(\bissue{1}),
\bfpage{293}
(\byear{2024})
\end{barticle}
\endbibitem

\bibitem[\protect\citeauthoryear{Chanussot et~al.}{2021}]{supp_chanussot2021open}
\begin{barticle}
\bauthor{\bsnm{Chanussot}, \binits{L.}},
\bauthor{\bsnm{Das}, \binits{A.}},
\bauthor{\bsnm{Goyal}, \binits{S.}},
\bauthor{\bsnm{Lavril}, \binits{T.}},
\bauthor{\bsnm{Shuaibi}, \binits{M.}},
\bauthor{\bsnm{Riviere}, \binits{M.}},
\bauthor{\bsnm{Tran}, \binits{K.}},
\bauthor{\bsnm{Heras-Domingo}, \binits{J.}},
\bauthor{\bsnm{Ho}, \binits{C.}},
\bauthor{\bsnm{Hu}, \binits{W.}}, \betal:
\batitle{Open catalyst 2020 (oc20) dataset and community challenges}.
\bjtitle{Acs Catalysis}
\bvolume{11}(\bissue{10}),
\bfpage{6059}--\blpage{6072}
(\byear{2021})
\end{barticle}
\endbibitem

\bibitem[\protect\citeauthoryear{Cerqueira et~al.}{2024}]{supp_dfttc}
\begin{barticle}
\bauthor{\bsnm{Cerqueira}, \binits{T.F.}},
\bauthor{\bsnm{Sanna}, \binits{A.}},
\bauthor{\bsnm{Marques}, \binits{M.A.}}:
\batitle{Sampling the materials space for conventional superconducting compounds}.
\bjtitle{Advanced Materials}
\bvolume{36}(\bissue{1}),
\bfpage{2307085}
(\byear{2024})
\end{barticle}
\endbibitem

\bibitem[\protect\citeauthoryear{Chen et~al.}{2024}]{supp_sodnet}
\begin{barticle}
\bauthor{\bsnm{Chen}, \binits{P.}},
\bauthor{\bsnm{Peng}, \binits{L.}},
\bauthor{\bsnm{Jiao}, \binits{R.}},
\bauthor{\bsnm{Mo}, \binits{Q.}},
\bauthor{\bsnm{Wang}, \binits{Z.}},
\bauthor{\bsnm{Huang}, \binits{W.}},
\bauthor{\bsnm{Liu}, \binits{Y.}},
\bauthor{\bsnm{Lu}, \binits{Y.}}:
\batitle{Learning superconductivity from ordered and disordered material structures}.
\bjtitle{Advances in Neural Information Processing Systems}
\bvolume{37},
\bfpage{108902}--\blpage{108928}
(\byear{2024})
\end{barticle}
\endbibitem

\bibitem[\protect\citeauthoryear{Group}{}]{supp_supercon}
\begin{botherref}
\oauthor{\bsnm{Group}, \binits{M.D.}}:
MDR SuperCon Datasheet Ver.220808.
National Institute for Materials Science
\end{botherref}
\endbibitem

\bibitem[\protect\citeauthoryear{Deng et~al.}{2023}]{supp_deng2023chgnet}
\begin{barticle}
\bauthor{\bsnm{Deng}, \binits{B.}},
\bauthor{\bsnm{Zhong}, \binits{P.}},
\bauthor{\bsnm{Jun}, \binits{K.}},
\bauthor{\bsnm{Riebesell}, \binits{J.}},
\bauthor{\bsnm{Han}, \binits{K.}},
\bauthor{\bsnm{Bartel}, \binits{C.J.}},
\bauthor{\bsnm{Ceder}, \binits{G.}}:
\batitle{Chgnet as a pretrained universal neural network potential for charge-informed atomistic modelling}.
\bjtitle{Nature Machine Intelligence}
\bvolume{5}(\bissue{9}),
\bfpage{1031}--\blpage{1041}
(\byear{2023})
\end{barticle}
\endbibitem

\bibitem[\protect\citeauthoryear{Batatia et~al.}{2025}]{supp_botnet}
\begin{barticle}
\bauthor{\bsnm{Batatia}, \binits{I.}},
\bauthor{\bsnm{Batzner}, \binits{S.}},
\bauthor{\bsnm{Kov{\'a}cs}, \binits{D.P.}},
\bauthor{\bsnm{Musaelian}, \binits{A.}},
\bauthor{\bsnm{Simm}, \binits{G.N.}},
\bauthor{\bsnm{Drautz}, \binits{R.}},
\bauthor{\bsnm{Ortner}, \binits{C.}},
\bauthor{\bsnm{Kozinsky}, \binits{B.}},
\bauthor{\bsnm{Cs{\'a}nyi}, \binits{G.}}:
\batitle{The design space of e (3)-equivariant atom-centred interatomic potentials}.
\bjtitle{Nature Machine Intelligence}
\bvolume{7}(\bissue{1}),
\bfpage{56}--\blpage{67}
(\byear{2025})
\end{barticle}
\endbibitem

\bibitem[\protect\citeauthoryear{Song et~al.}{2021}]{supp_ddim}
\begin{bchapter}
\bauthor{\bsnm{Song}, \binits{J.}},
\bauthor{\bsnm{Meng}, \binits{C.}},
\bauthor{\bsnm{Ermon}, \binits{S.}}:
\bctitle{Denoising diffusion implicit models}.
In: \bbtitle{International Conference on Learning Representations}
(\byear{2021}).
\burl{https://openreview.net/forum?id=St1giarCHLP}
\end{bchapter}
\endbibitem

\bibitem[\protect\citeauthoryear{weitao Du et~al.}{2023}]{supp_leftnet}
\begin{bchapter}
\bauthor{\bsnm{Du}},
\bauthor{\bsnm{Du}, \binits{Y.}},
\bauthor{\bsnm{Wang}, \binits{L.}},
\bauthor{\bsnm{Feng}, \binits{D.}},
\bauthor{\bsnm{Wang}, \binits{G.}},
\bauthor{\bsnm{Ji}, \binits{S.}},
\bauthor{\bsnm{Gomes}, \binits{C.P.}},
\bauthor{\bsnm{Ma}, \binits{Z.-M.}}:
\bctitle{A new perspective on building efficient and expressive 3d equivariant graph neural networks}.
In: \bbtitle{Thirty-seventh Conference on Neural Information Processing Systems}
(\byear{2023})
\end{bchapter}
\endbibitem

\bibitem[\protect\citeauthoryear{Gasteiger et~al.}{2020}]{supp_DimeNet}
\begin{bchapter}
\bauthor{\bsnm{Gasteiger}, \binits{J.}},
\bauthor{\bsnm{Groß}, \binits{J.}},
\bauthor{\bsnm{Günnemann}, \binits{S.}}:
\bctitle{Directional message passing for molecular graphs}.
In: \bbtitle{International Conference on Learning Representations}
(\byear{2020})
\end{bchapter}
\endbibitem

\bibitem[\protect\citeauthoryear{Liu et~al.}{2022}]{supp_spherenet}
\begin{bchapter}
\bauthor{\bsnm{Liu}, \binits{Y.}},
\bauthor{\bsnm{Wang}, \binits{L.}},
\bauthor{\bsnm{Liu}, \binits{M.}},
\bauthor{\bsnm{Lin}, \binits{Y.}},
\bauthor{\bsnm{Zhang}, \binits{X.}},
\bauthor{\bsnm{Oztekin}, \binits{B.}},
\bauthor{\bsnm{Ji}, \binits{S.}}:
\bctitle{Spherical message passing for 3d molecular graphs}.
In: \bbtitle{International Conference on Learning Representations}
(\byear{2022})
\end{bchapter}
\endbibitem

\bibitem[\protect\citeauthoryear{Wang et~al.}{2023}]{supp_Geoformer}
\begin{bchapter}
\bauthor{\bsnm{Wang}, \binits{Y.}},
\bauthor{\bsnm{Li}, \binits{S.}},
\bauthor{\bsnm{Wang}, \binits{T.}},
\bauthor{\bsnm{Shao}, \binits{B.}},
\bauthor{\bsnm{Zheng}, \binits{N.}},
\bauthor{\bsnm{Liu}, \binits{T.-Y.}}:
\bctitle{Geometric transformer with interatomic positional encoding}.
In: \bbtitle{Thirty-seventh Conference on Neural Information Processing Systems}
(\byear{2023})
\end{bchapter}
\endbibitem

\bibitem[\protect\citeauthoryear{Feng et~al.}{2023}]{supp_frad}
\begin{bchapter}
\bauthor{\bsnm{Feng}, \binits{S.}},
\bauthor{\bsnm{Ni}, \binits{Y.}},
\bauthor{\bsnm{Lan}, \binits{Y.}},
\bauthor{\bsnm{Ma}, \binits{Z.-M.}},
\bauthor{\bsnm{Ma}, \binits{W.-Y.}}:
\bctitle{Fractional denoising for 3d molecular pre-training}.
In: \bbtitle{International Conference on Machine Learning},
pp. \bfpage{9938}--\blpage{9961}
(\byear{2023}).
\bcomment{PMLR}
\end{bchapter}
\endbibitem

\bibitem[\protect\citeauthoryear{Jiao et~al.}{2026}]{supp_ept}
\begin{botherref}
\oauthor{\bsnm{Jiao}, \binits{R.}},
\oauthor{\bsnm{Kong}, \binits{X.}},
\oauthor{\bsnm{Zhang}, \binits{L.}},
\oauthor{\bsnm{Yu}, \binits{Z.}},
\oauthor{\bsnm{Ren}, \binits{F.}},
\oauthor{\bsnm{Tan}, \binits{W.}},
\oauthor{\bsnm{Huang}, \binits{W.}},
\oauthor{\bsnm{Liu}, \binits{Y.}}:
An equivariant pretrained transformer for unified 3d molecular representation learning.
Nature Communications
(2026)
\end{botherref}
\endbibitem

\bibitem[\protect\citeauthoryear{Ni et~al.}{2024}]{supp_slide}
\begin{bchapter}
\bauthor{\bsnm{Ni}, \binits{Y.}},
\bauthor{\bsnm{Feng}, \binits{S.}},
\bauthor{\bsnm{Ma}, \binits{W.-Y.}},
\bauthor{\bsnm{Ma}, \binits{Z.-M.}},
\bauthor{\bsnm{Lan}, \binits{Y.}}:
\bctitle{Sliced denoising: A physics-informed molecular pre-training method}.
In: \bbtitle{The Twelfth International Conference on Learning Representations}
(\byear{2024}).
\burl{https://openreview.net/forum?id=liKkG1zcWq}
\end{bchapter}
\endbibitem

\bibitem[\protect\citeauthoryear{Aykent and Xia}{2025}]{supp_gotennet}
\begin{bchapter}
\bauthor{\bsnm{Aykent}, \binits{S.}},
\bauthor{\bsnm{Xia}, \binits{T.}}:
\bctitle{Gotennet: Rethinking efficient 3d equivariant graph neural networks}.
In: \bbtitle{The Thirteenth International Conference on Learning Representations}
(\byear{2025})
\end{bchapter}
\endbibitem

\bibitem[\protect\citeauthoryear{Choudhary and DeCost}{2021}]{supp_alignn}
\begin{barticle}
\bauthor{\bsnm{Choudhary}, \binits{K.}},
\bauthor{\bsnm{DeCost}, \binits{B.}}:
\batitle{Atomistic line graph neural network for improved materials property predictions}.
\bjtitle{npj Computational Materials}
\bvolume{7}(\bissue{1}),
\bfpage{185}
(\byear{2021})
\end{barticle}
\endbibitem

\bibitem[\protect\citeauthoryear{De~Breuck et~al.}{2021}]{supp_MODNet}
\begin{barticle}
\bauthor{\bsnm{De~Breuck}, \binits{P.-P.}},
\bauthor{\bsnm{Evans}, \binits{M.L.}},
\bauthor{\bsnm{Rignanese}, \binits{G.-M.}}:
\batitle{Robust model benchmarking and bias-imbalance in data-driven materials science: a case study on modnet}.
\bjtitle{Journal of Physics: Condensed Matter}
\bvolume{33}(\bissue{40}),
\bfpage{404002}
(\byear{2021})
\end{barticle}
\endbibitem

\bibitem[\protect\citeauthoryear{Xie and Grossman}{2018}]{supp_CGCNN}
\begin{barticle}
\bauthor{\bsnm{Xie}, \binits{T.}},
\bauthor{\bsnm{Grossman}, \binits{J.C.}}:
\batitle{Crystal graph convolutional neural networks for an accurate and interpretable prediction of material properties}.
\bjtitle{Physical review letters}
\bvolume{120}(\bissue{14}),
\bfpage{145301}
(\byear{2018})
\end{barticle}
\endbibitem

\bibitem[\protect\citeauthoryear{Chen et~al.}{2019}]{supp_MEGNet}
\begin{barticle}
\bauthor{\bsnm{Chen}, \binits{C.}},
\bauthor{\bsnm{Ye}, \binits{W.}},
\bauthor{\bsnm{Zuo}, \binits{Y.}},
\bauthor{\bsnm{Zheng}, \binits{C.}},
\bauthor{\bsnm{Ong}, \binits{S.P.}}:
\batitle{Graph networks as a universal machine learning framework for molecules and crystals}.
\bjtitle{Chemistry of Materials}
\bvolume{31}(\bissue{9}),
\bfpage{3564}--\blpage{3572}
(\byear{2019})
\end{barticle}
\endbibitem

\bibitem[\protect\citeauthoryear{Sch{\"u}tt et~al.}{2017}]{supp_SchNet}
\begin{bchapter}
\bauthor{\bsnm{Sch{\"u}tt}, \binits{K.}},
\bauthor{\bsnm{Kindermans}, \binits{P.-J.}},
\bauthor{\bsnm{Sauceda~Felix}, \binits{H.E.}},
\bauthor{\bsnm{Chmiela}, \binits{S.}},
\bauthor{\bsnm{Tkatchenko}, \binits{A.}},
\bauthor{\bsnm{M{\"u}ller}, \binits{K.-R.}}:
\bctitle{Schnet: A continuous-filter convolutional neural network for modeling quantum interactions}.
In: \bbtitle{Advances in Neural Information Processing Systems},
vol. \bseriesno{30}
(\byear{2017})
\end{bchapter}
\endbibitem

\bibitem[\protect\citeauthoryear{Ruff et~al.}{2024}]{supp_coGN_coNGN}
\begin{barticle}
\bauthor{\bsnm{Ruff}, \binits{R.}},
\bauthor{\bsnm{Reiser}, \binits{P.}},
\bauthor{\bsnm{St{\"u}hmer}, \binits{J.}},
\bauthor{\bsnm{Friederich}, \binits{P.}}:
\batitle{Connectivity optimized nested line graph networks for crystal structures}.
\bjtitle{Digital Discovery}
\bvolume{3}(\bissue{3}),
\bfpage{594}--\blpage{601}
(\byear{2024})
\end{barticle}
\endbibitem

\bibitem[\protect\citeauthoryear{Gasteiger et~al.}{2022}]{supp_gemnet-oc}
\begin{botherref}
\oauthor{\bsnm{Gasteiger}, \binits{J.}},
\oauthor{\bsnm{Shuaibi}, \binits{M.}},
\oauthor{\bsnm{Sriram}, \binits{A.}},
\oauthor{\bsnm{G{\"u}nnemann}, \binits{S.}},
\oauthor{\bsnm{Ulissi}, \binits{Z.W.}},
\oauthor{\bsnm{Zitnick}, \binits{C.L.}},
\oauthor{\bsnm{Das}, \binits{A.}}:
Gemnet-{OC}: Developing graph neural networks for large and diverse molecular simulation datasets.
Transactions on Machine Learning Research
(2022)
\end{botherref}
\endbibitem

\bibitem[\protect\citeauthoryear{Musaelian et~al.}{2023}]{supp_Allegro}
\begin{barticle}
\bauthor{\bsnm{Musaelian}, \binits{A.}},
\bauthor{\bsnm{Batzner}, \binits{S.}},
\bauthor{\bsnm{Johansson}, \binits{A.}},
\bauthor{\bsnm{Sun}, \binits{L.}},
\bauthor{\bsnm{Owen}, \binits{C.J.}},
\bauthor{\bsnm{Kornbluth}, \binits{M.}},
\bauthor{\bsnm{Kozinsky}, \binits{B.}}:
\batitle{Learning local equivariant representations for large-scale atomistic dynamics}.
\bjtitle{Nature Communications}
\bvolume{14}(\bissue{1}),
\bfpage{579}
(\byear{2023})
\end{barticle}
\endbibitem

\bibitem[\protect\citeauthoryear{Xie et~al.}{2022}]{supp_cdvae}
\begin{bchapter}
\bauthor{\bsnm{Xie}, \binits{T.}},
\bauthor{\bsnm{Fu}, \binits{X.}},
\bauthor{\bsnm{Ganea}, \binits{O.-E.}},
\bauthor{\bsnm{Barzilay}, \binits{R.}},
\bauthor{\bsnm{Jaakkola}, \binits{T.S.}}:
\bctitle{Crystal diffusion variational autoencoder for periodic material generation}.
In: \bbtitle{International Conference on Learning Representations}
(\byear{2022})
\end{bchapter}
\endbibitem

\bibitem[\protect\citeauthoryear{Miller et~al.}{2024}]{supp_FlowMM}
\begin{bchapter}
\bauthor{\bsnm{Miller}, \binits{B.K.}},
\bauthor{\bsnm{Chen}, \binits{R.T.}},
\bauthor{\bsnm{Sriram}, \binits{A.}},
\bauthor{\bsnm{Wood}, \binits{B.M.}}:
\bctitle{Flowmm: Generating materials with riemannian flow matching}.
In: \bbtitle{International Conference on Machine Learning},
pp. \bfpage{35664}--\blpage{35686}
(\byear{2024}).
\bcomment{PMLR}
\end{bchapter}
\endbibitem

\bibitem[\protect\citeauthoryear{Luo et~al.}{2025}]{supp_CrystalFlow}
\begin{barticle}
\bauthor{\bsnm{Luo}, \binits{X.}},
\bauthor{\bsnm{Wang}, \binits{Z.}},
\bauthor{\bsnm{Wang}, \binits{Q.}},
\bauthor{\bsnm{Shao}, \binits{X.}},
\bauthor{\bsnm{Lv}, \binits{J.}},
\bauthor{\bsnm{Wang}, \binits{L.}},
\bauthor{\bsnm{Wang}, \binits{Y.}},
\bauthor{\bsnm{Ma}, \binits{Y.}}:
\batitle{Crystalflow: a flow-based generative model for crystalline materials}.
\bjtitle{Nature Communications}
\bvolume{16}(\bissue{1}),
\bfpage{9267}
(\byear{2025})
\end{barticle}
\endbibitem

\bibitem[\protect\citeauthoryear{Wu et~al.}{2025}]{supp_CrysBFN}
\begin{bchapter}
\bauthor{\bsnm{Wu}, \binits{H.}},
\bauthor{\bsnm{Song}, \binits{Y.}},
\bauthor{\bsnm{Gong}, \binits{J.}},
\bauthor{\bsnm{Cao}, \binits{Z.}},
\bauthor{\bsnm{Ouyang}, \binits{Y.}},
\bauthor{\bsnm{Zhang}, \binits{J.}},
\bauthor{\bsnm{Zhou}, \binits{H.}},
\bauthor{\bsnm{Ma}, \binits{W.-Y.}},
\bauthor{\bsnm{Liu}, \binits{J.}}:
\bctitle{A periodic bayesian flow for material generation}.
In: \bbtitle{The Thirteenth International Conference on Learning Representations}
(\byear{2025})
\end{bchapter}
\endbibitem

\bibitem[\protect\citeauthoryear{Ganin et~al.}{2006}]{supp_MoN_1}
\begin{barticle}
\bauthor{\bsnm{Ganin}, \binits{A.Y.}},
\bauthor{\bsnm{Kienle}, \binits{L.}},
\bauthor{\bsnm{Vajenine}, \binits{G.V.}}:
\batitle{Synthesis and characterisation of hexagonal molybdenum nitrides}.
\bjtitle{Journal of Solid State Chemistry}
\bvolume{179}(\bissue{8}),
\bfpage{2339}--\blpage{2348}
(\byear{2006})
\end{barticle}
\endbibitem

\bibitem[\protect\citeauthoryear{Tilley}{2019}]{supp_V3Pb}
\begin{botherref}
\oauthor{\bsnm{Tilley}, \binits{D.R.}}:
Superfluidity and superconductivity.
Routledge
(2019)
\end{botherref}
\endbibitem

\bibitem[\protect\citeauthoryear{Wang et~al.}{2015}]{supp_MoN_2}
\begin{barticle}
\bauthor{\bsnm{Wang}, \binits{S.}},
\bauthor{\bsnm{Antonio}, \binits{D.}},
\bauthor{\bsnm{Yu}, \binits{X.}},
\bauthor{\bsnm{Zhang}, \binits{J.}},
\bauthor{\bsnm{Cornelius}, \binits{A.L.}},
\bauthor{\bsnm{He}, \binits{D.}},
\bauthor{\bsnm{Zhao}, \binits{Y.}}:
\batitle{The hardest superconducting metal nitride}.
\bjtitle{Scientific reports}
\bvolume{5}(\bissue{1}),
\bfpage{13733}
(\byear{2015})
\end{barticle}
\endbibitem

\bibitem[\protect\citeauthoryear{Zhang et~al.}{2021}]{supp_Mo2C}
\begin{barticle}
\bauthor{\bsnm{Zhang}, \binits{J.}},
\bauthor{\bsnm{Cao}, \binits{Z.}},
\bauthor{\bsnm{He}, \binits{X.}},
\bauthor{\bsnm{Liu}, \binits{W.}},
\bauthor{\bsnm{Wen}, \binits{Y.}},
\bauthor{\bsnm{Cavallo}, \binits{L.}},
\bauthor{\bsnm{Ren}, \binits{W.}},
\bauthor{\bsnm{Cheng}, \binits{H.}},
\bauthor{\bsnm{Zhang}, \binits{X.}}:
\batitle{Superconductivity and high-pressure performance of 2d mo2c crystals}.
\bjtitle{The journal of physical chemistry letters}
\bvolume{12}(\bissue{9}),
\bfpage{2219}--\blpage{2225}
(\byear{2021})
\end{barticle}
\endbibitem

\bibitem[\protect\citeauthoryear{Gauzzi et~al.}{2008}]{supp_CaC6}
\begin{barticle}
\bauthor{\bsnm{Gauzzi}, \binits{A.}},
\bauthor{\bsnm{Bendiab}, \binits{N.}},
\bauthor{\bsnm{d’Astuto}, \binits{M.}},
\bauthor{\bsnm{Canny}, \binits{B.}},
\bauthor{\bsnm{Calandra}, \binits{M.}},
\bauthor{\bsnm{Mauri}, \binits{F.}},
\bauthor{\bsnm{Loupias}, \binits{G.}},
\bauthor{\bsnm{Emery}, \binits{N.}},
\bauthor{\bsnm{H{\'e}rold}, \binits{C.}},
\bauthor{\bsnm{Lagrange}, \binits{P.}}, \betal:
\batitle{Maximum t c at the verge of a simultaneous order-disorder and lattice-softening transition in superconducting cac 6}.
\bjtitle{Physical Review B—Condensed Matter and Materials Physics}
\bvolume{78}(\bissue{6}),
\bfpage{064506}
(\byear{2008})
\end{barticle}
\endbibitem

\bibitem[\protect\citeauthoryear{Yamaura et~al.}{2006}]{supp_KOs3O2}
\begin{barticle}
\bauthor{\bsnm{Yamaura}, \binits{J.-I.}},
\bauthor{\bsnm{Yonezawa}, \binits{S.}},
\bauthor{\bsnm{Muraoka}, \binits{Y.}},
\bauthor{\bsnm{Hiroi}, \binits{Z.}}:
\batitle{Crystal structure of the pyrochlore oxide superconductor kos2o6}.
\bjtitle{Journal of solid state chemistry}
\bvolume{179}(\bissue{1}),
\bfpage{336}--\blpage{340}
(\byear{2006})
\end{barticle}
\endbibitem

\bibitem[\protect\citeauthoryear{Kim et~al.}{2007}]{supp_SrC6}
\begin{barticle}
\bauthor{\bsnm{Kim}, \binits{J.}},
\bauthor{\bsnm{Boeri}, \binits{L.}},
\bauthor{\bsnm{O’Brien}, \binits{J.}},
\bauthor{\bsnm{Razavi}, \binits{F.}},
\bauthor{\bsnm{Kremer}, \binits{R.}}:
\batitle{Superconductivity in heavy alkaline-earth intercalated graphites}.
\bjtitle{Physical review letters}
\bvolume{99}(\bissue{2}),
\bfpage{027001}
(\byear{2007})
\end{barticle}
\endbibitem

\bibitem[\protect\citeauthoryear{McWhan et~al.}{1967}]{supp_CaSi2}
\begin{barticle}
\bauthor{\bsnm{McWhan}, \binits{D.}},
\bauthor{\bsnm{Compton}, \binits{V.}},
\bauthor{\bsnm{Silverman}, \binits{M.}},
\bauthor{\bsnm{Soulen}, \binits{J.}}:
\batitle{Crystal structure and superconductivity of a high-pressure phase of casi2}.
\bjtitle{Journal of the Less Common Metals}
\bvolume{12}(\bissue{1}),
\bfpage{75}--\blpage{76}
(\byear{1967})
\end{barticle}
\endbibitem

\bibitem[\protect\citeauthoryear{Zhou et~al.}{2020}]{supp_PrH9}
\begin{barticle}
\bauthor{\bsnm{Zhou}, \binits{D.}},
\bauthor{\bsnm{Semenok}, \binits{D.V.}},
\bauthor{\bsnm{Duan}, \binits{D.}},
\bauthor{\bsnm{Xie}, \binits{H.}},
\bauthor{\bsnm{Chen}, \binits{W.}},
\bauthor{\bsnm{Huang}, \binits{X.}},
\bauthor{\bsnm{Li}, \binits{X.}},
\bauthor{\bsnm{Liu}, \binits{B.}},
\bauthor{\bsnm{Oganov}, \binits{A.R.}},
\bauthor{\bsnm{Cui}, \binits{T.}}:
\batitle{Superconducting praseodymium superhydrides}.
\bjtitle{Science advances}
\bvolume{6}(\bissue{9}),
\bfpage{6849}
(\byear{2020})
\end{barticle}
\endbibitem

\bibitem[\protect\citeauthoryear{Kase and Akimitsu}{2009}]{supp_B3Ru7}
\begin{barticle}
\bauthor{\bsnm{Kase}, \binits{N.}},
\bauthor{\bsnm{Akimitsu}, \binits{J.}}:
\batitle{Superconducting state of the binary boride ru7b3 with the noncentrosymmetric crystal structure}.
\bjtitle{journal of the physical society of japan}
\bvolume{78}(\bissue{4}),
\bfpage{044710}
(\byear{2009})
\end{barticle}
\endbibitem

\bibitem[\protect\citeauthoryear{Tsukamoto et~al.}{1988}]{supp_VRu_TaRu_NbRu}
\begin{barticle}
\bauthor{\bsnm{Tsukamoto}, \binits{T.}},
\bauthor{\bsnm{Koyama}, \binits{K.}},
\bauthor{\bsnm{Oota}, \binits{A.}},
\bauthor{\bsnm{Noguchi}, \binits{S.}}:
\batitle{Superconductivity and transformation of near-equiatomic m-ru (m= v, nb and ta) alloys}.
\bjtitle{Cryogenics}
\bvolume{28}(\bissue{9}),
\bfpage{580}--\blpage{584}
(\byear{1988})
\end{barticle}
\endbibitem

\bibitem[\protect\citeauthoryear{Tang et~al.}{2015}]{supp_Rb2Cr3As3}
\begin{barticle}
\bauthor{\bsnm{Tang}, \binits{Z.-T.}},
\bauthor{\bsnm{Bao}, \binits{J.-K.}},
\bauthor{\bsnm{Liu}, \binits{Y.}},
\bauthor{\bsnm{Sun}, \binits{Y.-L.}},
\bauthor{\bsnm{Ablimit}, \binits{A.}},
\bauthor{\bsnm{Zhai}, \binits{H.-F.}},
\bauthor{\bsnm{Jiang}, \binits{H.}},
\bauthor{\bsnm{Feng}, \binits{C.-M.}},
\bauthor{\bsnm{Xu}, \binits{Z.-A.}},
\bauthor{\bsnm{Cao}, \binits{G.-H.}}:
\batitle{Unconventional superconductivity in quasi-one-dimensional rb 2 cr 3 as 3}.
\bjtitle{Physical Review B}
\bvolume{91}(\bissue{2}),
\bfpage{020506}
(\byear{2015})
\end{barticle}
\endbibitem

\bibitem[\protect\citeauthoryear{Li et~al.}{2002}]{supp_MgAlB4}
\begin{barticle}
\bauthor{\bsnm{Li}, \binits{J.}},
\bauthor{\bsnm{Li}, \binits{L.}},
\bauthor{\bsnm{Liu}, \binits{F.}},
\bauthor{\bsnm{Dong}, \binits{C.}},
\bauthor{\bsnm{Xiang}, \binits{J.}},
\bauthor{\bsnm{Zhao}, \binits{Z.}}:
\batitle{Superconductivity, superstructure, and structure anomalies in mg 1- x al x b 2}.
\bjtitle{Physical Review B}
\bvolume{65}(\bissue{13}),
\bfpage{132505}
(\byear{2002})
\end{barticle}
\endbibitem

\bibitem[\protect\citeauthoryear{Zheng et~al.}{2015}]{supp_TaBRu}
\begin{barticle}
\bauthor{\bsnm{Zheng}, \binits{Q.}},
\bauthor{\bsnm{Gumeniuk}, \binits{R.}},
\bauthor{\bsnm{Rosner}, \binits{H.}},
\bauthor{\bsnm{Schnelle}, \binits{W.}},
\bauthor{\bsnm{Prots}, \binits{Y.}},
\bauthor{\bsnm{Burkhardt}, \binits{U.}},
\bauthor{\bsnm{Grin}, \binits{Y.}},
\bauthor{\bsnm{Leithe-Jasper}, \binits{A.}}:
\batitle{Synthesis, crystal structure and properties of the new superconductors tarub and nbosb}.
\bjtitle{Journal of Physics: Condensed Matter}
\bvolume{27}(\bissue{41}),
\bfpage{415701}
(\byear{2015})
\end{barticle}
\endbibitem

\bibitem[\protect\citeauthoryear{Bull et~al.}{2004}]{supp_MoN_3}
\begin{barticle}
\bauthor{\bsnm{Bull}, \binits{C.L.}},
\bauthor{\bsnm{McMillan}, \binits{P.F.}},
\bauthor{\bsnm{Soignard}, \binits{E.}},
\bauthor{\bsnm{Leinenweber}, \binits{K.}}:
\batitle{Determination of the crystal structure of $\delta$-mon by neutron diffraction}.
\bjtitle{Journal of Solid state chemistry}
\bvolume{177}(\bissue{4-5}),
\bfpage{1488}--\blpage{1492}
(\byear{2004})
\end{barticle}
\endbibitem

\bibitem[\protect\citeauthoryear{Baublitz~Jr and Ruoff}{1983}]{supp_AlSb}
\begin{barticle}
\bauthor{\bsnm{Baublitz~Jr}, \binits{M.}},
\bauthor{\bsnm{Ruoff}, \binits{A.}}:
\batitle{X-ray diffraction data from the high pressure phase of alsb}.
\bjtitle{Journal of Applied Physics}
\bvolume{54}(\bissue{4}),
\bfpage{2109}--\blpage{2110}
(\byear{1983})
\end{barticle}
\endbibitem

\bibitem[\protect\citeauthoryear{Moodenbaugh et~al.}{1974}]{supp_ZrP}
\begin{barticle}
\bauthor{\bsnm{Moodenbaugh}, \binits{A.}},
\bauthor{\bsnm{Johnston}, \binits{D.}},
\bauthor{\bsnm{Viswanathan}, \binits{R.}}:
\batitle{Super conductivity in two nacl structure compounds: $\alpha$-zrp and scs1+ x}.
\bjtitle{Materials Research Bulletin}
\bvolume{9}(\bissue{12}),
\bfpage{1671}--\blpage{1675}
(\byear{1974})
\end{barticle}
\endbibitem

\bibitem[\protect\citeauthoryear{Lawson}{1971}]{supp_Re2W3C}
\begin{botherref}
\oauthor{\bsnm{Lawson}, \binits{A.}}:
Superconductivity of the fcc transition metals, and of their alloys, and fcc carbides.
J. Less-Common Met.;(Switzerland)
\textbf{23}
(1971)
\end{botherref}
\endbibitem

\bibitem[\protect\citeauthoryear{Tuleushev et~al.}{2003}]{supp_BeNb3}
\begin{barticle}
\bauthor{\bsnm{Tuleushev}, \binits{A.Z.}},
\bauthor{\bsnm{Volodin}, \binits{V.}},
\bauthor{\bsnm{Tuleushev}, \binits{Y.Z.}}:
\batitle{Novel superconducting niobium beryllide nb3be with a15 structure}.
\bjtitle{Journal of Experimental and Theoretical Physics Letters}
\bvolume{78}(\bissue{7}),
\bfpage{440}--\blpage{442}
(\byear{2003})
\end{barticle}
\endbibitem

\bibitem[\protect\citeauthoryear{Chen et~al.}{2021}]{supp_BaH12}
\begin{barticle}
\bauthor{\bsnm{Chen}, \binits{W.}},
\bauthor{\bsnm{Semenok}, \binits{D.V.}},
\bauthor{\bsnm{Kvashnin}, \binits{A.G.}},
\bauthor{\bsnm{Huang}, \binits{X.}},
\bauthor{\bsnm{Kruglov}, \binits{I.A.}},
\bauthor{\bsnm{Galasso}, \binits{M.}},
\bauthor{\bsnm{Song}, \binits{H.}},
\bauthor{\bsnm{Duan}, \binits{D.}},
\bauthor{\bsnm{Goncharov}, \binits{A.F.}},
\bauthor{\bsnm{Prakapenka}, \binits{V.B.}}, \betal:
\batitle{Synthesis of molecular metallic barium superhydride: pseudocubic bah12}.
\bjtitle{Nature communications}
\bvolume{12}(\bissue{1}),
\bfpage{273}
(\byear{2021})
\end{barticle}
\endbibitem

\bibitem[\protect\citeauthoryear{Iwasaki et~al.}{1982}]{supp_Nb3Si}
\begin{barticle}
\bauthor{\bsnm{Iwasaki}, \binits{H.}},
\bauthor{\bsnm{Wang}, \binits{W.}},
\bauthor{\bsnm{Toyota}, \binits{N.}},
\bauthor{\bsnm{Fukase}, \binits{T.}},
\bauthor{\bsnm{Fujimori}, \binits{H.}},
\bauthor{\bsnm{Akahama}, \binits{Y.}},
\bauthor{\bsnm{Endo}, \binits{S.}}:
\batitle{A15 nb3si produced by high-pressure annealing of amorphous sputter deposits}.
\bjtitle{Solid State Communications}
\bvolume{42}(\bissue{5}),
\bfpage{381}--\blpage{384}
(\byear{1982})
\end{barticle}
\endbibitem

\bibitem[\protect\citeauthoryear{Meisner}{1983}]{supp_HfAsRu}
\begin{barticle}
\bauthor{\bsnm{Meisner}, \binits{G.}}:
\batitle{Superconductivity and structural transformation in hfruas}.
\bjtitle{Physics Letters A}
\bvolume{96}(\bissue{9}),
\bfpage{483}--\blpage{486}
(\byear{1983})
\end{barticle}
\endbibitem

\bibitem[\protect\citeauthoryear{Petrunin et~al.}{1980}]{supp_Ta5N6}
\begin{botherref}
\oauthor{\bsnm{Petrunin}, \binits{V.}},
\oauthor{\bsnm{Sorokin}, \binits{N.}},
\oauthor{\bsnm{Borovinskaya}, \binits{I.}},
\oauthor{\bsnm{Pityulin}, \binits{A.}}:
Stability of cubic tantalum nitrides during heat treatment.
Poroshk. Metall.(Kiev);(Ukrainian SSR)
\textbf{3}
(1980)
\end{botherref}
\endbibitem

\bibitem[\protect\citeauthoryear{Juarez-Arellano et~al.}{2013}]{supp_Re7B3}
\begin{barticle}
\bauthor{\bsnm{Juarez-Arellano}, \binits{E.A.}},
\bauthor{\bsnm{Winkler}, \binits{B.}},
\bauthor{\bsnm{Friedrich}, \binits{A.}},
\bauthor{\bsnm{Bayarjargal}, \binits{L.}},
\bauthor{\bsnm{Morgenroth}, \binits{W.}},
\bauthor{\bsnm{Kunz}, \binits{M.}},
\bauthor{\bsnm{Milman}, \binits{V.}}:
\batitle{In situ study of the formation of rhenium borides from the elements at high-(p, t) conditions: Extreme incompressibility of re7b3 and formation of new phases}.
\bjtitle{Solid state sciences}
\bvolume{25},
\bfpage{85}--\blpage{92}
(\byear{2013})
\end{barticle}
\endbibitem

\bibitem[\protect\citeauthoryear{Gillan and Kaner}{1994}]{supp_NbN}
\begin{barticle}
\bauthor{\bsnm{Gillan}, \binits{E.G.}},
\bauthor{\bsnm{Kaner}, \binits{R.B.}}:
\batitle{Rapid solid-state synthesis of refractory nitrides}.
\bjtitle{Inorganic chemistry}
\bvolume{33}(\bissue{25}),
\bfpage{5693}--\blpage{5700}
(\byear{1994})
\end{barticle}
\endbibitem

\bibitem[\protect\citeauthoryear{Shang et~al.}{2018}]{supp_Nb5Re24}
\begin{barticle}
\bauthor{\bsnm{Shang}, \binits{T.}},
\bauthor{\bsnm{Smidman}, \binits{M.}},
\bauthor{\bsnm{Ghosh}, \binits{S.K.}},
\bauthor{\bsnm{Baines}, \binits{C.}},
\bauthor{\bsnm{Chang}, \binits{L.-J.}},
\bauthor{\bsnm{Gawryluk}, \binits{D.}},
\bauthor{\bsnm{Barker}, \binits{J.A.}},
\bauthor{\bsnm{Singh}, \binits{R.P.}},
\bauthor{\bsnm{Paul}, \binits{D.M.}},
\bauthor{\bsnm{Balakrishnan}, \binits{G.}}, \betal:
\batitle{Time-reversal symmetry breaking in re-based superconductors}.
\bjtitle{Physical review letters}
\bvolume{121}(\bissue{25}),
\bfpage{257002}
(\byear{2018})
\end{barticle}
\endbibitem

\bibitem[\protect\citeauthoryear{Flukiger et~al.}{1974}]{supp_Cr3Os}
\begin{barticle}
\bauthor{\bsnm{Flukiger}, \binits{R.}},
\bauthor{\bsnm{Paoli}, \binits{A.}},
\bauthor{\bsnm{Muller}, \binits{J.}}:
\batitle{Electronically ‘atypical’a 15-type compounds based on chromium and molybdenum}.
\bjtitle{Solid State Communications}
\bvolume{14}(\bissue{6}),
\bfpage{443}--\blpage{447}
(\byear{1974})
\end{barticle}
\endbibitem

\bibitem[\protect\citeauthoryear{Darby~Jr et~al.}{1965}]{supp_ThTc2}
\begin{barticle}
\bauthor{\bsnm{Darby~Jr}, \binits{J.}},
\bauthor{\bsnm{Berndt}, \binits{A.}},
\bauthor{\bsnm{Downey}, \binits{J.}}:
\batitle{Some intermediate phases in the thorium-technetium and uranium-technetium systems}.
\bjtitle{Journal of the Less Common Metals}
\bvolume{9}(\bissue{6}),
\bfpage{466}--\blpage{468}
(\byear{1965})
\end{barticle}
\endbibitem

\bibitem[\protect\citeauthoryear{Yosida and Oguro}{2007}]{supp_NbC}
\begin{barticle}
\bauthor{\bsnm{Yosida}, \binits{Y.}},
\bauthor{\bsnm{Oguro}, \binits{I.}}:
\batitle{Surface superconductivity in nanocrystals nbcx encapsulated in the multiwall carbon cages}.
\bjtitle{Physica C: Superconductivity}
\bvolume{453}(\bissue{1-2}),
\bfpage{52}--\blpage{56}
(\byear{2007})
\end{barticle}
\endbibitem

\bibitem[\protect\citeauthoryear{Meisner and Ku}{1983}]{supp_ZrPRu_2_ZrPOs}
\begin{barticle}
\bauthor{\bsnm{Meisner}, \binits{G.}},
\bauthor{\bsnm{Ku}, \binits{H.}}:
\batitle{The superconductivity and structure of equiatomic ternary transition metal pnictides}.
\bjtitle{Applied Physics A}
\bvolume{31}(\bissue{4}),
\bfpage{201}--\blpage{212}
(\byear{1983})
\end{barticle}
\endbibitem

\bibitem[\protect\citeauthoryear{Willens et~al.}{1967}]{supp_MoC}
\begin{barticle}
\bauthor{\bsnm{Willens}, \binits{R.}},
\bauthor{\bsnm{Buehler}, \binits{E.}},
\bauthor{\bsnm{Matthias}, \binits{B.}}:
\batitle{Superconductivity of the transition-metal carbides}.
\bjtitle{Physical Review}
\bvolume{159}(\bissue{2}),
\bfpage{327}
(\byear{1967})
\end{barticle}
\endbibitem

\bibitem[\protect\citeauthoryear{Li et~al.}{2017}]{supp_AlV2N}
\begin{barticle}
\bauthor{\bsnm{Li}, \binits{J.}},
\bauthor{\bsnm{Cui}, \binits{X.}},
\bauthor{\bsnm{Jin}, \binits{Y.}},
\bauthor{\bsnm{Yin}, \binits{X.}},
\bauthor{\bsnm{Cao}, \binits{S.}},
\bauthor{\bsnm{Feng}, \binits{Z.}},
\bauthor{\bsnm{Zhang}, \binits{J.}}:
\batitle{Appearance of superconductivity at 15.9 k in layered v2aln}.
\bjtitle{Journal of Superconductivity and Novel Magnetism}
\bvolume{30}(\bissue{1}),
\bfpage{1}--\blpage{4}
(\byear{2017})
\end{barticle}
\endbibitem

\bibitem[\protect\citeauthoryear{Wood et~al.}{1958}]{supp_V3Ga}
\begin{barticle}
\bauthor{\bsnm{Wood}, \binits{E.A.}},
\bauthor{\bsnm{Compton}, \binits{V.B.}},
\bauthor{\bsnm{Matthias}, \binits{B.}},
\bauthor{\bsnm{Corenzwit}, \binits{E.}}:
\batitle{$\beta$-wolfram structure of compounds between transition elements and aluminum, gallium and antimony}.
\bjtitle{Acta Crystallographica}
\bvolume{11}(\bissue{9}),
\bfpage{604}--\blpage{606}
(\byear{1958})
\end{barticle}
\endbibitem

\bibitem[\protect\citeauthoryear{Zhao et~al.}{2020}]{supp_Rb2Mo3As3}
\begin{barticle}
\bauthor{\bsnm{Zhao}, \binits{K.}},
\bauthor{\bsnm{Mu}, \binits{Q.-G.}},
\bauthor{\bsnm{Ruan}, \binits{B.-B.}},
\bauthor{\bsnm{Zhou}, \binits{M.-H.}},
\bauthor{\bsnm{Yang}, \binits{Q.-S.}},
\bauthor{\bsnm{Liu}, \binits{T.}},
\bauthor{\bsnm{Pan}, \binits{B.-J.}},
\bauthor{\bsnm{Zhang}, \binits{S.}},
\bauthor{\bsnm{Chen}, \binits{G.-F.}},
\bauthor{\bsnm{Ren}, \binits{Z.-A.}}:
\batitle{A new quasi-one-dimensional ternary molybdenum pnictide rb2mo3as3 with superconducting transition at 10.5 k}.
\bjtitle{Chinese Physics Letters}
\bvolume{37}(\bissue{9}),
\bfpage{097401}
(\byear{2020})
\end{barticle}
\endbibitem

\bibitem[\protect\citeauthoryear{Yen et~al.}{1967}]{supp_HfN}
\begin{barticle}
\bauthor{\bsnm{Yen}, \binits{C.}},
\bauthor{\bsnm{Toth}, \binits{L.}},
\bauthor{\bsnm{Shy}, \binits{Y.}},
\bauthor{\bsnm{Anderson}, \binits{D.}},
\bauthor{\bsnm{Rosner}, \binits{L.}}:
\batitle{Superconducting h c-j c and t c measurements in the nb--ti--n, nb--hf--n, and nb--v--n ternary systems}.
\bjtitle{Journal of Applied Physics}
\bvolume{38}(\bissue{5}),
\bfpage{2268}--\blpage{2271}
(\byear{1967})
\end{barticle}
\endbibitem

\bibitem[\protect\citeauthoryear{Kortan et~al.}{1994}]{supp_SrC10}
\begin{barticle}
\bauthor{\bsnm{Kortan}, \binits{A.}},
\bauthor{\bsnm{Kopylov}, \binits{N.}},
\bauthor{\bsnm{{\"O}zdas}, \binits{E.}},
\bauthor{\bsnm{Ramirez}, \binits{A.}},
\bauthor{\bsnm{Fleming}, \binits{R.}},
\bauthor{\bsnm{Haddon}, \binits{R.}}:
\batitle{Strontium doped fullerite compounds}.
\bjtitle{Chemical physics letters}
\bvolume{223}(\bissue{5-6}),
\bfpage{501}--\blpage{505}
(\byear{1994})
\end{barticle}
\endbibitem

\bibitem[\protect\citeauthoryear{Yu et~al.}{2012}]{supp_Nb3Al}
\begin{barticle}
\bauthor{\bsnm{Yu}, \binits{Z.}},
\bauthor{\bsnm{Li}, \binits{C.}},
\bauthor{\bsnm{Liu}, \binits{H.}}:
\batitle{Compressibility anomaly in the superconducting material nb3al under high pressure}.
\bjtitle{Physica B: Condensed Matter}
\bvolume{407}(\bissue{17}),
\bfpage{3635}--\blpage{3638}
(\byear{2012})
\end{barticle}
\endbibitem

\bibitem[\protect\citeauthoryear{Jorda et~al.}{1988}]{supp_ZrRh_1}
\begin{barticle}
\bauthor{\bsnm{Jorda}, \binits{J.}},
\bauthor{\bsnm{Graf}, \binits{T.}},
\bauthor{\bsnm{Schellenberg}, \binits{L.}},
\bauthor{\bsnm{Muller}, \binits{J.}},
\bauthor{\bsnm{Cenzual}, \binits{K.}},
\bauthor{\bsnm{Gachon}, \binits{J.}},
\bauthor{\bsnm{Hertz}, \binits{J.}}:
\batitle{Phase relations, thermochemistry and superconductivity in the zr rh system}.
\bjtitle{Journal of the Less Common Metals}
\bvolume{136}(\bissue{2}),
\bfpage{313}--\blpage{328}
(\byear{1988})
\end{barticle}
\endbibitem

\bibitem[\protect\citeauthoryear{Romero et~al.}{2013}]{supp_Nb2CS}
\begin{barticle}
\bauthor{\bsnm{Romero}, \binits{M.}},
\bauthor{\bsnm{Huerta}, \binits{L.}},
\bauthor{\bsnm{Akachi}, \binits{T.}},
\bauthor{\bsnm{Llamazares}, \binits{J.S.}},
\bauthor{\bsnm{Escamilla}, \binits{R.}}:
\batitle{X-ray photoelectron spectroscopy studies of the electronic structure of superconducting nb2snc and nb2sc}.
\bjtitle{Journal of alloys and compounds}
\bvolume{579},
\bfpage{516}--\blpage{520}
(\byear{2013})
\end{barticle}
\endbibitem

\bibitem[\protect\citeauthoryear{Narayan and Finnemore}{1978}]{supp_NbS_ScS}
\begin{barticle}
\bauthor{\bsnm{Narayan}, \binits{P.}},
\bauthor{\bsnm{Finnemore}, \binits{D.}}:
\batitle{Superconductivity in the niobium and scandium monosulfide systems at pressures up to 20 kbar}.
\bjtitle{Journal of the Less Common Metals}
\bvolume{61}(\bissue{2}),
\bfpage{231}--\blpage{235}
(\byear{1978})
\end{barticle}
\endbibitem

\bibitem[\protect\citeauthoryear{Uehara et~al.}{2006}]{supp_CdNi3C}
\begin{barticle}
\bauthor{\bsnm{Uehara}, \binits{M.}},
\bauthor{\bsnm{Amano}, \binits{T.}},
\bauthor{\bsnm{Takano}, \binits{S.}},
\bauthor{\bsnm{K{\^o}ri}, \binits{T.}},
\bauthor{\bsnm{Yamazaki}, \binits{T.}},
\bauthor{\bsnm{Kimishima}, \binits{Y.}}:
\batitle{Chemical pressure effect on the superconductor mgcni3}.
\bjtitle{Physica C: Superconductivity}
\bvolume{440}(\bissue{1-2}),
\bfpage{6}--\blpage{9}
(\byear{2006})
\end{barticle}
\endbibitem

\bibitem[\protect\citeauthoryear{Hui et~al.}{2014}]{supp_CuNi3N}
\begin{barticle}
\bauthor{\bsnm{Hui}, \binits{Z.}},
\bauthor{\bsnm{Tang}, \binits{X.}},
\bauthor{\bsnm{Shao}, \binits{D.}},
\bauthor{\bsnm{Lei}, \binits{H.}},
\bauthor{\bsnm{Yang}, \binits{J.}},
\bauthor{\bsnm{Song}, \binits{W.}},
\bauthor{\bsnm{Luo}, \binits{H.}},
\bauthor{\bsnm{Zhu}, \binits{X.}},
\bauthor{\bsnm{Sun}, \binits{Y.}}:
\batitle{Epitaxial antiperovskite superconducting cunni 3 thin films synthesized by chemical solution deposition}.
\bjtitle{Chemical Communications}
\bvolume{50}(\bissue{84}),
\bfpage{12734}--\blpage{12737}
(\byear{2014})
\end{barticle}
\endbibitem

\bibitem[\protect\citeauthoryear{Poineau et~al.}{2010}]{supp_Zr4Tc25}
\begin{barticle}
\bauthor{\bsnm{Poineau}, \binits{F.}},
\bauthor{\bsnm{Hartmann}, \binits{T.}},
\bauthor{\bsnm{Weck}, \binits{P.F.}},
\bauthor{\bsnm{Kim}, \binits{E.}},
\bauthor{\bsnm{Silva}, \binits{G.C.}},
\bauthor{\bsnm{Jarvinen}, \binits{G.D.}},
\bauthor{\bsnm{Czerwinski}, \binits{K.R.}}:
\batitle{Structural studies of technetium- zirconium alloys by x-ray diffraction, high-resolution electron microscopy, and first-principles calculations}.
\bjtitle{Inorganic chemistry}
\bvolume{49}(\bissue{4}),
\bfpage{1433}--\blpage{1438}
(\byear{2010})
\end{barticle}
\endbibitem

\bibitem[\protect\citeauthoryear{McWhan and Marezio}{1966}]{supp_InSb}
\begin{barticle}
\bauthor{\bsnm{McWhan}, \binits{D.}},
\bauthor{\bsnm{Marezio}, \binits{M.}}:
\batitle{Structure and superconductivity of the high-pressure phases of insb}.
\bjtitle{The Journal of Chemical Physics}
\bvolume{45}(\bissue{7}),
\bfpage{2508}--\blpage{2511}
(\byear{1966})
\end{barticle}
\endbibitem

\bibitem[\protect\citeauthoryear{Syu et~al.}{2013}]{supp_Zr2Co}
\begin{barticle}
\bauthor{\bsnm{Syu}, \binits{K.}},
\bauthor{\bsnm{Chen}, \binits{S.}},
\bauthor{\bsnm{Wu}, \binits{H.}},
\bauthor{\bsnm{Sung}, \binits{H.}},
\bauthor{\bsnm{Lee}, \binits{W.}}:
\batitle{Superconductivity in zr2 (co1- xcux)}.
\bjtitle{Physica C: Superconductivity}
\bvolume{495},
\bfpage{10}--\blpage{14}
(\byear{2013})
\end{barticle}
\endbibitem

\bibitem[\protect\citeauthoryear{Bao et~al.}{2015}]{supp_K2Cr3As3}
\begin{barticle}
\bauthor{\bsnm{Bao}, \binits{J.-K.}},
\bauthor{\bsnm{Liu}, \binits{J.-Y.}},
\bauthor{\bsnm{Ma}, \binits{C.-W.}},
\bauthor{\bsnm{Meng}, \binits{Z.-H.}},
\bauthor{\bsnm{Tang}, \binits{Z.-T.}},
\bauthor{\bsnm{Sun}, \binits{Y.-L.}},
\bauthor{\bsnm{Zhai}, \binits{H.-F.}},
\bauthor{\bsnm{Jiang}, \binits{H.}},
\bauthor{\bsnm{Bai}, \binits{H.}},
\bauthor{\bsnm{Feng}, \binits{C.-M.}}, \betal:
\batitle{Superconductivity in quasi-one-dimensional k 2 cr 3 as 3 with significant electron correlations}.
\bjtitle{Physical Review X}
\bvolume{5}(\bissue{1}),
\bfpage{011013}
(\byear{2015})
\end{barticle}
\endbibitem

\bibitem[\protect\citeauthoryear{Yan et~al.}{2021}]{supp_TaC}
\begin{barticle}
\bauthor{\bsnm{Yan}, \binits{D.Y.}},
\bauthor{\bsnm{Yang}, \binits{M.}},
\bauthor{\bsnm{Wang}, \binits{C.}},
\bauthor{\bsnm{Song}, \binits{P.}},
\bauthor{\bsnm{Yi}, \binits{C.}},
\bauthor{\bsnm{Shi}, \binits{Y.}}:
\batitle{Superconductivity in centrosymmetric topological superconductor candidate tac}.
\bjtitle{Superconductor Science and Technology}
\bvolume{34}(\bissue{3}),
\bfpage{035025}
(\byear{2021})
\end{barticle}
\endbibitem

\bibitem[\protect\citeauthoryear{Benndorf et~al.}{2017}]{supp_NbSiOs_TaSiOs}
\begin{barticle}
\bauthor{\bsnm{Benndorf}, \binits{C.}},
\bauthor{\bsnm{Heletta}, \binits{L.}},
\bauthor{\bsnm{Heymann}, \binits{G.}},
\bauthor{\bsnm{Huppertz}, \binits{H.}},
\bauthor{\bsnm{Eckert}, \binits{H.}},
\bauthor{\bsnm{P{\"o}ttgen}, \binits{R.}}:
\batitle{Nbossi and taossi--two new superconducting ternary osmium silicides}.
\bjtitle{Solid State Sciences}
\bvolume{68},
\bfpage{32}--\blpage{38}
(\byear{2017})
\end{barticle}
\endbibitem

\bibitem[\protect\citeauthoryear{Kayhan et~al.}{2012}]{supp_BW_B5W2}
\begin{barticle}
\bauthor{\bsnm{Kayhan}, \binits{M.}},
\bauthor{\bsnm{Hildebrandt}, \binits{E.}},
\bauthor{\bsnm{Frotscher}, \binits{M.}},
\bauthor{\bsnm{Senyshyn}, \binits{A.}},
\bauthor{\bsnm{Hofmann}, \binits{K.}},
\bauthor{\bsnm{Alff}, \binits{L.}},
\bauthor{\bsnm{Albert}, \binits{B.}}:
\batitle{Neutron diffraction and observation of superconductivity for tungsten borides, wb and w2b4}.
\bjtitle{Solid state sciences}
\bvolume{14}(\bissue{11-12}),
\bfpage{1656}--\blpage{1659}
(\byear{2012})
\end{barticle}
\endbibitem

\bibitem[\protect\citeauthoryear{Schneidmiller et~al.}{1997}]{supp_Mo3Se}
\begin{barticle}
\bauthor{\bsnm{Schneidmiller}, \binits{R.}},
\bauthor{\bsnm{Hornbostel}, \binits{M.D.}},
\bauthor{\bsnm{Johnson}, \binits{D.C.}}:
\batitle{Kinetics of formation of molybdenum selenides from modulated reactants and structure of the new compound mo3se}.
\bjtitle{Inorganic chemistry}
\bvolume{36}(\bissue{25}),
\bfpage{5894}--\blpage{5899}
(\byear{1997})
\end{barticle}
\endbibitem

\bibitem[\protect\citeauthoryear{Schaak and Cava}{2004}]{supp_ZrPRu_1}
\begin{barticle}
\bauthor{\bsnm{Schaak}, \binits{R.}},
\bauthor{\bsnm{Cava}, \binits{R.}}:
\batitle{Boron substitution in ternary metal phosphide superconductors}.
\bjtitle{Materials research bulletin}
\bvolume{39}(\bissue{9}),
\bfpage{1231}--\blpage{1235}
(\byear{2004})
\end{barticle}
\endbibitem

\bibitem[\protect\citeauthoryear{Borsa and Olcese}{1973}]{supp_LaBe13}
\begin{barticle}
\bauthor{\bsnm{Borsa}, \binits{F.}},
\bauthor{\bsnm{Olcese}, \binits{G.}}:
\batitle{Magnetic properties, hyperfine interactions, and spin dynamics of rare earth--beryllium intermetallic compounds}.
\bjtitle{physica status solidi (a)}
\bvolume{17}(\bissue{2}),
\bfpage{631}--\blpage{642}
(\byear{1973})
\end{barticle}
\endbibitem

\bibitem[\protect\citeauthoryear{Zhou et~al.}{2020}]{supp_NdH9}
\begin{barticle}
\bauthor{\bsnm{Zhou}, \binits{D.}},
\bauthor{\bsnm{Semenok}, \binits{D.V.}},
\bauthor{\bsnm{Xie}, \binits{H.}},
\bauthor{\bsnm{Huang}, \binits{X.}},
\bauthor{\bsnm{Duan}, \binits{D.}},
\bauthor{\bsnm{Aperis}, \binits{A.}},
\bauthor{\bsnm{Oppeneer}, \binits{P.M.}},
\bauthor{\bsnm{Galasso}, \binits{M.}},
\bauthor{\bsnm{Kartsev}, \binits{A.I.}},
\bauthor{\bsnm{Kvashnin}, \binits{A.G.}}, \betal:
\batitle{High-pressure synthesis of magnetic neodymium polyhydrides}.
\bjtitle{Journal of the American Chemical Society}
\bvolume{142}(\bissue{6}),
\bfpage{2803}--\blpage{2811}
(\byear{2020})
\end{barticle}
\endbibitem

\bibitem[\protect\citeauthoryear{Zhao et~al.}{1995}]{supp_La3InB}
\begin{barticle}
\bauthor{\bsnm{Zhao}, \binits{J.-T.}},
\bauthor{\bsnm{Dong}, \binits{Z.-C.}},
\bauthor{\bsnm{Vaughey}, \binits{J.}},
\bauthor{\bsnm{Ostenson}, \binits{J.E.}},
\bauthor{\bsnm{Corbett}, \binits{J.D.}}:
\batitle{Synthesis, structures and properties of cubic r3in and r3inz phases (r= y, la; z= b, c, n, o): the effect of interstitial z on the superconductivity of la3in}.
\bjtitle{Journal of alloys and compounds}
\bvolume{230}(\bissue{1}),
\bfpage{1}--\blpage{12}
(\byear{1995})
\end{barticle}
\endbibitem

\bibitem[\protect\citeauthoryear{Shirotani et~al.}{1997}]{supp_ZrP2Ru4}
\begin{barticle}
\bauthor{\bsnm{Shirotani}, \binits{I.}},
\bauthor{\bsnm{Iwasaki}, \binits{G.}},
\bauthor{\bsnm{Isamu}, \binits{K.}},
\bauthor{\bsnm{Sekine}, \binits{C.}},
\bauthor{\bsnm{Todo}, \binits{S.}},
\bauthor{\bsnm{Yagi}, \binits{T.}}:
\batitle{Electrical conductivity and superconductivity of zrni4p2 and zrru4p2 prepared at high pressure}.
\bjtitle{Solid state communications}
\bvolume{104}(\bissue{4}),
\bfpage{217}--\blpage{221}
(\byear{1997})
\end{barticle}
\endbibitem

\bibitem[\protect\citeauthoryear{Stolze et~al.}{2018}]{supp_ZrRh_2}
\begin{barticle}
\bauthor{\bsnm{Stolze}, \binits{K.}},
\bauthor{\bsnm{Tao}, \binits{J.}},
\bauthor{\bsnm{Von~Rohr}, \binits{F.O.}},
\bauthor{\bsnm{Kong}, \binits{T.}},
\bauthor{\bsnm{Cava}, \binits{R.J.}}:
\batitle{Sc--zr--nb--rh--pd and sc--zr--nb--ta--rh--pd high-entropy alloy superconductors on a cscl-type lattice}.
\bjtitle{Chemistry of Materials}
\bvolume{30}(\bissue{3}),
\bfpage{906}--\blpage{914}
(\byear{2018})
\end{barticle}
\endbibitem

\bibitem[\protect\citeauthoryear{Xie et~al.}{2015}]{supp_NbBRu}
\begin{barticle}
\bauthor{\bsnm{Xie}, \binits{W.}},
\bauthor{\bsnm{Luo}, \binits{H.}},
\bauthor{\bsnm{Baroudi}, \binits{K.}},
\bauthor{\bsnm{Krizan}, \binits{J.W.}},
\bauthor{\bsnm{Phelan}, \binits{B.F.}},
\bauthor{\bsnm{Cava}, \binits{R.J.}}:
\batitle{Fragment-based design of nbrub as a new metal-rich boride superconductor}.
\bjtitle{Chemistry of Materials}
\bvolume{27}(\bissue{4}),
\bfpage{1149}--\blpage{1152}
(\byear{2015})
\end{barticle}
\endbibitem

\bibitem[\protect\citeauthoryear{Lee et~al.}{2008}]{supp_CaAlSi}
\begin{barticle}
\bauthor{\bsnm{Lee}, \binits{M.H.}},
\bauthor{\bsnm{Bjorling}, \binits{T.}},
\bauthor{\bsnm{Hauback}, \binits{B.}},
\bauthor{\bsnm{Utsumi}, \binits{T.}},
\bauthor{\bsnm{Moser}, \binits{D.}},
\bauthor{\bsnm{Bull}, \binits{D.}},
\bauthor{\bsnm{Noreus}, \binits{D.}},
\bauthor{\bsnm{Sankey}, \binits{O.F.}},
\bauthor{\bsnm{Haussermann}, \binits{U.}}:
\batitle{Crystal structure, electronic structure, and vibrational properties of malsih (m= ca, sr, ba): Hydrogenation-induced semiconductors from the alb2-type alloys malsi}.
\bjtitle{Physical Review B Condensed Matter And Materials Physics}
\bvolume{78}(\bissue{19}),
\bfpage{195209}
(\byear{2008})
\end{barticle}
\endbibitem

\bibitem[\protect\citeauthoryear{Kordan et~al.}{2019}]{supp_AlV3}
\begin{barticle}
\bauthor{\bsnm{Kordan}, \binits{V.}},
\bauthor{\bsnm{Zhyshkovych}, \binits{O.}},
\bauthor{\bsnm{Zelinska}, \binits{O.}},
\bauthor{\bsnm{Tarasiuk}, \binits{I.}},
\bauthor{\bsnm{Pavlyuk}, \binits{V.}},
\bauthor{\bsnm{Serkiz}, \binits{R.}}:
\batitle{Pecularities of electrochemical lithiation of the binary intermetallics of the systems $\{$Ti, V$\}$--al, visnyk lviv univ}.
\bjtitle{Ser. Chem}
\bvolume{60}(\bissue{1}),
\bfpage{127}--\blpage{139}
(\byear{2019})
\end{barticle}
\endbibitem

\bibitem[\protect\citeauthoryear{G{\'o}rnicka et~al.}{2021}]{supp_LiGa2Ir}
\begin{barticle}
\bauthor{\bsnm{G{\'o}rnicka}, \binits{K.}},
\bauthor{\bsnm{Kuderowicz}, \binits{G.}},
\bauthor{\bsnm{Winiarski}, \binits{M.J.}},
\bauthor{\bsnm{Wiendlocha}, \binits{B.}},
\bauthor{\bsnm{Klimczuk}, \binits{T.}}:
\batitle{Superconductivity in liga2ir heusler type compound with vec= 16}.
\bjtitle{Scientific Reports}
\bvolume{11}(\bissue{1}),
\bfpage{16517}
(\byear{2021})
\end{barticle}
\endbibitem

\bibitem[\protect\citeauthoryear{Blokhin and Villars}{}]{supp_mpds}
\begin{botherref}
\oauthor{\bsnm{Blokhin}, \binits{E.}},
\oauthor{\bsnm{Villars}, \binits{P.}}:
MPDS: Materials Platform for Data Science.
\url{https://mpds.io/}
\end{botherref}
\endbibitem

\bibitem[\protect\citeauthoryear{Yan et~al.}{2022}]{supp_matformer}
\begin{barticle}
\bauthor{\bsnm{Yan}, \binits{K.}},
\bauthor{\bsnm{Liu}, \binits{Y.}},
\bauthor{\bsnm{Lin}, \binits{Y.}},
\bauthor{\bsnm{Ji}, \binits{S.}}:
\batitle{Periodic graph transformers for crystal material property prediction}.
\bjtitle{Advances in Neural Information Processing Systems}
\bvolume{35},
\bfpage{15066}--\blpage{15080}
(\byear{2022})
\end{barticle}
\endbibitem

\bibitem[\protect\citeauthoryear{Kresse and Hafner}{1993}]{supp_vasp1}
\begin{barticle}
\bauthor{\bsnm{Kresse}, \binits{G.}},
\bauthor{\bsnm{Hafner}, \binits{J.}}:
\batitle{Ab initio molecular dynamics for liquid metals}.
\bjtitle{Phys. Rev. B}
\bvolume{47},
\bfpage{558}--\blpage{561}
(\byear{1993})
\doiurl{10.1103/PhysRevB.47.558}
\end{barticle}
\endbibitem

\bibitem[\protect\citeauthoryear{Kresse and Furthm{\"u}ller}{1996}]{supp_vasp2}
\begin{barticle}
\bauthor{\bsnm{Kresse}, \binits{G.}},
\bauthor{\bsnm{Furthm{\"u}ller}, \binits{J.}}:
\batitle{Efficient iterative schemes for ab initio total-energy calculations using a plane-wave basis set}.
\bjtitle{Phys. Rev. B}
\bvolume{54},
\bfpage{11169}--\blpage{11186}
(\byear{1996})
\doiurl{10.1103/PhysRevB.54.11169}
\end{barticle}
\endbibitem

\bibitem[\protect\citeauthoryear{Perdew et~al.}{1996}]{supp_pbe}
\begin{barticle}
\bauthor{\bsnm{Perdew}, \binits{J.P.}},
\bauthor{\bsnm{Burke}, \binits{K.}},
\bauthor{\bsnm{Ernzerhof}, \binits{M.}}:
\batitle{Generalized gradient approximation made simple}.
\bjtitle{Phys. Rev. Lett.}
\bvolume{77},
\bfpage{3865}--\blpage{3868}
(\byear{1996})
\doiurl{10.1103/PhysRevLett.77.3865}
\end{barticle}
\endbibitem

\bibitem[\protect\citeauthoryear{Bl{\"o}chl}{1994}]{supp_paw1}
\begin{barticle}
\bauthor{\bsnm{Bl{\"o}chl}, \binits{P.E.}}:
\batitle{Projector augmented-wave method}.
\bjtitle{Phys. Rev. B}
\bvolume{50},
\bfpage{17953}--\blpage{17979}
(\byear{1994})
\doiurl{10.1103/PhysRevB.50.17953}
\end{barticle}
\endbibitem

\bibitem[\protect\citeauthoryear{Kresse and Joubert}{1999}]{supp_paw2}
\begin{barticle}
\bauthor{\bsnm{Kresse}, \binits{G.}},
\bauthor{\bsnm{Joubert}, \binits{D.}}:
\batitle{From ultrasoft pseudopotentials to the projector augmented-wave method}.
\bjtitle{Phys. Rev. B}
\bvolume{59},
\bfpage{1758}--\blpage{1775}
(\byear{1999})
\doiurl{10.1103/PhysRevB.59.1758}
\end{barticle}
\endbibitem

\bibitem[\protect\citeauthoryear{Monkhorst and Pack}{1976}]{supp_monkhorst_pack}
\begin{barticle}
\bauthor{\bsnm{Monkhorst}, \binits{H.J.}},
\bauthor{\bsnm{Pack}, \binits{J.D.}}:
\batitle{Special points for {B}rillouin-zone integrations}.
\bjtitle{Phys. Rev. B}
\bvolume{13},
\bfpage{5188}--\blpage{5192}
(\byear{1976})
\doiurl{10.1103/PhysRevB.13.5188}
\end{barticle}
\endbibitem

\bibitem[\protect\citeauthoryear{Baroni et~al.}{2001}]{supp_baroni_dfpt}
\begin{barticle}
\bauthor{\bsnm{Baroni}, \binits{S.}},
\bauthor{\bsnm{Gironcoli}, \binits{S.}},
\bauthor{\bsnm{Dal~Corso}, \binits{A.}},
\bauthor{\bsnm{Giannozzi}, \binits{P.}}:
\batitle{Phonons and related crystal properties from density-functional perturbation theory}.
\bjtitle{Rev. Mod. Phys.}
\bvolume{73},
\bfpage{515}--\blpage{562}
(\byear{2001})
\doiurl{10.1103/RevModPhys.73.515}
\end{barticle}
\endbibitem

\bibitem[\protect\citeauthoryear{Giannozzi et~al.}{2017}]{supp_qe}
\begin{barticle}
\bauthor{\bsnm{Giannozzi}, \binits{P.}},
\bauthor{\bsnm{Andreussi}, \binits{O.}},
\bauthor{\bsnm{Brumme}, \binits{T.}},
\bauthor{\bsnm{Bunau}, \binits{O.}},
\bauthor{\bsnm{Nardelli}, \binits{M.B.}},
\bauthor{\bsnm{Calandra}, \binits{M.}},
\bauthor{\bsnm{Car}, \binits{R.}},
\bauthor{\bsnm{Cavazzoni}, \binits{C.}},
\bauthor{\bsnm{Ceresoli}, \binits{D.}},
\bauthor{\bsnm{Cococcioni}, \binits{M.}}, \betal:
\batitle{Advanced capabilities for materials modelling with quantum espresso}.
\bjtitle{Journal of physics: Condensed matter}
\bvolume{29}(\bissue{46}),
\bfpage{465901}
(\byear{2017})
\end{barticle}
\endbibitem

\bibitem[\protect\citeauthoryear{Vanderbilt}{1990}]{supp_vanderbilt_uspp}
\begin{barticle}
\bauthor{\bsnm{Vanderbilt}, \binits{D.}}:
\batitle{Soft self-consistent pseudopotentials in a generalized eigenvalue formalism}.
\bjtitle{Phys. Rev. B}
\bvolume{41},
\bfpage{7892}--\blpage{7895}
(\byear{1990})
\doiurl{10.1103/PhysRevB.41.7892}
\end{barticle}
\endbibitem

\bibitem[\protect\citeauthoryear{Prozorov and Kogan}{2018}]{supp_demag}
\begin{barticle}
\bauthor{\bsnm{Prozorov}, \binits{R.}},
\bauthor{\bsnm{Kogan}, \binits{V.G.}}:
\batitle{Effective demagnetizing factors of diamagnetic samples of various shapes}.
\bjtitle{Physical review applied}
\bvolume{10}(\bissue{1}),
\bfpage{014030}
(\byear{2018})
\end{barticle}
\endbibitem

\end{thebibliography}
\end{document}